\documentclass{article}

\PassOptionsToPackage{numbers, compress}{natbib}

\usepackage[preprint]{neurips_2025}

\usepackage[utf8]{inputenc} 
\usepackage[T1]{fontenc}    
\usepackage{hyperref}       
\usepackage{url}            
\usepackage{booktabs}       
\usepackage{amsfonts}       
\usepackage{nicefrac}       
\usepackage{microtype}      
\usepackage{xcolor}         

\usepackage{diagbox}
\usepackage{amsmath, amsthm, amssymb} 
\usepackage{subcaption, graphicx, multirow, arydshln}
\usepackage{tikz}
\usepackage{float}
\usepackage{placeins}
\usetikzlibrary{arrows.meta, positioning}
\usepackage{enumitem}
\setlist[enumerate]{itemsep=1pt, parsep=0pt, topsep=2pt, partopsep=0pt}

\newtheorem{theorem}{Theorem} 
\newtheorem{lemma}[theorem]{Lemma}

\newcommand{\norm}[1]{{\|{#1}\|}}
\def \bR {\mathbb{R}}

\title{Computational Advantages of Multi-Grade Deep Learning: Convergence Analysis and Performance Insights}

\author{%
  Ronglong Fang, Yuesheng Xu \\
  Department of Mathematics and Statistics, Old Dominion University\\
  \texttt{\{rfang002, y1xu\}@odu.edu}
}

\begin{document}

\maketitle

\begin{abstract}
Multi-grade deep learning (MGDL) has been shown to significantly outperform the standard single-grade deep learning (SGDL) across various applications. This work aims to investigate the computational advantages of MGDL focusing on its performance in image regression, denoising, and deblurring tasks, and comparing it to  SGDL. We establish convergence results for the gradient descent (GD) method applied to these models and provide mathematical insights into MGDL’s improved performance. In particular, we demonstrate that MGDL is more robust to the choice of learning rate under GD than SGDL. Furthermore, we analyze the eigenvalue distributions of the Jacobian matrices associated with the iterative schemes arising from the GD iterations, offering an explanation for MGDL’s enhanced training stability.
The code used to generate the numerical results is available on GitHub: \href{https://anonymous.4open.science/r/Computational-Advantages-of-MGDL-C459}{\texttt{Computational Advantages of MGDL}}.
\end{abstract}



\section{Introduction}

Deep learning has achieved remarkable success across various domains, fundamentally transforming computer vision, natural language processing, speech recognition, medicine, and scientific computing.
In computer vision, deep convolutional neural networks (CNNs) have driven major breakthroughs in image recognition and classification \cite{he2016deep, krizhevsky2012imagenet, simonyan2015very}. In natural language processing, transformer-based architectures \cite{vaswani2017attention} have revolutionized tasks such as machine translation, chatbots, and sophisticated text generation. Speech recognition has advanced significantly through recurrent neural networks (RNNs) and long short-term memory (LSTM) networks \cite{hochreiter1997long, sherstinsky2020fundamentals}, enabling end-to-end learning from raw audio signals. Beyond these traditional domains, deep learning has profoundly impacted scientific computing, accelerating discoveries in physics, chemistry, biology, and medicine. Notable applications include protein structure prediction \cite{jumper2021highly}, drug discovery \cite{chen2018rise}, and solving complex partial differential equations \cite{raissi2019physics}. These achievements highlight deep learning’s transformative role across both theoretical and applied sciences.

Despite its success--driven by advances in neural network architectures, optimization techniques, and computational power--deep learning continues to face significant challenges in explainability, efficiency, and generalization. Training a deep neural network (DNN) involves solving a highly non-convex optimization problem, complicated by the large number of layers and parameters. Optimization algorithms often become trapped in local minima, and iterative training methods suffer from vanishing or exploding gradients due to the repeated application of the chain rule. Additionally, DNNs exhibit spectral bias \cite{rahaman2019spectral, xu2019training}, prioritizing low-frequency components during training, which can limit their ability to capture fine details. 
Moreover, the analysis in \cite{cohen2021gradient} shows that during gradient descent (GD), the maximum eigenvalue of the Hessian tends to hover just above $2/\text{(learning rate)}$, contributing to training instability.
Current training paradigms largely operate as black boxes, lacking interpretability, computational efficiency, and robust generalization.



To address these challenges, Xu \cite{xu2025multi} recently introduced multi-grade deep learning (MGDL), inspired
by human learning systems. Instead of optimizing the loss function end-to-end, MGDL decomposes
the learning process into a sequence of smaller optimization problems, each training a shallow neural
network with only a few hidden layers. These problems are structured into multiple grades, where
each grade learns from the residuals of the previous one. At each stage, a new shallow network is
trained while being composed with previously learned networks, whose parameters remain fixed
and serve as features or adaptive ‘basis’ functions. This iterative refinement process progressively
enhances learning while significantly reducing the complexity of the optimization problem.

MGDL has been shown to significantly outperform standard single-grade deep learning (SGDL) in applications such as regression \cite{FangXuaddressing, xu2025SAL}, numerical solutions of oscillatory Fredholm integral equations \cite{jiang2024deep}, and numerical solutions of partial differential equations \cite{xuzeng2023multi}. Notably, MGDL effectively mitigates spectral bias \cite{rahaman2019spectral, xu2019training}. For a general numerical study of this phenomenon, see \cite{FangXuaddressing}; its effectiveness in solving oscillatory Fredholm integral equations is demonstrated in \cite{jiang2024deep}, while \cite{xu2025multi} applies MGDL to solving the Helmholtz equation.
The goal of this paper is to investigate why MGDL outperforms SGDL by analyzing the computational advantages it offers. Since most optimization algorithms used in training deep neural networks (DNNs) are modifications of the GD method, we focus on studying GD applied to the optimization problems arising in both MGDL and SGDL. Specifically, we establish convergence theorems for GD in the MGDL framework and, based on these results, demonstrate that MGDL exhibits greater robustness to the learning rate compared to SGDL.

To analyze the convergence of MGDL and SGDL, we study a linear surrogate iterative scheme defined by the Jacobian matrix of the nonlinear map associated with the original iterative scheme. We show that the sequence generated by this surrogate scheme converges to the same limit as the sequence produced by the original scheme. Furthermore, we examine the eigenvalue distribution of the Jacobian matrices for the linear surrogate iterative schemes of MGDL and SGDL. Our numerical analysis reveals that the eigenvalues associated with MGDL are confined within the open interval $(-1,1)$, whereas those for SGDL may extend beyond this range. This confirms that, in the cases studied, the GD method for MGDL converges, whereas for SGDL, it does not. This explains why the loss values for SGDL exhibit oscillations, while those for MGDL are significantly more stable.

The Multi-scale Deep Neural Network (MSDL) was introduced in \cite{liu2020mscalednn, wang2020mscalednn} to mitigate the issue of spectral bias in deep learning. MSDL consists of multiple parallel standard sub-networks, each operating on a scaled version of the input. The outputs of these sub-networks are then combined to produce the final prediction. Numerical experiments reports in these studies demonstrate that MSDL converges rapidly across a range of applications, including regression problems, Poisson–Boltzmann equations, and oscillatory Stokes flows.
We conduct a numerical analysis of gradient descent applied to MSDL and compare its convergence behavior with that of MGDL. The analysis reveals that the eigenvalues of the iteration matrices associated with MSDL may lie outside the interval $(-1,1)$, potentially leading to unstable training dynamics. This observation provides a theoretical explanation for the highly oscillatory loss function values reported in the numerical experiments of \cite{liu2020mscalednn}.

The key contributions of this paper are as follows: 

(a) We provide a rigorous convergence analysis of the GD method for SGDL and MGDL, demonstrating the computational advantages of MGDL in solving the corresponding optimization problems.

(b) We conduct extensive numerical experiments, including image regression, denoising, and deblurring, showing that MGDL consistently outperforms SGDL and  exhibits greater stability.

(c) We study the impact of learning rate on SGDL and MGDL, showing that MGDL is more robust.

(d) We analyze a linear approximation of the GD dynamics to explain the convergence behavior of SGDL and MGDL. Furthermore, we numerically verify the superior stability of MGDL by studying the eigenvalue distribution of the associated Jacobian matrix.

\section{Standard Deep Learning Model}
In this section, we review the standard deep learning model and analyze the convergence behavior of the gradient descent (GD) method when applied to the resulting optimization problem.

We begin with a quick review of the definition of DNNs. 
A DNN is a composition of affine transformations and activation functions, consisting of an input layer, $D-1$ hidden layers, and an output layer. Let $d_0:=d$  (input dimension) and $d_D:=s$ (output dimension), and denote the number of neurons in the $j$-th layer by $d_j$. For $j\in \mathbb{N}_D:=\{1,2,\dots, D\}$, the weight matrices and bias vectors are $\mathbf{W}_j \in \mathbb{R}^{d_{j-1} \times d_{j}}$ and $\mathbf{b}_j \in \mathbb{R}^{d_j}$, respectively. The activation function $\sigma: \mathbb{R} \to \mathbb{R}$, applied componentwise, is taken to be ReLU: $\sigma(x)= (x)_{+}:=\max\{0,x\}$.

Given an input $\mathbf{x}\in \mathbb{R}^d$, the hidden layers are defined recursively by
$$
\mathcal{H}_1(\mathbf{x}) := \sigma\left(  \mathbf{W}_{1}^{\top}\mathbf{x}+\mathbf{b}_{1}\right), \mathcal{H}_{j+1}(\mathbf{x}) := \sigma\left(  \mathbf{W}_{j+1}^{\top}\mathcal{H}_{j}\left(\mathbf{x}\right)+\mathbf{b}_{j+1}\right), j=1,\dots, D-2.
$$
The network output is then given by
$\mathcal{N}_{D}\left(\{\mathbf{W}_j,\mathbf{b}_j\}_{j=1}^D; \mathbf{x}\right) =\mathcal{N}_{D}(\mathbf{x}) :=    \mathbf{W}_{D}^{\top}\mathcal{H}_{D-1}\left(\mathbf{x}\right)+\mathbf{b}_{D}.$
For a dataset $\mathbb{D}:=\left\{\mathbf{x}_{\ell}, \mathbf{y}_{\ell}\right\}_{\ell = 1}^{N}$, the loss function is 
\begin{equation}\label{DNNs loss}
    \mathcal{L}\left(\left\{\mathbf{W}_j, \mathbf{b}_j\right\}_{j=1}^{D}; \mathbb{D}\right):= \frac{1}{2N}\sum\nolimits_{\ell=1}^{N}\norm{\mathbf{y}_{\ell} - \mathcal{N}_D\left(\left\{\mathbf{W}_j, \mathbf{b}_j\right\}_{j=1}^{D}; \cdot\right)(\mathbf{x}_{\ell})}^2.
\end{equation}
The standard SGDL model minimizes this loss with respect to the network parameter $\Theta:=\{\mathbf{W}_j, \mathbf{b}_j\}_{j=1}^{D}$, yielding the optimal parameters $\Theta^*$ and the trained network  $\mathcal{N}_D\left(\Theta^*; \cdot\right)$.

Among commonly used methods for solving optimization problems for deep learning are stochastic gradient descent (SGD) \cite{kiefer1952stochastic, robbins1951stochastic} and adaptive moment estimation (Adam) \cite{kingma2015adam}. Most of them are based on GD. For this reason, we will study GD for minimizing the loss function defined in \eqref{DNNs loss}. To facilitate the convergence analysis, we stack all network parameters $\left\{\mathbf{W}_j, \mathbf{b}_j\right\}_{j=1}^{D}$ into a single column vector. For this purpose, we introduce the following notation.
Given a matrix or vector $\mathbf{A}$, we denote its vectorization by $A$: If $\mathbf{A}$ is a matrix, $A$ is the column vector obtained by stacking the columns of $\mathbf{A}$. If $\mathbf{A}$ is already a column vector, then $A = \mathbf{A}$. If $\mathbf{A}$ is a row vector, then $A = \mathbf{A}^\top$. We define the parameter vector
$W:= \left(W_1^{\top}, b_1^{\top}, \ldots, W_D^{\top}, b_D^{\top}\right)^{\top}$, where the total number of parameters is $M:= \sum_{j=1}^{D}(d_{j-1}+1)d_{j}$. We consider GD applied to a general objective function $\mathcal{F}: \mathbb{R}^M \to \mathbb{R}$, assumed to be nonnegative, twice continuously differentiable, and generally nonconvex. The GD iteration is given by 
\begin{equation}\label{general gradient descent} 
W^{k+1} := W^k - \eta \frac{\partial \mathcal{F}}{\partial W}(W^k), 
\end{equation} 
where $k$ denotes the iteration index and $\eta > 0$ is the learning rate.
In the context of this paper, we are particularly interested in the case where $\mathcal{F}$ is the loss function $\mathcal{L}$ defined in \eqref{DNNs loss}.



We now analyze the convergence of the GD method for minimizing the loss function \eqref{DNNs loss}. Assume there exists a compact, convex set $\mathcal{W} \subset \mathbb{R}^M$ such that for all $\eta \in (0, \eta_0)$ (for some $\eta_0 > 0$), the GD iterates $\{W^k\}_{k=0}^\infty$ from \eqref{general gradient descent} with $\mathcal{F} := \mathcal{L}$ remain in $\mathcal{W}$. Convergence depends on the Hessian of $\mathcal{L}$ over $\mathcal{W}$, for which we define
$\alpha := \sup_{W \in \mathcal{W}}\norm{\mathbf{H}_{\mathcal{L}}(W)}$,
where $\|\cdot\|$ denotes the spectral norm. Since $\mathbf{H}_{\mathcal{L}}(W) \in \mathbb{R}^{M \times M}$, the constant $\alpha$ reflects the network’s depth and size.
The following theorem, proved in Appendix \ref{appendix: convergence proof}, establishes the convergence of GD with $\mathcal{F} := \mathcal{L}$, extending the approach of Theorem 6 in \cite{xu2025multi}, which assumes all biases are zero.




\begin{theorem}\label{theorem: convergence SGDL}
    Let $\left\{W^{k}\right\}_{k=1}^{\infty}$ be a sequence generated by the gradient descent iteration \eqref{general gradient descent} with $\mathcal{F}:=\mathcal{L}$ and initial guess $W^{0}$. Suppose the activation function $\sigma$ is twice continuously differentiable, and there exists a convex compact set $\mathcal{W} \subset \mathbb{R}^M$ such that $\left\{ W^{k}\right\}_{k=1}^{\infty} \subset \mathcal{W}$. If the learning rate $\eta\in(0, 2/\alpha)$, then the following statements hold:
\begin{enumerate}  
 \item [(i)]  $\lim_{k \to \infty}\mathcal{L}(W^{k}) = L^*$ for some $L^*\geq 0$; 
    
   \item [(ii)]  $\lim_{k \to \infty}\frac{\partial \mathcal{L}}{\partial W}(W^{k}) = 0$; 
    
   \item [(iii)]  Every cluster point $\hat W$ of $\left\{W^k\right\}_{k=0}^{\infty}$ satisfies $\frac{\partial \mathcal{L}}{\partial W}(\hat W) = 0$.
\end{enumerate}
\end{theorem}

Deep neural networks are characterized by weight matrices and bias vectors, with parameter counts growing rapidly with depth—for instance, LeNet-5 \cite{lecun1998gradient} has 60K parameters, ResNet-152 \cite{he2016deep} 60.2M, and GPT-3 \cite{brown2020language} 175B. Traditional end-to-end training becomes increasingly difficult at scale due to both optimization and stability issues. First, deeper networks lead to highly nonconvex loss landscapes, often trapping solutions in poor local minima \cite{bengio2006greedy}. Second, training instability arises from vanishing or exploding gradients, which can hinder convergence \cite{glorot2010understanding, goodfellow2016deep, pascanu2013difficulty}. To address these challenges, the multi-grade deep learning (MGDL) model \cite{xu2025multi} mimics the human education system, where learning progresses in stages. MGDL sequentially trains shallow networks, each building on and refining the output of the previous grade to incrementally approximate the target function.

\vspace{-6mm}
\section{Multi-Grade Deep Learning}

This section reviews MGDL, analyzes GD convergence at each grade, and presents a convex program when each grade is a two-layer ReLU network.

Given data $\mathbb{D}:=\left\{\mathbf{x}_{\ell}, \mathbf{y}_{\ell}\right\}_{\ell=1}^{N}$, MGDL \cite{xu2025multi} decomposes learning a DNN of depth $D$ into $L<D$ sequential grades, each training a shallow neural network (SNN) $\mathcal{N}_{D_l}$ on the residuals from the previous grade. The depths satisfy $1<D_l<D$ and $\sum_{l=1}^{L}D_{l} = D+L-1$. Let $\Theta_l:=\left\{\mathbf{W}_{lj}, \mathbf{b}_{lj}\right\}_{j=1}^{D_l}$ be the parameters in grade $l$. The model is recursively defined by
\begin{small}
    $$
g_1(\Theta_1;\mathbf{x}):= \mathcal{N}_{D_1}\left(\Theta_1; \mathbf{x}\right), \ g_{l+1}(\Theta_{l+1}; \mathbf{x}):= \mathcal{N}_{D_{l+1}}\left(\Theta_{l+1};\cdot\right) \circ \mathcal{H}_{D_{l}-1}(\Theta_{l}^*;\cdot) \circ \ldots \circ \mathcal{H}_{D_{1}-1}(\Theta_1^*; \cdot)(\mathbf{x})
$$
\end{small}
for $l\in\mathbb{N}_{L-1}$, with loss function $
    \mathcal{L}_l\left(\Theta_l; \mathbb{D}\right):= \frac{1}{2N}\sum\nolimits_{\ell=1}^{N}\norm{\mathbf{e}_{l\ell} - g_l\left(\Theta_l; \mathbf{x}_{\ell}\right)}^2
$ where $\mathbf{e}_{1\ell} := \mathbf{y}_\ell$ and $\mathbf{e}_{(l+1)\ell} := \mathbf{e}_{l\ell} - g_l(\Theta_l^*; \mathbf{x}_\ell)$. Each $\Theta_l^*$ is obtained by minimizing $\mathcal{L}_l$ while fixing previous parameters $\{\Theta^*_j\}_{j=1}^{l-1}$, which act as feature extractors. After training all $L$ grades, the final MGDL output is the additive composition
$
\bar{g}_L(\left\{\Theta^*_{l}\right\}_{l=1}^L; \mathbf{x}):= \sum\nolimits_{l=1}^{L}g_l(\Theta_{l}^{*}; \mathbf{x}). 
$


We apply the GD method to optimize each grade in MGDL. For clarity, define $\mathbf{x}   _{1\ell} := \mathbf{x}_{\ell}$, and for $l = 2, \ldots, L$ and $\ell = 1, \ldots, N$, let
$
\mathbf{x}   _{l\ell} := \mathcal{H}_{D_{l-1}-1}(\Theta_{l-1}^*;\cdot) \circ \ldots \circ \mathcal{H}_{D_{1}-1}(\Theta_1^*; \cdot)(\mathbf{x}_{\ell})
$
and $\mathbb{D}_{l}:=\left\{\mathbf{x}   _{l\ell}, \mathbf{e}_{l\ell}\right\}_{\ell=1}^{N}$.
The loss function $\mathcal{L}_l\left(\Theta_l; \mathbb{D}\right)$ at grade $l$ becomes
\begin{small}
\begin{equation}\label{loss: grade l}
\mathcal{L}_l\left(\Theta_l; \mathbb{D}_{l}\right):= \frac{1}{2N}\sum\nolimits_{\ell=1}^{N}\norm{\mathbf{e}_{l\ell} - \mathcal{N}_{D_{l}}\left(\Theta_l; \mathbf{x}   _{l\ell}\right)}_2^2. 
\end{equation}
\end{small}
Each grade thus corresponds to a traditional shallow neural network whose input is the feature $\mathbf{x}   _{l\ell}$ from previous grades and whose output approximates the residual. We apply GD \eqref{general gradient descent} to minimize $\mathcal{L}_l$.


For grade $l$, define the parameter vector
$
W_{l}:= (
W_{l1}^{\top}, b_{l1}^{\top}, \ldots, W_{lD_l}^{\top}, b_{lD_l}^{\top}   
)^{\top} \in \mathbb{R}^{M_l},
$
where $M_l:= \sum_{j=1}^{D_l}(d_{l(j-1)}+1)d_{lj}
$. 
\begin{small}
The GD iteration is
\begin{equation}\label{gradient descent grade l}
W_{l}^{k+1} := W_l^{k}-\eta_l
\frac{\partial \mathcal{L}_l}{\partial W_{l}}( W_l^{k}).
\end{equation}
\end{small}
This is a special case of \eqref{general gradient descent} with $\mathcal{F}:=\mathcal{L}_{l}$ and $W:= W_{l}$. 
Let $\mathcal{W}_l \subset \mathbb{R}^{M_l}$ be a convex compact set containing the GD iterates, and define 
$
\alpha_{l} := \sup_{W_l \in \mathcal{W}_l} \left\| \mathbf{H}_{\mathcal{L}_l}(W_l)\right\|.
$
The following theorem proves convergence of the GD method \eqref{gradient descent grade l} for grade $l$ of MGDL. The proof is in Appendix \ref{appendix: convergence proof}.

\begin{theorem}\label{theorem: convergence grade l}
    Let $\left\{W_l^{k}\right\}_{k=1}^{\infty}$ be the sequence generated by \eqref{gradient descent grade l} from an initial guess $W_l^{0}$. Assume the dataset $\left\{\mathbf{x}_{l\ell}, \mathbf{e}_{l\ell}\right\}_{\ell=1}^N \subset \mathbb{R}^{d_{l0}} \times \mathbb{R}^{d}$ is bounded. Suppose the activation function $\sigma$ is twice continuously differentiable, and there exists a convex compact set $\mathcal{W}_l \subset \mathbb{R}^{M_l}$ such that $\left\{ W_l^{k}\right\}_{k=1}^{\infty} \subset \mathcal{W}_l$. If the learning rate $\eta_l$ is chosen from the interval $(0, 2/\alpha_l)$, then 
\begin{enumerate}
        \item [(i)] $\lim_{k \to \infty}\mathcal{L}_l(W_l^{k}) = L_l^*$ for some $L_l^*\geq 0$; 
        \item [(ii)] $\lim_{k \to \infty}\frac{\partial \mathcal{L}_l}{\partial W_l}(W_l^{k}) = 0$; 
        \item [(iii)] $\frac{\partial \mathcal{L}}{\partial W_l}(\hat W_l) = 0$ for any cluster point $\hat W_l$ of $\left\{W_l^k\right\}_{k=0}^{\infty}$.
\end{enumerate}


\end{theorem}

Theorem \ref{theorem: convergence SGDL} parallels Theorem \ref{theorem: convergence grade l}, with the main differences arising from the network depth. SGDL requires deep networks for expressiveness, while MGDL solves only a shallow network at each grade. This difference affects both the gradient behavior and the learning rate, which in turn influence the stability and performance of gradient descent.
The gradient $\frac{\partial \mathcal{L}}{\partial W}$, as detailed in the Appendix of \cite{xu2025multi} and the Supplementary Material, involves a chain of $D$ matrix products for depth-$D$ networks. For large $D$, this can lead to vanishing or exploding gradients, degrading convergence and stability. MGDL mitigates this by restricting optimization to shallow networks, reducing such risks.
Additionally, the allowable learning rate depends on the Hessian norm: SGDL requires $\eta \in (0, 2/\alpha)$, with $\alpha$ tied to a high-dimensional Hessian due to deep architecture. In contrast, MGDL only requires $\eta_l \in (0, 2/\alpha_l)$, where $\alpha_l$ is derived from a much smaller shallow network. Since $\alpha_l \ll \alpha$, MGDL supports a broader range of learning rates, enhancing robustness and training stability.

To close this section, we establish that when each grade of MGDL consists of one hidden layer ReLU network, the overall nonconvex optimization problem is reduced to solving a sequence of convex programming problems. 
It was shown in \cite{ergen2021convex,pilanci2020neural} that training two-layer ReLU networks can be exactly represented by a single convex program with the number of variables polynomial in the number of training samples and hidden neurons. 
Using this observation, MGDL with $L$ grades transforms a highly nonconvex problem into $L$ convex programs. We formulate the nonconvex problem for grade $l$ as \eqref{opt dnn} and its corresponding convex program as \eqref{finite-convex problem} in Appendix \ref{appendix: Convex Program Formulation for MGDL}. The following theorem establishes their relationship. Since each grade of MGDL is essentially a two-layer ReLU network, the proof of the theorem follows directly from \cite{pilanci2020neural}.

\begin{theorem}\label{thm: two-layer convex program}
The convex program \eqref{finite-convex problem} and the non-convex problem \eqref{opt dnn} with $m_l \geq m_l^*$ have identical optimal values. Moreover, an optimal solution to \eqref{opt dnn} with $m_l^*:=\sum\nolimits_{i=1}^{P_l} (1[\mathbf{u}_{li}^* \neq 0] + 1[\mathbf{v}_{li}^* \neq 0])$ neurons can be constructed from an optimal solution to \eqref{finite-convex problem} as follows
\begin{equation*}
\begin{aligned}
& \left(\mathbf{w}_{{lj}_{1i}}^*, \alpha_{{lj}_{1i}}^*\right)=\Big(\frac{\mathbf{u}_{li}^*}{\sqrt{\left\|\mathbf{u}_{li}^*\right\|}}, \sqrt{\left\|\mathbf{u}_{li}^*\right\|}\Big) \quad \text { if } \quad \mathbf{u}_{li}^* \neq 0 \\
&  \left(\mathbf{w}_{{lj}_{2i}}^*, \alpha_{{lj}_{2i}}^*\right)=\Big(\frac{\mathbf{v}_{li}^*}{\sqrt{\left\|\mathbf{v}_{li}^*\right\|}}, -\sqrt{\left\|\mathbf{v}_{li}^*\right\|}\Big) \quad \text { if } \quad \mathbf{v}_{li}^* \neq 0,
\end{aligned}
\end{equation*}
where $\{\mathbf{u}_{li}^*, \mathbf{v}_{li}^*\}_{i=1}^{P_l}$ are the optimal solutions to \eqref{finite-convex problem}.
\end{theorem}

\section{Comparison of SGDL and MGDL for Image Reconstruction}\label{section: comparsion image reconstruction}

In this section, we apply SGDL and MGDL to image reconstruction tasks, including regression, denoising, and deblurring. Using the ReLU activation and PSNR \eqref{PSNR} as the evaluation metric, we show that SGDL suffers from training instability and yields lower reconstruction accuracy than MGDL. TrPSNR and TePSNR denote PSNR values on the training and testing sets, respectively.


\textbf{Image regression.}
We model grayscale images as 2D functions 
$f: \mathbb{R}^2 \to \mathbb{R}$, from pixel coordinates to intensity values. The training set comprises a regularly spaced grid covering 1/4 of the pixels, while the test set includes all pixels. We apply SGDL and MGDL to six images of various sizes (Figure \ref{fig:testing images}). For images (b)–(f), we use structures \eqref{SGDL-2} and \eqref{MGDL-2}; for image (g), we use \eqref{SGDL-3} and \eqref{MGDL-3}.

We analyze the results for the `Cameraman' image separately, as SGDL behaves differently for this image. The loss for SGDL and MGDL are shown in Figures \ref{fig:SGDL MGDL cameraman} (a) and (b). SGDL exhibits extreme oscillations, leading to unstable predictions, as seen in Figures (c)-(f) for iterations 9800, 9850, 9900, and 9950, with oscillating PSNR. In contrast, the MGDL loss function steadily decreases (b), and its predictions (g)-(j) show consistent improvement, reflecting stable performance across iterations.

Numerical results for images (b)-(g) in Figure \ref{fig:testing images} are presented in Table \ref{tab: 2d image regression} and Figure \ref{fig:SGDL MGDL image regression}. Table \ref{tab: 2d image regression} compares the PSNR values from SGDL and MGDL. MGDL outperforms SGDL by $0.42$ to $3.94$ dB in testing PSNR, with improvements for all images. Figure \ref{fig:SGDL MGDL image regression} shows the loss function for both methods. SGDL shows consistent oscillations across all images, while MGDL's behavior varies. For images like Barbara, Butterfly, and Walnut, MGDL oscillates initially but stabilizes in later stages. For Pirate and Chest, MGDL oscillates more in earlier stages before stabilizing. These results suggest MGDL stabilizes or steadily decreases over time, while SGDL maintains persistent oscillations.


\begin{table*}[ht]
\centering
\begin{minipage}{0.48\textwidth}
\centering
\caption{PSNR comparison for image regression.}
\resizebox{\textwidth}{!}{
\begin{tabular}{llll}
\hline
Image & Method & TrPSNR & TePSNR \\
\hline
Cameraman & SGDL & $27.05$ & $24.79$ \\
                           & MGDL & $\mathbf{31.80}$ & $\mathbf{25.21}$ \\
Barbara   & SGDL & $23.14$ & $22.75$ \\
                           & MGDL & $\mathbf{24.36}$ & $\mathbf{23.84}$ \\
Butterfly & SGDL & $26.22$ & $24.87$ \\
                           & MGDL & $\mathbf{28.23}$ & $\mathbf{27.06}$ \\
Pirate      & SGDL & $24.20$ & $24.34$ \\
                           & MGDL & $\mathbf{27.40}$ & $\mathbf{26.45}$ \\
Chest     & SGDL & $34.77$ & $34.56$ \\
                           & MGDL & $\mathbf{39.44}$ & $\mathbf{38.50}$ \\
Walnut    & SGDL & $19.94$ & $20.05$ \\
                           & MGDL & $\mathbf{21.83}$ & $\mathbf{21.31}$ \\
\hline
\end{tabular}
\label{tab: 2d image regression}
}

\end{minipage}
\hfill
\begin{minipage}{0.48\textwidth}
\centering
\caption{PSNR comparison for image denoising.}
\resizebox{\textwidth}{!}{
\begin{tabular}{lllll}
\hline
Noise  & Method & Butterfly &  Pirate & Chest\\
\hline
10
 & SGDL & $27.53$ & $25.13$ & $36.20$  \\
 & MGDL & $\mathbf{31.67}$ & $\mathbf{29.36}$ & $\mathbf{38.58}$  \\
20
 & SGDL & $26.73$ & $25.02$ & $35.34$  \\
 & MGDL & $\mathbf{28.39}$ & $\mathbf{27.74}$ & $\mathbf{36.89}$  \\
30 
 & SGDL & $26.05$ & $24.63$ & $34.30$  \\
 & MGDL & $\mathbf{27.09}$ & $\mathbf{27.20}$ & $\mathbf{35.48}$  \\
40 
 & SGDL & $25.54$ & $24.47$ & $33.55$  \\
 & MGDL & $\mathbf{26.37}$ & $\mathbf{26.25}$ & $\mathbf{34.61}$  \\
50 
 & SGDL & $24.65$ & $24.01$ & $33.51$  \\
 & MGDL & $\mathbf{25.84}$ & $\mathbf{25.77}$ & $\mathbf{33.94}$  \\
60 
 & SGDL & $24.30$ & $23.82$ & $32.90$  \\
 & MGDL & $\mathbf{25.21}$ & $\mathbf{25.32}$ & $\mathbf{33.06}$  \\ 
\hline
\end{tabular}
\label{tab: 2d image denoising}
}

\end{minipage}
\end{table*}

\textbf{Image denoising.} We consider the image denoising problem. For a given noisy image 
$
\mathbf{\hat{f}}:= \mathbf{f} + \boldsymbol{\epsilon}
$
where $\mathbf{\hat{f}} \in \bR^{n \times n}$ is the observed image, $\mathbf{f} \in \bR^{n \times n}$ is the ground truth image, and $[\boldsymbol{\epsilon}]_{i,j} \sim \mathcal{N}(0, \sigma^2)
$ is the Gaussian noise with mean $0$ and standard deviation $\sigma$. The task is to use SGDL and MGDL to recover the ground truth image $\mathbf{f}$ from the observed noisy image $\mathbf{\hat{f}}$. The optimization problem is provided in Appendix \ref{Optimization Problem for Image Denoising} with the transform operator $\mathbf{A}$ chosen as the identity operator. 

The structures of SGDL and MGDL are given by \eqref{SGDL-3} and \eqref{MGDL-3}, respectively. We test the models under six noise levels: $\sigma = 10, 20, 30, 40, 50, 60$, as shown in Figure \ref{fig: noisy image}. The results for image denoising are reported in Table \ref{tab: 2d image denoising} and Figures \ref{fig:SGDL MGDL butterfly image denoising}-\ref{fig:SGDL MGDL medical image denoising}. Table \ref{tab: 2d image denoising} compares the PSNR values achieved by SGDL and MGDL across various images and noise levels. MGDL consistently outperforms SGDL, with improvements ranging from $0.16$ to $4.23$ dB in PSNR values. 
Figures \ref{fig:SGDL MGDL butterfly image denoising}-\ref{fig:SGDL MGDL medical image denoising} display the PSNR values during training. Across all cases, the PSNR values for SGDL exhibit strong oscillations throughout the training process, while those for MGDL increase steadily—especially for grades 2 to 4.

\textbf{Image deblurring.} We address the image deblurring problem, where the blurred image is modeled as $\mathbf{\hat{f}} := \mathbf{K}\mathbf{f} + \boldsymbol{\epsilon}$, with $\mathbf{\hat{f}}$ as the observed image, $\mathbf{K}$ as the Gaussian blurring operator, $\mathbf{f}$ as the ground truth image, and $[\boldsymbol{\epsilon}]_{i,j} \sim \mathcal{N}(0, \sigma^2)$ with $\sigma=3$. The task is to recover $\mathbf{f}$ from $\mathbf{\hat{f}}$ using SGDL and MGDL. The optimization problem and operator $\mathbf{A}$ (Gaussian blurring operator) are detailed in Appendix \ref{Optimization Problem for Image Denoising}. 

\begin{figure}[htbp]
    \centering
    \begin{minipage}{0.48\textwidth}
        \centering
        \captionof{table}{PSNR comparison for image deblurring.}
        \footnotesize
        \begin{tabular}{lllll}
        \hline
          image & method & $3$ & $5$ & $7$
        \\
        \hline
        Butterfly & SGDL& $25.43$ &  $24.20$ & $22.70$\\
        &  MGDL & $\textbf{27.06}$ & $\textbf{25.19}$ & $\textbf{23.65}$\\
        Pirate & SGDL & $24.72$ & $23.79$ & $23.13$ \\
        & MGDL & $\textbf{26.47}$ & $\textbf{24.95}$ & $\textbf{23.98}$\\
        Chest& SGDL & $35.40$ & $34.61$ & $33.69$ \\
        & MGDL& $\textbf{38.24}$  & $\textbf{36.51}$ & $\textbf{35.14}$\\
        \hline
        \end{tabular}
        \label{tab: 2d image deburring}
    \end{minipage}
    \hfill
    \begin{minipage}{0.51\textwidth}
        \centering
        \begin{subfigure}{0.49\linewidth}
            \includegraphics[width=\linewidth]{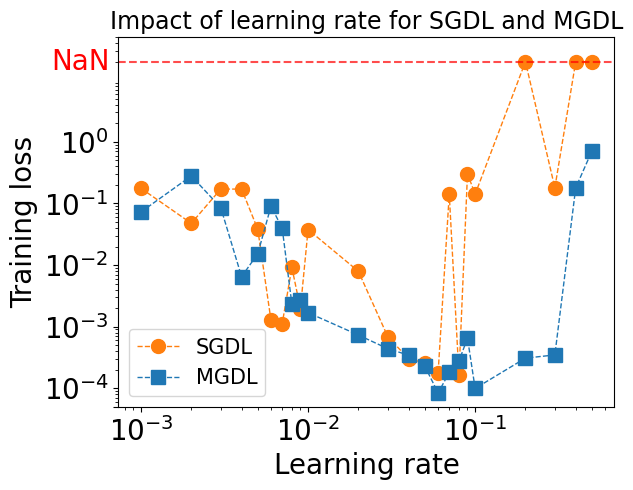}
        \end{subfigure}
        \begin{subfigure}{0.49\linewidth}
            \includegraphics[width=\linewidth]{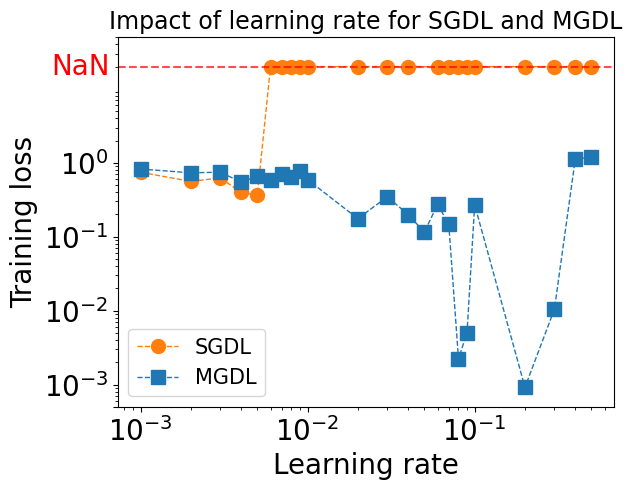}
        \end{subfigure}
        \caption{Impact of learning rate.}
        \label{fig:learning rate effectiveness synthetic data}
    \end{minipage}
\end{figure}

The structures of SGDL and MGDL are given by \eqref{SGDL-3} and \eqref{MGDL-3}, respectively. We test the models under three blurring levels: $\hat{\sigma} = 3$, $\hat{\sigma} = 5$, and $\hat{\sigma} = 7$, as shown in Figure \ref{fig:blurred images}. Results, including PSNR values, are reported in Table \ref{tab: 2d image deburring} and Figures \ref{fig:SGDL MGDL butterfly image deblurring}-\ref{fig:SGDL MGDL Chest image deblurring}. MGDL outperforms SGDL by $0.85$ to $2.84$ dB in PSNR. While SGDL shows strong oscillations in PSNR during training, MGDL shows steady improvement, particularly from grades 2 to 4.

Results on image regression, denoising, and deblurring show that MGDL consistently outperforms SGDL. While SGDL exhibits strong oscillations in loss or PSNR during training, MGDL shows a steady decrease in loss or an increase in PSNR. The reasons for this will be explored in Section \ref{section: Eigenvalue Analysis}.

\section{Impact of Learning Rate on SGDL and MGDL}\label{section: Impact of learning rate}

This section explores the effect of learning rate on  SGDL and MGDL. Both models are optimized by using the GD method. We evaluate various learning rates for both models to assess their performance.


\textbf{Synthetic data regression.} 
We approximate function $g:[0,1] \to \mathbb{R}$ defined by
\begin{small}
\begin{equation}\label{regression function}
    g(\mathbf{x}) := \sum\nolimits_{j=1}^{M} \sin\left(2 \pi \kappa_j \mathbf{x} + \varphi_j\right), \mathbf{x} \in [0,1],
\end{equation}
\end{small}
where $\varphi_j \sim \mathcal{U}(0,2 \pi)$, and $\mathcal{U}$ denotes the uniform distribution. We consider two settings: (1) $M=3, \kappa = [1, 5,5, 10]$; (2) $M=5, \kappa=[1, 8.25, 15.5, 22.75, 30]$. The training set contains $1024$ equally spaced points in $[0, 1]$, and  validation set contains $1000$ points sampled uniformly from $[0, 1]$.


The structures of SGDL and MGDL follow \eqref{SGDL-1} and \eqref{MGDL-1}, respectively. For both models, the learning rate is selected from the interval $[0.001, 0.5]$, with training run for $1\times 10^{6}$ epochs. The impact of the learning rate on performance is illustrated in Figure \ref{fig:learning rate effectiveness synthetic data} (left: Setting 1, right: Setting 2). In the figure, `NaN' indicates that the training loss diverged, resulting in an invalid value. In Setting 1, where the target function has a lower frequency and is easier to approximate, both SGDL and MGDL perform well. However, MGDL is robust across a wider range of learning rates. Specifically, SGDL achieves a loss below $0.001$ when the learning rate $\eta$ lies in $[0.03, 0.08]$, whereas MGDL maintains this performance over a broader interval of $[0.01, 0.3]$. SGDL fails to converge when $\eta>0.3$. In Setting 2, involving a higher-frequency target function, SGDL struggles with convergence for $\eta > 0.006$ and performs best at $\eta = 0.005$. By contrast,  MGDL remains stable and effective across a significantly wider range, with the loss staying below $0.01$ for $\eta$ in $0.08$ and $0.3$.

\textbf{Image regression.} The problem setting follows that of Section \ref{section: comparsion image reconstruction}. SGDL and MGDL are implemented using structures \eqref{SGDL-2} and \eqref{MGDL-2}, respectively. The learning rate is selected from the interval $[0.001, 1]$, with training performed for $1 \times 10^{5}$ epochs.


Figure \ref{fig:learning rate effectiveness image} illustrates the effect of learning rate on performance across four images: `Resolution Chart', `Cameraman',  `Barbara', and  `Pirate'. In all cases, MGDL consistently achieves higher accuracy than SGDL. Notably,  SGDL fails to train on the `Cameraman' and `Pirate' when the learning rate is near $1$, while MGDL remains stable and continues to produce accurate results under these learning rates.

Numerical results on synthetic data regression and image regression demonstrate that MGDL is more robust to the choice of learning rate and performs well with larger learning rates compared to SGDL.

\section{Eigenvalue Analysis for SGDL and MGDL}
\label{section: Eigenvalue Analysis}

This section analyzes the GD method \eqref{general gradient descent} applied to both SGDL and MGDL, as it is the most widely used approach for these optimization problems and serves as the foundation for many other methods.

GD can be viewed as a Picard iteration:
$W^{k+1} = \left(
\mathbf{I} - \eta \frac{\partial \mathcal{F}}{\partial W}
\right)(W^k)$, $k=0,1, \dots$,
whose convergence is well-studied in fixed-point theory under nonexpansiveness assumptions \cite{Bauschke2011ConvexAA}. However, when applied to deep learning, the resulting operator is typically expansive. To address this, we linearize $\frac{\partial \mathcal{F}}{\partial W}$ to facilitate the analysis of gradient descent behavior in deep networks.


Using the Taylor expansion at $W^{k-1}$, the gradient can be approximated as
$$
\frac{\partial \mathcal{F}}{\partial W}(W^k)  = \mathbf{H}_{\mathcal{F}}(W^{k-1})W^k + u^{k-1}+ \frac{1}{2}(W^{k} - W^{k-1})^{\top}\mathbf{T}_{\mathcal{F}}(\bar W)(W^{k} - W^{k-1}),
$$
where $u^{k}:=\frac{\partial \mathcal{F}}{\partial W}(W^{k}) - \mathbf{H}_{\mathcal{F}}(W^{k})W^{k}$, $\mathbf{T}_{\mathcal{F}}(\bar W)$ is the third derivative of $\mathcal{F}$ at $\bar W$, and $\bar W$ lies between $W^k$ and $W^{k-1}$. Substituting into the GD update yields
\begin{equation}\label{GD matrix form}
W^{k+1} = \mathbf{A}^{k-1} W^k - \eta u^{k-1} + r^{k-1}    \end{equation}
where $\mathbf{A}^{k-1}:= \mathbf{I} - \eta\mathbf{H}_{\mathcal{F}}(W^{k-1})$ and $r^{k-1} :=  -\frac{\eta}{2}(W^{k} - W^{k-1})^{\top}\mathbf{T}_{\mathcal{F}}(\bar W)(W^{k} - W^{k-1})
$. Neglecting the remainder term $r^{k-1}$ gives a linear approximation:
\begin{equation}\label{GD matrix form linear approxi}
\tilde W^{k+1} = \mathbf{A}^{k-1} \tilde W^{k} - \eta u^{k-1}. 
\end{equation}
We next consider the convergence of the linearized GD \eqref{GD matrix form linear approxi}, which depends on the spectral norm $\tau:=\text{sup}_{W \in \Omega}\|\mathbf{I} - \eta \mathbf{H}_{\mathcal{F}}(W)\|$ over a compact domain $\Omega$. The proof is provided in Appendix \ref{appendix: convergence proof}.



\begin{theorem}\label{convergence of gradient descent approxi} 
Suppose $\mathcal{F}: \mathbb{R}^M \to \mathbb{R}$ is a nonnegative, twice continuously differentiable and $\Omega \in \mathbb{R}^M$ is a convex, compact set. Let $\{\tilde{W}^{k}\}_{k=0}^{\infty}$ be generated by \eqref{GD matrix form linear approxi} and $\{W^{k}\}_{k=0}^{\infty}$ be generated by \eqref{GD matrix form} and assume that $\{W^{k}\}_{k=0}^{\infty} \subset \Omega$. If $\tau<1$, then the sequence $\{\tilde{W}^{k}\}_{k=0}^{\infty}$ converges.
\end{theorem}

We establish the connection between the GD \eqref{GD matrix form} and its linearization \eqref{GD matrix form linear approxi}; see Appendix \ref{appendix: convergence proof} for proof.



\begin{theorem}\label{relation between gradient descent and its linearized form} 
Suppose $\mathcal{F}: \mathbb{R}^M \to \mathbb{R}$ is a nonnegative, three times continuously differentiable function and $\Omega \in \mathbb{R}^M$ is a convex, compact set. Let the sequence $\{ W^k \}_{k=0}^{\infty}$ be generated by \eqref{GD matrix form}, $\{ \tilde{W}^k\}_{k=0}^{\infty}$ be generated by \eqref{GD matrix form linear approxi} with $\tilde{W}^0 = W^0$ and $\tilde{W}^1 = W^1$, and assume $\{ W^k \}_{k=0}^{\infty} \subset \Omega$. If $\tau < 1$, then both $\{ \tilde{W}^k \}_{k=0}^{\infty}$ and $\{ W^k\}_{k=0}^{\infty}$ converge to the same point. 
\end{theorem}


Theorem \ref{relation between gradient descent and its linearized form} shows that the convergence of sequence $\left\{W^k\right\}_{k=1}^{\infty}$ is mainly governed by $\mathbf{I} - \eta\mathbf{H}_{\mathcal{F}}(W)$. A sufficient and practically verifiable condition for convergence is that all eigenvalues of this matrix lie within $(-1, 1)$.
For numerical analysis, we provide the explicit Hessian of $\mathcal{F}$ (with $\mathcal{F}:=\mathcal{L}$ for SGDL and $\mathcal{F}:=\mathcal{L}_l$ for MGDL) under ReLU activation in the Supplementary Material.



We next track the eigenvalues of $\mathbf{I} - \eta\mathbf{H}_{\mathcal{F}}(W^{k})$ during training to
study convergence. In deep networks like SGDL, these eigenvalues often fall outside $(-1, 1)$, leading to oscillatory loss. In contrast, MGDL’s shallow structure keeps them within $(-1, 1)$, yielding smooth loss decay.



\begin{figure}[htbp]
  \centering

   \begin{subfigure}{0.265\linewidth}
\includegraphics[width=\linewidth]{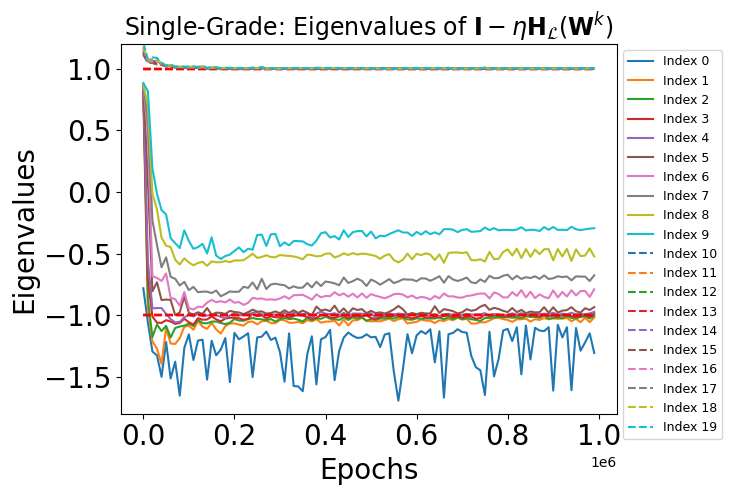}
    \end{subfigure}
   \begin{subfigure}{0.265\linewidth}
\includegraphics[width=\linewidth]{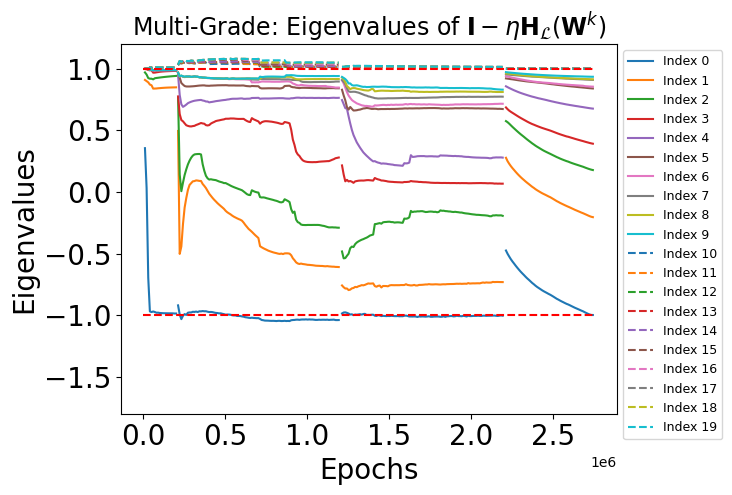}
    \end{subfigure}
   \begin{subfigure}{0.225\linewidth}
\includegraphics[width=\linewidth]{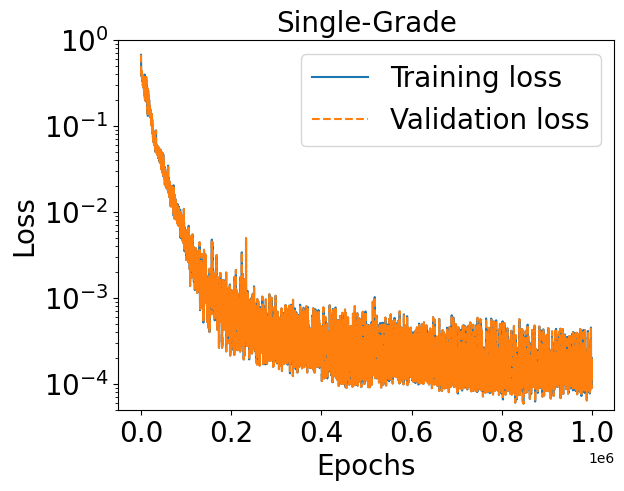}
    \end{subfigure}
   \begin{subfigure}{0.225\linewidth}
\includegraphics[width=\linewidth]{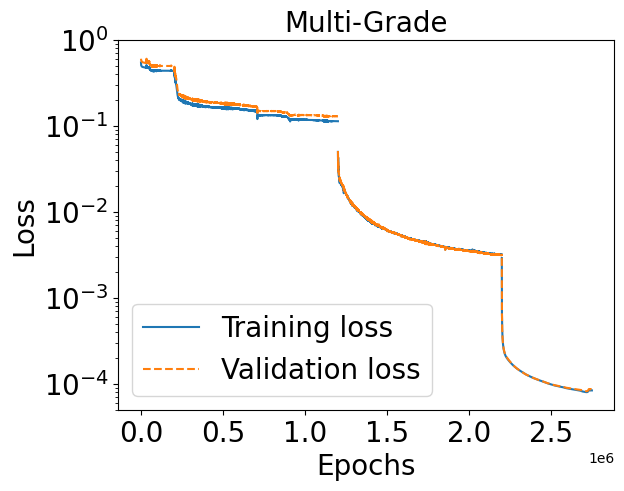}
    \end{subfigure}
\caption{Training process of SGDL ($\eta = 0.08$) and MGDL ($\eta = 0.06$) for Setting 1.}
	\label{fig: Numerical analysis sin SGDL MGDL setting 1}
\end{figure}

\textbf{Synthetic data regression.} The problem settings—including target function, training/validation data, network structures, and activation function—are the same as in Section~\ref{section: Impact of learning rate}. We use gradient descent to optimize both models, selecting the learning rate $\eta$ from the interval $[0.001, 0.5]$ based on the lowest validation loss. Numerical results are shown in Figure~\ref{fig: Numerical analysis sin SGDL MGDL setting 1} (Setting 1) and Figure~\ref{fig: Numerical analysis sin SGDL MGDL setting 2} (Setting 2).

We begin by analyzing SGDL under Setting 1. In Figure~\ref{fig: Numerical analysis sin SGDL MGDL setting 1}, the first subfigure shows the ten smallest (solid) and ten largest (dashed) eigenvalues of $\mathbf{I} - \eta\mathbf{H}_{\mathcal{L}}(W^k)$ over the training process (up to $10^6$ epochs). The smallest eigenvalue (index 0) remains significantly below $-1$, while eigenvalues with indices $1$ through $5$ stay below or near $-1$. All ten largest eigenvalues remain slightly above $1$, indicating limited influence on training dynamics.
The third subfigure plots the loss function over training. While the loss decreases overall, it shows moderate oscillations in the first $10^5$ epochs and stronger oscillations thereafter. These patterns align with the behavior of the smallest eigenvalues: moderate oscillations correspond to three eigenvalues below or near $-1$, while stronger oscillations occur as more eigenvalues fall into this range. These results indicate that oscillations in the loss are closely linked to the number of eigenvalues of $\mathbf{I} - \eta\mathbf{H}_{\mathcal{L}}(W^k)$ that are less than or near $-1$.

We next analyze MGDL for Setting 1. In Figure \ref{fig: Numerical analysis sin SGDL MGDL setting 1}, the second subfigure shows the ten smallest (solid) and ten largest (dashed) eigenvalues of $\mathbf{I} - \eta\mathbf{H}_{\mathcal{L}}(W^k)$ during training. Nearly all of the smallest eigenvalues stay within $(-1, 1)$ across grades 1-4, while the largest eigenvalues remain slightly above $1$. This behavior supports the stable loss reduction observed in the fourth subfigure.


Training behavior of both methods is influenced by the eigenvalues of $\mathbf{I} - \eta\mathbf{H}_{\mathcal{L}}(W^k)$. SGDL's eigenvalue drops below or near $-1$, causing loss oscillation, while MGDL's eigenvalues remain within $(-1, 1)$, resulting in stable loss reduction. This shows that the MGDL improves training stability. As shown in the first two subfigures of Figure \ref{fig: appendix synthetic data regression}, both methods achieve similar accuracy.



We now turn to the numerical analysis of Setting 2, as shown in Figure \ref{fig: Numerical analysis sin SGDL MGDL setting 2}. The target function in Setting 2 is similar to that in Setting 1, but it includes a higher-frequency component, making it more challenging to learn. For SGDL, the first subfigure shows eigenvalues of  $\mathbf{I} - \eta\mathbf{H}_{\mathcal{L}}(W^k)$. At the initial $2\times 10^5$ epochs, all eigenvalues fall in $(-1, 1)$. Therefore, the loss function decreases stably during this period, as shown in the third subfigure. After $2 \times 10^5$ epochs, the smallest eigenvalue (index 0), drops to $-1$, and the second smallest eigenvalue approaches $-1$. This causes the strong oscillations of the loss function during the period $2\times 10^5$ to $10^6$. The oscillations become even stronger after $6\times 10^5$ as the smallest eigenvalue drops $-1$ significantly and more eigenvalues approach or fall below $-1$. For MGDL, the behavior of eigenvalues and the loss values, as shown in the second and fourth subfigures, are consistent with those in Setting 1. The third and fourth subfigures in Figure \ref{fig: appendix synthetic data regression} illustrate MGDL achieves much better prediction accuracy than SGDL.


Both Settings 1 and 2 demonstrate that MGDL is more stable than SGDL.  The instability of SGDL can be attributed to the eigenvalues of $\mathbf{I} - \eta\mathbf{H}_{\mathcal{L}}(W^k)$, where the eigenvalue drops below or approaches $-1$ during the training.
For the same reason, the stability of MGDL is also explained by the eigenvalues of $\mathbf{I} - \eta\mathbf{H}_{\mathcal{L}}(W^k)$ where nearly all eigenvalues remain within the range $(-1, 1)$ throughout the training. Moreover, MGDL achieves significantly better accuracy in the more challenging Setting 2.


Across Settings 1 and 2 for both SGDL and MGDL, the smallest eigenvalue of $\mathbf{I} - \eta\mathbf{H}_{\mathcal{L}}(W^k)$ has a greater impact on the behavior of the loss function compared to the largest eigenvalue. This is because the largest eigenvalue is close to $1$, contributing to the stability of the iteration. In contrast, when the smallest eigenvalue approaches or drops below $-1$, the iteration tends to oscillate.




\textbf{Image regression.} The problem setting follows Section \ref{section: comparsion image reconstruction}. To facilitate Hessian computation for SGDL and MGDL, we adopt a shallow and narrow neural network. SGDL and MGDL follow the architectures in \eqref{SGDL-4} and \eqref{MGDL-4}, respectively, and are optimized via gradient descent.

\begin{figure}[htbp]
  \centering
\begin{subfigure}{0.265\linewidth}
\includegraphics[width=\linewidth]{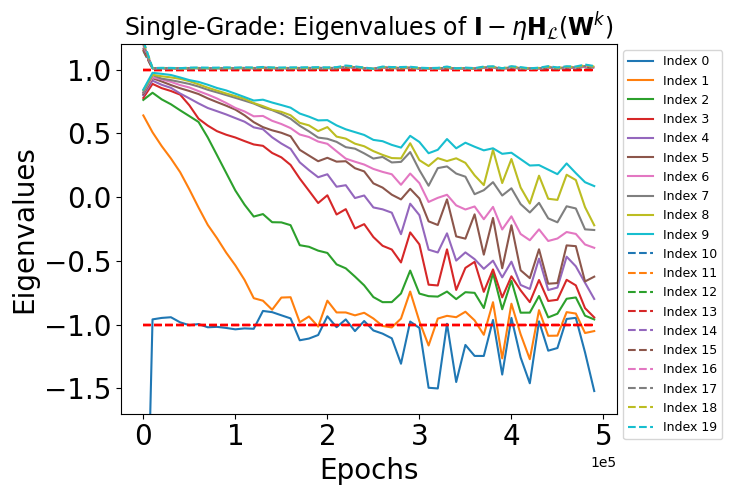}
    \end{subfigure}
  \begin{subfigure}{0.265\linewidth}
\includegraphics[width=\linewidth]{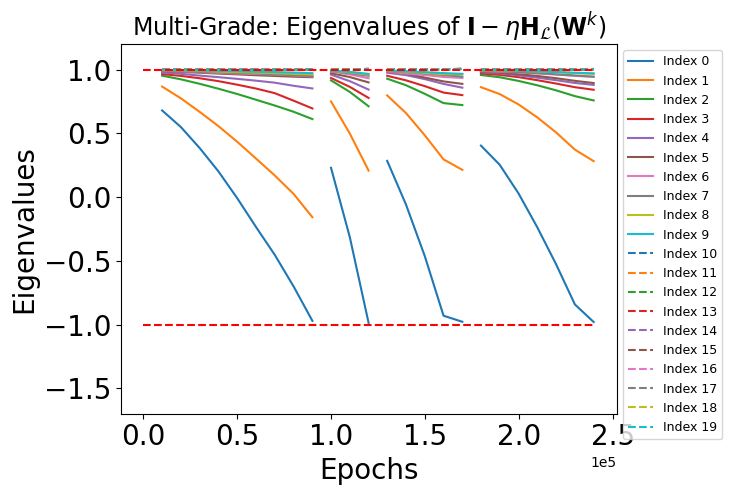}
\end{subfigure}
\begin{subfigure}{0.225\linewidth}
\includegraphics[width=\linewidth]{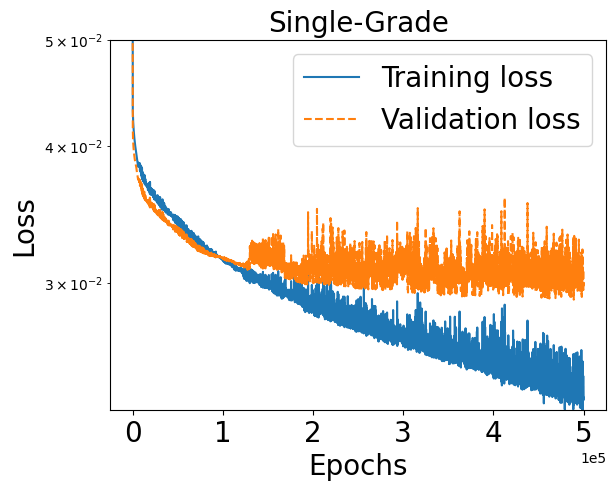}
    \end{subfigure}
  \begin{subfigure}{0.225\linewidth}
\includegraphics[width=\linewidth]{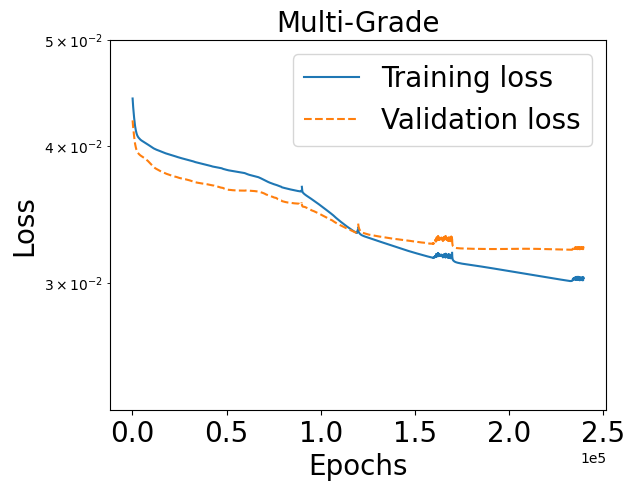}
\end{subfigure}
    
\caption{Training process of SGDL ($\eta = 0.02$) and MGDL ($\eta = 0.2$) for `Resolution chart'. 
}
	\label{fig: image regression resolutionchart EigStop}
\end{figure}

We conduct an eigenvalue analysis of SGDL and MGDL on the image regression task, using the images `Resolution chart', `Cameraman', `Barbara', and `Butterfly'. Computing the Hessian matrix of the loss becomes computationally expensive when using all the pixels from images. To mitigate this, we train the models on a selected subset of pixels. Results are reported in Figures \ref{fig: image regression resolutionchart EigStop} and \ref{fig: image regression cameraman EigStop}-\ref{fig: image regression butterfly EigStop}. The performance of SGDL and MGDL remains similar across the four images. For SGDL, when only the smallest eigenvalue of $\mathbf{I} - \eta \mathbf{H}_{\mathcal{L}}(\mathbf
W^k)$ is close to $-1$, the loss exhibits slight oscillations, for example, during the initial $1.3 \times 10^5$, $5 \times 10^4$, $8 \times 10^4$, and $2 \times 10^5$ epochs for the four images, respectively.
When multiple eigenvalues approach $-1$, the oscillations become more pronounced. 
For MGDL, 
all the eigenvalue of $\mathbf{I} - \eta \mathbf{H}_{\mathcal{L}}(\mathbf
W^k)$ fall in $(-1, 1)$ across the four images, resulting in a stable decrease in loss across all images, as shown in Figures \ref{fig: image regression resolutionchart EigStop} and \ref{fig: image regression cameraman EigStop}-\ref{fig: image regression butterfly EigStop}. 

\textbf{Image denoising.} We consider the same problem setting in Section \ref{section: comparsion image reconstruction}. SGDL and MGDL follow the architectures in \eqref{SGDL-4} and \eqref{MGDL-4}, respectively, and are optimized via gradient descent.


The numerical results are reported in Figures \ref{fig: image regression butterfly image denoisng10}-\ref{fig: image regression barbara image denoisng30}. We observe that, for SGDL, the smallest eigenvalue approaches $-1$, causing instability and oscillations in the loss function. In contrast, MGDL maintains all eigenvalues within $(-1, 1)$, ensuring a stable and steadily decreasing loss.

Numerical results, including synthetic data regression, image regression, and image denoising, consistently show that the eigenvalues of $\mathbf{I} - \eta \mathbf{H}_{\mathcal{L}}(\mathbf{W}^k)$ for SGDL tend to approach or fall below $-1$, while those for MGDL are more likely to remain stable within the interval $(-1, 1)$. This behavior explains why MGDL exhibits greater stability compared to SGDL.


\section{Comparison of MSDL and MGDL}\label{section:MGDL and MSDL}

This section compares multi-scale deep learning (MSDL) \cite{liu2020mscalednn, wang2020mscalednn} and MGDL on the synthetic data regression task, using the same problem setup described in Section \ref{section: Impact of learning rate}. MGDL adopts the architecture defined in \eqref{MGDL-1}, while MSDL’s architecture—configured to match the total number of neurons—is provided in Appendix \ref{appendix: architecture}.  Both models employ ReLU activation functions and the shallow, narrow network are used to simplify the computation of  Hessian eigenvalues. Training is performed using gradient descent, with the learning rate $\eta$ selected from $[0.001, 0.5]$ to minimize validation loss.

Numerical comparisons between MSDL and MGDL are shown in Figures \ref{fig: Numerical analysis sin MSDL MGDL setting 1}-\ref{fig: Numerical analysis sin MSDL MGDL setting 2} for Settings 1 and 2. In both cases, MSDL exhibits more frequent occurrences of eigenvalues near or below $-1$, causing oscillations in the training loss. MSDL’s predictions appear in Figure \ref{fig: MSDL prediction}, and MGDL’s in Figure \ref{fig: appendix synthetic data regression}. Figure \ref{fig:learning rate effectiveness MSDL MGDL} highlights MSDL's greater sensitivity to learning rate. Improved stability for MSDL may be possible with compactly supported activations and deeper, wider networks, as suggested in \cite{liu2020mscalednn}.





\section{Conclusion}

We conducted extensive numerical experiments on synthetic regression and image reconstruction tasks. 
The results consistently show that MGDL outperforms SGDL, with MGDL exhibiting stable training dynamics, while SGDL displays oscillatory behavior. To explain this, we analyzed the eigenvalues of the iteration matrices and established a convergence theorem for the GD method. The theorem confirms that the training dynamics are primarily governed by the eigenvalues of the iteration matrices. The numerical analysis reveals that SGDL's eigenvalues frequently lie outside the interval $(-1, 1)$, leading to oscillatory training behavior, whereas MGDL's remain within this range, contributing to its stable training dynamics. We also investigated the effect of learning rate and found that MGDL is more robust and performs well with a larger learning rate when compared with SGDL. These comparisons highlight the computational advantages and practical robustness of MGDL.

\noindent
\textbf{Limitation}: We established convergence results for SGDL and MGDL within the gradient descent framework. However, widely used optimization algorithms such as SGD and Adam were not analyzed. While our numerical experiments demonstrate that MGDL achieves superior performance, improved training stability, and robustness to learning rate choices, the theoretical foundations of these advantages remain to be fully understood. 

\newpage

\bibliographystyle{plainnat} 
\bibliography{ref2025} 

\appendix

\section{Convex Program Formulation for MGDL}\label{appendix: Convex Program Formulation for MGDL}

We briefly formulate the convex program for MGDL, where each grade is a two-layer ReLU network. For detailed discussions, see \cite{pilanci2020neural}.

A two-layer ReLU network function at grade $l$, $\mathcal{N}_{2_l}: \mathbb{R}^{d_l} \to \mathbb{R}$, with $m_l$ neurons, is defined as
$$
\mathcal{N}_{2_l}(\mathbf{x}_l) = \sum_{j=1}^{m_l}\alpha_{lj} ( \mathbf{w}_{lj}^{\top}\mathbf{x}_l)_{+},
$$
where $\mathbf{x}_l$ is the input feature for grade $l$, $\mathbf{w}_{lj}$ and $\alpha_{lj}$ are paramaters at the first and second layers.  
Let data matrix be $\mathbf{X}_l:=[\mathbf{x}_{l1}, \mathbf{x}_{l2}, \ldots, \mathbf{x}_{lN}]^{\top} \in \mathbb{R}^{N \times d_l}$, the residual vector be $\mathbf{e}_{l}:= [\mathbf{e}_{l1}, \mathbf{e}_{l2}, \ldots, \mathbf{e}_{lN}]^{\top}\in \mathbb{R}^{N}$, and let $\beta_l>0$ be a regularization parameter. Consider minimizing the squared loss objective and squared $\ell_2$-norm of all parameters
\begin{equation}\label{opt dnn}
\min_{\{\mathbf{w}_{lj}, \alpha_{lj}\}_{j=1}^{m_l}} \frac{1}{2}\|\sum_{j=1}^{m_l} \alpha_{lj}(\mathbf{X}_l \mathbf{w}_{lj})_{+}-\mathbf{e}_l\|^2 +\frac{\beta_l}{2} \sum_{j=1}^{m_l}\left(\left\|\mathbf{w}_{lj}\right\|^2+\alpha_{lj}^2\right).
\end{equation}

We next state the corresponding convex program. Consider diagonal matrices Diag$(1[ \mathbf{X}_l \mathbf{w}_l \geq 0])$ where $\mathbf{w}_l \in\mathbb{R}^{d_l}$ is arbitrary and $1[\mathbf{X}_l \mathbf{w}_l \geq 0] \in\{0,1\}^N$ is an indicator vector with Boolean elements $\left[1[ \mathbf{x}_{l1}^{\top} \mathbf{w}_l \geq 0], \ldots, 1[\mathbf{x}_{lN}^{\top} \mathbf{w}_l  \geq 0]\right]$. Let us enumerate all such distinct diagonal matrices that can be obtained for all possible $\mathbf{w}_l \in \mathbb{R}^{d_l}$, and denote them as $\mathbf{D}_{l1}, \ldots, \mathbf{D}_{lP_l}$. $P_l$ is the number of regions in a partition of $\mathbb{R}^{d_l}$ by hyperplanes passing through the origin, and are perpendicular to the rows of $\mathbf{X}_l$. The convex program is
\begin{equation}\label{finite-convex problem}
\begin{aligned}
& \min_{\{\mathbf{u}_{li}, \mathbf{v}_{li}\}_{i=1}^{P_l}} \frac{1}{2}\|\sum_{i=1}^{P_l} \mathbf{D}_{li} \mathbf{X}_l(\mathbf{u}_{li}-\mathbf{v}_{li})-\mathbf{e}_l\|^2 +\beta_l \sum_{i=1}^{P_l}(\|\mathbf{u}_{li}\|+\|\mathbf{v}_{li}\|) \\ \\
& \text { s.t. }(2 \mathbf{D}_{li}-\mathbf{I}_N) \mathbf{X}_l \mathbf{u}_{li} \geq 0,(2 \mathbf{D}_{li}-\mathbf{I}_N) \mathbf{X}_l \mathbf{v}_{li} \geq 0,  i\in \mathbb{N}_{P_l} .
\end{aligned}
\end{equation}

\section{Convergence Proof}\label{appendix: convergence proof}

\textbf{Proofs of Theorem \ref{theorem: convergence SGDL} and Theorem \ref{theorem: convergence grade l}}

We begin by establishing the convergence of the general gradient descent iteration \ref{general gradient descent}, which serves as the foundation for the proofs of Theorems \ref{theorem: convergence SGDL} and \ref{theorem: convergence grade l}. For a compact convex set $\Omega \in \mathbb{R}^M$, we let
\begin{equation}\label{general alpha}
    \alpha := \sup_{W \in \Omega}\norm{\mathbf{H}_{\mathcal{F}}(W)}
\end{equation}
where $\left\|\cdot\right\|$ is the spectral norm of a matrix. 

\begin{theorem}\label{theorem: general convergence SGDL}
     Suppose $\mathcal{F}: \mathbb{R}^M \to \mathbb{R}$ is a nonnegative, twice continuously differentiable function and $\Omega \in \mathbb{R}^M$ is a convex, compact set. Let $\left\{W^{k}\right\}_{k=1}^{\infty}$ be a sequence generate from \eqref{general gradient descent} for a given initial guess $W^{0}$ and assume that $\left\{W^{k}\right\}_{k=1}^{\infty} \subset \Omega$. If the learning rate $\eta \in (0, 2/\alpha)$, then the following statements hold:
    \begin{enumerate}
        \item [(i)] $\lim_{k \to \infty}\mathcal{F}(W^{k}) = F^*$ for some $F^*\geq 0$;

        \item [(ii)] $\lim_{k \to \infty}\frac{\partial \mathcal{F}}{\partial W}(W^{k}) = 0$ and $\lim_{k\to \infty} \|W^{k+1} - W^{k}\| = 0$;

        \item [(iii)] Every cluster point $\hat W$ of $\left\{W^k\right\}_{k=0}^{\infty}$ satisfies $\frac{\partial \mathcal{F}}{\partial W}(\hat W) = 0$ .
    \end{enumerate}
\end{theorem}

\begin{proof}
Since $\mathcal{F}$ is twice continuously differentiable, we can expand $\mathcal{F}(W^{k+1})$ at $W^{k}$ yields 
\begin{equation*}
\begin{aligned}
&\mathcal{F}(W^{k+1}) =\mathcal{F}(W^{k}) + 
\left(\frac{\partial \mathcal{F}}{\partial W}\right)^{\top}(W^k)
 \Delta W^k
 + r_k
\end{aligned}
\end{equation*}
with an error term
$$
r_k = \frac{1}{2}(\Delta W^k)^{\top}\mathbf{H}_{\mathcal{F}}(\bar W)\Delta W^k
$$
where $\Delta W^k = W^{k+1} - W^k$ and $\bar W$ is a point between $W^k$ and $W^{k+1}$. By using equation \eqref{general gradient descent}, we have that
\begin{equation}\label{gradient Delta W}
\frac{\partial \mathcal{F}}{\partial W}
\left(W^k   
\right)= -\frac{1}{\eta}\Delta  W^k.    
\end{equation}
Therefore,
\begin{equation}\label{eq: estimate 1}
\mathcal{F}(W^{k+1}) =\mathcal{F}(W^{k}) - \frac{1}{\eta}\|\Delta  W^k\|^2 + r_k
\end{equation}

We next estimate $r_k$. Since $\mathcal{F}$ is twice continuously differentiable, $\mathbf{H}_{\mathcal{F}}$ is continuous. As $\Omega$ is compact, $\mathbf{H}_{\mathcal{F}}$ is also bounded on $\Omega$. Moreover, since both $W^{k+1}$ and $W^{k}$ are in the convex set $\Omega$, we have that $\bar W \in \Omega$. It follows from \eqref{general alpha} and compactness of $\Omega$ that
$$
r_k \leq \frac{\alpha}{2}  \|
\Delta W^k
\|^2.
$$
Substituting the above inequality into the right-hand side of equation \eqref{eq: estimate 1}, we have that
\begin{equation}\label{eq: estimate 2}
\mathcal{F}(W^{k+1}) \leq \mathcal{F}(W^k)  - (\frac{1}{\eta} - \frac{\alpha}{2}) \|
\Delta W^k
\|^2.
\end{equation}
Since $\eta \in (0, \frac{2}{\alpha})$, we have that $\frac{1}{\eta} - \frac{\alpha}{2} > 0$. 
The nonnegative of $\frac{1}{\eta} - \frac{\alpha}{2}$ yields 
$$
0\leq \mathcal{F}(W^{k+1}) \leq \mathcal{F}(W^{k}), \text{ for }k=0, 2, \ldots.
$$
This guarantees that $\left\{\mathcal{F}(W^{k}) \right\}_{k=0}^{\infty}$ is a convergent sequence, thereby establishing Item (i).

We next prove Item (ii). For any positive integer $K$, summing inequality \eqref{eq: estimate 2} over $k=0, 1, \ldots, K$ and then we get
$$
\sum_{k=0}^K (\frac{1}{\eta} - \frac{\alpha}{2}) \|
\Delta W^k
\|^2 \leq \mathcal{F}(W^{0}) -   \mathcal{F}(W^{k+1}) \leq\mathcal{F}(W^{0}).
$$
Since $\frac{1}{\eta} - \frac{\alpha}{2}$ is positive, the above inequality implies that
$$
\sum_{k=0}^{\infty} \|
\Delta W^k
\|^2< \infty.
$$
Therefore, 
\begin{equation}\label{eq: diff W 0}
\lim_{k \to \infty} \|
\Delta  W^k
\|  = 0.
\end{equation}
Equation \eqref{gradient Delta W} yields that
\begin{equation}\label{eq: grad 0}
\lim_{k \to \infty} \|\frac{\partial \mathcal{F}}{\partial W}(W^k)\| = 0
\end{equation}
and
$$
\lim_{k\to \infty} \|W^{k+1} - W^{k}\| = 0
$$
which estimates Item (ii).

We next show Item (iii). Let $\hat W$ be a cluster point of $\left\{W^k\right\}_{k=0}^{\infty}$. Then there exists a subsequence $\left\{ W^{k_i}\right\}_{i=0}^{\infty}$ of $\left\{W^k\right\}_{k=0}^{\infty}$ such that $\lim_{i \to \infty}W^{k_i} = \hat{W}$. The continuous of the gradient with Item (ii) implies that
$$
\frac{\partial \mathcal{F}}{\partial W}(\hat W) = \lim_{i \to \infty} \frac{\partial \mathcal{F}}{\partial \tilde W}(W^{k_i}) = 0,
$$
which proves Item (iii).
\end{proof}

\begin{lemma}\label{lemma: continuous}
Suppose that the activation function $\sigma$ is twice continuously differentiable and the loss function $\mathcal{L}$ is defined by \eqref{DNNs loss}, then the gradient $\frac{\partial \mathcal{L}}{\partial  W}$ and hessian $\mathbf{H}_{\mathcal{L}}$ are continuous.
\end{lemma}
\begin{proof}
The key point of the proof is that a polynomial of a continuous function is continuous and so is a composition of continuous function.

    It follows from Lemma 3 of \cite{xu2025multi} that the componetwise of $\frac{\partial \mathcal{L}}{\partial W}$ and $\mathbf{H}_{\mathcal{L}}$ are polynomials of $\sigma, \sigma', \left\{\mathbf{x}_{\ell}, \mathbf{y}_{\ell}\right\}_{\ell=1}^N$ and the composition of $\sigma, \sigma', \sigma''$. These ensure that $\frac{\partial \mathcal{L}}{\partial W}$ and $\mathbf{H}_{\mathcal{L}}$ are continuous.
\end{proof}


\begin{proof}[proof of Theorem \ref{theorem: convergence SGDL}]
    We apply Theorem \ref{theorem: general convergence SGDL} with $\mathcal{F}:=\mathcal{L}$. Under the hypothesis that $\sigma$ is twice continuously differentiable, we have shown in Lemma \ref{lemma: continuous} that the gradient $\frac{\partial \mathcal{L}}{\partial W}$ and $\mathbf{H}_{\mathcal{L}}$ are continuous. Therefore, $\mathcal{L}$ is twice continuously differentiable. It follows from the continuity of $\mathbf{H}_{\mathcal{L}}$ and the compactness of the domain $\mathcal{W}$ that $\alpha$ is finite. Thus, the hypothesis of Theorem \ref{theorem: general convergence SGDL} is satisfied with $\mathcal{F}:=\mathcal{L}$. Theorem \ref{theorem: convergence SGDL} is a direct consequence of Theorem \ref{theorem: general convergence SGDL}.
\end{proof}

\begin{lemma}\label{lemma: continuous grade l}
Suppose that the activation function $\sigma$ is twice continuously differentiable and the loss function $\mathcal{L}_l$ is defined by \eqref{loss: grade l} with  $\left\{\mathbf{x}_{l\ell}, \mathbf{e}_{l\ell}\right\}_{\ell=1}^N$ being bounded, then the gradient $\frac{\partial \mathcal{L}_l}{\partial  W_l}$ and hessian $\mathbf{H}_{\mathcal{L}_l}$ are continuous.
\end{lemma}

\begin{proof}
Since grade 
$l$ in MGDL is essentially a traditional shallow neural network with the only change being the training data, which is replaced by  $\left\{\mathbf{x}_{l\ell}, \mathbf{e}_{l\ell}\right\}_{\ell=1}^N$. We further assume that  $\left\{\mathbf{x}   _{l\ell}, \mathbf{e}_{l\ell}\right\}_{\ell=1}^N$ is bounded. This change does not affect the continuity of the gradient and Hessian. Consequently, this lemma follows directly from Lemma \ref{lemma: continuous}.
\end{proof}

\begin{proof}[Proof of Theorem \ref{theorem: convergence grade l}]
    We apply Theorem \ref{theorem: general convergence SGDL} with $\mathcal{F}:=\mathcal{L}_l$. Under the hypothesis that $\sigma$ is twice continuously differentiable, we have shown in Lemma \ref{lemma: continuous grade l} that the gradient $\frac{\partial \mathcal{L}_l}{\partial W_l}$ and $\mathbf{H}_{\mathcal{L}_l}$ are continuous. Therefore, $\mathcal{L}_l$ is twice continuously differentiable. It follows from the continuity of $\mathbf{H}_{\mathcal{L}_l}$ and the compactness of the domain $\mathcal{W}_l$ that $\alpha$ is finite. Thus, the hypothesis of Theorem \ref{theorem: general convergence SGDL} is satisfied with $\mathcal{F}:=\mathcal{L}_l$. Theorem \ref{theorem: convergence grade l} is a direct consequence of Theorem \ref{theorem: general convergence SGDL}.
\end{proof}

\textbf{Proofs of Theorem \ref{convergence of gradient descent approxi} and Theorem \ref{relation between gradient descent and its linearized form}}




\begin{proof}[Proof of Theorem \ref{convergence of gradient descent approxi}]

The linear approximation of the gradient descent method \eqref{GD matrix form linear approxi} can be expressed as  
\begin{equation}\label{iter: linearization of gradient descent}
\tilde{W}^{k+1}= (\prod_{j=0}^{k-1} \mathbf{A}^j) \tilde{W}^1-\eta\sum_{m=0}^{k-1}(\prod_{j=m+1}^{k-1} \mathbf{A}^j) u^m.
\end{equation}
To establish the convergence of $\tilde{W}^{k+1}$, it suffices to demonstrate that the two terms on the right-hand side of \eqref{iter: linearization of gradient descent} converge separately.

We first show the first term. By the definition of $\tau$, we have that $\|A^k\|\leq \tau$ for each $k$.
By using the property of norm, we have that
$$
\|(\prod_{j=0}^{k-1} \mathbf{A}^j) \tilde{W}^1\| \leq (\prod_{j=0}^{k-1} \|\mathbf{A}^j\|) \|\tilde W^1\| \leq \tau^{k-1} \|\tilde W^1\|. 
$$
 The first term of the right-hand side of \eqref{iter: linearization of gradient descent} converges to zero as $k \to \infty$.

We next show the convergence of the second term of the right-hand side of \eqref{iter: linearization of gradient descent}. We first show that $u^k$ is bound for each $k$. Since $\mathcal{F}$ is twice continuously differentiable, we have that $\frac{\partial \mathcal{F}}{\partial W}(W)$ and $\mathbf{H}_{\mathcal{F}}(W)$ are continuous. By using the condition $\left\{W^k\right\} \subset \Omega$ and $\Omega$ is compact, we obtain that $u^k:=\frac{\partial \mathcal{F}}{\partial W}(W^k) - \mathbf{H}_{\mathcal{F}}(W^k)W^k$ is also bounded for each $k$. Suppose $u^k$ is bounded by $C$ for all $k$.
Then we have that
$$
\|(\prod_{j=m+1}^{k-1} \mathbf{A}^j) u^m\| \leq \tau^{k-1 - m} C.
$$
Hence,
$$
 \sum_{m=0}^{k-1} \| (\prod_{j=m+1}^{k-1} \mathbf{A}^j) u^m \| \leq  C\sum_{m=0}^{k-1} \tau^{k-1 - m}  = C \frac{1 - \tau^{k}}{1 - \tau}.
$$
As $k \to \infty$, we have that
$$
\| \sum_{m=0}^{\infty}(\prod_{j=m+1}^{\infty} \mathbf{A}^j) u^m \| \leq  \sum_{m=0}^{\infty} \|(\prod_{j=m+1}^{\infty} \mathbf{A}^j) u^m \| \leq \frac{C}{1 - \tau},
$$
which proves the second term of the right-hand side of \eqref{iter: linearization of gradient descent} converges. 
\end{proof}

\begin{lemma}\label{lemma: tau less 1}
Suppose $\mathcal{F}: \mathbb{R}^M \to \mathbb{R}$ is a nonnegative, twice continuously differentiable function and $\Omega \in \mathbb{R}^M$ is a convex, compact set. If $\tau < 1$ for all $W \in \Omega$, then $\eta \in (0, 2/\alpha)$ with $\alpha$ defined in \eqref{general alpha}. 
\end{lemma}

\begin{proof}
    Let $W \in \Omega$, and let $\lambda_1(W), \ldots, \lambda_M(W)$ denote the eigenvalue of $\mathbf{H}_{\mathcal{F}}(W)$. By the definition of $\tau$, we have
    $$
    |1 - \eta \lambda_j(W)|\leq \tau, \text{ for all } W \in \Omega, \text{ and } j \in \mathbb{N}_M.
    $$
    Since $\tau < 1$, it follows that $\lambda_j(W) > 0$ and
    $$
     \eta \lambda_j(W) \leq 1+\tau, \text{ for all } W \in \Omega, \text{ and } j \in \mathbb{N}_M.
    $$
     Recalling the definition of $\alpha$, we obtain
    $$
    \eta \alpha \leq 1+\tau.
    $$
    Since $\tau < 1$, we have that $\eta \in (0, 2/\alpha)$, which completes the proof.

\end{proof}

\begin{proof}[Proof of Theorem \ref{relation between gradient descent and its linearized form}]

    The gradient descent iteration \eqref{GD matrix form} can be expressed as 
    \begin{equation}\label{iter: gradient descent full}
    W^{k+1}= (\prod_{j=0}^{k-1} \mathbf{A}^j) W^1- \eta \sum_{m=0}^{k-1}(\prod_{j=m+1}^{k-1} \mathbf{A}^j)u^m + \sum_{m=0}^{k-1}(\prod_{j=m+1}^{k-1} \mathbf{A}^j)r^m .
    \end{equation}
    To establish the theorem, it suffices to demonstrate that the last term on the right-hand side of \eqref{iter: gradient descent full} converges to zero.

    We first show that $\norm{r^k}$ converges to zero. From Lemma \ref{lemma: tau less 1}, $\tau < 1$ includes $\eta \in (0, 2/\alpha)$. Thus, the hypothesis of Theorem \ref{theorem: general convergence SGDL} is satisfied. From the Item (ii) of Theorem \ref{theorem: general convergence SGDL}, we have that $\lim_{k \to \infty }\|W^{k+1} - W^k\| = 0$. Since $\mathcal{F}$ is three times continuously differentiable, $\mathbf{T}_{\mathcal{F}}$ is continuous. As $\Omega$ is compact, $\mathbf{T}_{\mathcal{F}}$ is also bounded on $\Omega$; let its spectral norm be bounded by $C$. Moreover, since both $W^{k+1}$ and $W^{k}$ are in the convex set $\Omega$, we have that $\bar W \in \Omega$. Therefore
$$
\|r^k\| \leq \frac{\eta C}{2}  \|
 W^{k+1} - W^k
\|^2.
$$ 
By using $\lim_{k \to \infty }\|W^{k+1} - W^k\| = 0$, we have $\lim_{k \to 0} \norm{r^k} = 0$.

Since $\norm{r^k}$ converges to zero, for any $\epsilon>0$, there exists an $N$ such that for all $k>N$, $\norm{r^k}<\epsilon$. We split the last term on the right-hand side of \eqref{iter: gradient descent full} into two parts: from $m=0$ to $N-1$ and from $N$ to $k-1$.

For the first part ($m<N$), we have that
$$
\|\sum_{m=0}^{N-1}(\prod_{j=m+1}^{k-1} \mathbf{A}^j)r^m\| \leq  \sum_{m=0}^{N-1}\tau^{k-1 -m}\norm{r^k}.
$$
As $k \to \infty$, $\tau^{k-1-m}$ converges to zero for each $m$, so the entire sum converges to zero. Therefore, we can find a integer $K$ such that for all $k \geq K$
$$
\|\sum_{m=0}^{N-1}(\prod_{j=m+1}^{k-1} \mathbf{A}^j)r^m\| < \epsilon.
$$
For the second part ($m\geq N$):
$$
\|\sum_{m=N}^{k-1}(\prod_{j=m+1}^{k-1} \mathbf{A}^j)r^m\| \leq   \sum_{m=N}^{k-1}\tau^{k-1-m}\norm{r^k} \leq \epsilon \sum_{m=N}^{k-1}\tau^{k-1-m} \leq \frac{\epsilon}{1 - \tau}.
$$

Therefore, for all $k>K$, we have that
$$
\|\sum_{m=0}^{k-1}(\prod_{j=m+1}^{k-1} \mathbf{A}^j)r^m\|  \leq \epsilon + \frac{\epsilon}{1 - \tau}.
$$
As $\epsilon$ is any positive number, the last term on the right-hand side of \eqref{iter: gradient descent full} converges to zero, which completes the proof of the theorem.

\end{proof}

\section{Optimization Problem for Image Reconstruction}\label{Optimization Problem for Image Denoising}


This appendix provides the optimization problems for solving image reconstruction tasks, including image denoising and deblurring.

Image denoising and deblurring are fundamental problems in image processing and have been extensively studied in the literature, as demonstrated in works such as \cite{li2015multi, micchelli2011proximity, micchelli2013proximity}. These problems can be modeled as
$$
\mathbf{\hat{f}} := \mathbf{A} \mathbf{f} + \boldsymbol{\epsilon}
$$
where  $\mathbf{\hat{f}} \in \mathbb{R}^{n \times n}$ denotes the observed corrupted image, $\mathbf{A}$ is a transform operator, $\mathbf{f} \in \mathbb{R}^{n \times n}$ is the ground truth image, and $\boldsymbol{\epsilon}$ is the noise involved in the observed image.  When  $\mathbf{A}$ is the identity transform, the model reduces to an image denoising problem \cite{micchelli2011proximity, micchelli2013proximity}; when $\mathbf{A}$ represents a blurring operator, it corresponds to an image deblurring problem \cite{fang2024inexact}. The statistical nature of the noise $\boldsymbol{\epsilon}$ depends on the specific application: for instance, Gaussian noise is commonly used for natural images \cite{micchelli2011proximity, micchelli2013proximity}, while Poisson noise is typical in medical imaging \cite{guo2022fast}. In this paper, we focus on the case where $\mathbf{A}$ is a Gaussian blurring operator and $\boldsymbol{\epsilon}$ is additive Gaussian noise.

We adopt the Rudin–Osher–Fatemi (ROF) total variation model \cite{rudin1992nonlinear}, one of the most widely used models in image processing, to formalize the optimization problem. We view the gray scale image as a two-dimensional function $f:\mathbb{R}^2 \to \mathbb{R}$ from pixel coordinates to intensity values. The goal is to recover the underlying function $f$ from the observed corrupted image $\mathbf{\hat{f}}$ using neural network-based function approximators. We begin by employing the SGDL framework to reconstruct the function $f$. 

\subsection{Problem Formalization for SGDL}

Let $\mathcal{N}_D(\Theta; \mathbf{x})$ denote the SGDL function, we define the corresponding image matrix $\mathbf{N}_{\Theta} \in \bR^{n \times n}$ be given by $[\mathbf{N}_{\Theta}]_{i, j}:=  \mathcal{N}_D(\Theta; \mathbf{x}_{i, j})$. The objective function is
\begin{equation}\label{ROF model}
\mathcal{G}(\Theta): = \frac{1}{2}\norm{\mathbf{
\hat{f}} - \mathbf{A}\mathbf{N}_{\Theta}}_{\text{F}}^2 + \lambda \norm{\mathbf{B}\mathbf{N}_{\Theta}}_{1, 1}  
\end{equation}
where $\mathbf{B}$ denotes the first-order difference operator used to produce total-variation, as described in \cite{micchelli2011proximity}. The norm $\|\cdot\|_{\text{F}}$ denotes the Frobenius norm of a matrix, which is the $\ell_2$ norm of the vectorized form. The norm $\|\cdot\|_{1,1}$ denotes the entrywise $\ell_1$ norm of a matrix, defined as the $\ell_1$ norm of its vectorized form. The regularization parameter $\lambda > 0$ balances the trade-off between data fidelity and regularization. Since the objective function \eqref{ROF model} involves a non-differentiable term $\norm{\cdot}_{1, 1}$ composed with the operator $\mathbf{B}$, minimizing \eqref{ROF model} is challenging, as discussed in \cite{fang2024inexact}. To address this issue, following the approach in \cite{fang2024inexact, shen2016wavelet, wu2022inverting}, we introduce an auxiliary variable $\mathbf{u}$ to free $\mathbf{B}\mathbf{N}_{\Theta}$ for the non-differentiable norm $\norm{\cdot}_{1, 1}$ and add the difference between $\mathbf{u}$ and $\mathbf{B}\mathbf{N}_{\Theta}$ as a penalized term, resulting the following objective function:
$$
\mathcal{L}(\Theta, \mathbf{u}): = \frac{1}{2}\norm{\mathbf{
\hat{f}} - \mathbf{A}\mathbf{N}_{\Theta}}_{\text{F}}^2 + \frac{\beta}{2}\norm{\mathbf{u} - \mathbf{B}\mathbf{N}_{\Theta}}_{\text{F}}^2 +\lambda\norm{\mathbf{u}}_{1, 1}.    
$$
The optimization problem to reconstruct the underlying function can be formulated as
\begin{equation}\label{regression: SGDL}
    \text{argmin} \left\{ \mathcal{L}(\Theta, \mathbf{u}): \Theta \in \bR^{M_D}, \mathbf{u}\in \bR^{2n \times n} \right\}.
\end{equation}
Since $\mathcal{L}$ is non-differentiable with respect to $\mathbf{u}$ and differentiable with respect to $\Theta$, we use the proximity operator to update $\mathbf{u}$ and use gradient-based method to update $\Theta$. The resulting proximity-gradient algorithm is
\begin{align}
    &\mathbf{u}^{k+1} = \text{prox}_{\alpha \lambda /\beta \norm{\cdot}_{1, 1}}\left( \alpha \mathbf{B}\mathbf{N}_{\Theta^{k}} + (1 - \alpha) \mathbf{u}^k\right), \label{regression SGDL: prox update u}\\
    &\text{Using Adam optimizer to minimize }  \mathcal{L}(\Theta, \mathbf{u}^{k+1})  \text{ and obtain } \Theta^{k+1}. \label{regression SGDL: adam theta}
\end{align}

\subsection{Problem Formalization for MGDL}

We now employ the MGDL framework to reconstruct the function $f$. Grade 1 for MGDL is the same as for SGDL; the only difference is that the hidden layer is relatively smaller. We let 
\begin{equation}\label{regression: MGDL grade1 loss function}
\mathcal{L}_1(\Theta_1, \mathbf{u}): = \frac{1}{2}\norm{\mathbf{
    \hat{f}} - \mathbf{A}\mathbf{N}_{\Theta_1}}_{\text{F}}^2 + \frac{\beta}{2}\norm{\mathbf{u} - \mathbf{B}\mathbf{N}_{\Theta_1}}_{\text{F}}^2 +\lambda \norm{\mathbf{u}}_{1, 1}.  
\end{equation}
The optimization problem can be formulated as
\begin{equation}\label{regression: MGDL grade 1}
    \text{argmin} \left\{ \mathcal{L}_1(\Theta_1, \mathbf{u}): \Theta_1 \in \bR^{M_{D_1}}, \mathbf{u}\in \bR^{2n \times n} \right\}.
\end{equation}

For grade $l$ with $l \geq 2$, we will learn a new function $\mathcal{N}_{D_l}$ which composite with $\mathcal{H}_{D_{l-1}-1}(\Theta_{l-1}^*; \cdot) \circ \ldots \circ \mathcal{H}_{D_{1}-1}(\Theta_1^*; \cdot) $. Let $[\mathbf{g}_{\Theta^*_{1}}]_{i, j}:= \mathcal{N}_{D_{1}}(\Theta_1^*; \mathbf{x}_{i, j})$ and 
$$
[\mathbf{g}_{\Theta_l}]_{i, j}:= \epsilon_l\mathcal{N}_{D_l}(\Theta_l; \cdot) \circ \mathcal{H}_{D_{l-1}-1}(\Theta_{l-1}^*; \cdot) \circ \ldots \circ \mathcal{H}_{D_{1}-1}(\Theta_1^*; \mathbf{x}_{i, j}) + [\mathbf{g}_{\Theta^*_{l-1}}]_{i, j}
$$
where $\epsilon_l$ is a given normalization parameter,
then the loss for grade $l$ is
\begin{equation}\label{regression: MGDL grade l loss function}
\mathcal{L}_l(\Theta_l, \mathbf{u}): = \frac{1}{2}\norm{\mathbf{
\hat{f}} - \mathbf{A}\mathbf{g}_{\Theta_l}}_{\text{F}}^2 + \frac{\beta}{2}\norm{\mathbf{u} - \mathbf{B}\mathbf{g}_{\Theta_l}}_{\text{F}}^2 +\lambda \norm{\mathbf{u}}_{1, 1}.  
\end{equation}
The minimization problem for grade $l$ can be formulated as
\begin{equation}\label{regression: MGDL grade l}
    \text{argmin} \left\{ \mathcal{L}_l(\Theta_l, \mathbf{u}): \Theta_l \in \bR^{M_{D_l}}, \mathbf{u}\in \bR^{2n \times n} \right\}.
\end{equation}

Both optimization problems \eqref{regression: MGDL grade 1} and \eqref{regression: MGDL grade l} are solved using the proximity-gradient algorithm \eqref{regression SGDL: prox update u}–\eqref{regression SGDL: adam theta}, with the objective function replaced accordingly.



\section{Network Structures}\label{appendix: architecture}

This Appendix provides the network structures used in the paper. 

For SGDL, we use the following network structures:
\begin{equation}\label{SGDL-1}
[1] \to [32] \times 4 \to [1],    
\end{equation}
\begin{equation}\label{SGDL-4}
[2] \to [48] \times 4 \to [1],
\end{equation}
\begin{equation}\label{SGDL-2}
[2] \to [128] \times 8 \to [1],
\end{equation}
\begin{equation}\label{SGDL-3}
[2] \to [128] \times 12 \to [1],
\end{equation}
where $[n]\times N$ indicates $N$ hidden layers, each with $n$ neurons. The corresponding network structures for MGDL are: 
\begin{equation}\label{MGDL-1}
\begin{aligned}
    &\text{Grade 1}: [1] \rightarrow[32]  \rightarrow[1]\\
&\text{Grade 2}: [1] \rightarrow[32]_F  \rightarrow[32] \rightarrow[1]\\
&\text{Grade 3}: [1] \rightarrow[32]_F \times 2 \rightarrow[32]  \rightarrow[1]\\
&\text{Grade 4}: [1] \rightarrow[32]_F \times 3 \rightarrow[32]  \rightarrow[1].
\end{aligned}
\end{equation}

\begin{equation}\label{MGDL-4}
\begin{aligned}
    &\text{Grade 1}: [2] \rightarrow[48]  \rightarrow[1]\\
&\text{Grade 2}: [2] \rightarrow[48]_F  \rightarrow[48] \rightarrow[1]\\
&\text{Grade 3}: [2] \rightarrow[48]_F \times 2 \rightarrow[48]  \rightarrow[1]\\
&\text{Grade 4}: [2] \rightarrow[48]_F \times 3 \rightarrow[48]  \rightarrow[1].
\end{aligned}
\end{equation}

\begin{equation}\label{MGDL-2}
\begin{aligned}
    &\text{Grade 1}: [2] \rightarrow[128] \times 2  \rightarrow[1]\\
&\text{Grade 2}: [2] \rightarrow[128]_F \times 2 \rightarrow[128] \times 2 \rightarrow[1]\\
&\text{Grade 3}: [2] \rightarrow[128]_F \times 4 \rightarrow[128] \times 2 \rightarrow[1]\\
&\text{Grade 4}: [2] \rightarrow[128]_F \times 6 \rightarrow[128] \times 2 \rightarrow[1].
\end{aligned}
\end{equation}

\begin{equation}\label{MGDL-3}
\begin{aligned}
    &\text{Grade 1}: [2] \rightarrow[128] \times 3  \rightarrow[1]\\
&\text{Grade 2}: [2] \rightarrow[128]_F \times 3 \rightarrow[128] \times 3 \rightarrow[1]\\
&\text{Grade 3}: [2] \rightarrow[128]_F \times 6 \rightarrow[128] \times 3 \rightarrow[1]\\
&\text{Grade 4}: [2] \rightarrow[128]_F \times 9 \rightarrow[128] \times 3 \rightarrow[1].
\end{aligned}
\end{equation}
Here, $[n]_F$ indicates a layer having $n$ neurons with parameters, trained in the previous grades, remaining fixed during the training of the current grade.

For a multi-scale neural network \cite{liu2020mscalednn} (MSDL), we consider the network with four subnetworks, each with the structure
$$
[1] \to [8] \times 4 \to [1]
$$
where the scale coefficients is $\left\{1, 2, 4, 8\right\}$.

Throughout the paper, ReLU is used as the activation function for SGDL, MSDL, and MGDL.

\section{Supporting material for Sections \ref{section: comparsion image reconstruction}-\ref{section:MGDL and MSDL}}\label{appendix: comparsion image reconstruction}

The experiments conducted in  Sections \ref{section: comparsion image reconstruction}-\ref{section:MGDL and MSDL} were performed on X86\_64 server equipped with AMD 7543 @ 2.8GHz (64 slots) and AVX512, 2 x Nvidia Ampere A100 GPU.

The quality of the reconstructed image is evaluated by the peak signal-to-noise ratio (PSNR) defined by
\begin{equation}\label{PSNR}
\text { PSNR }:=10 \log _{10}\left((n \times 255^2)/\left\| \mathbf{v}- \mathbf{\hat v}\right\|_{\text{F}}^2\right)
\end{equation}
where $\mathbf{v}$ is the ground truth image, $\mathbf{\hat v}$ is reconstructed image, $n$ is the number of pixels in $\mathbf{v}$, and $\norm{\mathbf{\cdot}}_{\text{F}}$ denotes the Frobenius norm of a matrix.

\subsection{Section \ref{section: comparsion image reconstruction}}

\begin{figure}[H]
  \centering
   
\begin{subfigure}{0.21\linewidth}
\includegraphics[width=\linewidth]{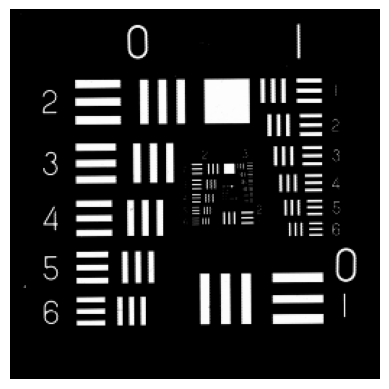}
\caption{}
    \end{subfigure}
    \hfill
   \begin{subfigure}{0.21\linewidth}
\includegraphics[width=\linewidth]{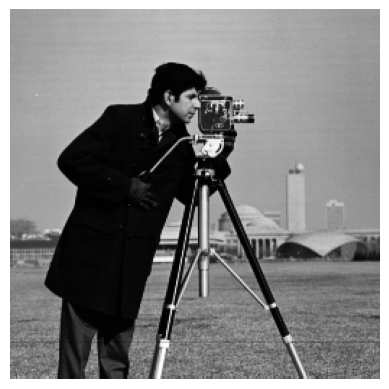}
\caption{}
    \end{subfigure}
    \hfill
   \begin{subfigure}{0.21\linewidth}
\includegraphics[width=\linewidth]{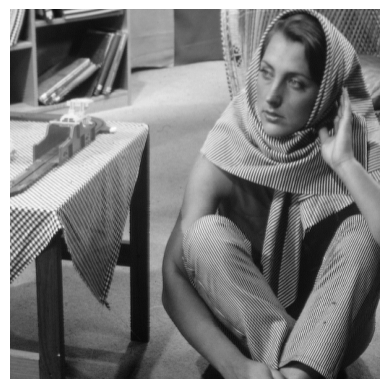}
\caption{}
    \end{subfigure}
    \hfill
\begin{subfigure}{0.21\linewidth}
\includegraphics[width=\linewidth]{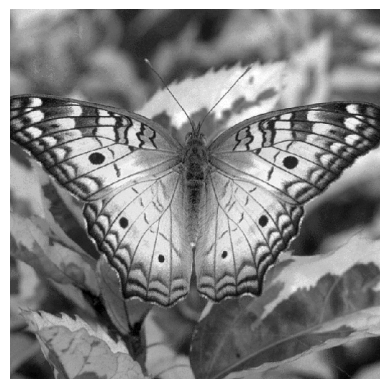}
\caption{}
    \end{subfigure}
    
\begin{subfigure}{0.21\linewidth}
\includegraphics[width=\linewidth]{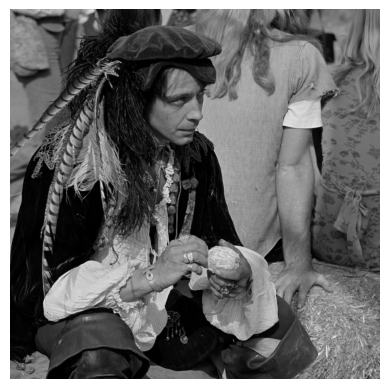}
\caption{}
    \end{subfigure}
    \hfill
\begin{subfigure}{0.21\linewidth}
\includegraphics[width=\linewidth]{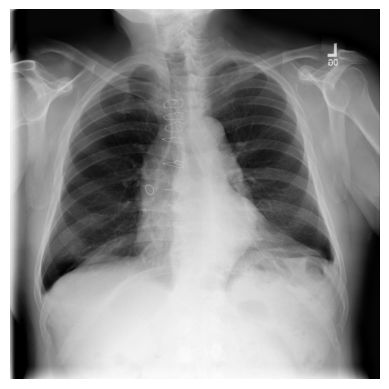}
\caption{}
    \end{subfigure}
    \hfill
\begin{subfigure}{0.3\linewidth}
\includegraphics[width=\linewidth]{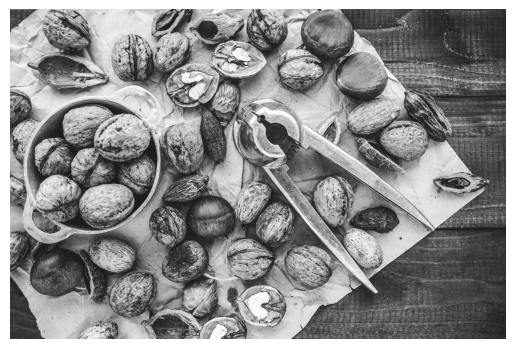}
\caption{}
    \end{subfigure}

\caption{Clean images: (a) `Resolution chart', (b) `Cameraman', (c) `Barbara', (d) `Butterfly', (e) `Pirate', (f) `Chest', (g) `Walnut'. 
}
	\label{fig:testing images}
\end{figure}

\begin{figure}[H]
    \centering
    \begin{subfigure}{0.4\linewidth}
\includegraphics[width=\linewidth]{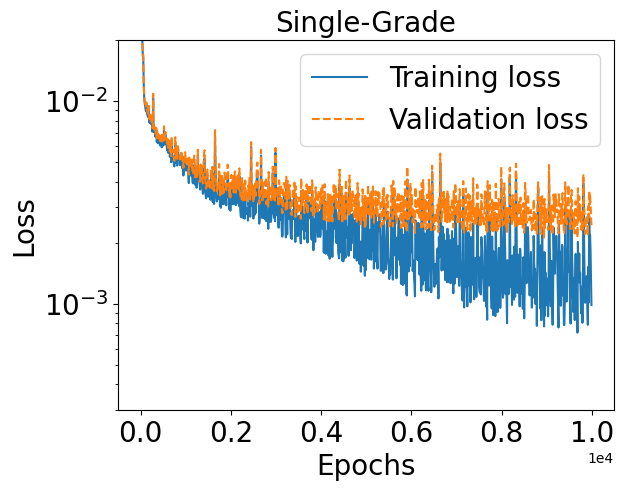}
\caption{}
    \end{subfigure}
    \begin{subfigure}{0.4\linewidth}
\includegraphics[width=\linewidth]{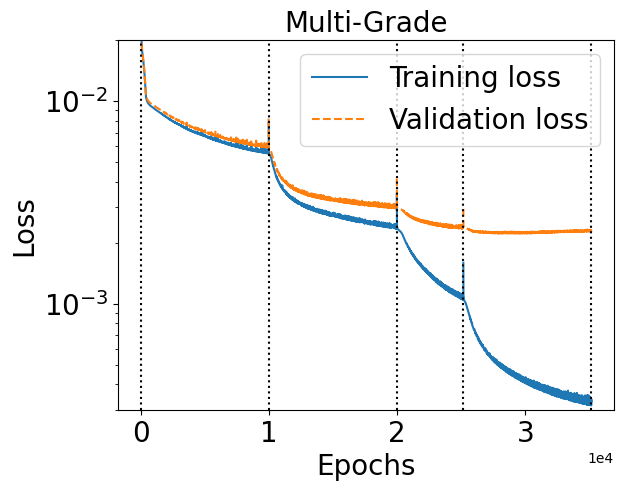}
\caption{}
    \end{subfigure}

    \begin{subfigure}{0.2\linewidth}
\includegraphics[width=\linewidth]{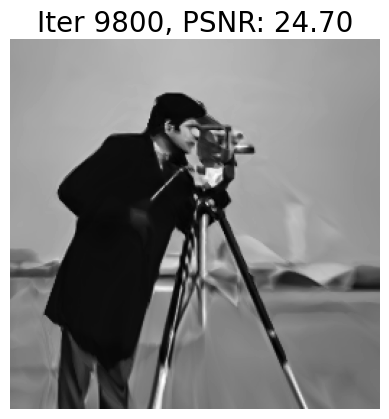}
\caption{}
    \end{subfigure}
\begin{subfigure}{0.2\linewidth}
\includegraphics[width=\linewidth]{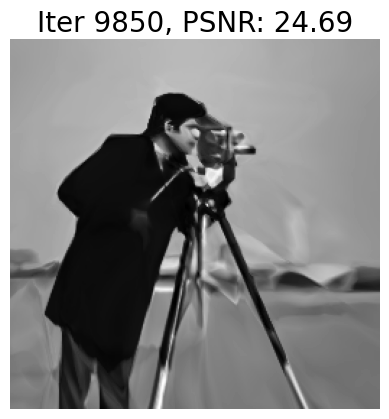}
\caption{}
    \end{subfigure}
    \begin{subfigure}{0.2\linewidth}
\includegraphics[width=\linewidth]{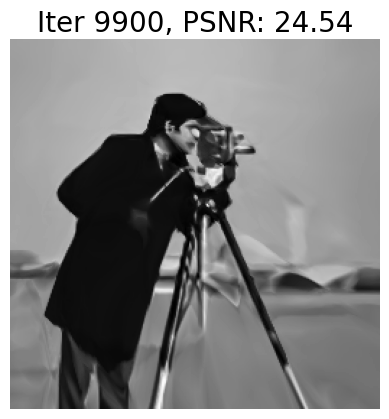}
\caption{}
    \end{subfigure}
    \begin{subfigure}{0.2\linewidth}
\includegraphics[width=\linewidth]{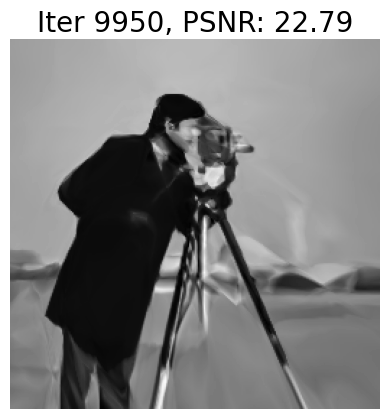}
\caption{}
    \end{subfigure}

   \begin{subfigure}{0.2\linewidth}
\includegraphics[width=\linewidth]{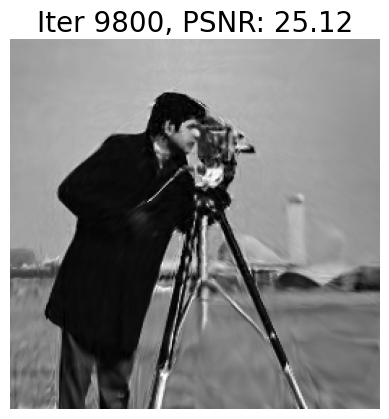}
\caption{}
    \end{subfigure}
\begin{subfigure}{0.2\linewidth}
\includegraphics[width=\linewidth]{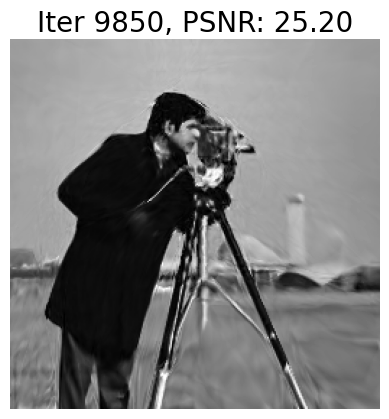}
\caption{}
    \end{subfigure}
    \begin{subfigure}{0.2\linewidth}
\includegraphics[width=\linewidth]{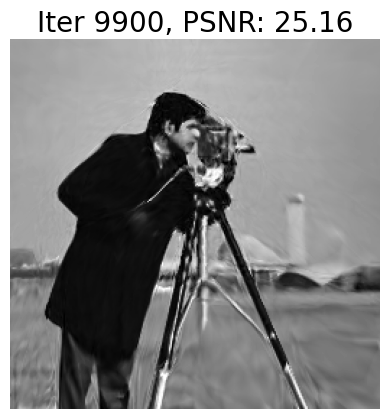}
\caption{}
    \end{subfigure}
    \begin{subfigure}{0.2\linewidth}
\includegraphics[width=\linewidth]{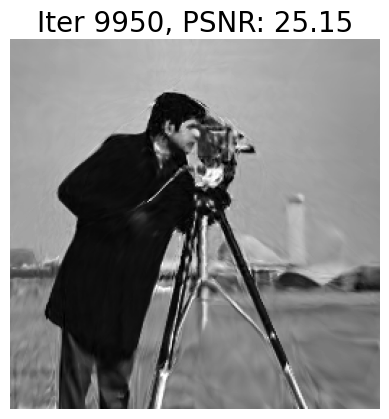}
\caption{}
    \end{subfigure}
    \caption{Comparison of SGDL and MGDL for the `Cameraman' image. (a)-(b) show the loss curves for SGDL and MGDL. (c)-(f) present the SGDL predictions at iterations 9800, 9850, 9900, and 9950. (g)-(j) display the MGDL predictions at iterations 9800, 9850, 9900, and 9950 for grade 4. The PSNR values are provided in the titles of (c)-(j).}
    \label{fig:SGDL MGDL cameraman}
\end{figure}

\begin{figure}[H]
    \centering
    \begin{subfigure}{0.28\linewidth}
\includegraphics[width=\linewidth]{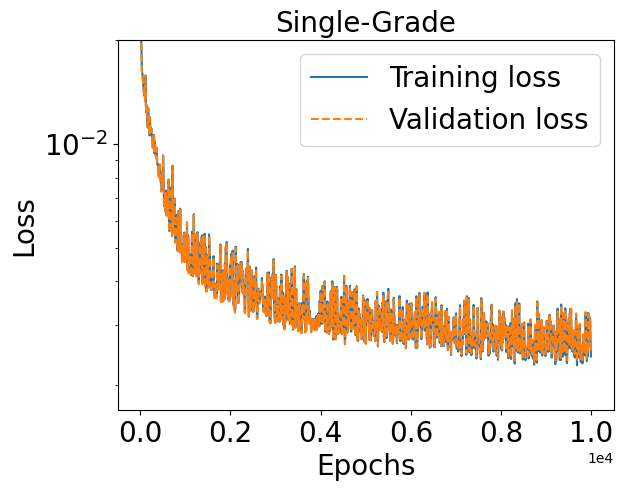}
\caption{}
    \end{subfigure}
    \begin{subfigure}{0.28\linewidth}
\includegraphics[width=\linewidth]{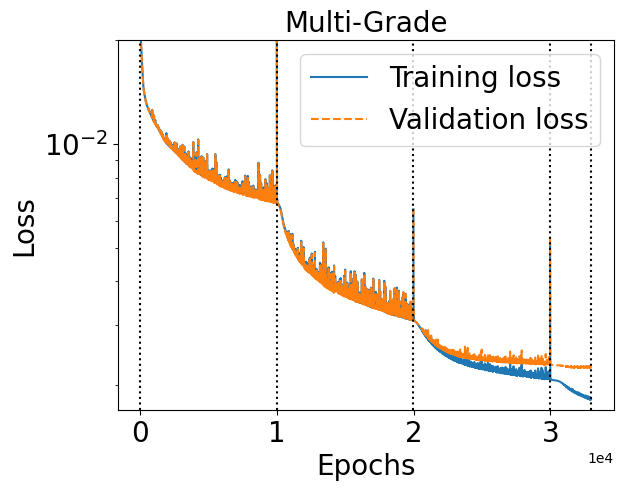}
\caption{}
    \end{subfigure}
    \begin{subfigure}{0.21\linewidth}
\includegraphics[width=\linewidth]{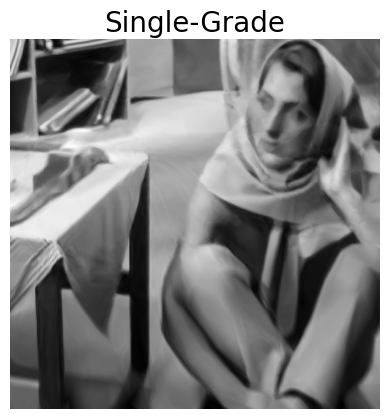}
\caption{PSNR: 22.75}
    \end{subfigure}
    \begin{subfigure}{0.21\linewidth}
\includegraphics[width=\linewidth]{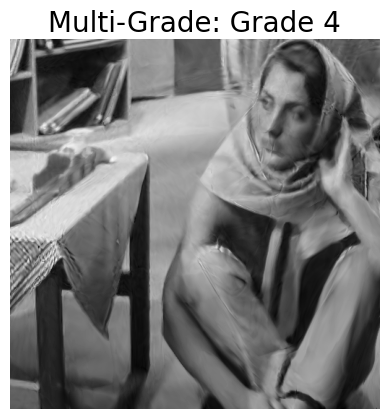}
\caption{PSNR: 23.84}
    \end{subfigure}

    \begin{subfigure}{0.28\linewidth}
\includegraphics[width=\linewidth]{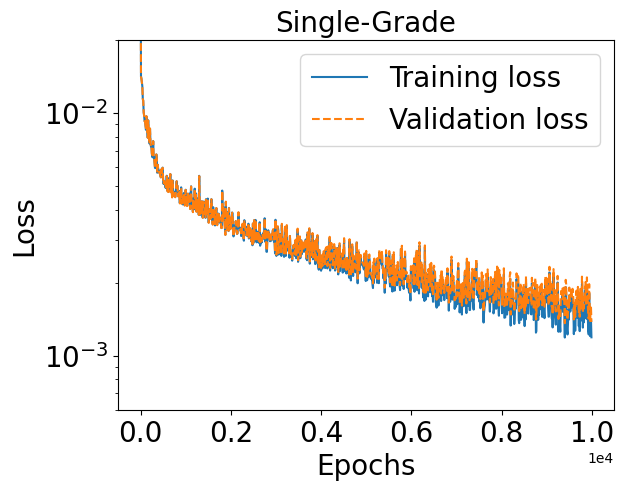}
\caption{}
    \end{subfigure}
    \begin{subfigure}{0.28\linewidth}
\includegraphics[width=\linewidth]{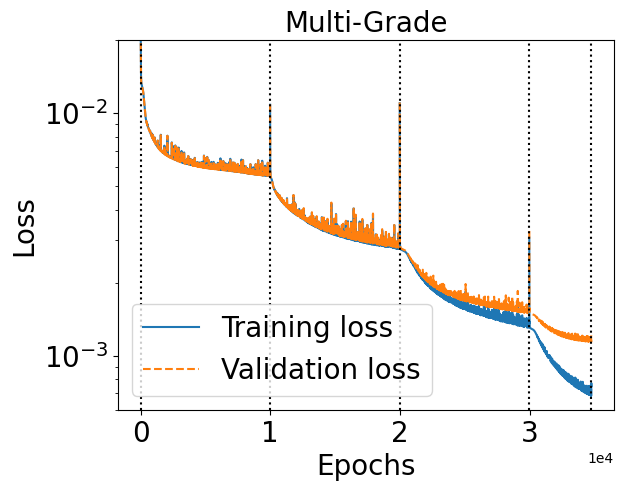}
\caption{}
    \end{subfigure}
    \begin{subfigure}{0.21\linewidth}
\includegraphics[width=\linewidth]{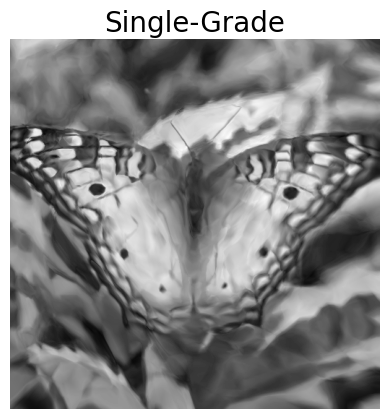}
\caption{PSNR: 24.87}
    \end{subfigure}
    \begin{subfigure}{0.21\linewidth}
\includegraphics[width=\linewidth]{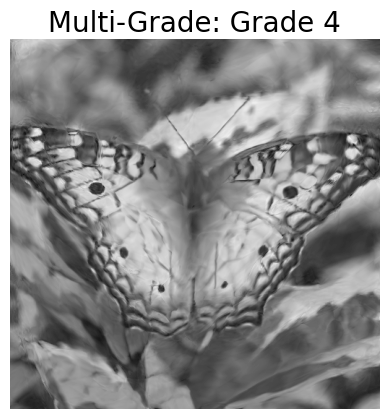}
\caption{PSNR: 27.06}
    \end{subfigure}

    \begin{subfigure}{0.28\linewidth}
\includegraphics[width=\linewidth]{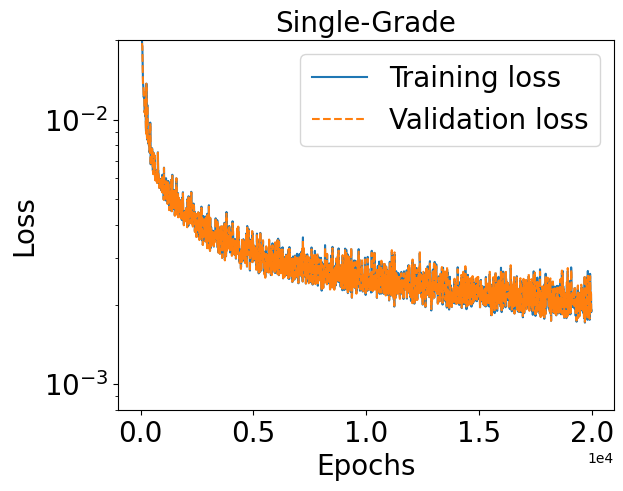}
\caption{}
    \end{subfigure}
    \begin{subfigure}{0.28\linewidth}
\includegraphics[width=\linewidth]{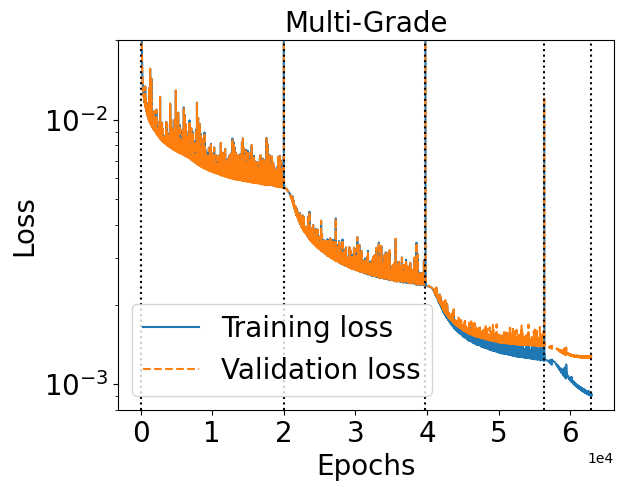}
\caption{}
    \end{subfigure}
    \begin{subfigure}{0.21\linewidth}
\includegraphics[width=\linewidth]{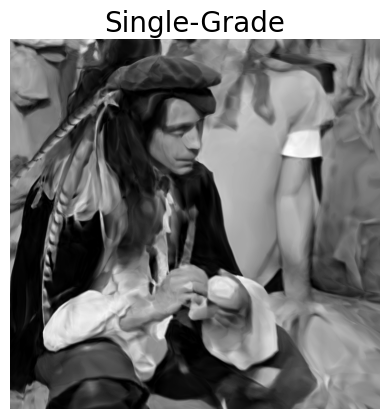}
\caption{PSNR: 24.34}
    \end{subfigure}
    \begin{subfigure}{0.21\linewidth}
\includegraphics[width=\linewidth]{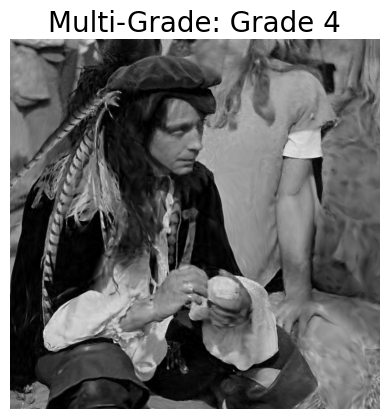}
\caption{PSNR: 26.45}
    \end{subfigure}

   \begin{subfigure}{0.28\linewidth}
\includegraphics[width=\linewidth]{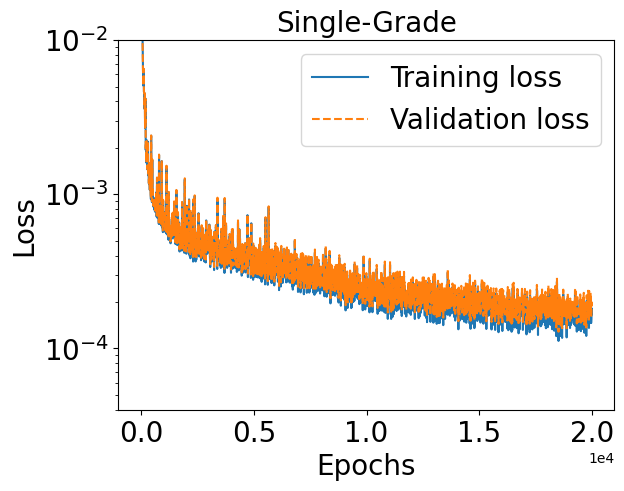}
\caption{}
    \end{subfigure}
    \begin{subfigure}{0.28\linewidth}
\includegraphics[width=\linewidth]{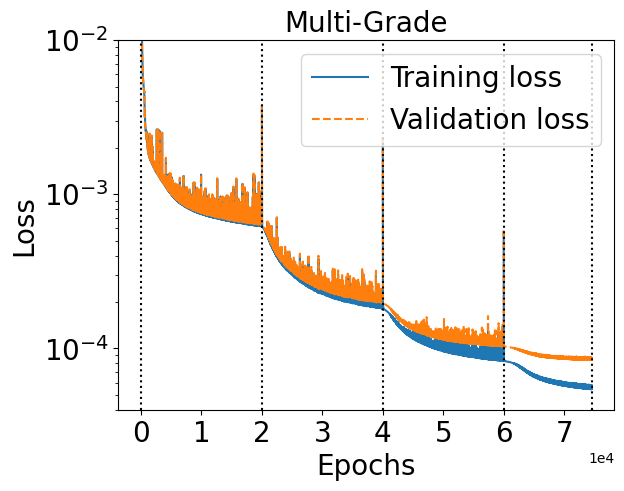}
\caption{}
    \end{subfigure}
    \begin{subfigure}{0.21\linewidth}
\includegraphics[width=\linewidth]{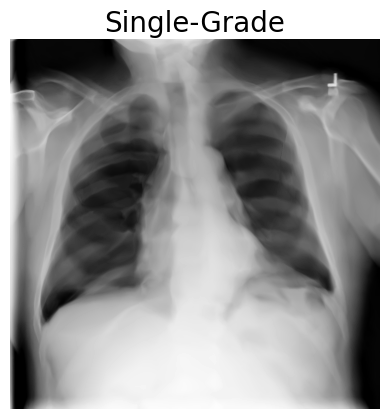}
\caption{PSNR: 35.56}
    \end{subfigure}
    \begin{subfigure}{0.21\linewidth}
\includegraphics[width=\linewidth]{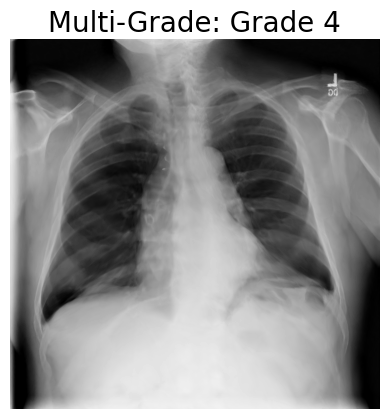}
\caption{PSNR: 38.49}
    \end{subfigure}

   \begin{subfigure}{0.28\linewidth}
\includegraphics[width=\linewidth]{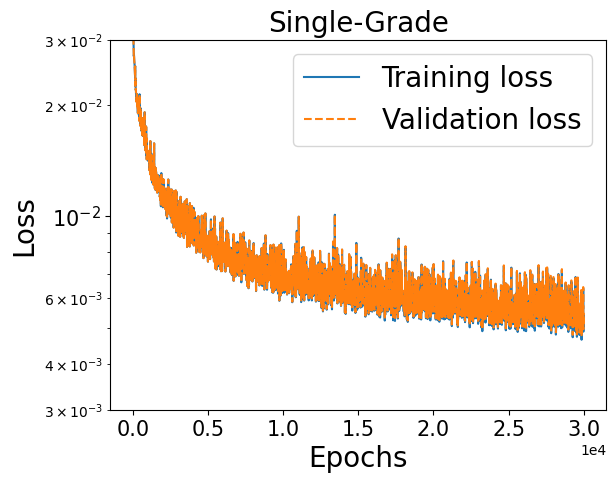}
\caption{}
    \end{subfigure}
    \begin{subfigure}{0.28\linewidth}
\includegraphics[width=\linewidth]{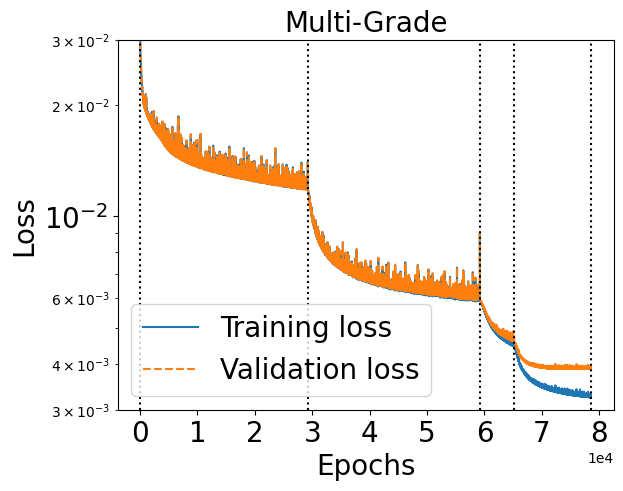}
\caption{}
    \end{subfigure}
    \begin{subfigure}{0.21\linewidth}
\includegraphics[width=\linewidth]{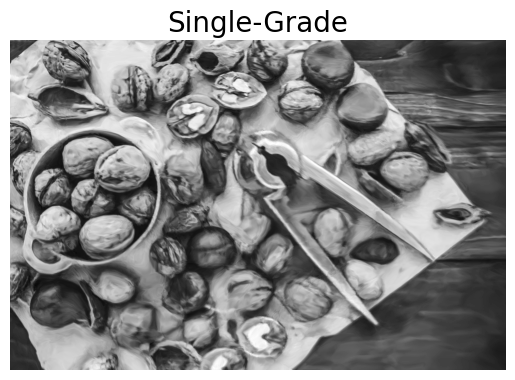}
\caption{PSNR: 20.05}
    \end{subfigure}
    \begin{subfigure}{0.21\linewidth}
\includegraphics[width=\linewidth]{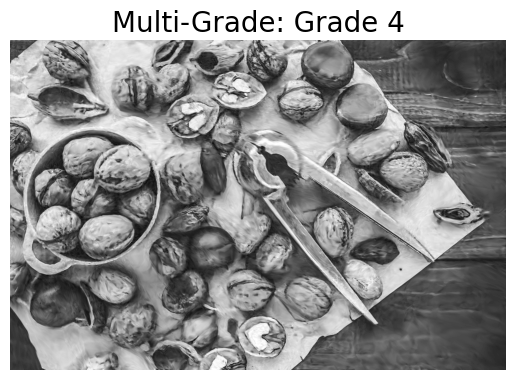}
\caption{PSNR: 21.31}
    \end{subfigure}
    
    \caption{Comparison of SGDL and MGDL for image regression. (a)-(d): `Barbara'; (e)-(h): `Butterfly'; (i)-(l): `Pirate'; (m)-(p): `Chest'; (q)-(t): `Walnut'. 
    }
    \label{fig:SGDL MGDL image regression}
\end{figure}

\begin{figure}[H]
  \centering

   \begin{subfigure}{0.24\linewidth}
\includegraphics[width=\linewidth]{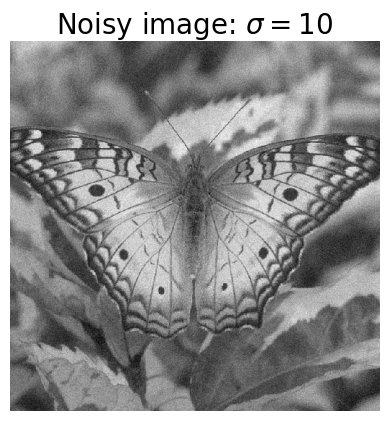}
\caption{PSNR: 28.12}
    \end{subfigure}
   \begin{subfigure}{0.24\linewidth}
\includegraphics[width=\linewidth]{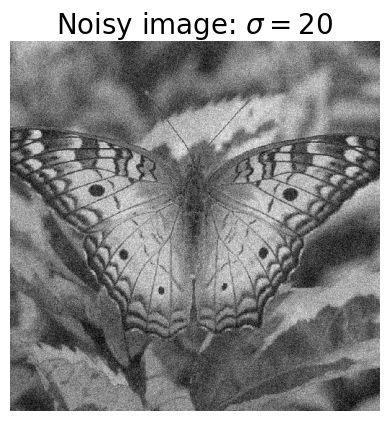}
\caption{PSNR: 22.10}
    \end{subfigure}
\begin{subfigure}{0.24\linewidth}
\includegraphics[width=\linewidth]{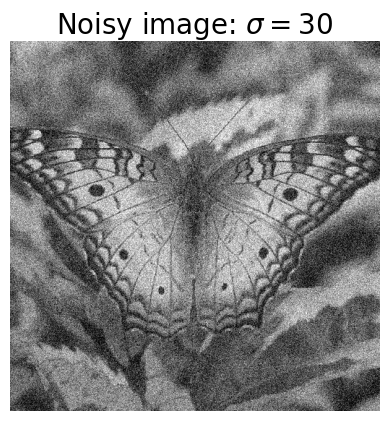}
\caption{PSNR: 18.57}
    \end{subfigure}
\begin{subfigure}{0.24\linewidth}
\includegraphics[width=\linewidth]{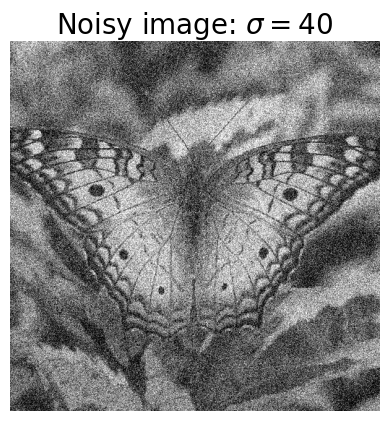}
\caption{PSNR: 16.08}
    \end{subfigure}

\begin{subfigure}{0.24\linewidth}
\includegraphics[width=\linewidth]{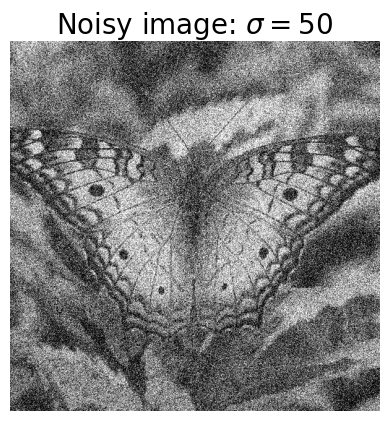}
\caption{PSNR: 14.14}
    \end{subfigure}
\begin{subfigure}{0.24\linewidth}
\includegraphics[width=\linewidth]{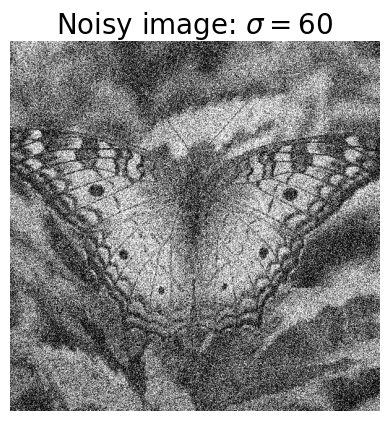}
\caption{PSNR: 12.55}
    \end{subfigure}
   \begin{subfigure}{0.24\linewidth}
\includegraphics[width=\linewidth]{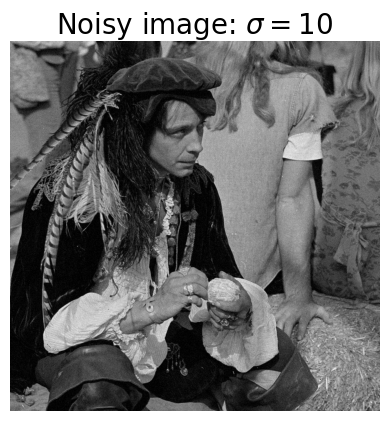}
\caption{PSNR: 28.14}
    \end{subfigure}
   \begin{subfigure}{0.24\linewidth}
\includegraphics[width=\linewidth]{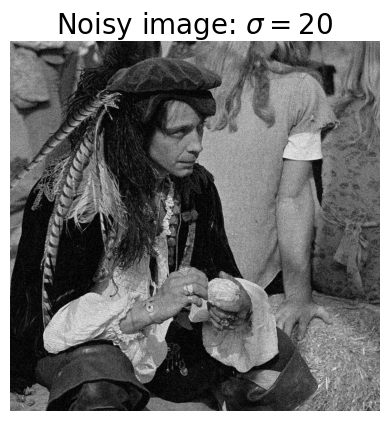}
\caption{PSNR: 22.12}
    \end{subfigure}

\begin{subfigure}{0.24\linewidth}
\includegraphics[width=\linewidth]{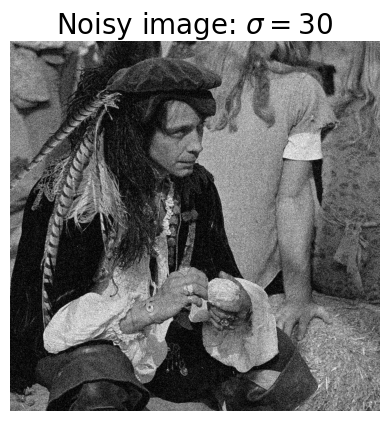}
\caption{PSNR: 18.60}
    \end{subfigure}
\begin{subfigure}{0.24\linewidth}
\includegraphics[width=\linewidth]{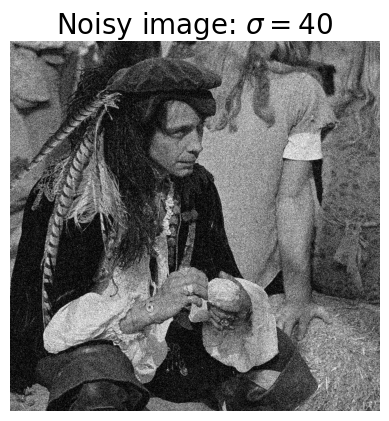}
\caption{PSNR: 16.10}
    \end{subfigure}
\begin{subfigure}{0.24\linewidth}
\includegraphics[width=\linewidth]{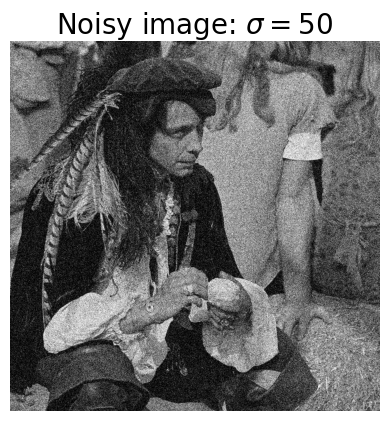}
\caption{PSNR: 14.16}
    \end{subfigure}
\begin{subfigure}{0.24\linewidth}
\includegraphics[width=\linewidth]{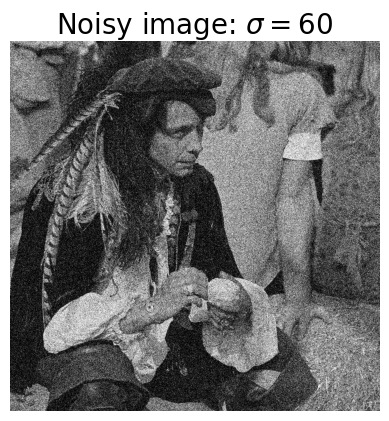}
\caption{PSNR: 12.57}
    \end{subfigure}

   \begin{subfigure}{0.24\linewidth}
\includegraphics[width=\linewidth]{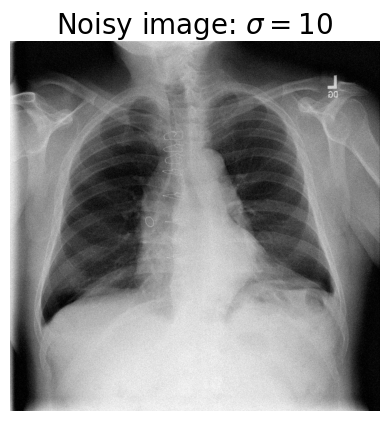}
\caption{PSNR: 28.14}
    \end{subfigure}
   \begin{subfigure}{0.24\linewidth}
\includegraphics[width=\linewidth]{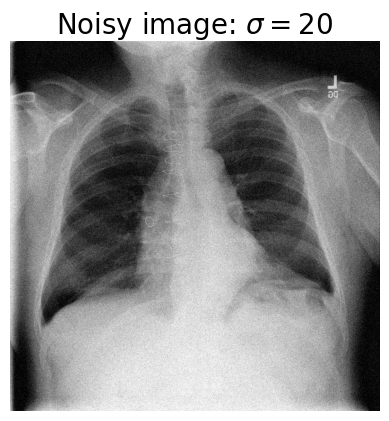}
\caption{PSNR: 22.12}
    \end{subfigure}
\begin{subfigure}{0.24\linewidth}
\includegraphics[width=\linewidth]{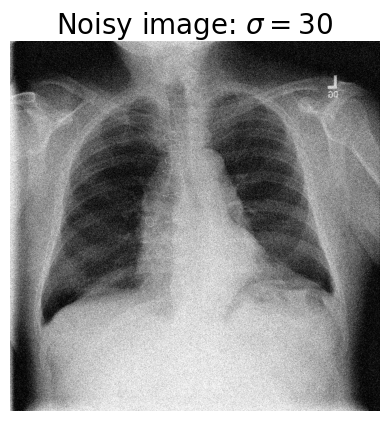}
\caption{PSNR: 18.60}
    \end{subfigure}
\begin{subfigure}{0.24\linewidth}
\includegraphics[width=\linewidth]{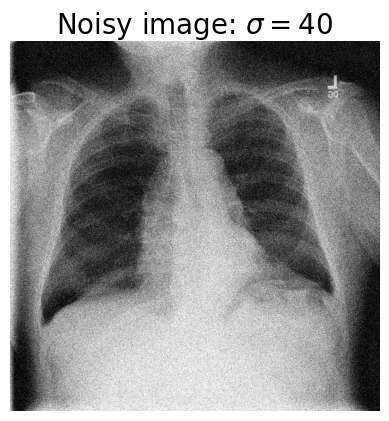}
\caption{PSNR: 16.10}
    \end{subfigure}
    
\begin{subfigure}{0.24\linewidth}
\includegraphics[width=\linewidth]{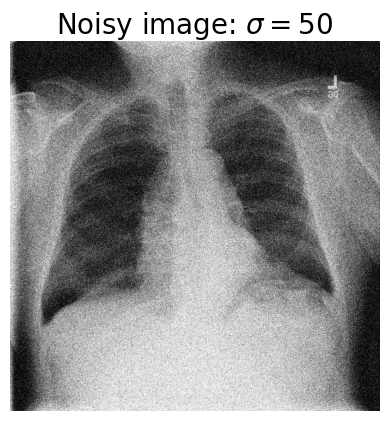}
\caption{PSNR: 14.16}
    \end{subfigure}
\begin{subfigure}{0.24\linewidth}
\includegraphics[width=\linewidth]{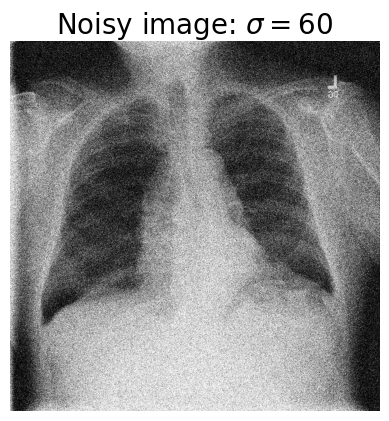}
\caption{PSNR: 12.57}
    \end{subfigure}

\caption{Noisy images: `Butterfly' (a)-(f), `Pirate' (g)-(l), and `Chest' (m)-(r). The PSNR value is shown in each title.
}
	\label{fig: noisy image}
\end{figure}

\begin{figure}[H]
    \centering
    
    \begin{subfigure}{0.27\linewidth}
\includegraphics[width=\linewidth]{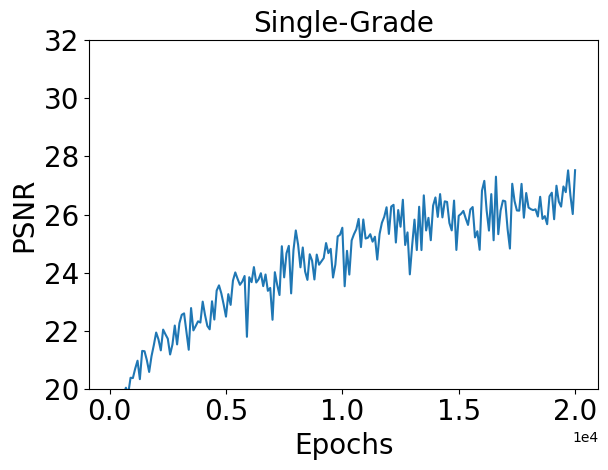}
\caption{}
    \end{subfigure}
    \begin{subfigure}{0.27\linewidth}
\includegraphics[width=\linewidth]{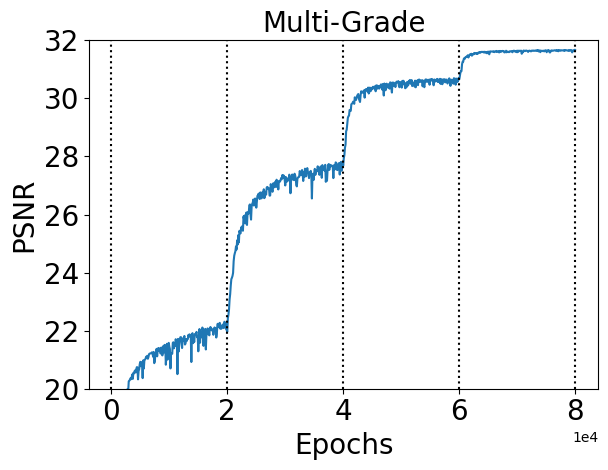}
\caption{}
    \end{subfigure}
    \begin{subfigure}{0.20\linewidth}
\includegraphics[width=\linewidth]{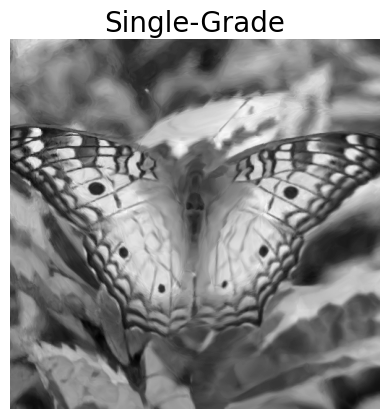}
\caption{PSNR: 27.53}
    \end{subfigure}
    \begin{subfigure}{0.20\linewidth}
\includegraphics[width=\linewidth]{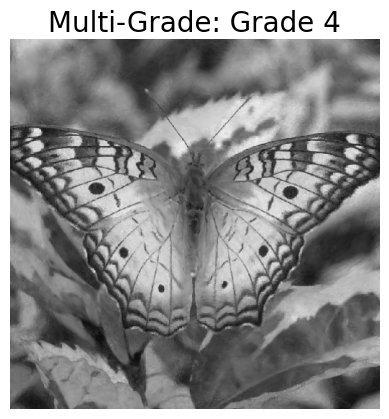}
\caption{PSNR: 31.67}
    \end{subfigure}

    \begin{subfigure}{0.27\linewidth}
\includegraphics[width=\linewidth]{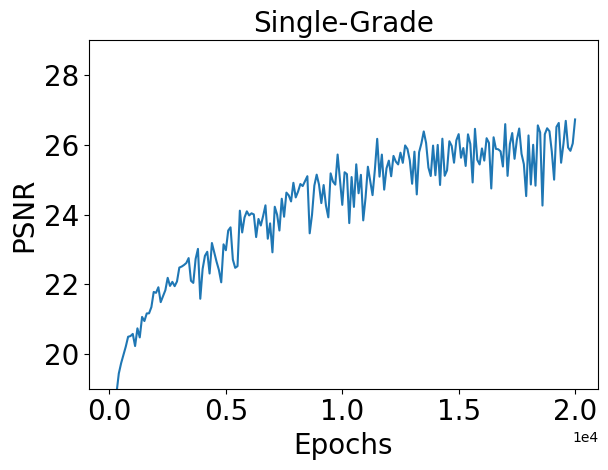}
\caption{}
    \end{subfigure}
    \begin{subfigure}{0.27\linewidth}
\includegraphics[width=\linewidth]{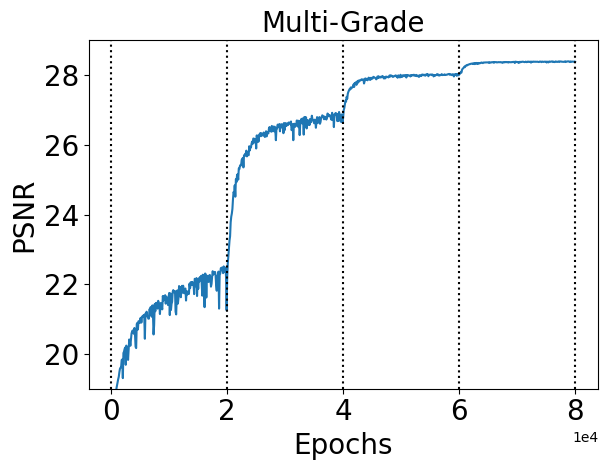}
\caption{}
    \end{subfigure}
    \begin{subfigure}{0.20\linewidth}
\includegraphics[width=\linewidth]{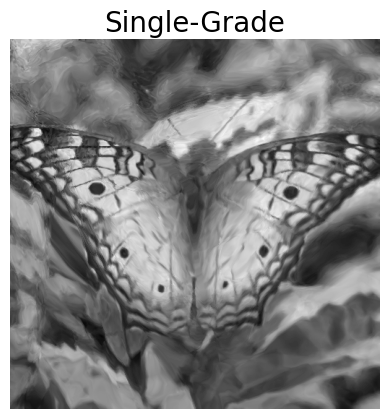}
\caption{PSNR: 26.73}
    \end{subfigure}
    \begin{subfigure}{0.20\linewidth}
\includegraphics[width=\linewidth]{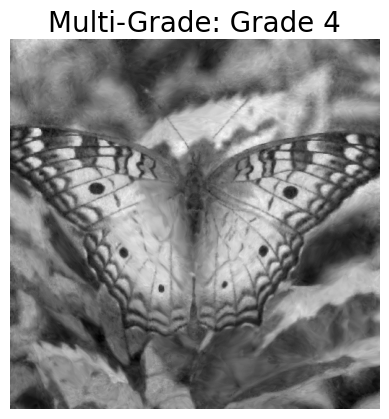}
\caption{PSNR: 28.39}
    \end{subfigure}

    \begin{subfigure}{0.27\linewidth}
\includegraphics[width=\linewidth]{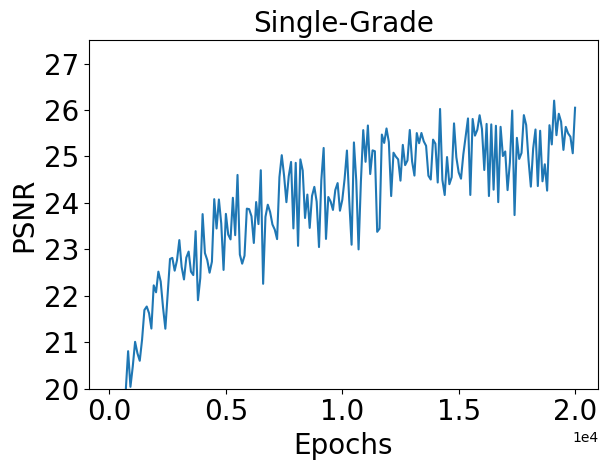}
\caption{}
    \end{subfigure}
    \begin{subfigure}{0.27\linewidth}
\includegraphics[width=\linewidth]{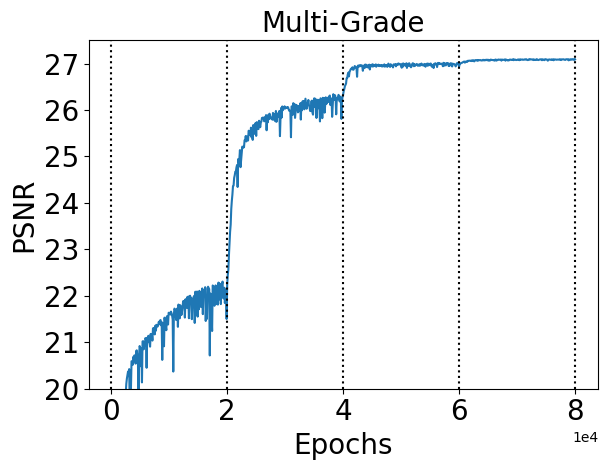}
\caption{}
    \end{subfigure}
    \begin{subfigure}{0.20\linewidth}
\includegraphics[width=\linewidth]{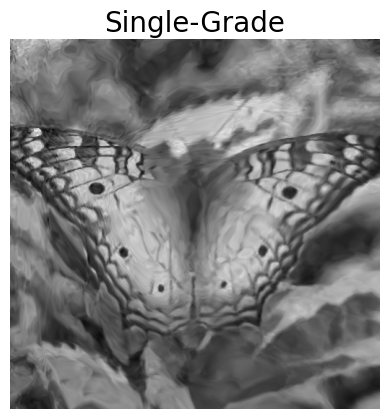}
\caption{PSNR: 26.05}
    \end{subfigure}
    \begin{subfigure}{0.20\linewidth}
\includegraphics[width=\linewidth]{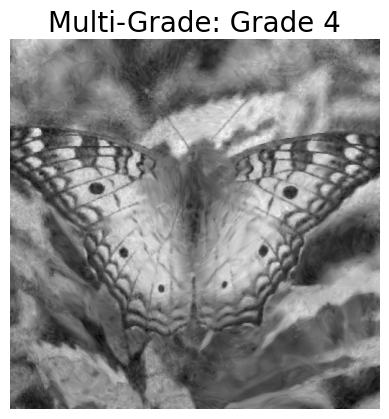}
\caption{PSNR: 27.09}
    \end{subfigure}

    \begin{subfigure}{0.27\linewidth}
\includegraphics[width=\linewidth]{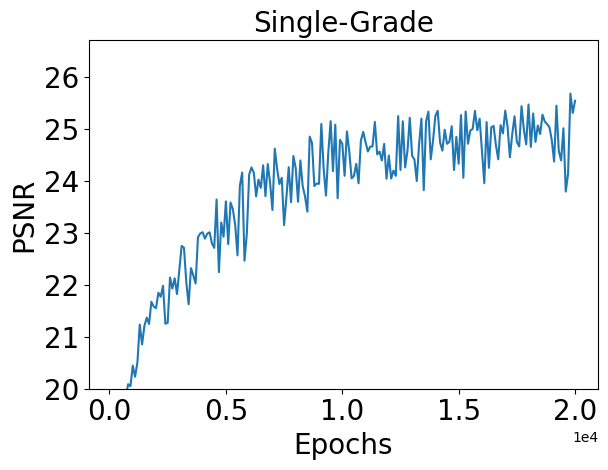}
\caption{}
    \end{subfigure}
    \begin{subfigure}{0.27\linewidth}
\includegraphics[width=\linewidth]{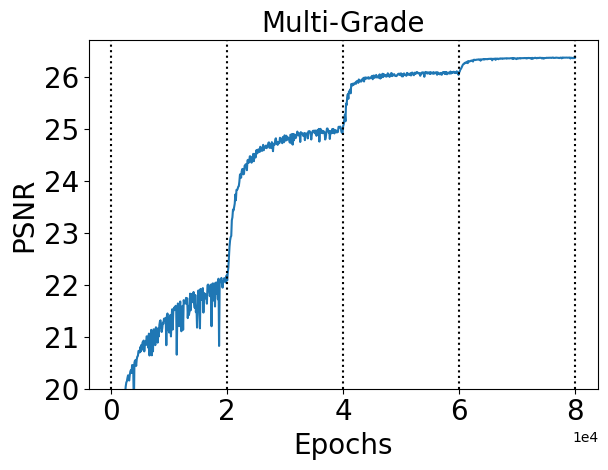}
\caption{}
    \end{subfigure}
    \begin{subfigure}{0.20\linewidth}
\includegraphics[width=\linewidth]{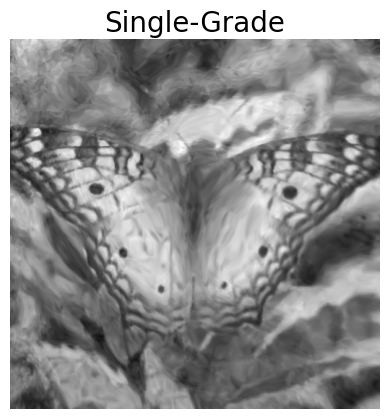}
\caption{PSNR: 25.54}
    \end{subfigure}
    \begin{subfigure}{0.20\linewidth}
\includegraphics[width=\linewidth]{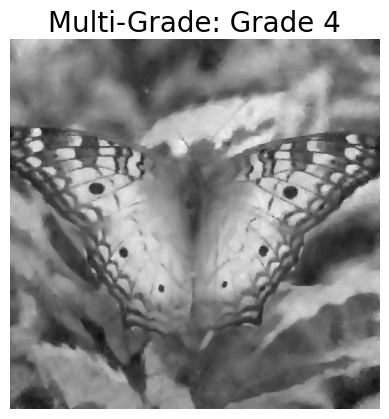}
\caption{PSNR: 26.37}
    \end{subfigure}

    \begin{subfigure}{0.27\linewidth}
\includegraphics[width=\linewidth]{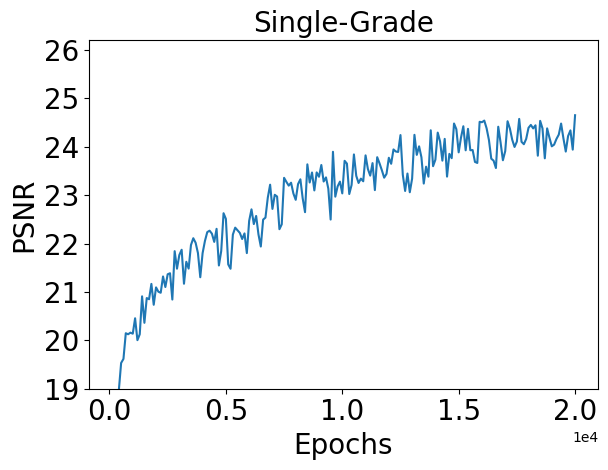}
\caption{}
    \end{subfigure}
    \begin{subfigure}{0.27\linewidth}
\includegraphics[width=\linewidth]{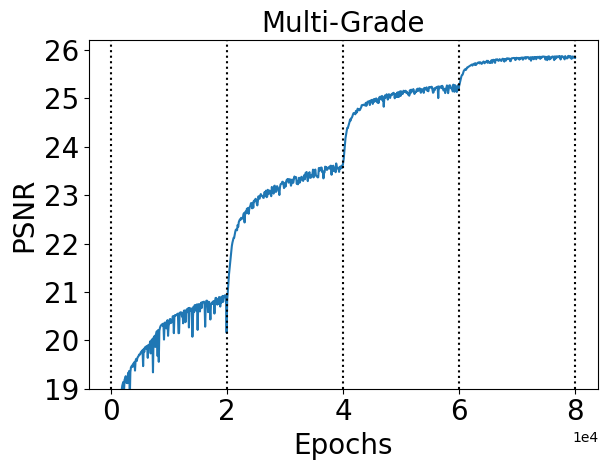}
\caption{}
    \end{subfigure}
    \begin{subfigure}{0.20\linewidth}
\includegraphics[width=\linewidth]{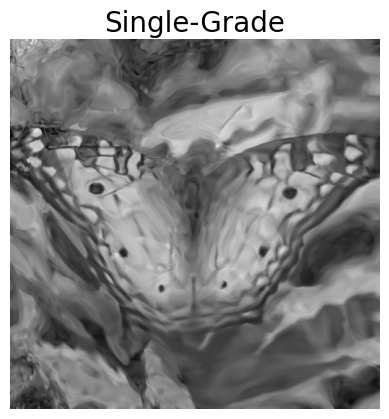}
\caption{PSNR: 24.65}
    \end{subfigure}
    \begin{subfigure}{0.20\linewidth}
\includegraphics[width=\linewidth]{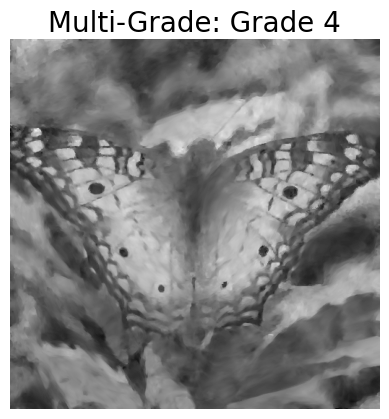}
\caption{PSNR: 25.84}
    \end{subfigure}

    \begin{subfigure}{0.27\linewidth}
\includegraphics[width=\linewidth]{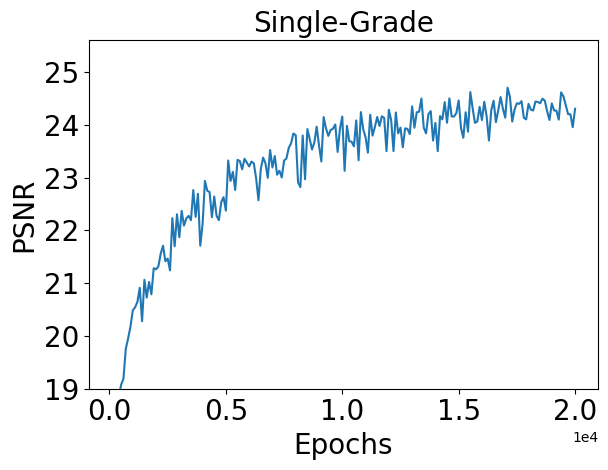}
\caption{}
    \end{subfigure}
    \begin{subfigure}{0.27\linewidth}
\includegraphics[width=\linewidth]{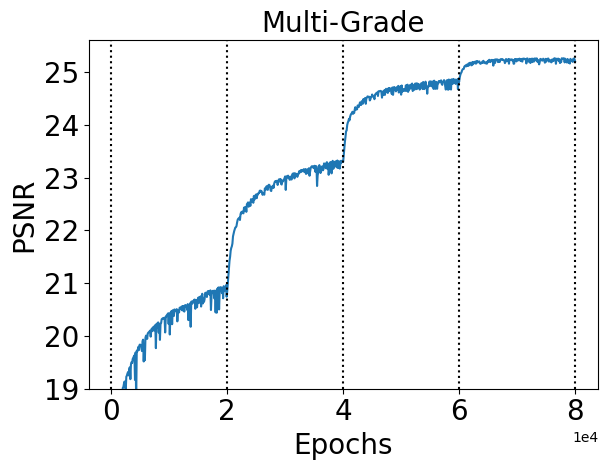}
\caption{}
    \end{subfigure}
    \begin{subfigure}{0.20\linewidth}
\includegraphics[width=\linewidth]{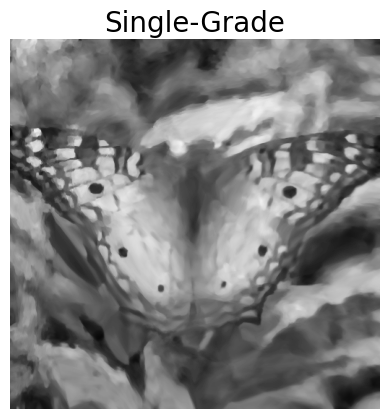}
\caption{PSNR: 24.30}
    \end{subfigure}
    \begin{subfigure}{0.20\linewidth}
\includegraphics[width=\linewidth]{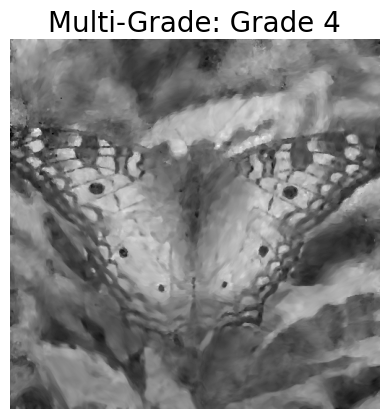}
\caption{PSNR: 25.21}
    \end{subfigure}

    \caption{Comparison of SGDL \eqref{SGDL-3} and MGDL \eqref{MGDL-3} denoising results for the `Butterfly' image. Rows one to six show noise levels $\sigma = 10, 20, 30, 40, 50, 60$, with PSNR during training and reconstructed images displayed with PSNR values in the subtitles.
    }
    \label{fig:SGDL MGDL butterfly image denoising}
\end{figure}

\begin{figure}[H]
    \centering
    
    \begin{subfigure}{0.27\linewidth}
\includegraphics[width=\linewidth]{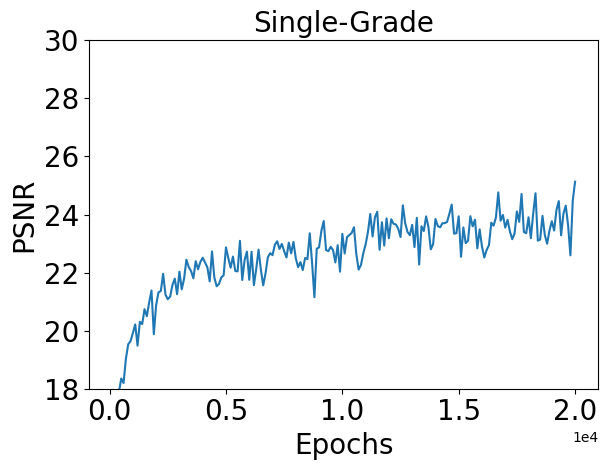}
\caption{}
    \end{subfigure}
    \begin{subfigure}{0.27\linewidth}
\includegraphics[width=\linewidth]{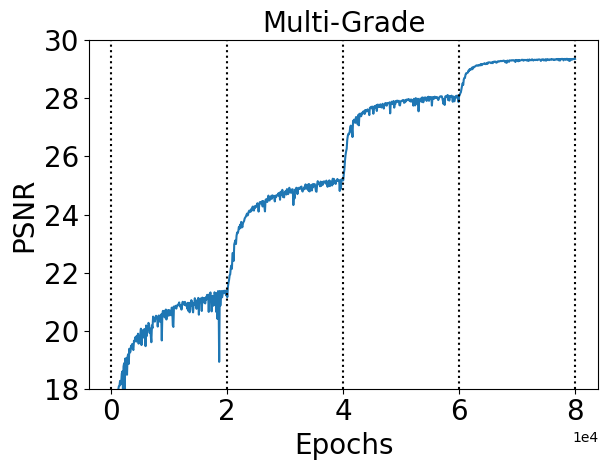}
\caption{}
    \end{subfigure}
    \begin{subfigure}{0.20\linewidth}
\includegraphics[width=\linewidth]{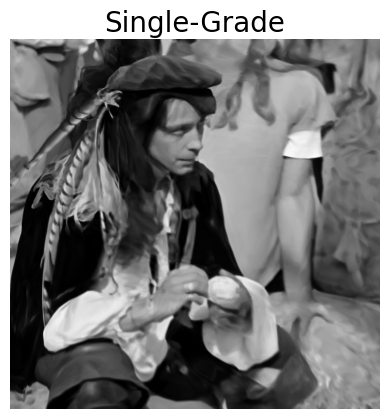}
\caption{PSNR: 25.12}
    \end{subfigure}
    \begin{subfigure}{0.20\linewidth}
\includegraphics[width=\linewidth]{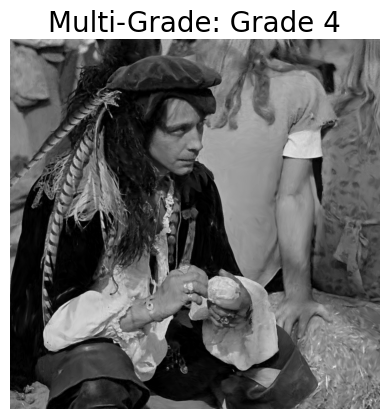}
\caption{PSNR: 29.28}
    \end{subfigure}

    \begin{subfigure}{0.27\linewidth}
\includegraphics[width=\linewidth]{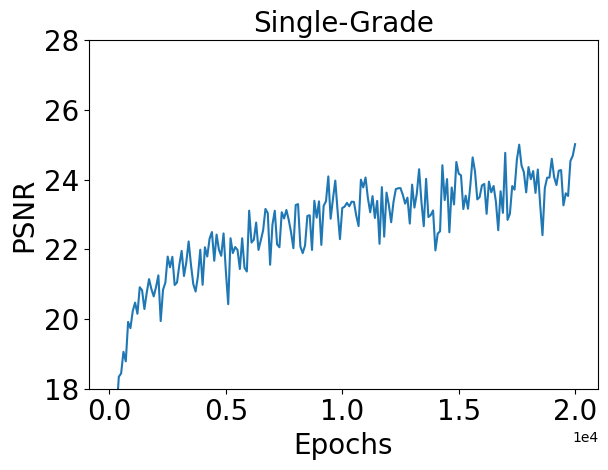}
\caption{}
    \end{subfigure}
    \begin{subfigure}{0.27\linewidth}
\includegraphics[width=\linewidth]{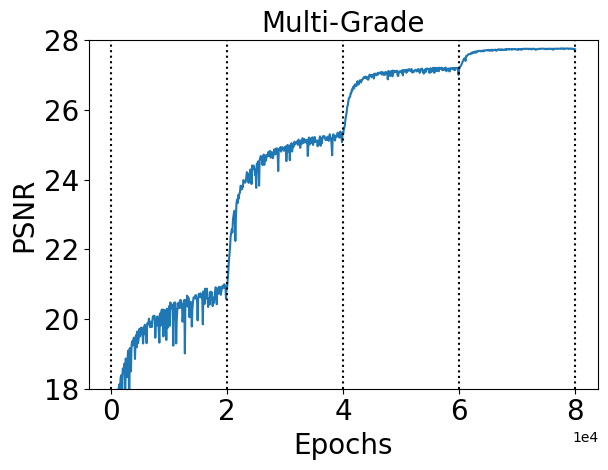}
\caption{}
    \end{subfigure}
    \begin{subfigure}{0.20\linewidth}
\includegraphics[width=\linewidth]{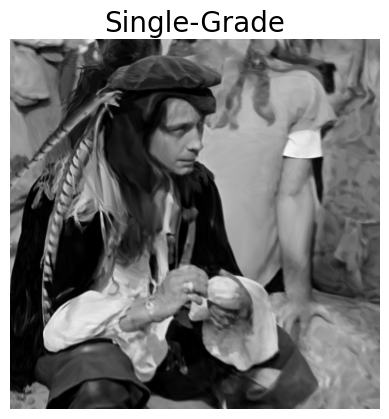}
\caption{PSNR: 25.02}
    \end{subfigure}
    \begin{subfigure}{0.20\linewidth}
\includegraphics[width=\linewidth]{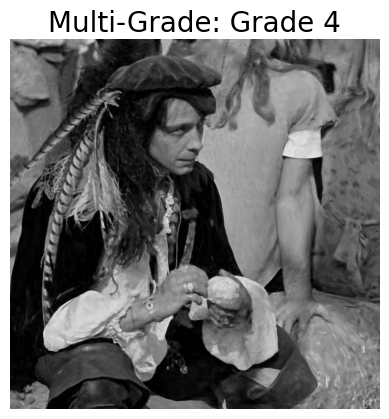}
\caption{PSNR: 27.74}
    \end{subfigure}

    \begin{subfigure}{0.27\linewidth}
\includegraphics[width=\linewidth]{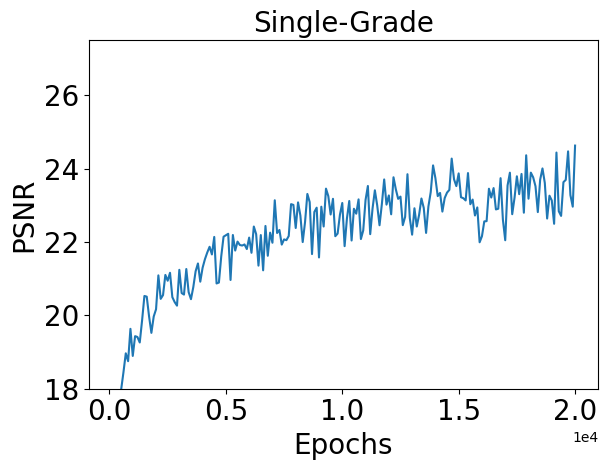}
\caption{}
    \end{subfigure}
    \begin{subfigure}{0.27\linewidth}
\includegraphics[width=\linewidth]{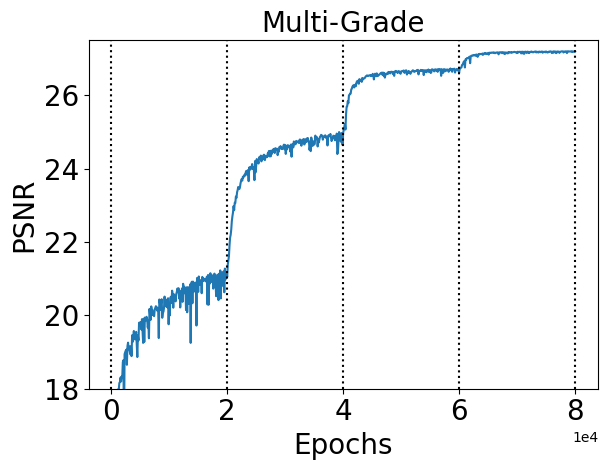}
\caption{}
    \end{subfigure}
    \begin{subfigure}{0.20\linewidth}
\includegraphics[width=\linewidth]{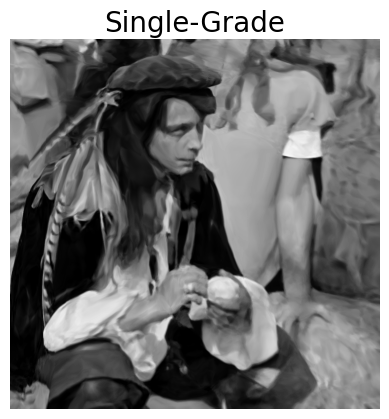}
\caption{PSNR: 24.63}
    \end{subfigure}
    \begin{subfigure}{0.20\linewidth}
\includegraphics[width=\linewidth]{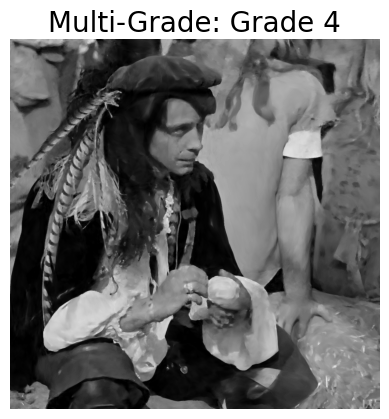}
\caption{PSNR: 27.20}
    \end{subfigure}

    \begin{subfigure}{0.27\linewidth}
\includegraphics[width=\linewidth]{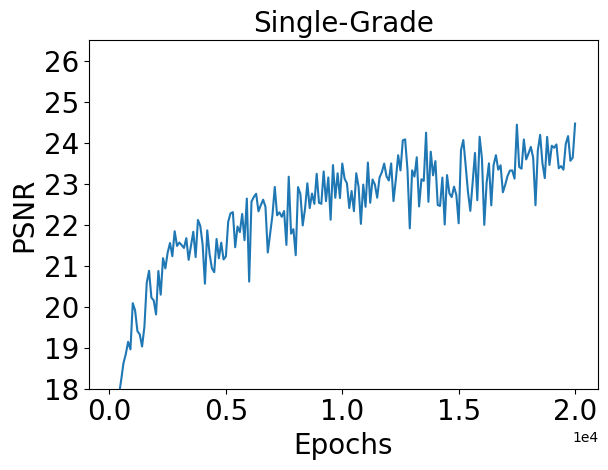}
\caption{}
    \end{subfigure}
    \begin{subfigure}{0.27\linewidth}
\includegraphics[width=\linewidth]{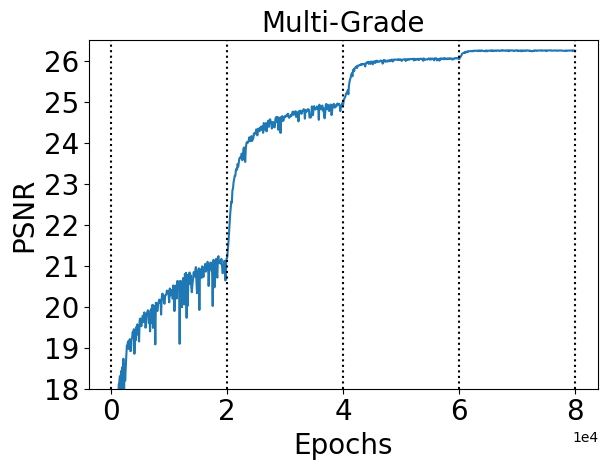}
\caption{}
    \end{subfigure}
    \begin{subfigure}{0.20\linewidth}
\includegraphics[width=\linewidth]{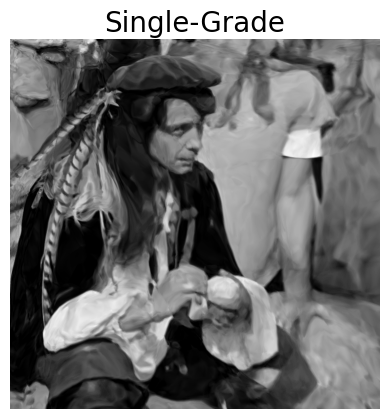}
\caption{PSNR: 24.15}
    \end{subfigure}
    \begin{subfigure}{0.20\linewidth}
\includegraphics[width=\linewidth]{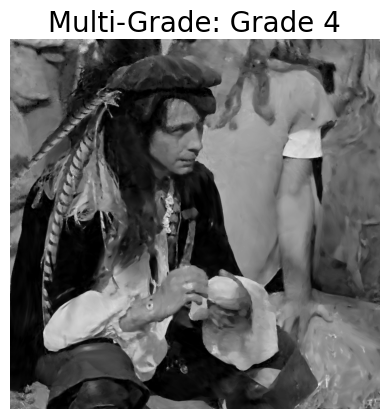}
\caption{PSNR: 26.25}
    \end{subfigure}

    \begin{subfigure}{0.27\linewidth}
\includegraphics[width=\linewidth]{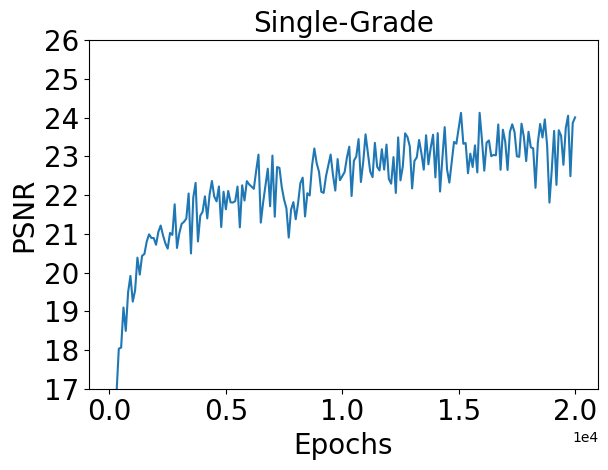}
\caption{}
    \end{subfigure}
    \begin{subfigure}{0.27\linewidth}
\includegraphics[width=\linewidth]{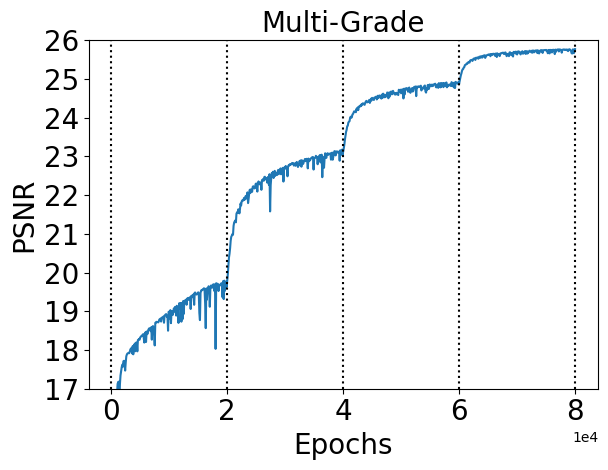}
\caption{}
    \end{subfigure}
    \begin{subfigure}{0.20\linewidth}
\includegraphics[width=\linewidth]{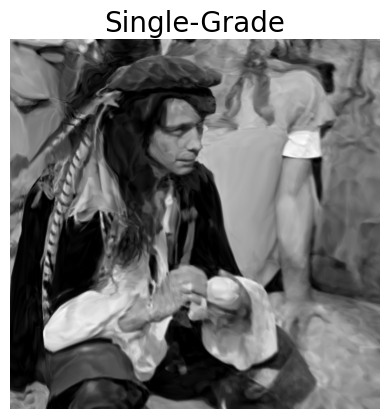}
\caption{PSNR: 24.65}
    \end{subfigure}
    \begin{subfigure}{0.20\linewidth}
\includegraphics[width=\linewidth]{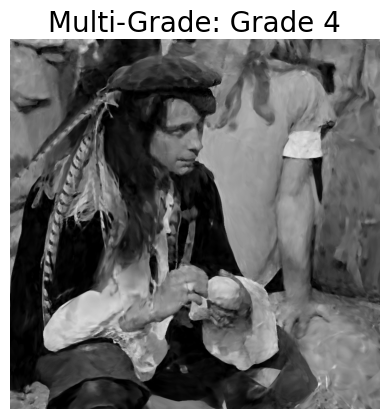}
\caption{PSNR: 25.77}
    \end{subfigure}

    \begin{subfigure}{0.27\linewidth}
\includegraphics[width=\linewidth]{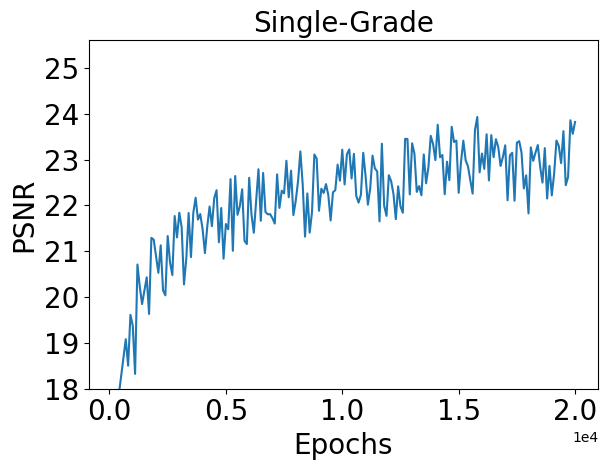}
\caption{}
    \end{subfigure}
    \begin{subfigure}{0.27\linewidth}
\includegraphics[width=\linewidth]{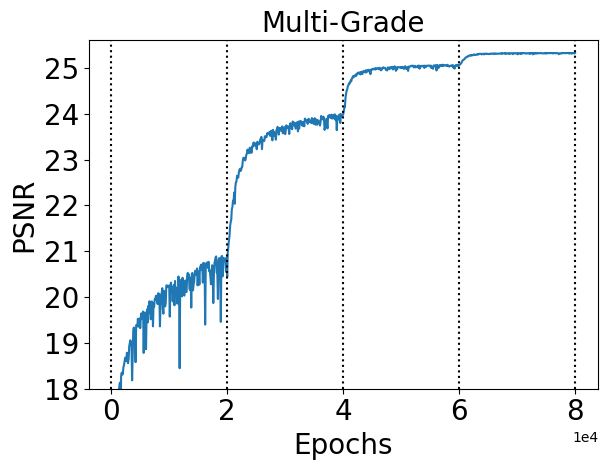}
\caption{}
    \end{subfigure}
    \begin{subfigure}{0.20\linewidth}
\includegraphics[width=\linewidth]{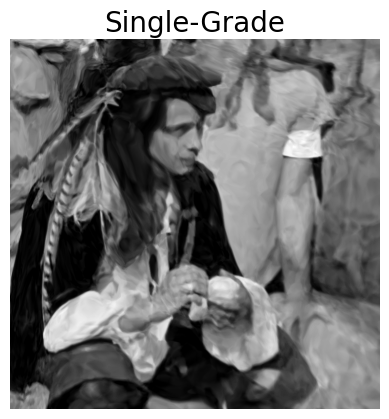}
\caption{PSNR: 23.82}
    \end{subfigure}
    \begin{subfigure}{0.20\linewidth}
\includegraphics[width=\linewidth]{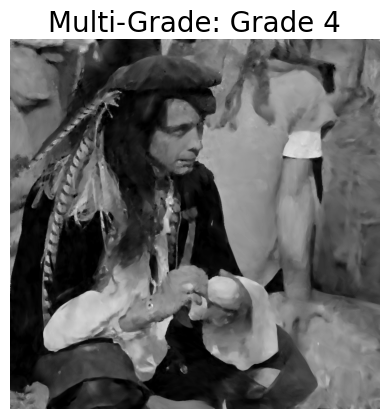}
\caption{PSNR: 25.32}
    \end{subfigure}

    \caption{Comparison of SGDL \eqref{SGDL-3} and MGDL \eqref{MGDL-3} denoising results for the `Butterfly' image. Rows one to six show noise levels $\sigma = 10, 20, 30, 40, 50, 60$, with PSNR during training and reconstructed images displayed with PSNR values in the subtitles.
    }
    \label{fig:SGDL MGDL male image denoising}
\end{figure}

\begin{figure}[H]
    \centering
    
    \begin{subfigure}{0.27\linewidth}
\includegraphics[width=\linewidth]{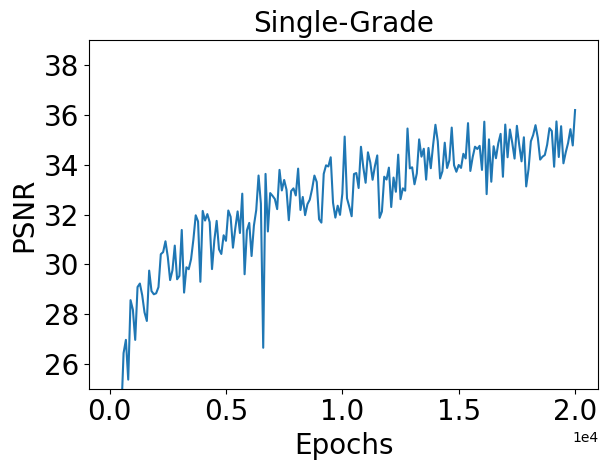}
\caption{}
    \end{subfigure}
    \begin{subfigure}{0.27\linewidth}
\includegraphics[width=\linewidth]{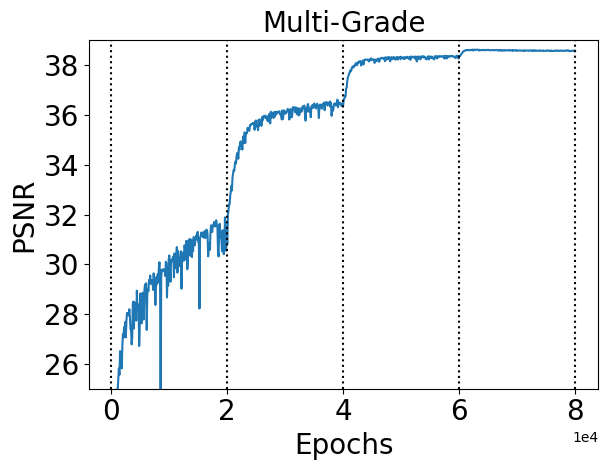}
\caption{}
    \end{subfigure}
    \begin{subfigure}{0.20\linewidth}
\includegraphics[width=\linewidth]{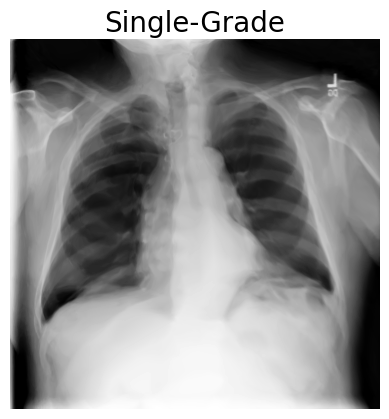}
\caption{PSNR: 36.20}
    \end{subfigure}
    \begin{subfigure}{0.20\linewidth}
\includegraphics[width=\linewidth]{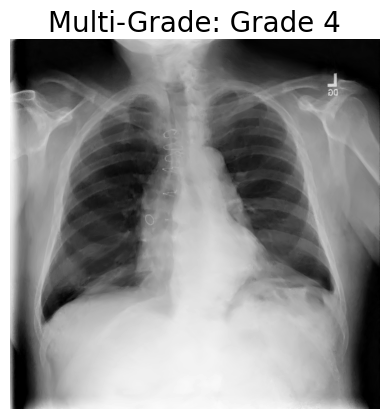}
\caption{PSNR: 38.58}
    \end{subfigure}

    \begin{subfigure}{0.27\linewidth}
\includegraphics[width=\linewidth]{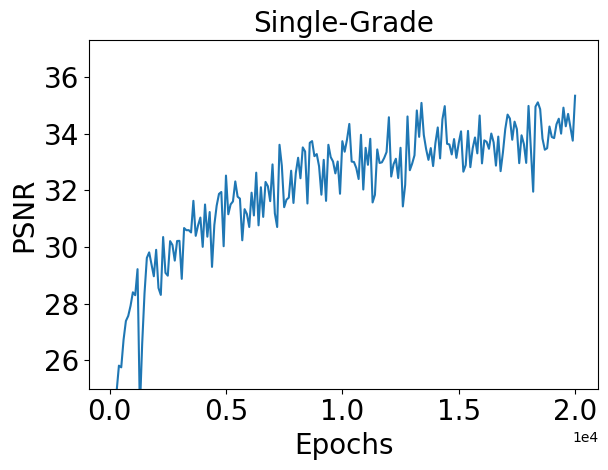}
\caption{}
    \end{subfigure}
    \begin{subfigure}{0.27\linewidth}
\includegraphics[width=\linewidth]{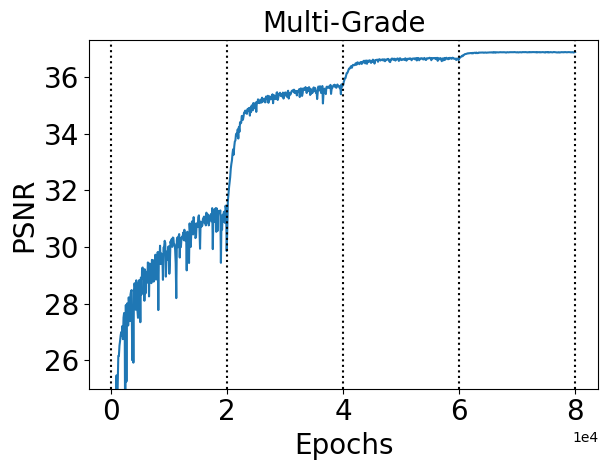}
\caption{}
    \end{subfigure}
    \begin{subfigure}{0.20\linewidth}
\includegraphics[width=\linewidth]{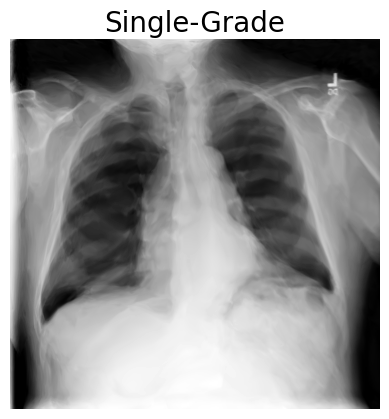}
\caption{PSNR: 35.34}
    \end{subfigure}
    \begin{subfigure}{0.20\linewidth}
\includegraphics[width=\linewidth]{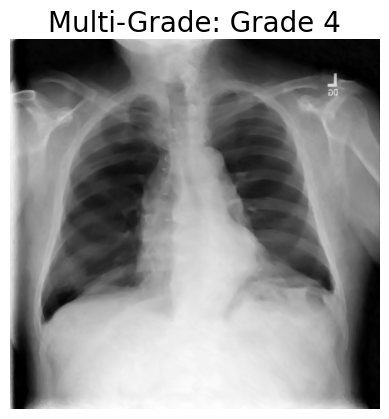}
\caption{PSNR: 36.89}
    \end{subfigure}

    \begin{subfigure}{0.27\linewidth}
\includegraphics[width=\linewidth]{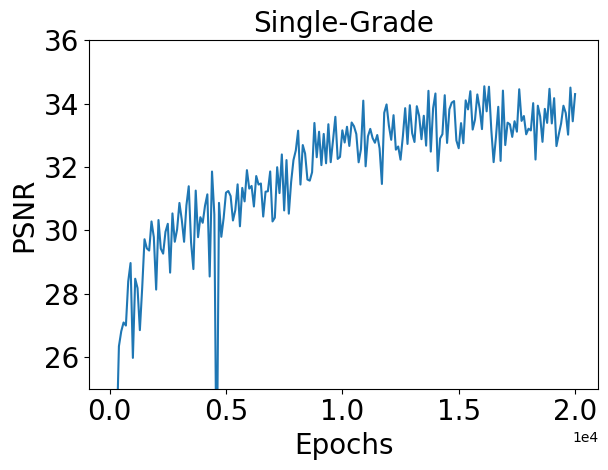}
\caption{}
    \end{subfigure}
    \begin{subfigure}{0.27\linewidth}
\includegraphics[width=\linewidth]{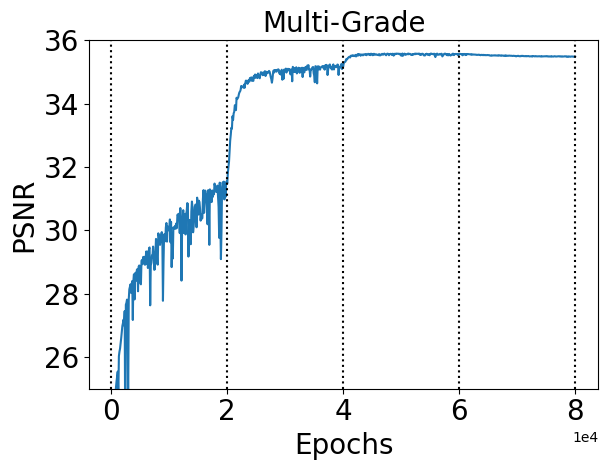}
\caption{}
    \end{subfigure}
    \begin{subfigure}{0.20\linewidth}
\includegraphics[width=\linewidth]{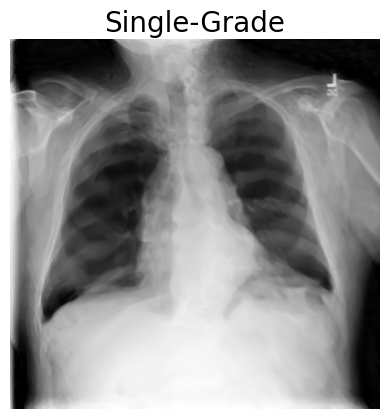}
\caption{PSNR: 34.30}
    \end{subfigure}
    \begin{subfigure}{0.20\linewidth}
\includegraphics[width=\linewidth]{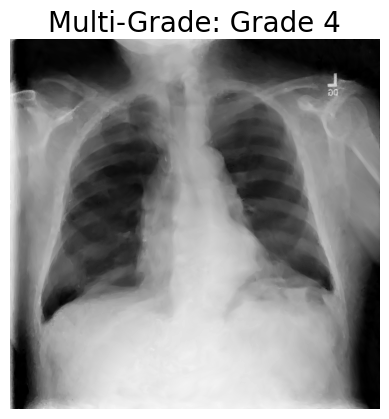}
\caption{PSNR: 35.48}
    \end{subfigure}

    \begin{subfigure}{0.27\linewidth}
\includegraphics[width=\linewidth]{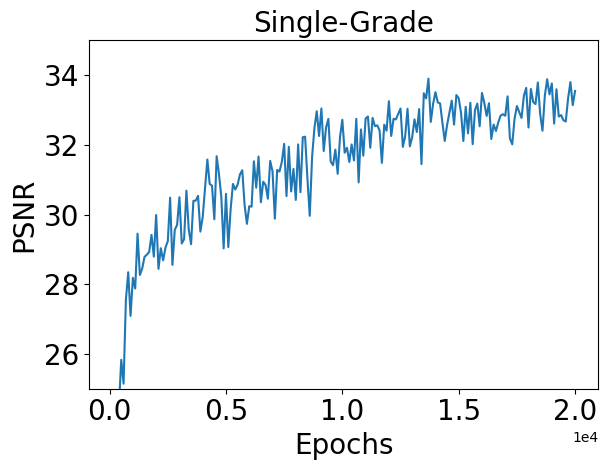}
\caption{}
    \end{subfigure}
    \begin{subfigure}{0.27\linewidth}
\includegraphics[width=\linewidth]{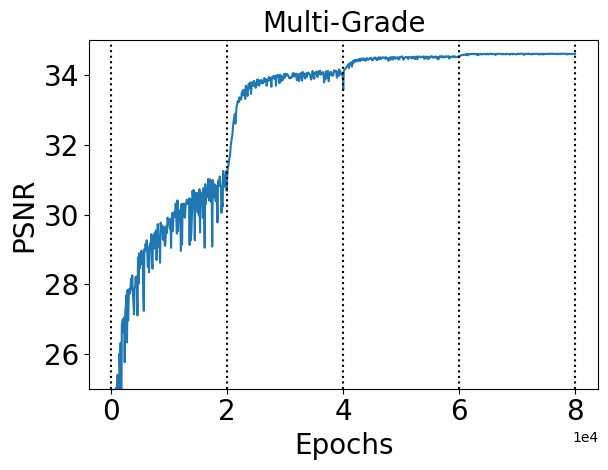}
\caption{}
    \end{subfigure}
    \begin{subfigure}{0.20\linewidth}
\includegraphics[width=\linewidth]{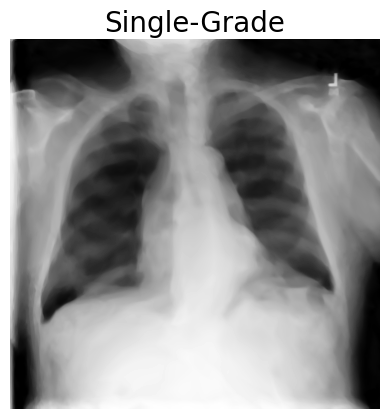}
\caption{PSNR: 33.55}
    \end{subfigure}
    \begin{subfigure}{0.20\linewidth}
\includegraphics[width=\linewidth]{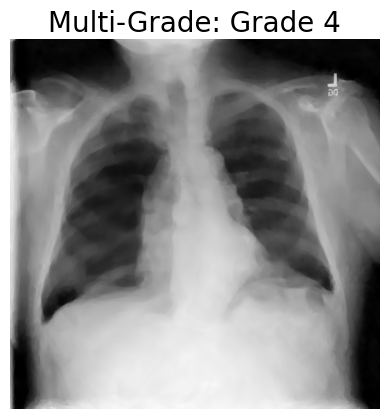}
\caption{PSNR: 34.61}
    \end{subfigure}

    \begin{subfigure}{0.27\linewidth}
\includegraphics[width=\linewidth]{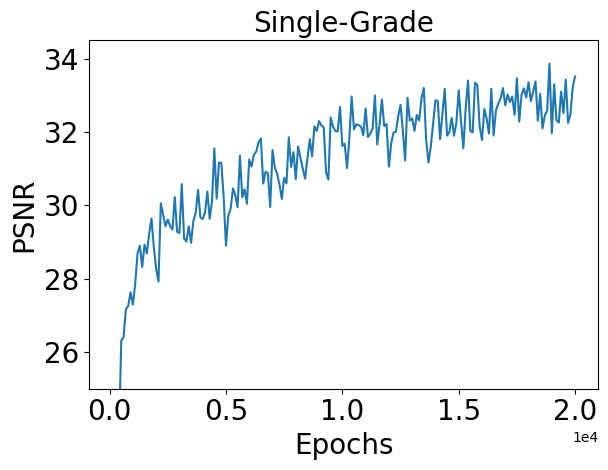}
\caption{}
    \end{subfigure}
    \begin{subfigure}{0.27\linewidth}
\includegraphics[width=\linewidth]{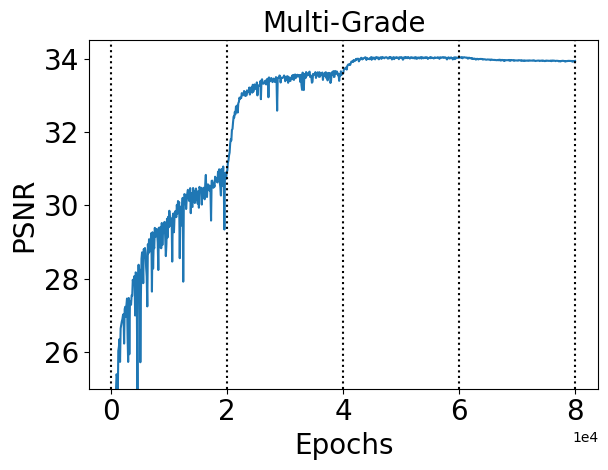}
\caption{}
    \end{subfigure}
    \begin{subfigure}{0.20\linewidth}
\includegraphics[width=\linewidth]{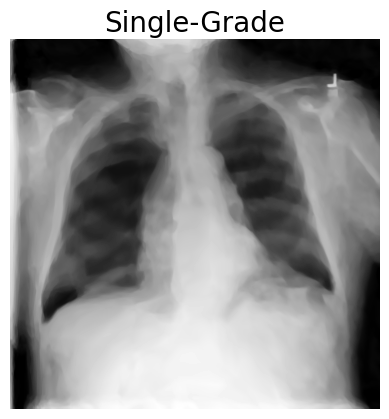}
\caption{PSNR: 33.51}
    \end{subfigure}
    \begin{subfigure}{0.20\linewidth}
\includegraphics[width=\linewidth]{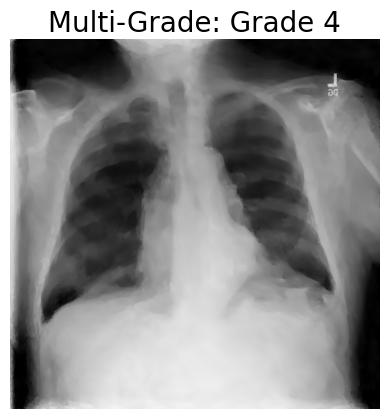}
\caption{PSNR: 33.94}
    \end{subfigure}

    \begin{subfigure}{0.27\linewidth}
\includegraphics[width=\linewidth]{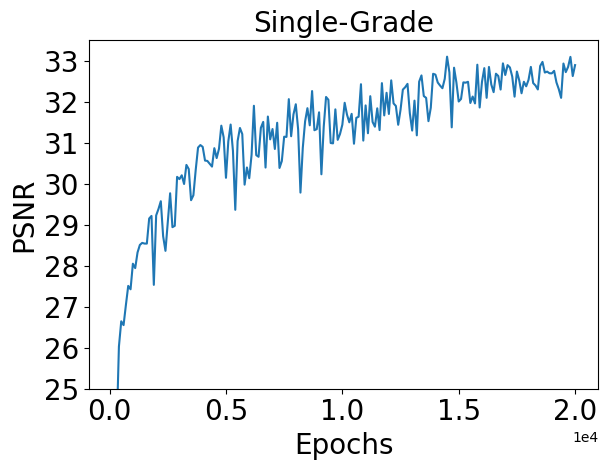}
\caption{}
    \end{subfigure}
    \begin{subfigure}{0.27\linewidth}
\includegraphics[width=\linewidth]{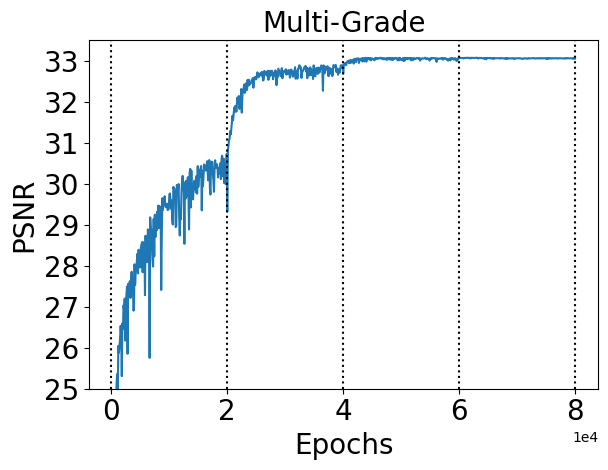}
\caption{}
    \end{subfigure}
    \begin{subfigure}{0.20\linewidth}
\includegraphics[width=\linewidth]{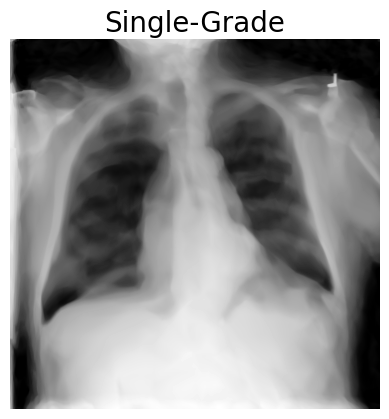}
\caption{PSNR: 32.90}
    \end{subfigure}
    \begin{subfigure}{0.20\linewidth}
\includegraphics[width=\linewidth]{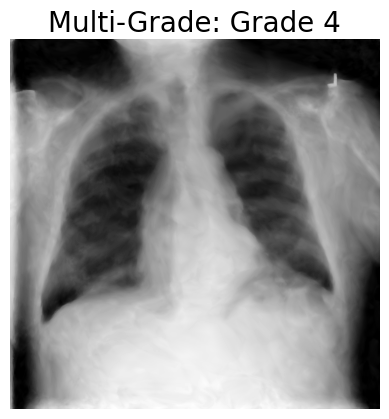}
\caption{PSNR: 33.06}
    \end{subfigure}

    \caption{Comparison of SGDL \eqref{SGDL-3} and MGDL \eqref{MGDL-3} denoising results for the `Chest' image. Rows one to six show noise levels $\sigma = 10, 20, 30, 40, 50, 60$, with PSNR during training and reconstructed images displayed with PSNR values in the subtitles.
    }
    \label{fig:SGDL MGDL medical image denoising}
\end{figure}

\begin{figure}[H]
  \centering

   \begin{subfigure}{0.25\linewidth}
\includegraphics[width=\linewidth]{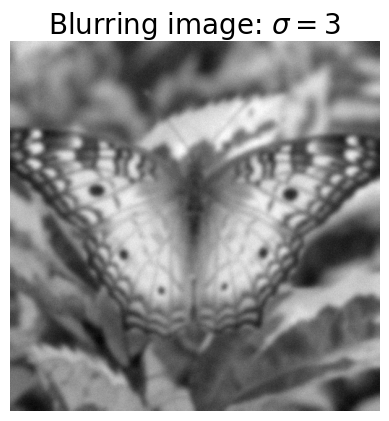}
\caption{PSNR: 24.61}
    \end{subfigure}
   \begin{subfigure}{0.25\linewidth}
\includegraphics[width=\linewidth]{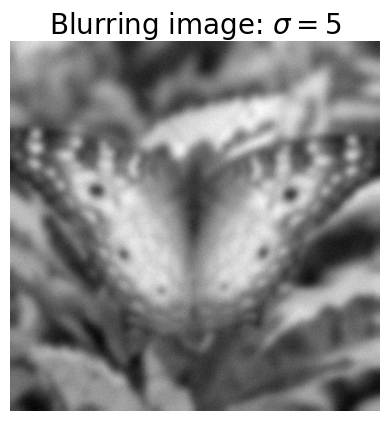}
\caption{PSNR: 22.61}
    \end{subfigure}
\begin{subfigure}{0.25\linewidth}
\includegraphics[width=\linewidth]{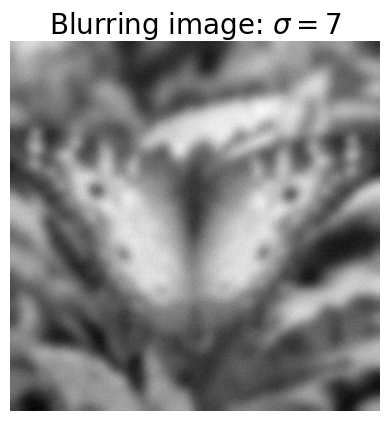}
\caption{PSNR: 21.54}
    \end{subfigure}

   \begin{subfigure}{0.25\linewidth}
\includegraphics[width=\linewidth]{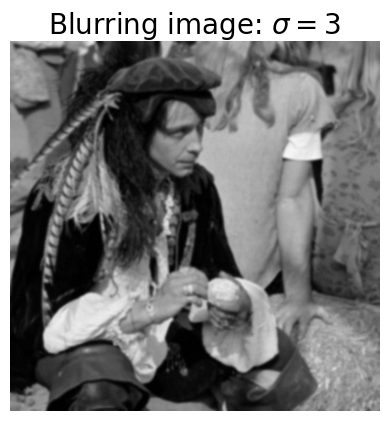}
\caption{PSNR: 24.94}
    \end{subfigure}
   \begin{subfigure}{0.25\linewidth}
\includegraphics[width=\linewidth]{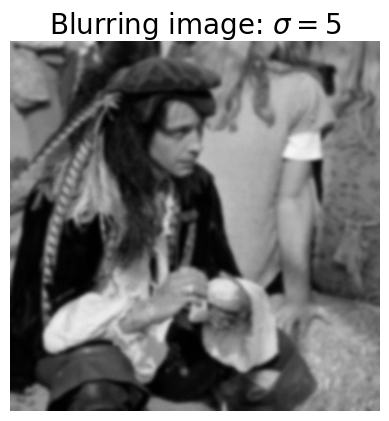}
\caption{PSNR: 22.99}
    \end{subfigure}
\begin{subfigure}{0.25\linewidth}
\includegraphics[width=\linewidth]{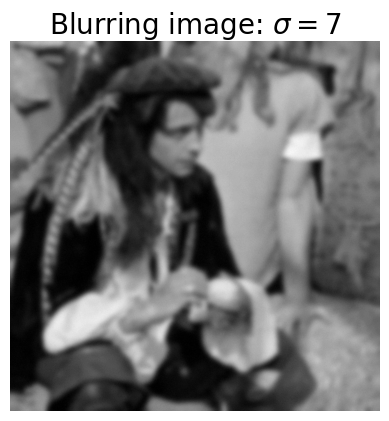}
\caption{PSNR: 21.85}
    \end{subfigure}

   \begin{subfigure}{0.25\linewidth}
\includegraphics[width=\linewidth]{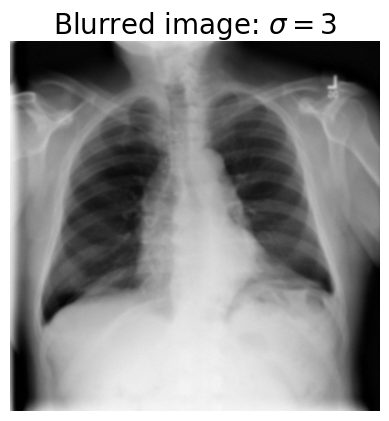}
\caption{PSNR: 34.72}
    \end{subfigure}
   \begin{subfigure}{0.25\linewidth}
\includegraphics[width=\linewidth]{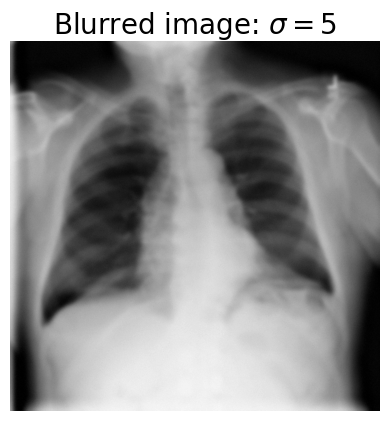}
\caption{PSNR: 33.00}
    \end{subfigure}
\begin{subfigure}{0.25\linewidth}
\includegraphics[width=\linewidth]{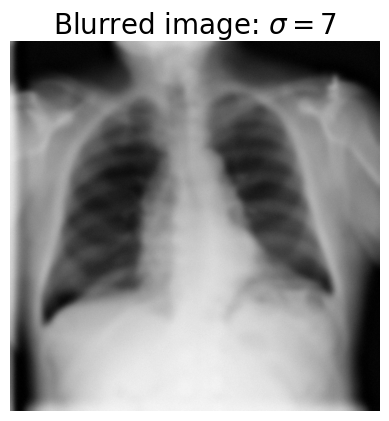}
\caption{PSNR: 31.61}
    \end{subfigure}
    
\caption{Blurred images: `Butterfly' (a)-(c); `Pirate' (d)-(f); `Chest' (g)-(i). The PSNR value is shown in each title.
}
	\label{fig:blurred images}
\end{figure}

\begin{figure}[H]
    \centering
    
    \begin{subfigure}{0.25\linewidth}
\includegraphics[width=\linewidth]{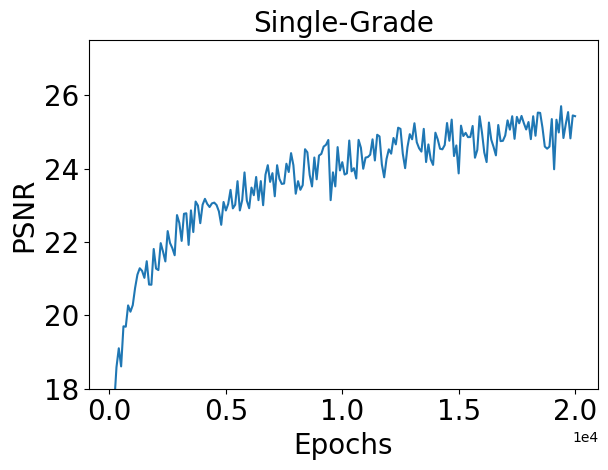}
\caption{}
    \end{subfigure}
    \begin{subfigure}{0.25\linewidth}
\includegraphics[width=\linewidth]{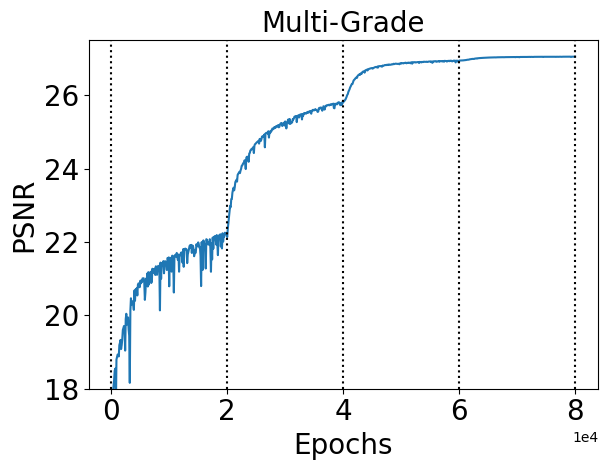}
\caption{}
    \end{subfigure}
    \begin{subfigure}{0.18\linewidth}
\includegraphics[width=\linewidth]{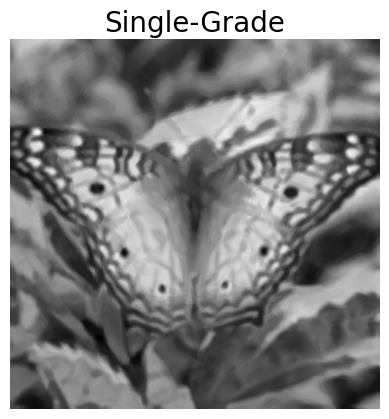}
\caption{PSNR: 25.43}
    \end{subfigure}
    \begin{subfigure}{0.18\linewidth}
\includegraphics[width=\linewidth]{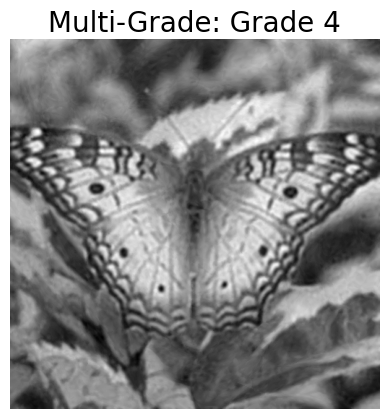}
\caption{PSNR: 27.06}
    \end{subfigure}

    \begin{subfigure}{0.25\linewidth}
\includegraphics[width=\linewidth]{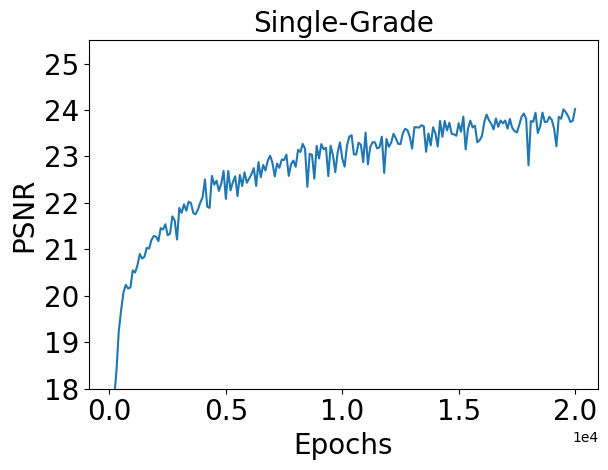}
\caption{}
    \end{subfigure}
    \begin{subfigure}{0.25\linewidth}
\includegraphics[width=\linewidth]{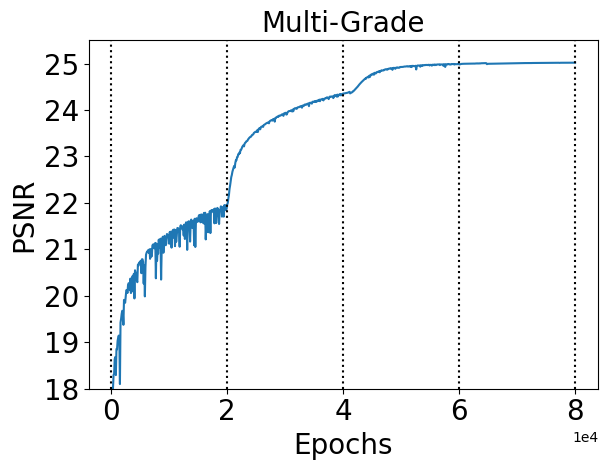}
\caption{}
    \end{subfigure}
    \begin{subfigure}{0.18\linewidth}
\includegraphics[width=\linewidth]{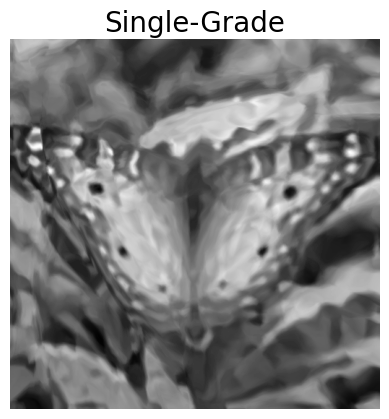}
\caption{PSNR: 24.20}
    \end{subfigure}
    \begin{subfigure}{0.18\linewidth}
\includegraphics[width=\linewidth]{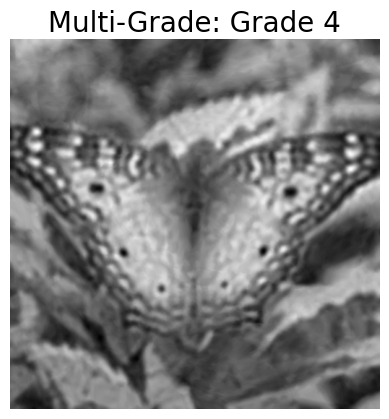}
\caption{PSNR: 25.19}
    \end{subfigure}

    \begin{subfigure}{0.25\linewidth}
\includegraphics[width=\linewidth]{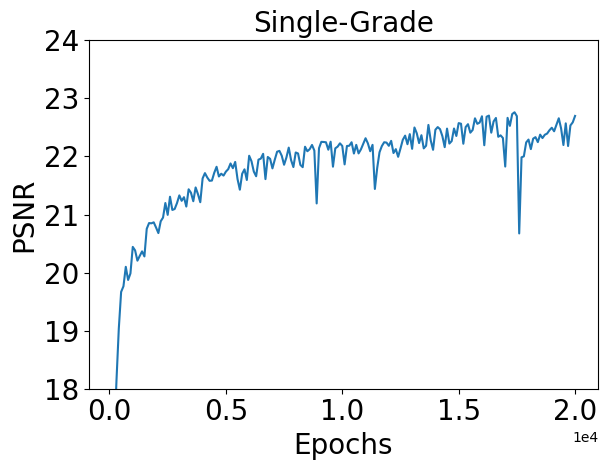}
\caption{}
    \end{subfigure}
    \begin{subfigure}{0.25\linewidth}
\includegraphics[width=\linewidth]{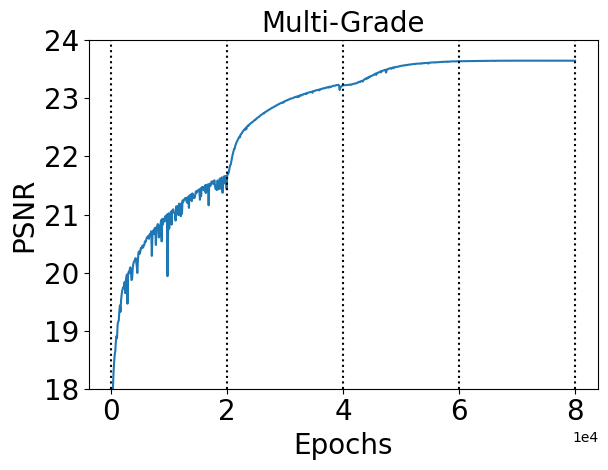}
\caption{}
    \end{subfigure}
    \begin{subfigure}{0.18\linewidth}
\includegraphics[width=\linewidth]{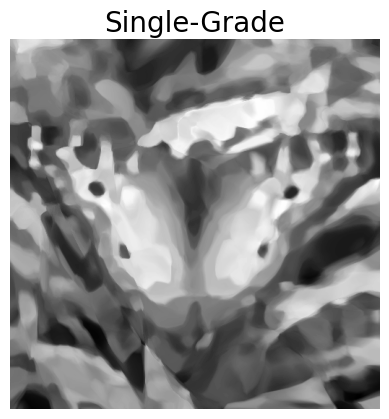}
\caption{PSNR: 22.70}
    \end{subfigure}
    \begin{subfigure}{0.18\linewidth}
\includegraphics[width=\linewidth]{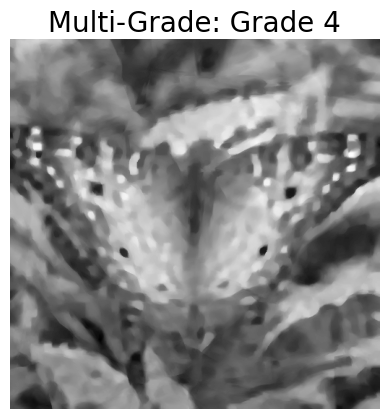}
\caption{PSNR: 23.65}
    \end{subfigure}

    \caption{Comparison of SGDL \eqref{SGDL-3} and MGDL \eqref{MGDL-3} deblurring results for the `Butterfly' image. Rows one to three correspond to blurring kernel standard deviations $\hat{\sigma} = 3, 5, 7$. 
    }
    \label{fig:SGDL MGDL butterfly image deblurring}
\end{figure}

\begin{figure}[H]
    \centering
    
    \begin{subfigure}{0.25\linewidth}
\includegraphics[width=\linewidth]{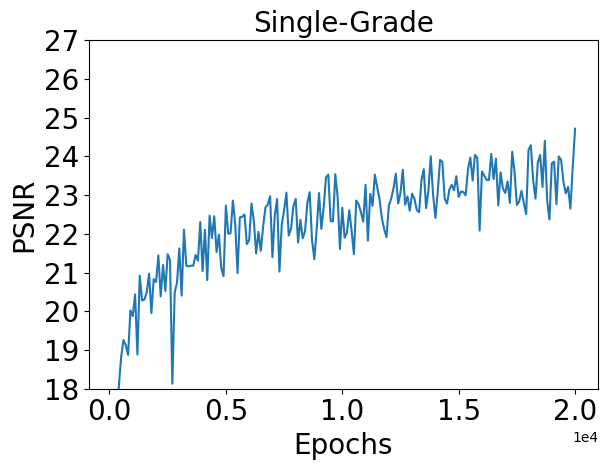}
\caption{}
    \end{subfigure}
    \begin{subfigure}{0.25\linewidth}
\includegraphics[width=\linewidth]{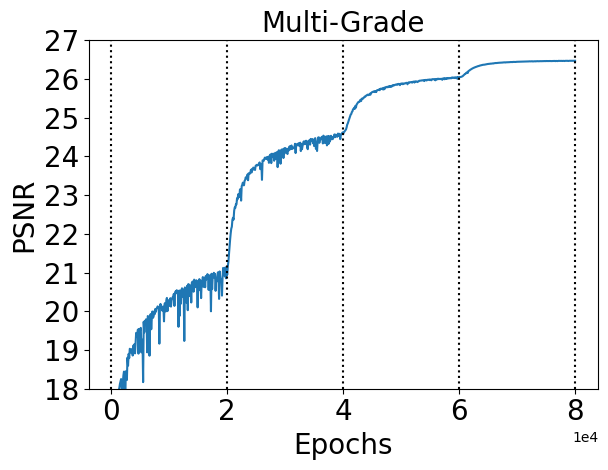}
\caption{}
    \end{subfigure}
    \begin{subfigure}{0.18\linewidth}
\includegraphics[width=\linewidth]{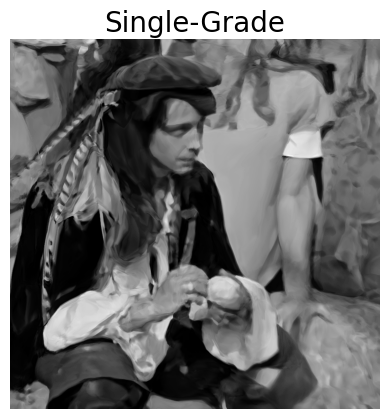}
\caption{PSNR: 24.72}
    \end{subfigure}
    \begin{subfigure}{0.18\linewidth}
\includegraphics[width=\linewidth]{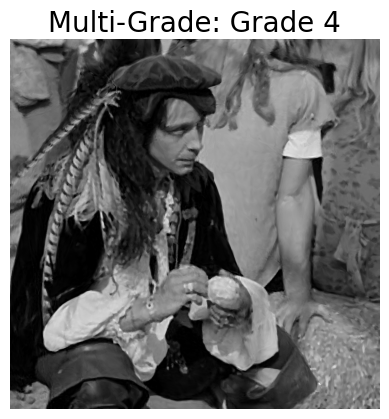}
\caption{PSNR: 26.47}
    \end{subfigure}

    \begin{subfigure}{0.25\linewidth}
\includegraphics[width=\linewidth]{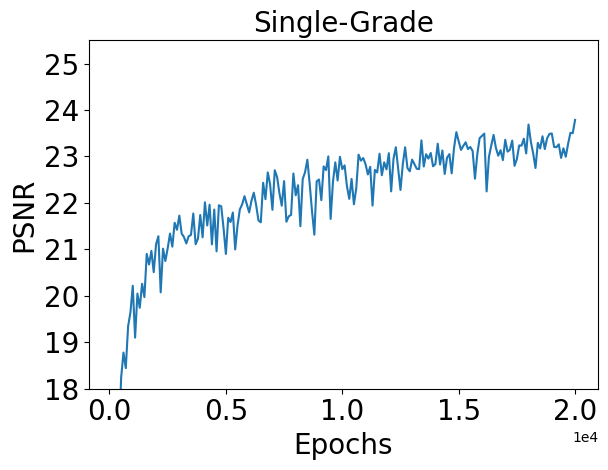}
\caption{}
    \end{subfigure}
    \begin{subfigure}{0.25\linewidth}
\includegraphics[width=\linewidth]{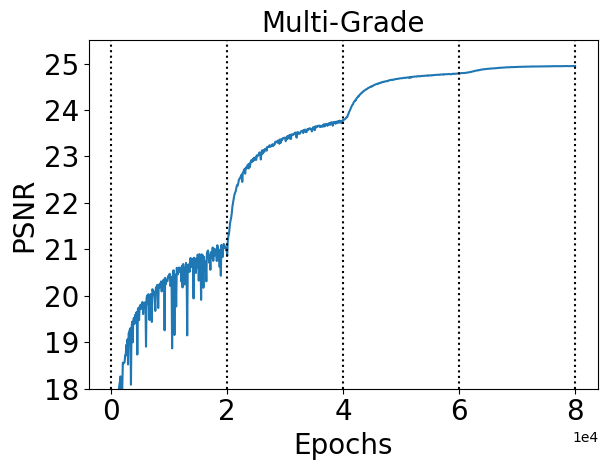}
\caption{}
    \end{subfigure}
    \begin{subfigure}{0.18\linewidth}
\includegraphics[width=\linewidth]{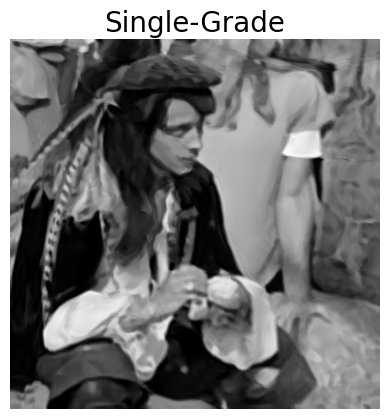}
\caption{PSNR: 23.79}
    \end{subfigure}
    \begin{subfigure}{0.18\linewidth}
\includegraphics[width=\linewidth]{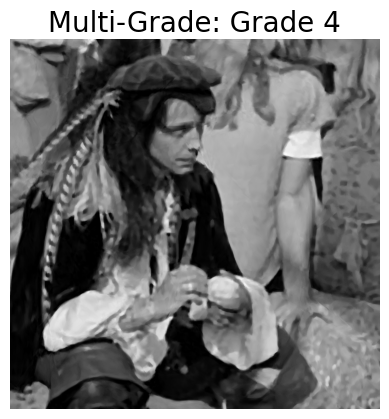}
\caption{PSNR: 24.95}
    \end{subfigure}

    \begin{subfigure}{0.25\linewidth}
\includegraphics[width=\linewidth]{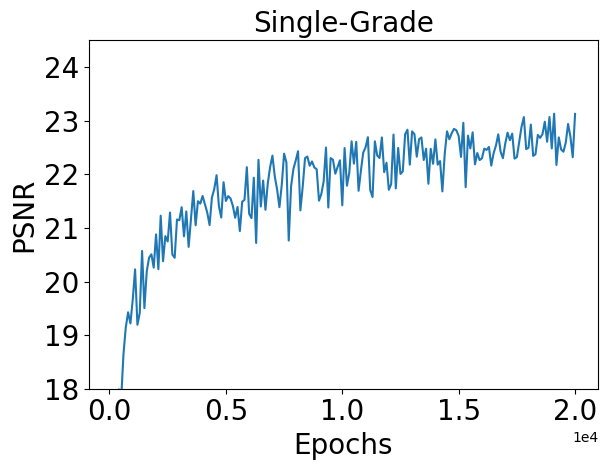}
\caption{}
    \end{subfigure}
    \begin{subfigure}{0.25\linewidth}
\includegraphics[width=\linewidth]{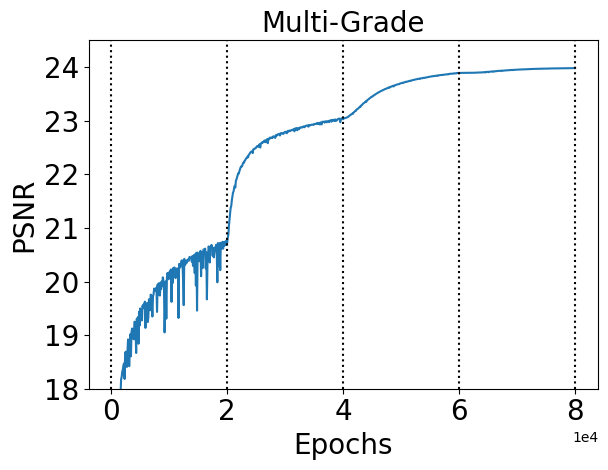}
\caption{}
    \end{subfigure}
    \begin{subfigure}{0.18\linewidth}
\includegraphics[width=\linewidth]{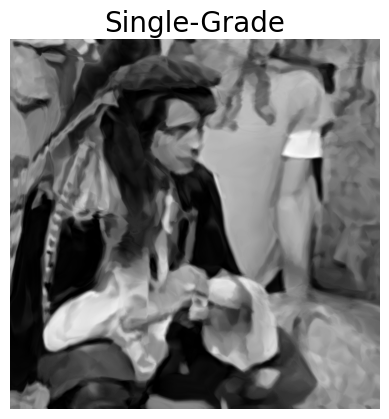}
\caption{PSNR: 23.13}
    \end{subfigure}
    \begin{subfigure}{0.18\linewidth}
\includegraphics[width=\linewidth]{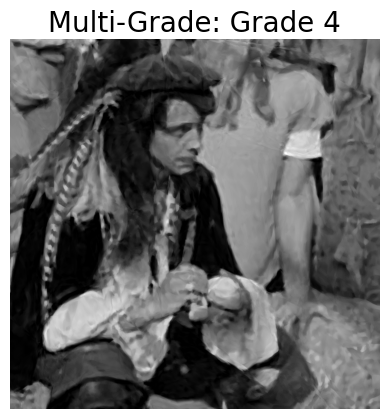}
\caption{PSNR: 23.98}
    \end{subfigure}

    \caption{Comparison of SGDL \eqref{SGDL-3} and MGDL \eqref{MGDL-3} deblurring results for the `Pirate' image. Rows one to three correspond to blurring kernel standard deviations $\hat{\sigma} = 3, 5, 7$. 
    }
    \label{fig:SGDL MGDL Male image deblurring}
\end{figure}

\begin{figure}[H]
    \centering
    
    \begin{subfigure}{0.25\linewidth}
\includegraphics[width=\linewidth]{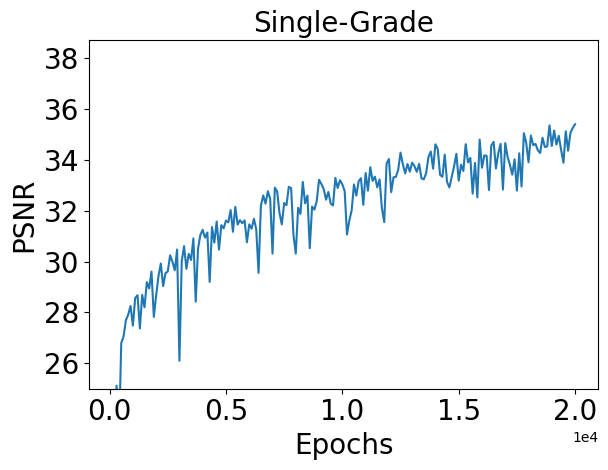}
\caption{}
    \end{subfigure}
    \begin{subfigure}{0.25\linewidth}
\includegraphics[width=\linewidth]{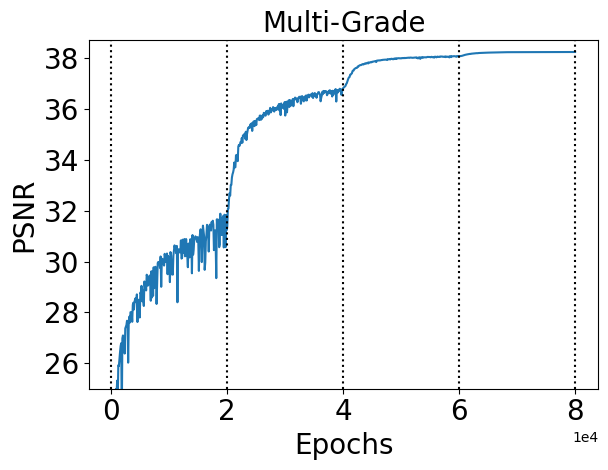}
\caption{}
    \end{subfigure}
    \begin{subfigure}{0.18\linewidth}
\includegraphics[width=\linewidth]{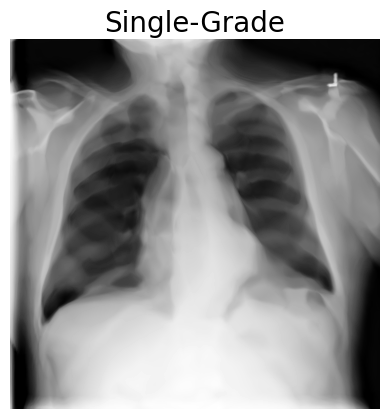}
\caption{PSNR: 35.40}
    \end{subfigure}
    \begin{subfigure}{0.18\linewidth}
\includegraphics[width=\linewidth]{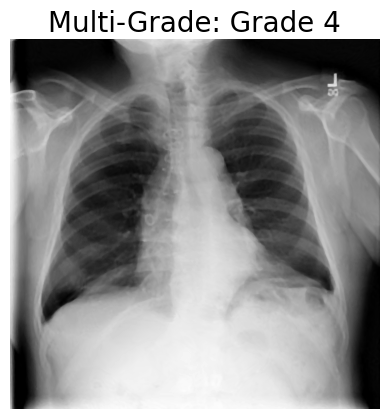}
\caption{PSNR: 38.24}
    \end{subfigure}

    \begin{subfigure}{0.25\linewidth}
\includegraphics[width=\linewidth]{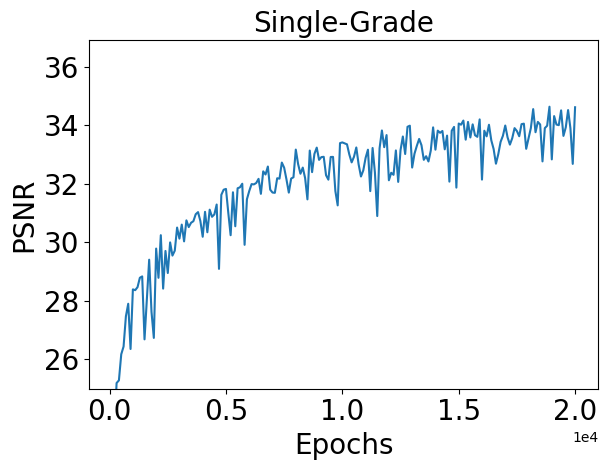}
\caption{}
    \end{subfigure}
    \begin{subfigure}{0.25\linewidth}
\includegraphics[width=\linewidth]{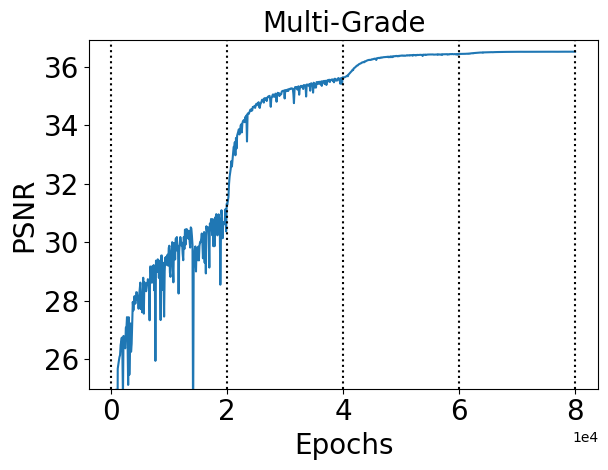}
\caption{}
    \end{subfigure}
    \begin{subfigure}{0.18\linewidth}
\includegraphics[width=\linewidth]{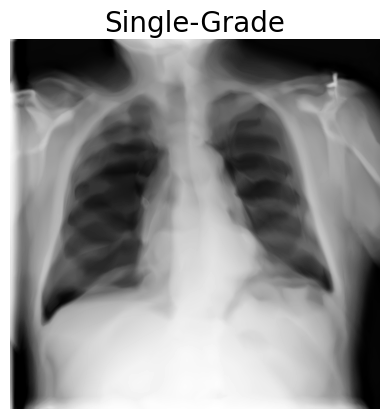}
\caption{PSNR: 34.61}
    \end{subfigure}
    \begin{subfigure}{0.18\linewidth}
\includegraphics[width=\linewidth]{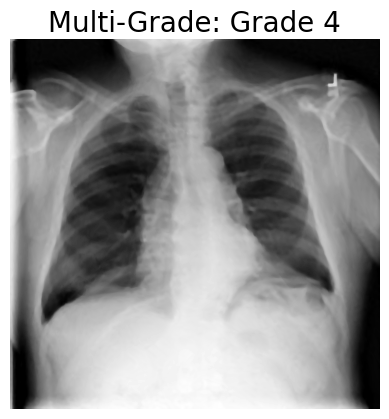}
\caption{PSNR: 36.51}
    \end{subfigure}

    \begin{subfigure}{0.25\linewidth}
\includegraphics[width=\linewidth]{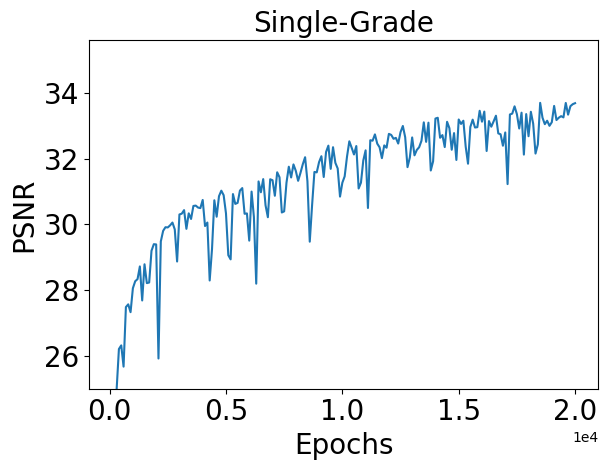}
\caption{}
    \end{subfigure}
    \begin{subfigure}{0.25\linewidth}
\includegraphics[width=\linewidth]{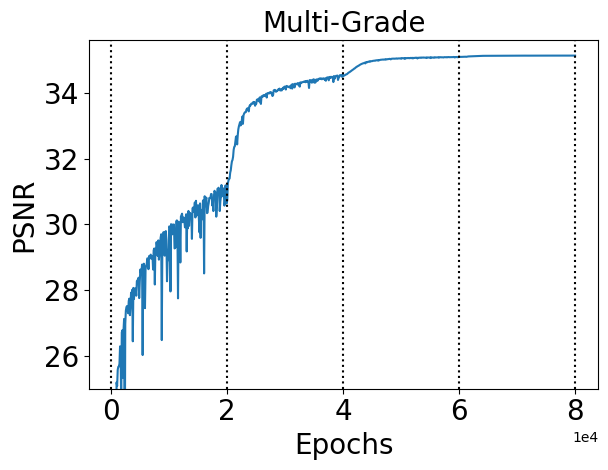}
\caption{}
    \end{subfigure}
    \begin{subfigure}{0.18\linewidth}
\includegraphics[width=\linewidth]{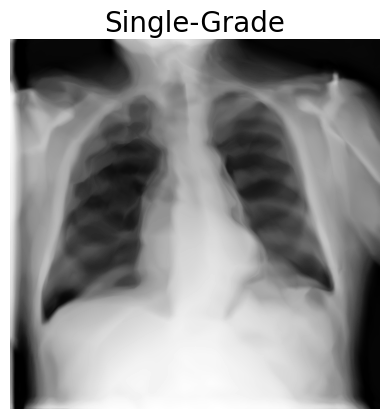}
\caption{PSNR: 33.69}
    \end{subfigure}
    \begin{subfigure}{0.18\linewidth}
\includegraphics[width=\linewidth]{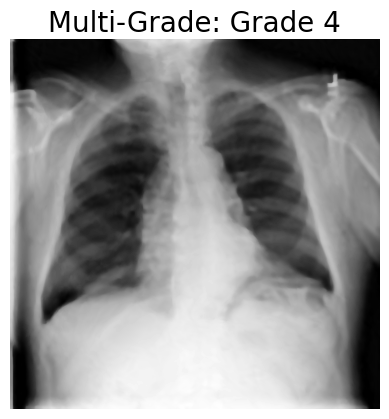}
\caption{PSNR: 35.14}
    \end{subfigure}

    \caption{Comparison of SGDL \eqref{SGDL-3} and MGDL \eqref{MGDL-3} deblurring results for the `Chest' image. Rows one to three correspond to blurring kernel standard deviations $\hat{\sigma} = 3, 5, 7$. 
    }
    \label{fig:SGDL MGDL Chest image deblurring}
\end{figure}

Supporting figures referenced in Section \ref{section: comparsion image reconstruction} include:

\begin{enumerate}
    \item Figure \ref{fig:testing images}: Clean images used in the paper.

    \item Figures \ref{fig:SGDL MGDL cameraman}--\ref{fig:SGDL MGDL image regression}: Results for the `Image Regression' in Section \ref{section: comparsion image reconstruction}.
    
    \item Figure \ref{fig: noisy image}: Noisy image used in the experiments. Figures \ref{fig:SGDL MGDL butterfly image denoising}--\ref{fig:SGDL MGDL medical image denoising}: PSNR values during training and the denoised images produced by SGDL and MGDL. These figures correspond to the `Image Denoising' in Section \ref{section: comparsion image reconstruction}.
    
    \item Figure \ref{fig:blurred images}: Blurred image used in the experiments. Figures \ref{fig:SGDL MGDL butterfly image deblurring}--\ref{fig:SGDL MGDL Chest image deblurring}: PSNR values during training and the deblurred images produced by SGDL and MGDL. These figures correspond to the `Image Deblurring' in Section \ref{section: comparsion image reconstruction}.

\end{enumerate}

\subsection{Section \ref{section: Impact of learning rate}}

\begin{figure}[H]
    \centering
    \begin{subfigure}{0.24\linewidth}
\includegraphics[width=\linewidth]{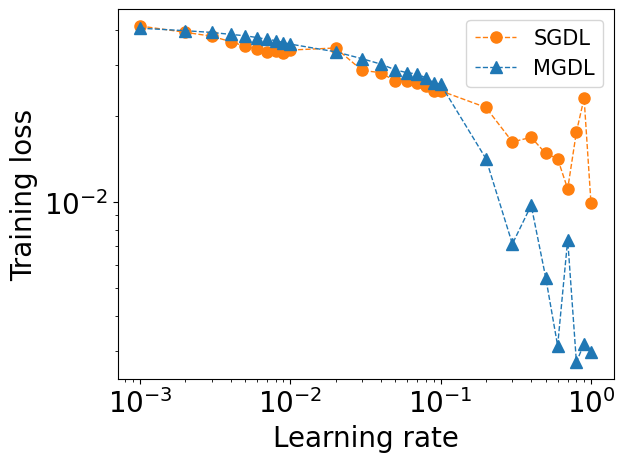}
    \end{subfigure}
    \begin{subfigure}{0.24\linewidth}
\includegraphics[width=\linewidth]{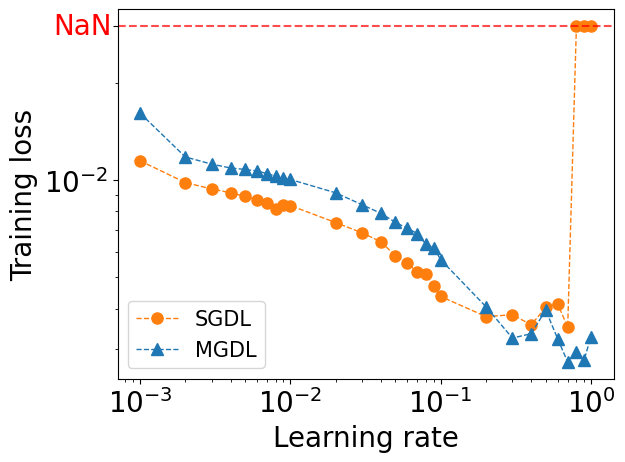}
    \end{subfigure}
   \begin{subfigure}{0.24\linewidth}
\includegraphics[width=\linewidth]{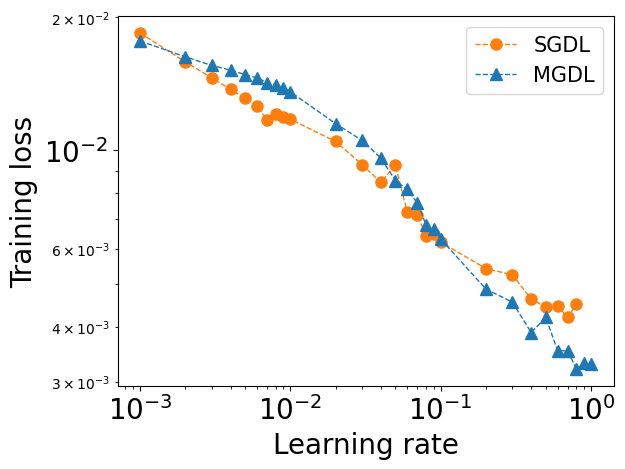}
    \end{subfigure}
   \begin{subfigure}{0.24\linewidth}
\includegraphics[width=\linewidth]{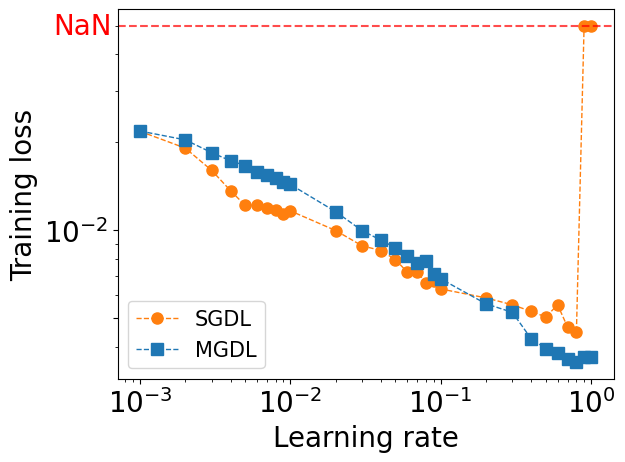}
    \end{subfigure}
    \caption{Impact of learning rate.
    }
    \label{fig:learning rate effectiveness image}
\end{figure}

Figure \ref{fig:learning rate effectiveness image} is the supporting figure referenced in Section \ref{section: Impact of learning rate}, which presents the results for `Image Regression'.

\subsection{Section \ref{section: Eigenvalue Analysis}}\label{appendix: Hessian eig sin function}

\begin{figure}[H]
  \centering

   \begin{subfigure}{0.265\linewidth}
\includegraphics[width=\linewidth]{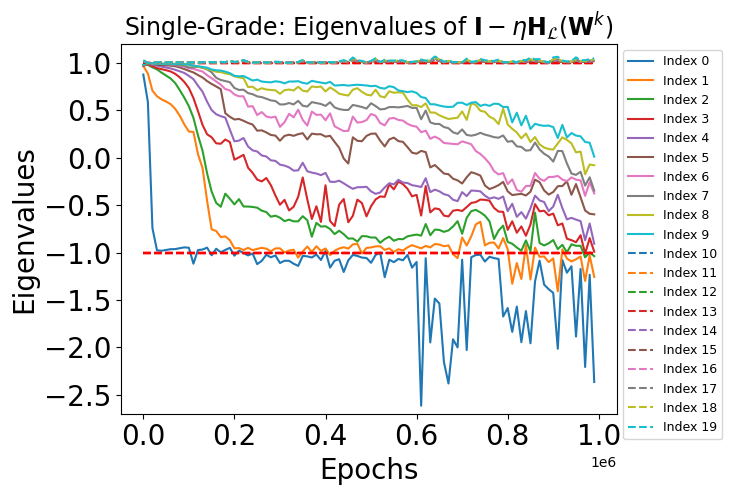}
    \end{subfigure}
   \begin{subfigure}{0.265\linewidth}
\includegraphics[width=\linewidth]{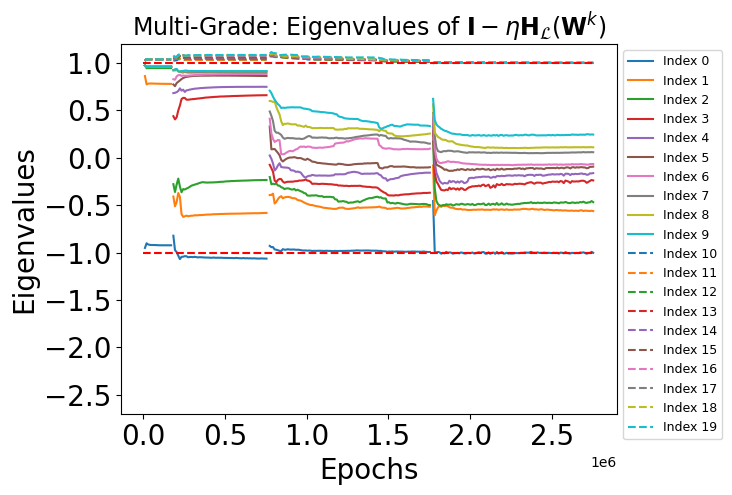}
    \end{subfigure}
   \begin{subfigure}{0.225\linewidth}
\includegraphics[width=\linewidth]{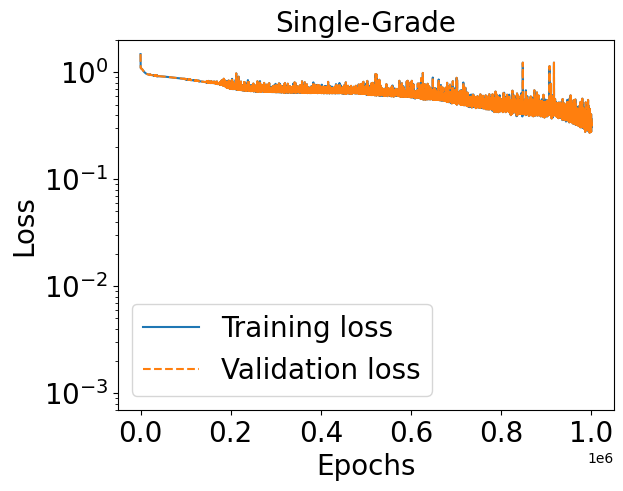}
    \end{subfigure}
   \begin{subfigure}{0.225\linewidth}
\includegraphics[width=\linewidth]{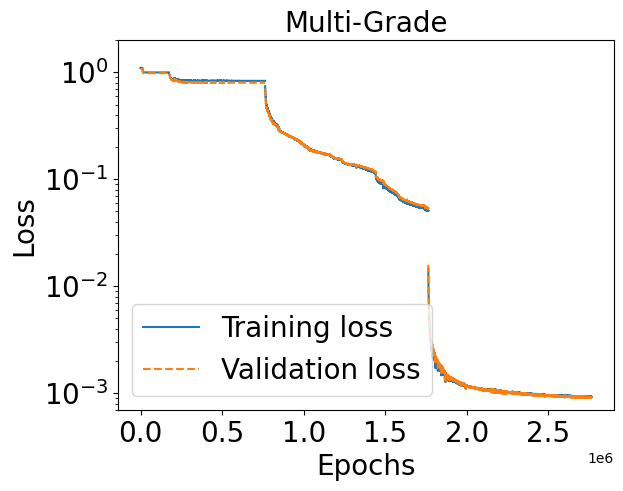}
    \end{subfigure}
    

\caption{Training process of SGDL ($\eta=0.005$) and MGDL ($0.2$) for Setting 2.
}
	\label{fig: Numerical analysis sin SGDL MGDL setting 2}
\end{figure}

\begin{figure}[H]
    \centering
   \begin{subfigure}{0.245\linewidth}
\includegraphics[width=\linewidth]{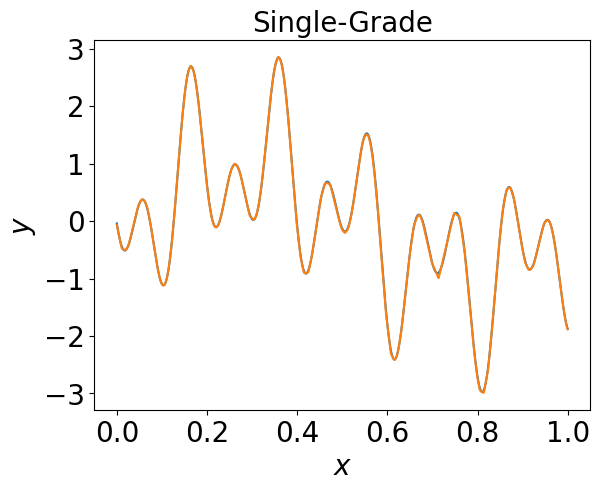}
    \end{subfigure}
   \begin{subfigure}{0.245\linewidth}
\includegraphics[width=\linewidth]{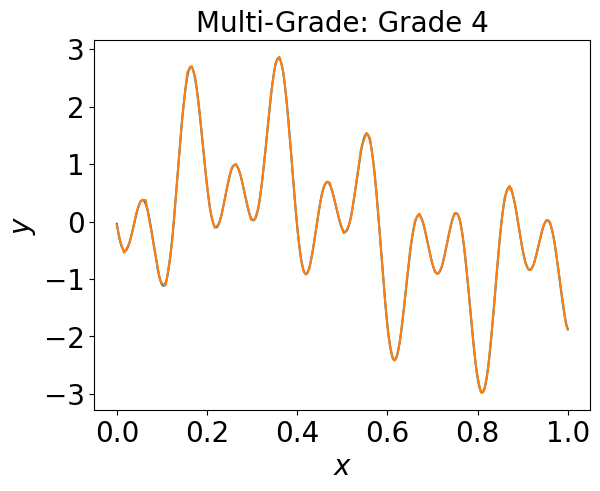}
    \end{subfigure}
   \begin{subfigure}{0.245\linewidth}
\includegraphics[width=\linewidth]{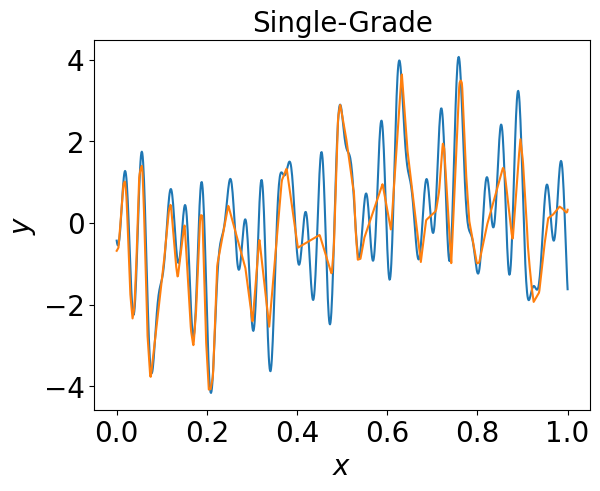}
    \end{subfigure}
   \begin{subfigure}{0.245\linewidth}
\includegraphics[width=\linewidth]{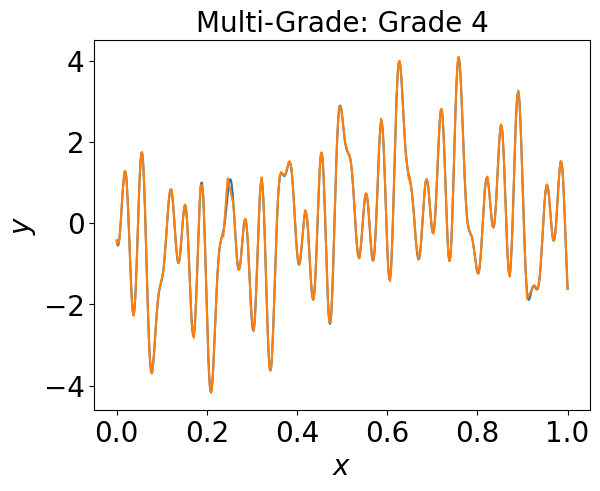}
    \end{subfigure}
    \caption{SGDL and MGDL predictions on synthetic data regression: Setting 1 (subfigures 1–2), Setting 2 (subfigures 3–4).}
    \label{fig: appendix synthetic data regression}
\end{figure}

\begin{figure}[H]
  \centering

\begin{subfigure}{0.265\linewidth}
\includegraphics[width=\linewidth]{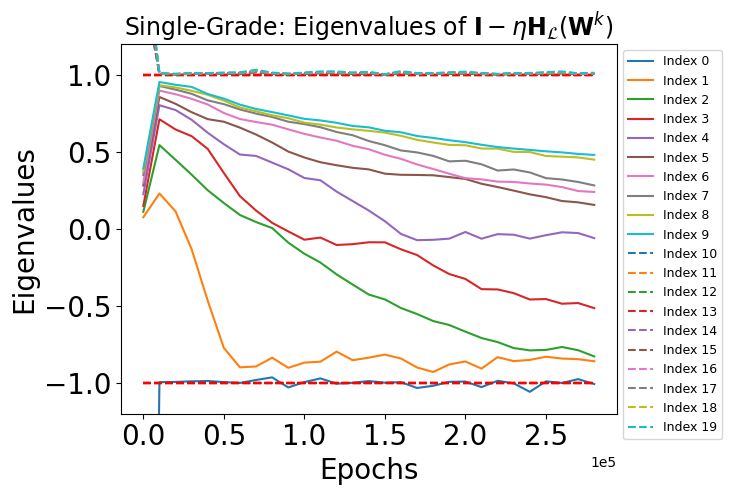}
    \end{subfigure}
  \begin{subfigure}{0.265\linewidth}
\includegraphics[width=\linewidth]{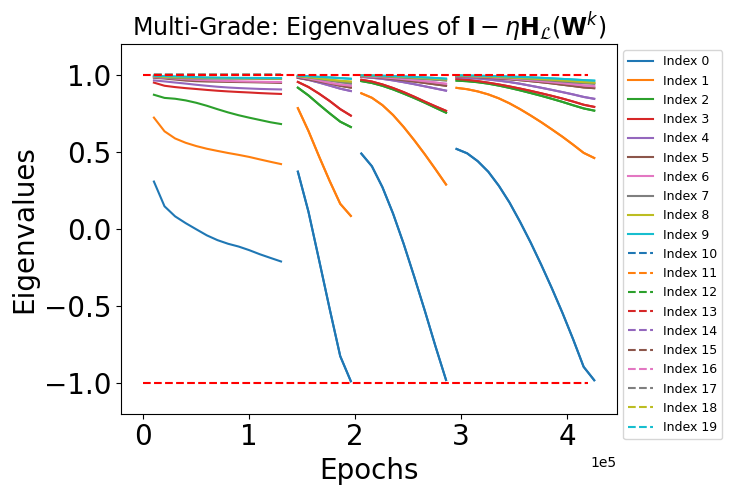}
\end{subfigure}
\begin{subfigure}{0.225\linewidth}
\includegraphics[width=\linewidth]{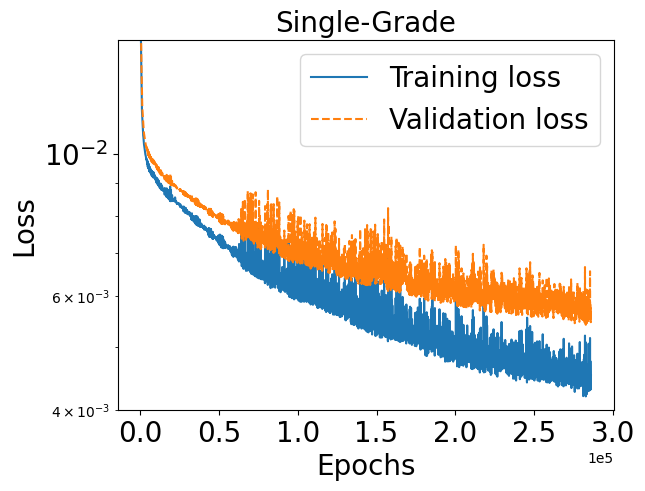}
    \end{subfigure}
  \begin{subfigure}{0.225\linewidth}
\includegraphics[width=\linewidth]{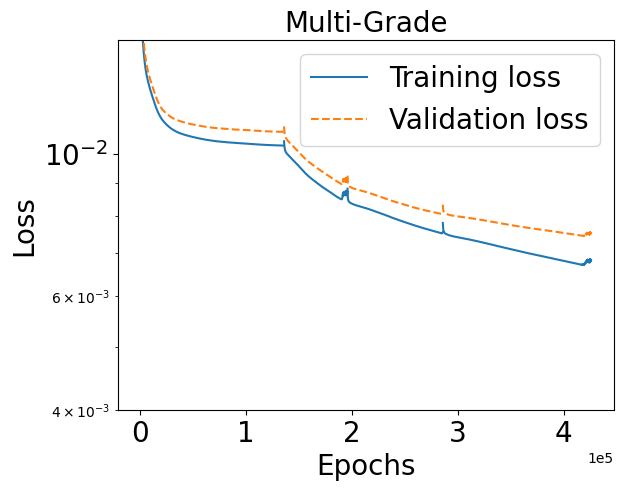}
\end{subfigure}
\caption{Training process of SGDL ($\eta = 0.1$) and MGDL ($\eta=0.2$) for image `Cameraman'. 
}
	\label{fig: image regression cameraman EigStop}
\end{figure}

\begin{figure}[H]
  \centering

\begin{subfigure}{0.265\linewidth}
\includegraphics[width=\linewidth]{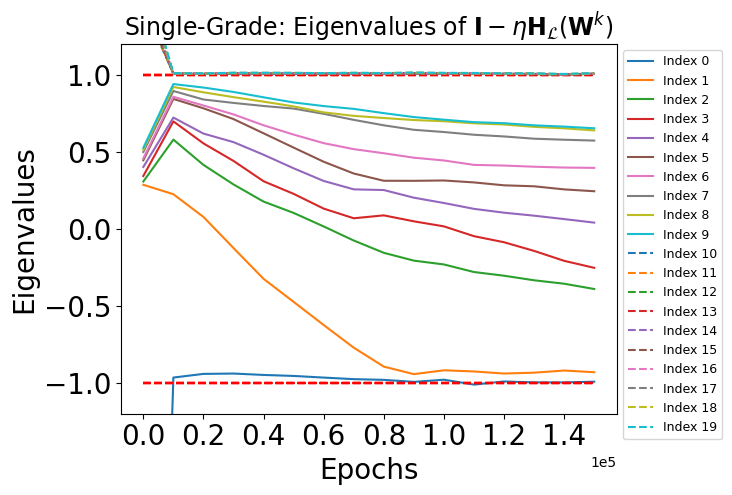}
    \end{subfigure}
  \begin{subfigure}{0.265\linewidth}
\includegraphics[width=\linewidth]{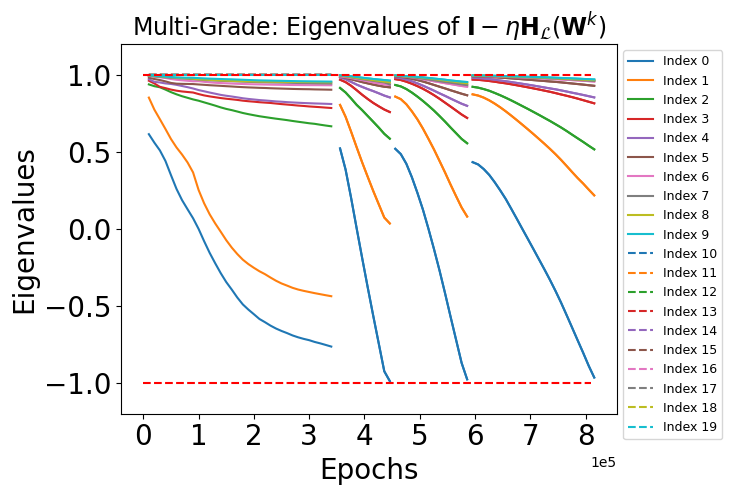}
\end{subfigure}
\begin{subfigure}{0.225\linewidth}
\includegraphics[width=\linewidth]{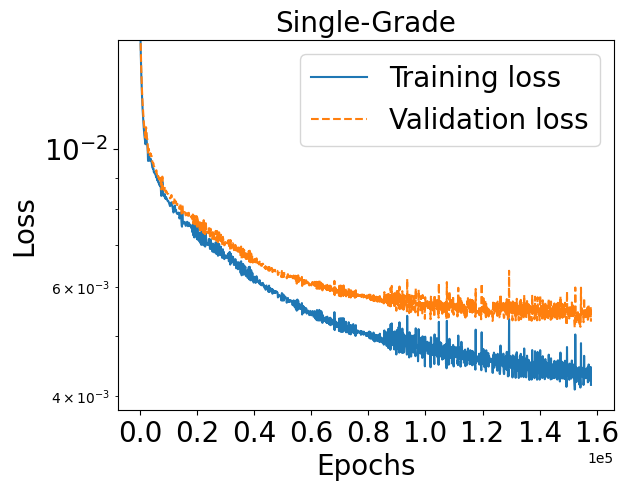}
    \end{subfigure}
  \begin{subfigure}{0.225\linewidth}
\includegraphics[width=\linewidth]{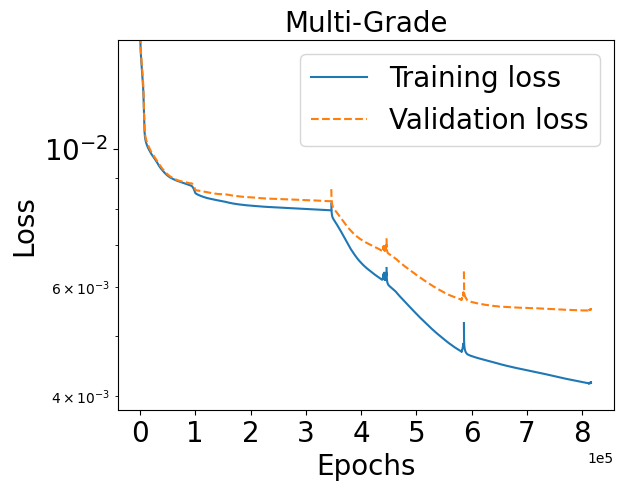}
\end{subfigure}
\caption{Training process of SGDL ($\eta = 0.08$) and MGDL ($\eta = 0.2$) for image `Barbara'.
}
	\label{fig: image regression barabra EigStop}
\end{figure}

\begin{figure}[H]
  \centering

\begin{subfigure}{0.265\linewidth}
\includegraphics[width=\linewidth]{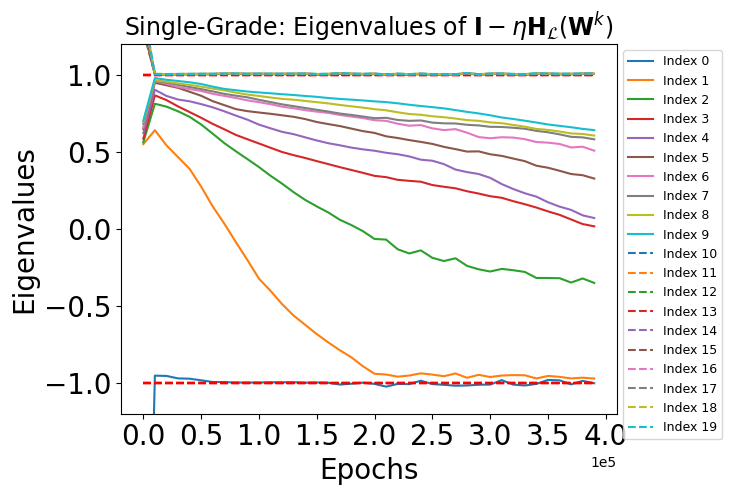}
    \end{subfigure}
  \begin{subfigure}{0.265\linewidth}
\includegraphics[width=\linewidth]{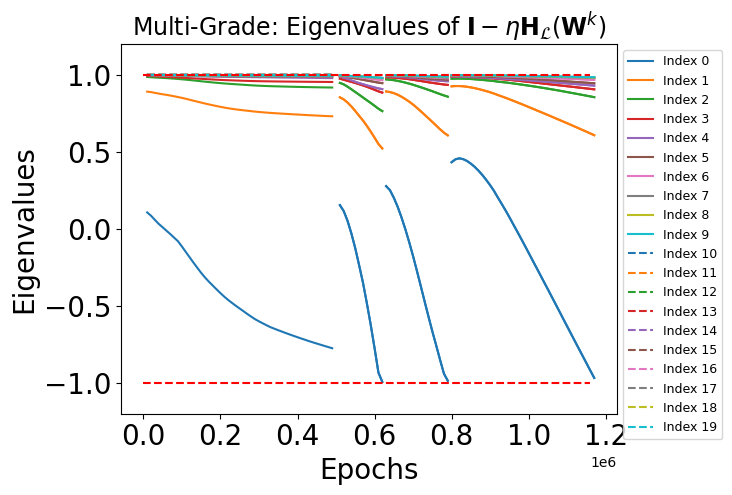}
\end{subfigure}
\begin{subfigure}{0.225\linewidth}
\includegraphics[width=\linewidth]{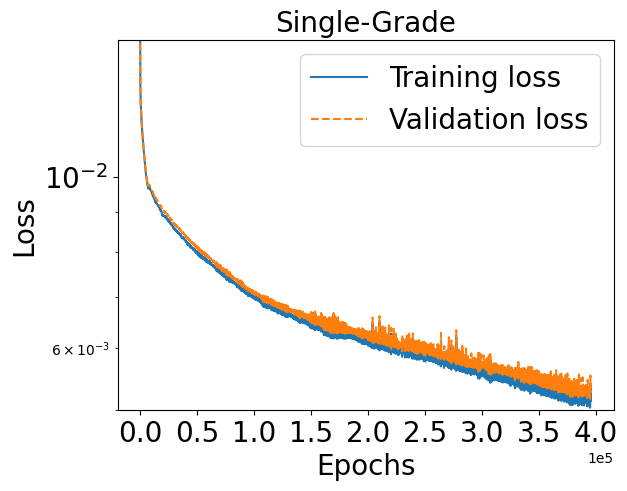}
    \end{subfigure}
  \begin{subfigure}{0.225\linewidth}
\includegraphics[width=\linewidth]{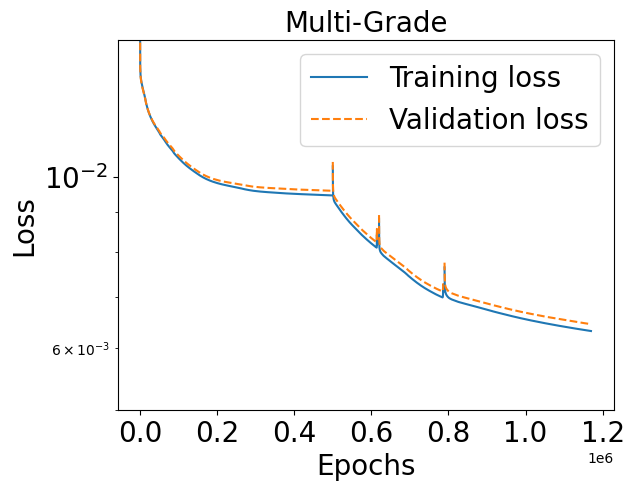}
\end{subfigure}
    
\caption{Training process of SGDL ($\eta = 0.05$) and MGDL ($\eta = 0.05$) for image `Butterfly'. 
}
	\label{fig: image regression butterfly EigStop}
\end{figure}

\begin{figure}[ht]
  \centering
\begin{subfigure}{0.265\linewidth}
\includegraphics[width=\linewidth]{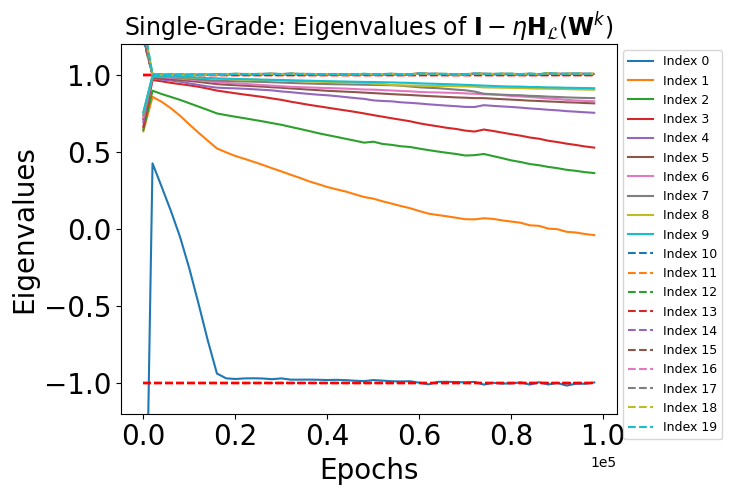}
    \end{subfigure}
  \begin{subfigure}{0.265\linewidth}
\includegraphics[width=\linewidth]{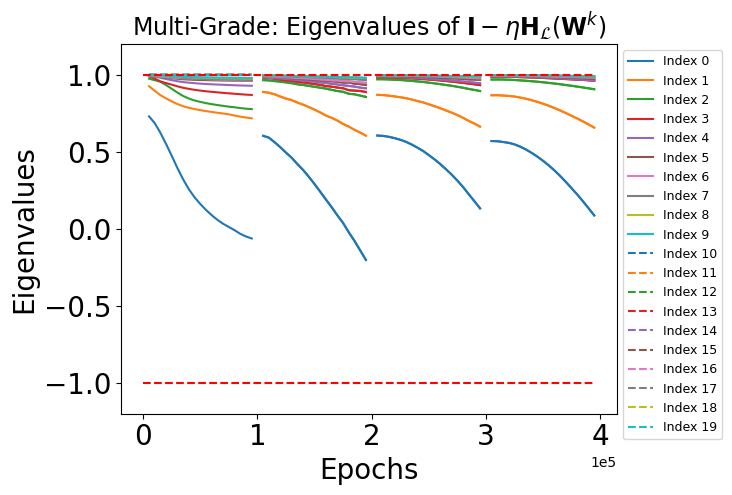}
\end{subfigure}
\begin{subfigure}{0.225\linewidth}
\includegraphics[width=\linewidth]{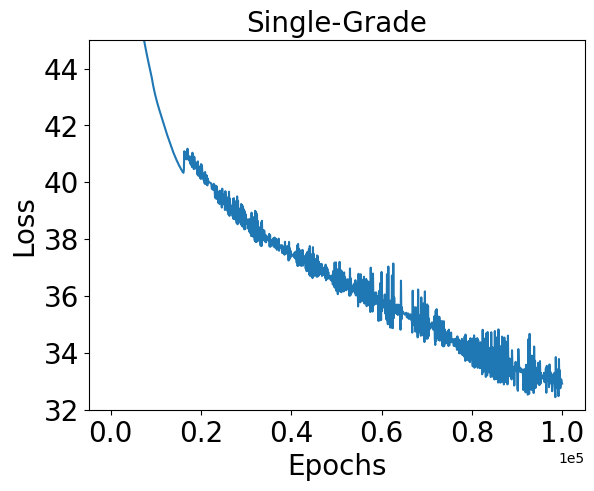}
    \end{subfigure}
  \begin{subfigure}{0.225\linewidth}
\includegraphics[width=\linewidth]{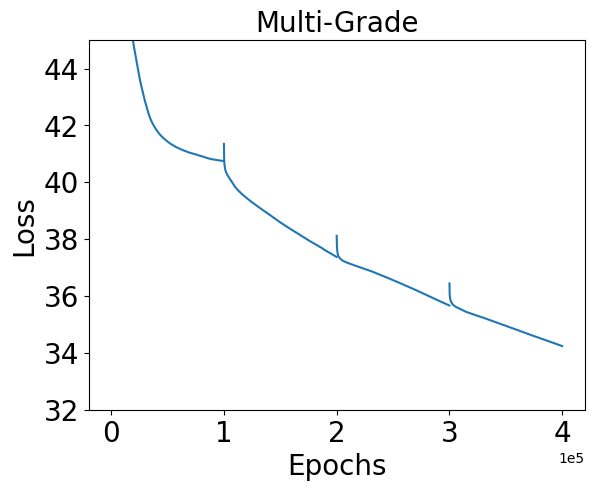}
\end{subfigure}

\caption{Training processes of SGDL and MGDL on the `Butterfly' image (noise level 10).
}
	\label{fig: image regression butterfly image denoisng10}
\end{figure}

\begin{figure}[H]
  \centering
\begin{subfigure}{0.265\linewidth}
\includegraphics[width=\linewidth]{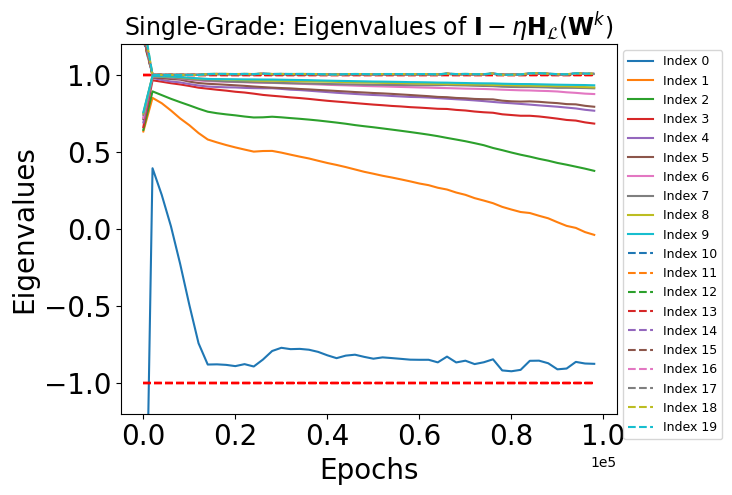}
    \end{subfigure}
  \begin{subfigure}{0.265\linewidth}
\includegraphics[width=\linewidth]{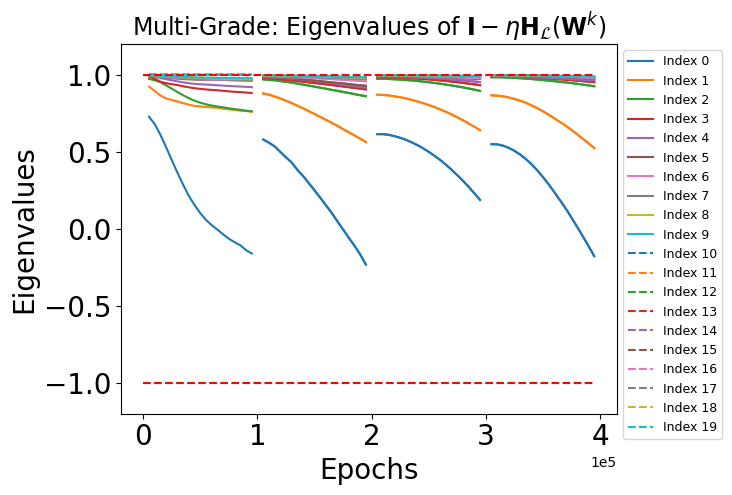}
\end{subfigure}
\begin{subfigure}{0.225\linewidth}
\includegraphics[width=\linewidth]{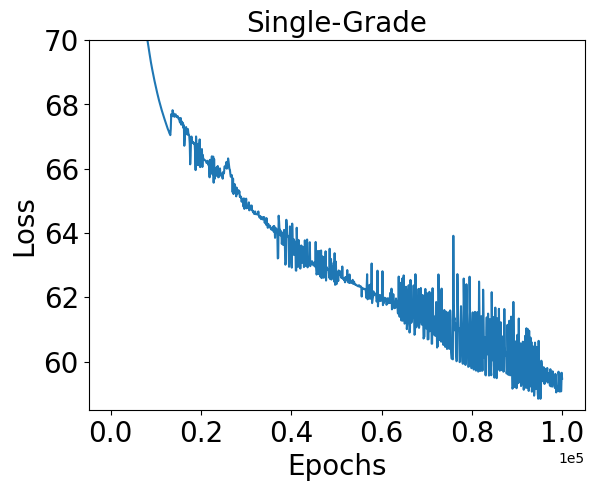}
    \end{subfigure}
  \begin{subfigure}{0.225\linewidth}
\includegraphics[width=\linewidth]{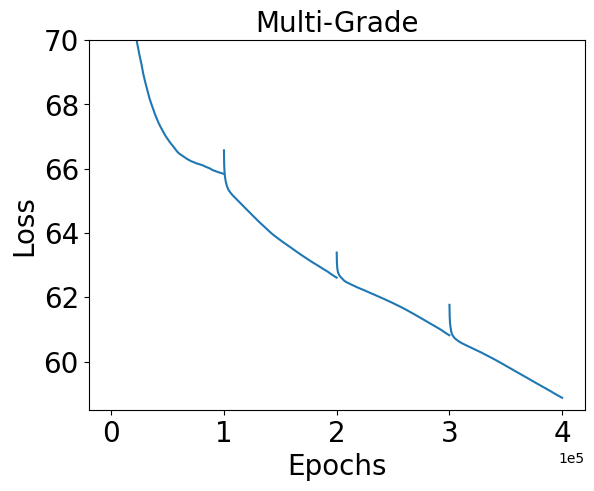}
\end{subfigure}
\caption{Training processes of SGDL and MGDL on the `Butterfly' image (noise level 30).
}
	\label{fig: image regression butterfly image denoisng30}
\end{figure}

\begin{figure}[H]
  \centering
\begin{subfigure}{0.265\linewidth}
\includegraphics[width=\linewidth]{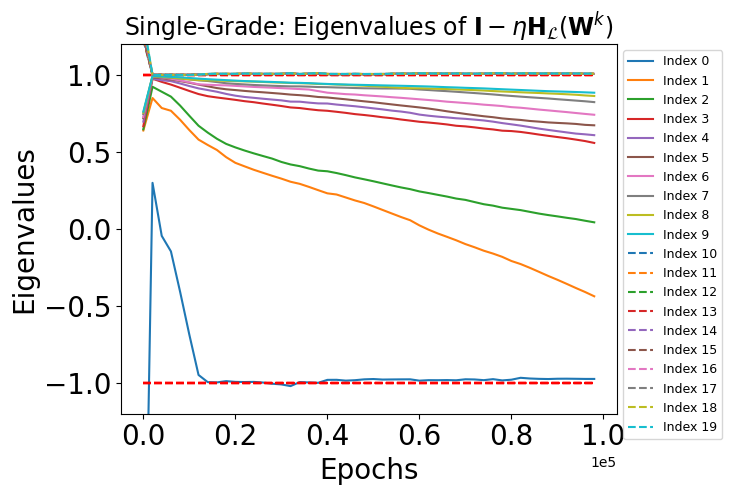}
    \end{subfigure}
  \begin{subfigure}{0.265\linewidth}
\includegraphics[width=\linewidth]{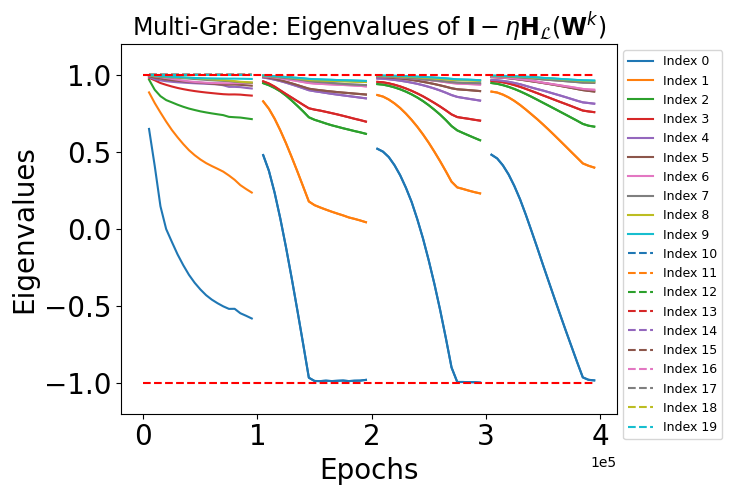}
\end{subfigure}
\begin{subfigure}{0.225\linewidth}
\includegraphics[width=\linewidth]{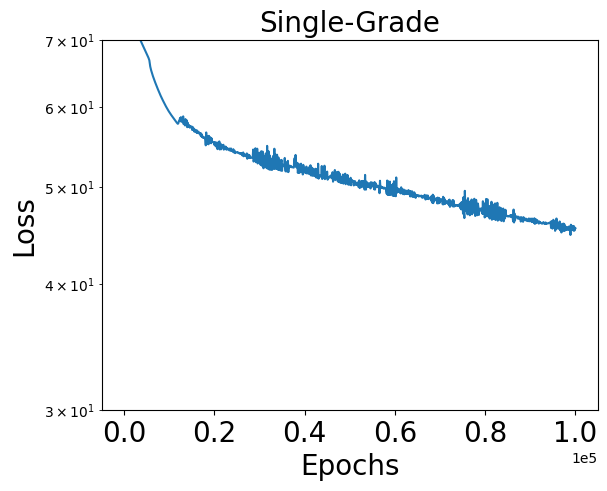}
    \end{subfigure}
  \begin{subfigure}{0.225\linewidth}
\includegraphics[width=\linewidth]{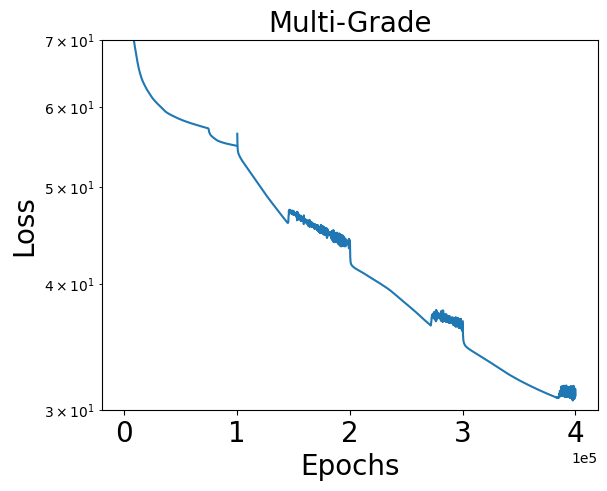}
\end{subfigure}

\caption{Training processes of SGDL and MGDL on the `Barbara' image (noise level 10).
}
	\label{fig: image regression barbara image denoisng10}
\end{figure}

\begin{figure}[H]
  \centering

\begin{subfigure}{0.265\linewidth}
\includegraphics[width=\linewidth]{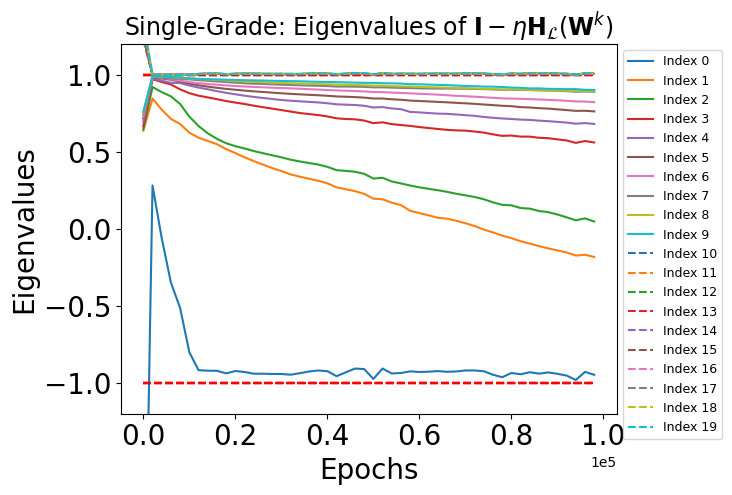}
    \end{subfigure}
  \begin{subfigure}{0.265\linewidth}
\includegraphics[width=\linewidth]{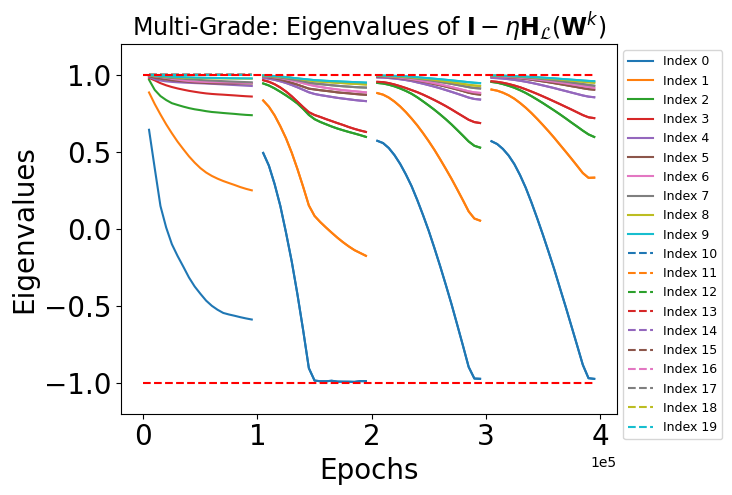}
\end{subfigure}
\begin{subfigure}{0.225\linewidth}
\includegraphics[width=\linewidth]{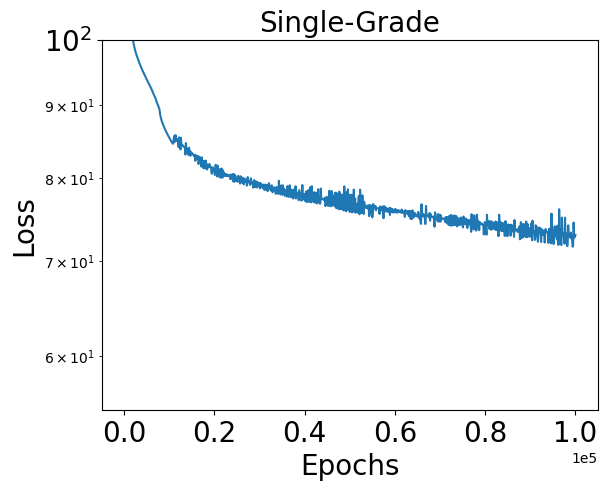}
    \end{subfigure}
  \begin{subfigure}{0.225\linewidth}
\includegraphics[width=\linewidth]{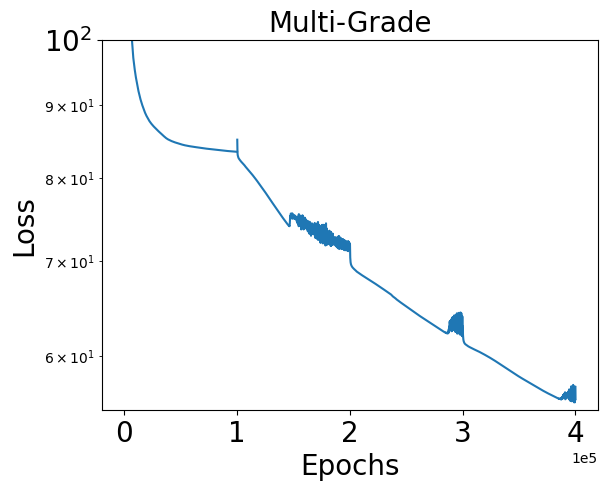}
\end{subfigure}

\caption{Training processes of SGDL and MGDL on the `Barbara' image (noise level 30).
}
	\label{fig: image regression barbara image denoisng30}
\end{figure}

The supporting figures referenced in Section \ref{section: Eigenvalue Analysis} include:

\begin{enumerate}
    \item Figures \ref{fig: Numerical analysis sin SGDL MGDL setting 2}-\ref{fig: appendix synthetic data regression}: Results for the `Synthetic data regression' in Section \ref{section: Eigenvalue Analysis}.

    \item Figures \ref{fig: image regression cameraman EigStop}-\ref{fig: image regression butterfly EigStop}: Results for the `Image regression' in  Section \ref{section: Eigenvalue Analysis}.

    \item Figures \ref{fig: image regression butterfly image denoisng10}-\ref{fig: image regression barbara image denoisng30}: Results for the `Image denoising' in  Section \ref{section: Eigenvalue Analysis}.
\end{enumerate}

\subsection{Section \ref{section:MGDL and MSDL}}

\begin{figure}[htbp]
  \centering

   \begin{subfigure}{0.265\linewidth}
\includegraphics[width=\linewidth]{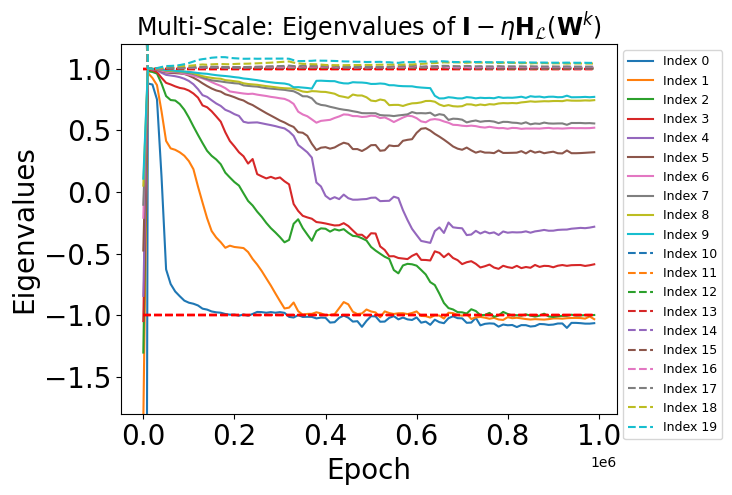}
    \end{subfigure}
   \begin{subfigure}{0.265\linewidth}
\includegraphics[width=\linewidth]{MGDLEXAMPLE/Frequency/MultiGrade_FrequencyFunction1_Eig.png}
    \end{subfigure}
   \begin{subfigure}{0.225\linewidth}
\includegraphics[width=\linewidth]{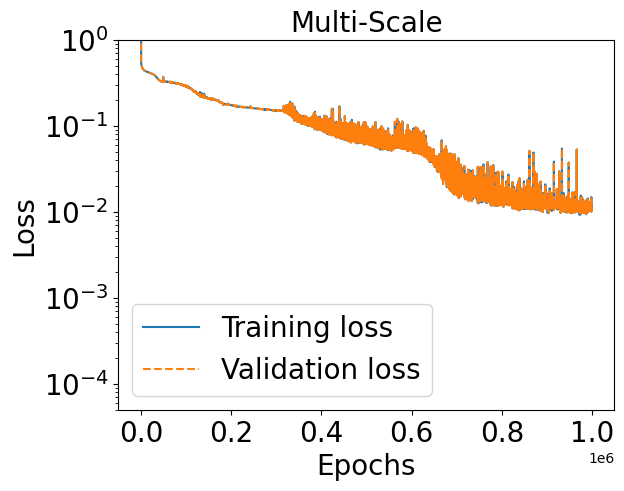}
    \end{subfigure}
   \begin{subfigure}{0.225\linewidth}
\includegraphics[width=\linewidth]{MGDLEXAMPLE/Frequency/MultiGrade_FrequencyFunction1_Loss.png}
    \end{subfigure}

\caption{Training process of MSDL ($\eta = 0.004$) and MGDL ($\eta = 0.06$) for Setting 1. 
}
	\label{fig: Numerical analysis sin MSDL MGDL setting 1}
\end{figure}

\begin{figure}[H]
  \centering

   \begin{subfigure}{0.265\linewidth}
\includegraphics[width=\linewidth]{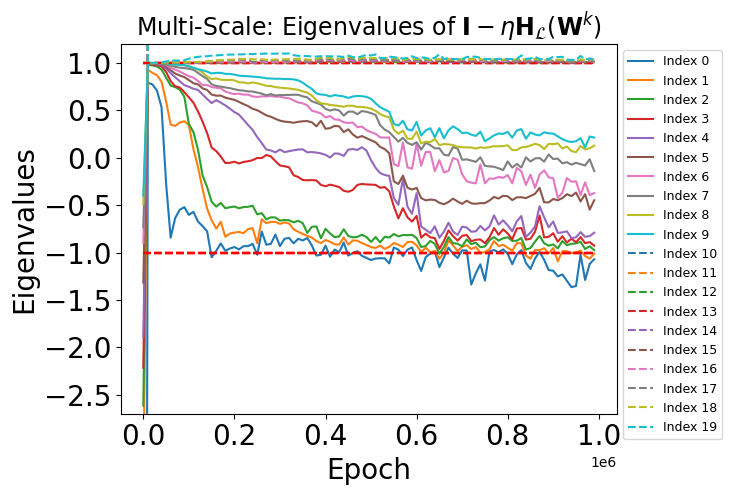}
    \end{subfigure}
   \begin{subfigure}{0.265\linewidth}
\includegraphics[width=\linewidth]{MGDLEXAMPLE/Frequency/MultiGrade_FrequencyFunction2_Eig.png}
    \end{subfigure}
   \begin{subfigure}{0.225\linewidth}
\includegraphics[width=\linewidth]{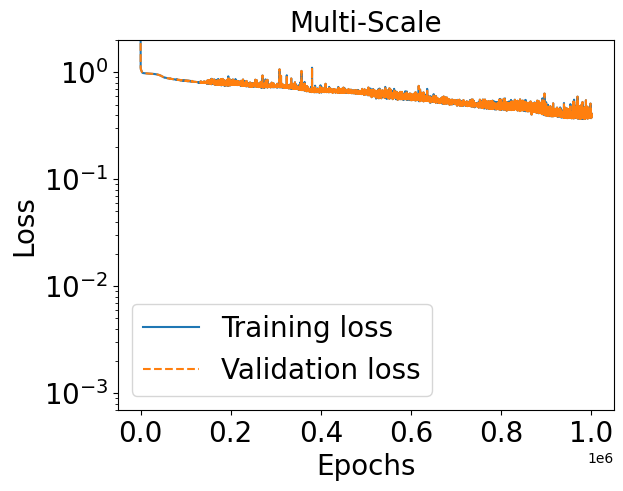}
    \end{subfigure}    
   \begin{subfigure}{0.225\linewidth}
\includegraphics[width=\linewidth]{MGDLEXAMPLE/Frequency/MultiGrade_FrequencyFunction2_Loss.png}
    \end{subfigure}

\caption{Training process of MSDL ($\eta=0.006$) and MGDL($\eta = 0.2$) for Setting 2. 
}
	\label{fig: Numerical analysis sin MSDL MGDL setting 2}
\end{figure}

\begin{figure}[H]
    \centering
    \begin{minipage}{0.49\linewidth}
        \centering
        \includegraphics[width=0.49\linewidth]{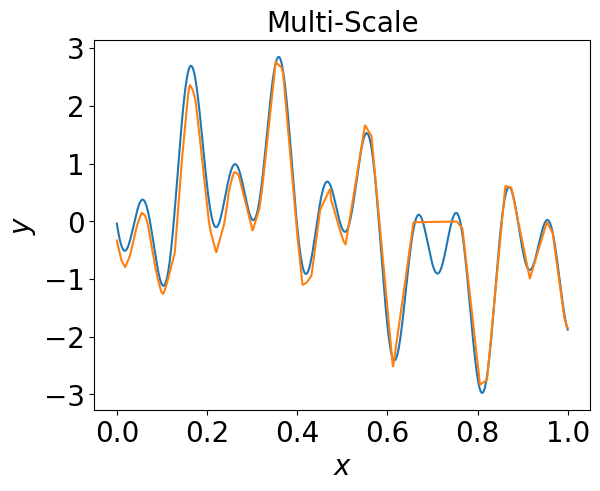}
        \includegraphics[width=0.49\linewidth]{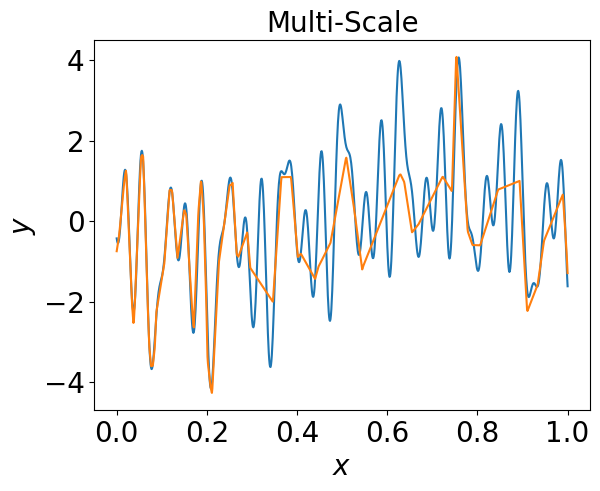}
        \caption{Prediction for MSDL.  
        }
        \label{fig: MSDL prediction}
    \end{minipage}
    \hfill
    \begin{minipage}{0.49\linewidth}
        \centering
        \includegraphics[width=0.49\linewidth]{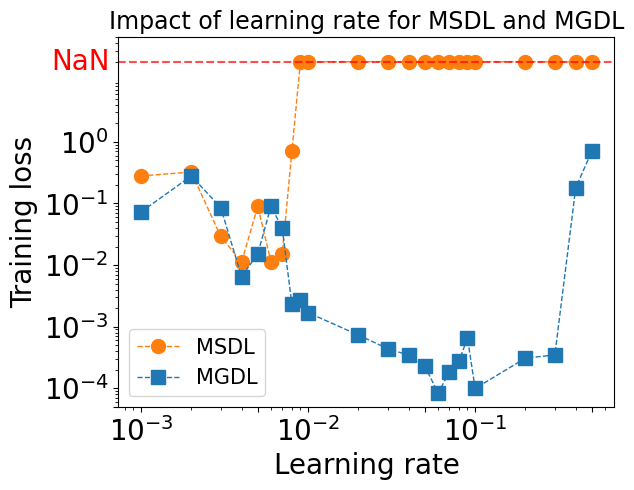}
        \includegraphics[width=0.49\linewidth]{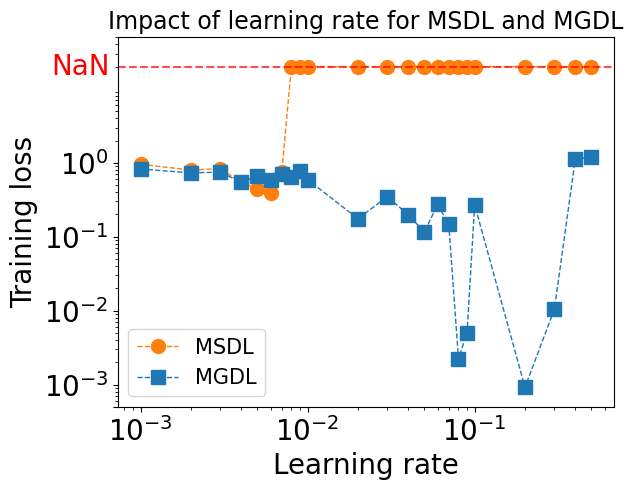}
        \caption{Impact of learning rate. 
        }
        \label{fig:learning rate effectiveness MSDL MGDL}
    \end{minipage}
\end{figure}

The supporting Figures for Section \ref{section:MGDL and MSDL} include Figures \ref{fig: Numerical analysis sin MSDL MGDL setting 1}-\ref{fig:learning rate effectiveness MSDL MGDL}.

\end{document}


\maketitle

\section*{Gradient and Hessian Matrix Computation}\label{appendix: hessian computation}

This supplementary material focuses on computing the gradient and the Hessian matrix of $\mathcal{L}$. 

We define some notations.
For $\ell \in \mathbb{N}_N, j \in \mathbb{N}_{D-1}$, we let
\begin{equation}\label{a_j}
\mathbf{a}_{j\ell} := \mathcal{H}_{j}(\mathbf{x}_{\ell}), \quad \mathbf{a}_{0\ell} := \mathbf{x}_{\ell}
\end{equation}
and the diagonal index matrix
\begin{equation}\label{notation I Wj bj x}
[\mathbb{I}_{\mathbf{W}_j, \mathbf{b}_j, \mathbf{a}_{(j-1)\ell}}]_{ii}:= \begin{cases}
    1, & [\mathbf{W}_j^{\top} \mathbf{a}_{(j-1)\ell} + \mathbf{b}_j]_{ii} \geq 0\\
    0, & [\mathbf{W}_j^{\top} \mathbf{a}_{(j-1)\ell} + \mathbf{b}_j]_{ii} < 0
\end{cases}, \ \ i\in \mathbb{N}_{d_j}
\end{equation}
and
$$
\mathbf{e}_{D\ell}:=\mathcal{N}_{D}\left(\{\mathbf{W}_j,\mathbf{b}_j\}_{j=1}^D; \mathbf{x}_{\ell}\right) - \mathbf{y}_{\ell}.
$$
From the definition, $\mathbb{I}_{\mathbf{W}_j, \mathbf{b}_j, \mathbf{a}_{(j-1)\ell}}$ is a piecewise constant function with respect to $\mathbf{W}_j$ and $\mathbf{b}_j$, and therefore its gradient with respective to $W_j$ and $b_j$ are zero. Note that we do not consider the boundaries of each piece, as they have zero measure.

We will consider the network functions with one and four hidden layers, which will be used in the main paper.


\textbf{A single hidden layer}

We consider the network with a single hidden layer. 
In this case, $D=2$. The network function is
\begin{equation}\label{layer 2 network function}
\mathcal{N}_2\left( \left\{\mathbf{W}_j, \mathbf{b}_j\right\}_{j=1}^2; \mathbf{x} \right) := \mathbf{W}_2^{\top}\sigma\left(\mathbf{W}_1^{\top} \mathbf{x} + \mathbf{b}_1\right) + \mathbf{b}_2.
\end{equation}

We first compute the gradient. Since we use the vectorization of the parameters in our computation, it is important to review the Kronecker product \cite{schacke2004kronecker, van2000ubiquitous}. If $\mathbf{A}$ is an $m \times n$ matrix and $\mathbf{B}$ is a $p \times q$ matrix, then the Kronecker product $\mathbf{A} \otimes \mathbf{B}$ is the $p m \times q n$ block matrix:
$$
\mathbf{A} \otimes \mathbf{B}=\left[\begin{array}{ccc}
a_{11} \mathbf{B} & \cdots & a_{1 n} \mathbf{B} \\
\vdots & \ddots & \vdots \\
a_{m 1} \mathbf{B} & \cdots & a_{m n} \mathbf{B}
\end{array}\right].
$$
Let $\mathbf{A}, \mathbf{B}$ and $\mathbf{V}$ be three matrices. The mixed Kronecker matrix-vector product can be written as:
$$
(\mathbf{A} \otimes \mathbf{B}) \operatorname{vec}(\mathbf{V})=\operatorname{vec}(\mathbf{B V} \mathbf{A}^T)
$$
where the operator $\operatorname{vec}$ applied on a matrix denotes the vectorization of the matrix.
For convenience, we let
$
\mathbf{z}_{\ell} = \mathbb{I}_{\mathbf{W}_1, \mathbf{b}_1, \mathbf{x}_{\ell}} \mathbf{W}_2
$ for $\ell \in \mathbb{N}_{N}$.

\begin{lemma}\label{lemma: graident}
Let $\mathcal{N}_2$ be the network function defined as \eqref{layer 2 network function} and $\mathcal{L}$ be the corresponding loss function for this network. Then, we have that
$$
    \frac{\partial \mathcal{L}}{\partial W_1} = \frac{1}{N}\sum_{\ell=1}^N (\mathbf{z}_{\ell} \otimes \mathbf{x}_{\ell} )\mathbf{e}_{2\ell}, 
    \quad
    \frac{\partial \mathcal{L}}{\partial b_1} = \frac{1}{N}\sum_{\ell=1}^N \mathbf{z}_{\ell}  \mathbf{e}_{2\ell},
    $$
    $$
    \frac{\partial \mathcal{L}}{\partial W_2} = \frac{1}{N}\sum_{\ell=1}^N  \mathbf{a}_{1\ell} \mathbf{e}_{2\ell}, \quad
    \frac{\partial \mathcal{L}}{\partial b_2} = \frac{1}{N}\sum_{\ell=1}^N   \mathbf{e}_{2\ell}.
    $$
\end{lemma}
\begin{proof}
    Using the notations \eqref{notation I Wj bj x} and \eqref{a_j}, we have rewrite function $\mathcal{N}_2$ as
    $$
    \begin{aligned}
    \mathcal{N}_2\left( \left\{\mathbf{W}_j, \mathbf{b}_j\right\}_{j=1}^2; \mathbf{x} \right) &=  \mathbf{W}_2^{\top} \mathbb{I}_{\mathbf{W}_1, \mathbf{b}_1, \mathbf{x}} \mathbf{W}_1^{\top} \mathbf{x} + \mathbf{W}_2^{\top} \mathbb{I}_{\mathbf{W}_1, \mathbf{b}_1, \mathbf{x}} \mathbf{b}_1 + \mathbf{b}_2 \\
    &=\left(\mathbf{z}^{\top} \otimes \mathbf{x}^{\top}\right)W_1 + \mathbf{z}^{\top}b_1 + b_2
    \end{aligned}
    $$
    and 
    $$
    \mathcal{N}_2\left( \left\{\mathbf{W}_j, \mathbf{b}_j\right\}_{j=1}^2; \mathbf{x} \right) = \mathbf{W}_2^{\top} \mathbf{a}_1 + \mathbf{b}_2.
    $$
    Therefore,
    $$
    \frac{\partial \mathcal{N}_2}{\partial W_1
    } =  \mathbf{z} \otimes \mathbf{x}, \ \ 
    \frac{\partial \mathcal{N}_2}{\partial b_1
    } =  \mathbf{z}, \ \
    \frac{\partial \mathcal{N}_2}{\partial W_2
    } = \mathbf{a}_1, \ \ \frac{\partial \mathcal{N}_2}{\partial b_2
    } = 1. 
    $$
    The lemma can be proved by using the chain's rule of gradient.
\end{proof}

The hessian of  $\mathcal{L}$ at $W$ is given by
$$
\mathbf{H}_{\mathcal{L}}(W) := \begin{bmatrix}
    \frac{\partial^2 \mathcal{L}}{\partial W_1^2} & \frac{\partial^2 \mathcal{L}}{\partial b_1 \partial W_1} & \frac{\partial^2 \mathcal{L}}{\partial W_2 \partial W_1} & \frac{\partial^2 \mathcal{L}}{\partial b_2 \partial W_1} \\
    \frac{\partial^2 \mathcal{L}}{\partial W_1 \partial b_1}  & \frac{\partial^2 \mathcal{L}}{\partial b_1^2} & \frac{\partial^2 \mathcal{L}}{\partial W_2 \partial b_1 } &  \frac{\partial^2 \mathcal{L}}{ \partial b_2 \partial b_1}\\
    \frac{\partial^2 \mathcal{L}}{\partial W_1 \partial W_2} & \frac{\partial^2 \mathcal{L}}{ \partial b_1 \partial b_2} & \frac{\partial^2 \mathcal{L}}{ \partial W_2^2} &  \frac{\partial^2 \mathcal{L}}{\partial b_2 \partial W_2 }\\
    \frac{\partial^2 \mathcal{L}}{\partial W_1 \partial b_2} & \frac{\partial^2 \mathcal{L}}{\partial b_1 \partial b_2} & \frac{\partial^2 \mathcal{L}}{ \partial W_2 \partial b_2} &  \frac{\partial^2 \mathcal{L}}{\partial b_2^2}
\end{bmatrix}.
$$
Since the Hessian matrix $\mathbf{H}_{\mathcal{L}}$ is symmetric, we compute only the upper triangular elements.

\begin{lemma}\label{lemma: hessian}
Let $\mathcal{N}_2$ be the network function defined as \eqref{layer 2 network function} and $\mathcal{L}$ be the corresponding loss function for this network. Then, we have that
$$
\frac{\partial^2 \mathcal{L}}{\partial W_1^2} = \frac{1}{N } \sum_{\ell=1}^N \left( \mathbf{z}_{\ell} \otimes \mathbf{x}_{\ell}   \right)  \left( \mathbf{z}_{\ell} \otimes \mathbf{x}_{\ell}   \right)^{\top}, \quad
\frac{\partial^2 \mathcal{L}}{\partial b_1 \partial W_1} = \frac{1}{N} \sum_{\ell=1}^N \left( \mathbf{z}_{\ell} \otimes \mathbf{x}_{\ell}   \right)   \mathbf{z}_{\ell}^{\top}
$$
$$
\begin{aligned}
\frac{\partial^2 \mathcal{L}}{\partial W_2 \partial W_1}  =& \frac{1}{N}\sum_{\ell=1}^{N} \left(\mathbb{I}_{\mathbf{W}_1, \mathbf{b}_1, \mathbf{x}_{\ell}} \otimes \mathbf{x}_{\ell}\right) \mathbf{e}_{2\ell} +   \left(\mathbf{z}_{\ell}\otimes \mathbf{x}_{\ell} \right) \mathbf{a}_{1\ell}^{\top}, \quad \frac{\partial^2 \mathcal{L}}{\partial b_2 \partial W_1}  = \frac{1}{N} \sum_{\ell=1}^N \mathbf{z}_{\ell} \otimes \mathbf{x}_{\ell}.
\end{aligned}
$$
$$
\frac{\partial^2 \mathcal{L}}{ \partial b_1^2} =  \frac{1}{N} \sum_{\ell=1}^{N} \mathbf{z}_{\ell} \mathbf{z}_{\ell}^{\top}, \quad
\frac{\partial^2 \mathcal{L}}{ \partial W_2 \partial b_1} = \frac{1}{N}\sum_{\ell=1}^{N}
\mathbb{I}_{\mathbf{W}_1, \mathbf{b}_1, \mathbf{x}_{\ell}} \mathbf{e}_{2\ell} + \mathbf{z}_{\ell}\mathbf{a}_{1\ell}^{\top}
,\quad
\frac{\partial^2 \mathcal{L}}{ \partial b_2 \partial b_1}  = \frac{1}{N} \sum_{\ell=1}^{N}\mathbf{z}_{\ell}
$$
$$
\frac{\partial^2 \mathcal{L}}{\partial W_2^2} = \frac{1}{N}\sum_{\ell=1}^{N}\mathbf{a}_{1\ell}\mathbf{a}_{1\ell}^{\top}
, \quad
\frac{\partial^2 \mathcal{L}}{\partial b_2 \partial W_2} = \frac{1}{N}\sum_{\ell=1}^{N} \mathbf{a}_{1\ell}, \quad
\frac{\partial^2 \mathcal{L}}{\partial b_2^2} = 1.
$$
\end{lemma}
\begin{proof}
We only prove $\frac{\partial^2 \mathcal{L}}{\partial W_1^2}$ and $\frac{\partial^2 \mathcal{L}}{\partial W_2 \partial W_1}$, as the proofs for the other terms follow in the same manner.

We first compute $\frac{\partial^2 \mathcal{L}}{\partial W_1^2}$.  From Lemma \ref{lemma: graident}, we have that
$$
\frac{\partial^2 \mathcal{L}}{\partial W_1^2} = \frac{1}{N}\sum_{\ell=1}^N \frac{\partial (\mathbf{z}_{\ell} \otimes \mathbf{x}_{\ell} )\mathbf{e}_{2\ell}}{\partial W_1} =  \frac{1}{N}\sum_{\ell=1}^N \frac{\partial (\mathbf{z}_{\ell} \otimes \mathbf{x}_{\ell} )}{\partial W_1}\mathbf{e}_{2\ell} + (\mathbf{z}_{\ell} \otimes \mathbf{x}_{\ell} )\left(\frac{\partial \mathbf{e}_{2\ell}}{\partial W_1}\right)^{\top}.
$$
Since 
$$
\frac{\partial (\mathbf{z}_{\ell} \otimes \mathbf{x}_{\ell} )}{\partial W_1} = 0, \ \ \frac{\partial \mathbf{e}_{2\ell}}{\partial W_1} = \mathbf{z}_{\ell} \otimes \mathbf{x}_{\ell},
$$
we have that
$$
\frac{\partial^2 \mathcal{L}}{\partial W_1^2} = \frac{1}{N } \sum_{\ell=1}^N \left( \mathbf{z}_{\ell} \otimes \mathbf{x}_{\ell}   \right)  \left( \mathbf{z}_{\ell} \otimes \mathbf{x}_{\ell}   \right)^{\top}.
$$

We now compute $\frac{\partial^2 \mathcal{L}}{\partial W_2 \partial W_1}$. Again from Lemma \ref{lemma: graident}, we have that
$$
\frac{\partial^2 \mathcal{L}}{\partial W_2 \partial W_1} = \frac{1}{N}\sum_{\ell=1}^N \frac{\partial (\mathbf{z}_{\ell} \otimes \mathbf{x}_{\ell} )\mathbf{e}_{2\ell}}{\partial W_2} =  \frac{1}{N}\sum_{\ell=1}^N \frac{\partial (\mathbf{z}_{\ell} \otimes \mathbf{x}_{\ell} )}{\partial W_2}\mathbf{e}_{2\ell} + (\mathbf{z}_{\ell} \otimes \mathbf{x}_{\ell} )\left(\frac{\partial \mathbf{e}_{2\ell}}{\partial W_2}\right)^{\top}.
$$
As $\mathbf{z}_{\ell} = \mathbb{I}_{\mathbf{W}_1, \mathbf{b}_1, \mathbf{x}_{\ell}} \mathbf{W}_2$, we have that
$$
\frac{\partial (\mathbf{z}_{\ell} \otimes \mathbf{x}_{\ell} )}{\partial W_2} = \mathbb{I}_{\mathbf{W}_1, \mathbf{b}_1, \mathbf{x}_{\ell}} \otimes \mathbf{x}_{\ell}.
$$
The definition of $\mathbf{e}_{2\ell}$ givens that $
\frac{\partial \mathbf{e}_{2\ell}}{\partial W_2} = \mathbf{a}_{1\ell}$.
Therefore, 
$$
\frac{\partial^2 \mathcal{L}}{\partial W_2 \partial W_1}  = \frac{1}{N}\sum_{\ell=1}^{N} \left(\mathbb{I}_{\mathbf{W}_1, \mathbf{b}_1, \mathbf{x}_{\ell}} \otimes \mathbf{x}_{\ell}\right) \mathbf{e}_{2\ell} +   \left(\mathbf{z}_{\ell}\otimes \mathbf{x}_{\ell} \right) \mathbf{a}_{1\ell}^{\top}.
$$ 
\end{proof}

\textbf{Four hidden layers}

We consider a network with four hidden layers. In this case, $D=5$. The network function is

\begin{equation}\label{layer 5 network function}
\mathcal{N}_5\left( \left\{\mathbf{W}_j, \mathbf{b}_j\right\}_{j=1}^5; \mathbf{x} \right) := \mathbf{W}_5^{\top}\sigma\left(\mathbf{W}_4^{\top}\sigma\left(\mathbf{W}_3^{\top} \sigma\left(\mathbf{W}_2^{\top}\sigma\left(\mathbf{W}_1^{\top} \mathbf{x} + \mathbf{b}_1\right) + \mathbf{b}_2\right)+\mathbf{b}_3\right) + \mathbf{b}_4\right) + \mathbf{b}_5.
\end{equation}

The Hessian $\mathbf{H}_{\mathcal{L}}(W)$ is given by
\footnotesize{
$$
\begin{bmatrix}
    \frac{\partial^2 \mathcal{L}}{\partial W_1^2} & \frac{\partial^2 \mathcal{L}}{ \partial b_1 \partial W_1} & \frac{\partial^2 \mathcal{L}}{ \partial W_2 \partial W_1} &  \frac{\partial^2 \mathcal{L}}{ \partial b_2 \partial W_1} & \frac{\partial^2 \mathcal{L}}{ \partial W_3 \partial W_1} & \frac{\partial^2 \mathcal{L}}{ \partial b_3 \partial W_1} & \frac{\partial^2 \mathcal{L}}{ \partial W_4 \partial W_1} & \frac{\partial^2 \mathcal{L}}{ \partial b_4 \partial W_1} & \frac{\partial^2 \mathcal{L}}{ \partial W_5 \partial W_1} & \frac{\partial^2 \mathcal{L}}{ \partial b_5 \partial W_1}
    \\
    \frac{\partial^2 \mathcal{L}}{ \partial W_1 \partial b_1} & \frac{\partial^2 \mathcal{L}}{\partial b_1^2} & \frac{\partial^2 \mathcal{L}}{\partial W_2 \partial b_1 } &  \frac{\partial^2 \mathcal{L}}{\partial b_2\partial b_1 } &  \frac{\partial^2 \mathcal{L}}{ \partial W_3\partial b_1} &  \frac{\partial^2 \mathcal{L}}{ \partial b_3\partial b_1}&  \frac{\partial^2 \mathcal{L}}{ \partial W_4\partial b_1} &  \frac{\partial^2 \mathcal{L}}{ \partial b_4\partial b_1} & \frac{\partial^2 \mathcal{L}}{ \partial W_5 \partial b_1} & \frac{\partial^2 \mathcal{L}}{ \partial b_5 \partial b_1}
    \\
    \frac{\partial^2 \mathcal{L}}{\partial W_1\partial W_2 } & \frac{\partial^2 \mathcal{L}}{ \partial b_1\partial W_2} & \frac{\partial^2 \mathcal{L}}{ \partial W_2^2} &  \frac{\partial^2 \mathcal{L}}{\partial b_2\partial W_2} & \frac{\partial^2 \mathcal{L}}{\partial W_3\partial W_2} & \frac{\partial^2 \mathcal{L}}{\partial b_3 \partial W_2}& \frac{\partial^2 \mathcal{L}}{\partial W_4\partial W_2} & \frac{\partial^2 \mathcal{L}}{\partial b_4 \partial W_2} & \frac{\partial^2 \mathcal{L}}{ \partial W_5 \partial W_2} & \frac{\partial^2 \mathcal{L}}{ \partial b_5 \partial W_2}
    \\
    \frac{\partial^2 \mathcal{L}}{\partial W_1\partial b_2 } & \frac{\partial^2 \mathcal{L}}{\partial b_1\partial b_2} & \frac{\partial^2 \mathcal{L}}{ \partial W_2\partial b_2} &  \frac{\partial^2 \mathcal{L}}{\partial b_2^2} & \frac{\partial^2 \mathcal{L}}{ \partial W_3\partial b_2} & \frac{\partial^2 \mathcal{L}}{ \partial b_3\partial b_2}  & \frac{\partial^2 \mathcal{L}}{ \partial W_4\partial b_2} & \frac{\partial^2 \mathcal{L}}{ \partial b_4\partial b_2} & \frac{\partial^2 \mathcal{L}}{ \partial W_5 \partial b_2} & \frac{\partial^2 \mathcal{L}}{ \partial b_5 \partial b_2}
    \\
     \frac{\partial^2 \mathcal{L}}{ \partial W_1\partial W_3} & \frac{\partial^2 \mathcal{L}}{ \partial b_1\partial W_3} & \frac{\partial^2 \mathcal{L}}{ \partial W_2\partial W_3} &  \frac{\partial^2 \mathcal{L}}{ \partial b_2\partial W_3 } & \frac{\partial^2 \mathcal{L}}{\partial W_3^2} & \frac{\partial^2 \mathcal{L}}{ \partial b_3\partial W_3} & \frac{\partial^2 \mathcal{L}}{\partial W_4\partial W_3} & \frac{\partial^2 \mathcal{L}}{ \partial b_4\partial W_3} & \frac{\partial^2 \mathcal{L}}{ \partial W_5 \partial W_3} & \frac{\partial^2 \mathcal{L}}{ \partial b_5 \partial W_3}
     \\
     \frac{\partial^2 \mathcal{L}}{ \partial W_1\partial b_3} & \frac{\partial^2 \mathcal{L}}{ \partial b_1\partial b_3} & \frac{\partial^2 \mathcal{L}}{ \partial W_2\partial b_3} &  \frac{\partial^2 \mathcal{L}}{  \partial b_2\partial b_3} & \frac{\partial^2 \mathcal{L}}{ \partial W_3\partial b_3} & \frac{\partial^2 \mathcal{L}}{\partial b_3^2} & \frac{\partial^2 \mathcal{L}}{ \partial W_4\partial b_3} & \frac{\partial^2 \mathcal{L}}{\partial b_4\partial b_3}  & \frac{\partial^2 \mathcal{L}}{ \partial W_5 \partial b_3} & \frac{\partial^2 \mathcal{L}}{ \partial b_5 \partial b_3}
     \\
 \frac{\partial^2 \mathcal{L}}{ \partial W_1\partial W_4} & \frac{\partial^2 \mathcal{L}}{ \partial b_1\partial W_4} & \frac{\partial^2 \mathcal{L}}{ \partial W_2\partial W_4} &  \frac{\partial^2 \mathcal{L}}{  \partial b_2\partial W_4} & \frac{\partial^2 \mathcal{L}}{ \partial W_3\partial W_4} & \frac{\partial^2 \mathcal{L}}{\partial b_3 \partial W_4} & \frac{\partial^2 \mathcal{L}}{ \partial W_4^2} & \frac{\partial^2 \mathcal{L}}{\partial b_4\partial W_4}  & \frac{\partial^2 \mathcal{L}}{ \partial W_5 \partial W_4} & \frac{\partial^2 \mathcal{L}}{ \partial b_5 \partial W_4}
 \\
\frac{\partial^2 \mathcal{L}}{ \partial W_1\partial b_4} & \frac{\partial^2 \mathcal{L}}{ \partial b_1\partial b_4} & \frac{\partial^2 \mathcal{L}}{ \partial W_2\partial b_4} &  \frac{\partial^2 \mathcal{L}}{  \partial b_2\partial b_4} & \frac{\partial^2 \mathcal{L}}{ \partial W_3\partial b_4} & \frac{\partial^2 \mathcal{L}}{\partial b_3 \partial b_4} & \frac{\partial^2 \mathcal{L}}{ \partial W_4 \partial b_4} & \frac{\partial^2 \mathcal{L}}{\partial b_4^2} & \frac{\partial^2 \mathcal{L}}{ \partial W_5 \partial b_4} & \frac{\partial^2 \mathcal{L}}{ \partial b_5 \partial b_4}
\\
 \frac{\partial^2 \mathcal{L}}{ \partial W_1\partial W_5} & \frac{\partial^2 \mathcal{L}}{ \partial b_1\partial W_5} & \frac{\partial^2 \mathcal{L}}{ \partial W_2\partial W_5} &  \frac{\partial^2 \mathcal{L}}{  \partial b_2\partial W_5} & \frac{\partial^2 \mathcal{L}}{ \partial W_3\partial W_5} & \frac{\partial^2 \mathcal{L}}{\partial b_3 \partial W_5} & \frac{\partial^2 \mathcal{L}}{ \partial W_4 \partial W_5} & \frac{\partial^2 \mathcal{L}}{\partial b_4\partial W_5}  & \frac{\partial^2 \mathcal{L}}{ \partial W_5^2} & \frac{\partial^2 \mathcal{L}}{ \partial b_5 \partial W_5}
 \\
\frac{\partial^2 \mathcal{L}}{ \partial W_1\partial b_5} & \frac{\partial^2 \mathcal{L}}{ \partial b_1\partial b_5} & \frac{\partial^2 \mathcal{L}}{ \partial W_2\partial b_5} &  \frac{\partial^2 \mathcal{L}}{  \partial b_2\partial b_5} & \frac{\partial^2 \mathcal{L}}{ \partial W_3\partial b_5} & \frac{\partial^2 \mathcal{L}}{\partial b_3 \partial b_5} & \frac{\partial^2 \mathcal{L}}{ \partial W_4 \partial b_5} & \frac{\partial^2 \mathcal{L}}{\partial b_4 \partial b_5} & \frac{\partial^2 \mathcal{L}}{ \partial W_5 \partial b_5} & \frac{\partial^2 \mathcal{L}}{ \partial b_5^2}
\end{bmatrix}.
$$
}

For convenience, we let 
$$
\mathbf{z}_{\ell} := \mathbb{I}_{\mathbf{W}_1, \mathbf{b}_1, \mathbf{x}_{\ell}} 
 \mathbf{W}_2 \mathbb{I}_{\mathbf{W}_2, \mathbf{b}_2, \mathbf{a}_{1\ell}}\mathbf{W}_3 \mathbb{I}_{\mathbf{W}_3, \mathbf{b}_3, \mathbf{a}_{2\ell}}\mathbf{W}_4\mathbb{I}_{\mathbf{W}_4, \mathbf{b}_4, \mathbf{a}_{3\ell}}\mathbf{W}_5
 $$
 and
 $$
\mathbf{z}_{1\ell}:=\mathbb{I}_{\mathbf{W}_1, \mathbf{b}_1, \mathbf{x}_{\ell}}\mathbf{W}_2,\quad \mathbf{z}_{2\ell}:=\mathbb{I}_{\mathbf{W}_2, \mathbf{b}_2, \mathbf{a}_{1\ell}}\mathbf{W}_3
,\quad
\mathbf{z}_{3\ell}:=\mathbb{I}_{\mathbf{W}_3, \mathbf{b}_3, \mathbf{a}_{2\ell}}\mathbf{W}_4, \quad 
\mathbf{z}_{4\ell}:=\mathbb{I}_{\mathbf{W}_4, \mathbf{b}_4, \mathbf{a}_{3\ell}}\mathbf{W}_5
$$
and 
$$
\widetilde{\mathbf{z}}_{5\ell} = \mathbf{z}_{3\ell}\mathbf{z}_{4\ell}, \quad \widetilde{\mathbf{z}}_{6\ell} = \mathbf{z}_{2\ell}\mathbf{z}_{3\ell}\mathbf{z}_{4\ell}, \quad \widetilde{\mathbf{z}}_{7\ell} = \mathbf{z}_{1\ell}\mathbf{z}_{2\ell}, \quad \widetilde{\mathbf{z}}_{8\ell} = \mathbf{z}_{1\ell}\mathbf{z}_{2\ell}\mathbf{z}_{3\ell}, \quad \widetilde{\mathbf{z}}_{9\ell} = \mathbf{z}_{2\ell}\mathbf{z}_{3\ell}.
$$

We first compute the gradient of $\mathcal{L}$.
\small{
$$
    \frac{\partial \mathcal{L}}{\partial W_1} = \frac{1}{N}\sum_{\ell=1}^N(\mathbf{z}_{\ell} \otimes \mathbf{x}_{\ell} )\mathbf{e}_{5\ell}
,\quad
    \frac{\partial \mathcal{L}}{\partial b_1} = \frac{1}{N}\sum_{\ell=1}^N \mathbf{z}_{\ell}\mathbf{e}_{5\ell}
,\quad
    \frac{\partial \mathcal{L}}{\partial W_2} = \frac{1}{N}\sum_{\ell=1}^N(\widetilde{\mathbf{z}}_{6\ell} \otimes \mathbf{a}_{1\ell} )\mathbf{e}_{5\ell}
$$

$$    
\frac{\partial \mathcal{L}}{\partial b_2} = \frac{1}{N}\sum_{\ell=1}^N\widetilde{\mathbf{z}}_{6\ell}\mathbf{e}_{5\ell}, \quad
    \frac{\partial \mathcal{L}}{\partial W_3} = \frac{1}{N}\sum_{\ell=1}^N (\widetilde{\mathbf{z}}_{5\ell} \otimes \mathbf{a}_{2\ell} )\mathbf{e}_{5\ell},
    \quad
    \frac{\partial \mathcal{L}}{\partial b_3} = \frac{1}{N}\sum_{\ell=1}^N \widetilde{\mathbf{z}}_{5\ell}\mathbf{e}_{5\ell}
$$

$$
 \frac{\partial \mathcal{L}}{\partial W_4} = \frac{1}{N}\sum_{\ell=1}^N (\mathbf{z}_{4\ell} \otimes \mathbf{a}_{3\ell} )\mathbf{e}_{5\ell},
 \quad
    \frac{\partial \mathcal{L}}{\partial b_4} = \frac{1}{N}\sum_{\ell=1}^N \mathbf{z}_{4\ell}\mathbf{e}_{5\ell},
    \quad 
    \frac{\partial \mathcal{L}}{\partial W_5} = \frac{1}{N}\sum_{\ell=1}^N  \mathbf{a}_{4\ell} \mathbf{e}_{5\ell}
,\quad
    \frac{\partial \mathcal{L}}{\partial b_5} = \frac{1}{N}\sum_{\ell=1}^N\mathbf{e}_{5\ell}
$$
}

We then compute the Hessian of $\mathcal{L}$.
\small{
$$
\frac{\partial^2 \mathcal{L}}{\partial W_1^2} = \frac{1}{N}\sum_{\ell=1}^{N}(\mathbf{z}_{\ell}\otimes \mathbf{x}_{\ell}
) \otimes  (\mathbf{z}_{\ell}\otimes \mathbf{x}_{\ell}
)^{\top}, \quad \frac{\partial^2 \mathcal{L}}{ \partial b_1 \partial W_1} = \frac{1}{N}\sum_{\ell=1}^{N}(\mathbf{z}_{\ell}\otimes \mathbf{x}_{\ell}
) \otimes  \mathbf{z}_{\ell}^{\top}
$$
$$
\frac{\partial^2 \mathcal{L}}{ \partial W_2 \partial W_1}  = \frac{1}{N}\sum_{\ell=1}^N\left(\widetilde{\mathbf{z}}_{6\ell}^{\top}\otimes \mathbb{I}_{\mathbf{W}_1, \mathbf{b}_1, \mathbf{x}_{\ell}} \otimes \mathbf{x}_{\ell}\right) \mathbf{e}_{5\ell} + \left(\mathbf{z}_{\ell} \otimes \mathbf{x}_{\ell}\right)\left(\widetilde{\mathbf{z}}_{6\ell} \otimes \mathbf{a}_{1\ell}\right)^{\top}
$$
$$
\frac{\partial^2 \mathcal{L}}{ \partial b_2 \partial W_1} = \frac{1}{N}\sum_{\ell=1}^N\left(\mathbf{z}_{\ell} \otimes \mathbf{x}_{\ell}\right)\widetilde{\mathbf{z}}_{6\ell}^{\top}
$$
$$
\frac{\partial^2 \mathcal{L}}{ \partial W_3 \partial W_1} = \frac{1}{N}\sum_{\ell=1}^N\left(\widetilde{\mathbf{z}}_{5\ell}^{\top}\otimes \left(\mathbf{z}_{1\ell}\mathbb{I}_{\mathbf{W}_2, \mathbf{b}_2, \mathbf{a}_{1\ell}}\right) \otimes \mathbf{x}_{\ell}\right) \mathbf{e}_{5\ell} + \left(\mathbf{z}_{\ell} \otimes \mathbf{x}_{\ell}\right)\left(\widetilde{\mathbf{z}}_{5\ell} \otimes \mathbf{a}_{2\ell}\right)^{\top}
$$
$$
\frac{\partial^2 \mathcal{L}}{ \partial b_3 \partial W_1} = \frac{1}{N}\sum_{\ell=1}^N\left(\mathbf{z}_{\ell} \otimes \mathbf{x}_{\ell}\right)\widetilde{\mathbf{z}}_{5\ell}^{\top}
$$
$$
\frac{\partial^2 \mathcal{L}}{ \partial W_4 \partial W_1} = \frac{1}{N}\sum_{\ell=1}^N\left(\mathbf{z}_{4\ell}^{\top}\otimes \left(\widetilde{\mathbf{z}}_{7\ell}\mathbb{I}_{\mathbf{W}_3, \mathbf{b}_3, \mathbf{a}_{2\ell}}\right) \otimes \mathbf{x}_{\ell}\right) \mathbf{e}_{5\ell} + \left(\mathbf{z}_{\ell} \otimes \mathbf{x}_{\ell}\right)\left(\mathbf{z}_{4\ell} \otimes \mathbf{a}_{3\ell}\right)^{\top}
$$
$$
\frac{\partial^2 \mathcal{L}}{ \partial b_4 \partial W_1} = \frac{1}{N}\sum_{\ell=1}^N \left(\mathbf{z}_{\ell} \otimes \mathbf{x}_{\ell}\right)\mathbf{z}_{4\ell}^{\top}
,\quad
\frac{\partial^2 \mathcal{L}}{ \partial W_5 \partial W_1} = \frac{1}{N}\sum_{\ell=1}^N\left( \left(\widetilde{\mathbf{z}}_{8\ell}\mathbb{I}_{\mathbf{W}_4, \mathbf{b}_4, \mathbf{a}_{3\ell}}\right) \otimes \mathbf{x}_{\ell}\right) \mathbf{e}_{5\ell} + \left(\mathbf{z}_{\ell} \otimes \mathbf{x}_{\ell}\right) \mathbf{a}_{4\ell}^{\top}
$$
$$
\frac{\partial^2 \mathcal{L}}{ \partial b_5 \partial W_1} = \frac{1}{N}\sum_{\ell=1}^N \mathbf{z}_{\ell} \otimes \mathbf{x}_{\ell}, \quad \frac{\partial^2 \mathcal{L}}{\partial b_1^2} = \frac{1}{N}\sum_{\ell=1}^{N}\mathbf{z}_{\ell} \mathbf{z}_{\ell}^{\top}
$$
$$
\frac{\partial^2 \mathcal{L}}{ \partial W_2 \partial b_1}  = \frac{1}{N}\sum_{\ell=1}^N\left(\widetilde{\mathbf{z}}_{6\ell}^{\top}\otimes \mathbb{I}_{\mathbf{W}_1, \mathbf{b}_1, \mathbf{x}_{\ell}}\right) \mathbf{e}_{5\ell} + \mathbf{z}_{\ell}\left(\widetilde{\mathbf{z}}_{6\ell} \otimes \mathbf{a}_{1\ell}\right)^{\top}
$$
$$
\frac{\partial^2 \mathcal{L}}{ \partial b_2 \partial b_1}  = \frac{1}{N}\sum_{\ell=1}^N\mathbf{z}_{\ell}\widetilde{\mathbf{z}}_{6\ell}^{\top}
,\quad
\frac{\partial^2 \mathcal{L}}{ \partial W_3 \partial b_1}  = \frac{1}{N}\sum_{\ell=1}^N\left(\widetilde{\mathbf{z}}_{5\ell}^{\top}\otimes \left(\mathbf{z}_{1\ell}\mathbb{I}_{\mathbf{W}_2, \mathbf{b}_2, \mathbf{a}_{1\ell}} \right)\right) \mathbf{e}_{5\ell} + \mathbf{z}_{\ell}\left(\widetilde{\mathbf{z}}_{5\ell} \otimes \mathbf{a}_{2\ell}\right)^{\top}
$$
$$
\frac{\partial^2 \mathcal{L}}{ \partial b_3 \partial b_1}  = \frac{1}{N}\sum_{\ell=1}^N\mathbf{z}_{\ell}\widetilde{\mathbf{z}}_{5\ell}^{\top}
,\quad
\frac{\partial^2 \mathcal{L}}{ \partial W_4 \partial b_1}  = \frac{1}{N}\sum_{\ell=1}^N\left(\mathbf{z}_{4\ell}^{\top}\otimes \left(\widetilde{\mathbf{z}}_{7\ell}\mathbb{I}_{\mathbf{W}_3, \mathbf{b}_3, \mathbf{a}_{2\ell}} \right)\right) \mathbf{e}_{5\ell} + \mathbf{z}_{\ell}\left(\mathbf{z}_{4\ell} \otimes \mathbf{a}_{3\ell}\right)^{\top}
$$
$$
\frac{\partial^2 \mathcal{L}}{ \partial b_4 \partial b_1}  = \frac{1}{N}\sum_{\ell=1}^N\mathbf{z}_{\ell}\mathbf{z}_{4\ell}^{\top}, \quad
\frac{\partial^2 \mathcal{L}}{ \partial W_5 \partial b_1}  = \frac{1}{N}\sum_{\ell=1}^N\left(\widetilde{\mathbf{z}}_{8\ell}\mathbb{I}_{\mathbf{W}_4, \mathbf{b}_4, \mathbf{a}_{3\ell}} \right)\mathbf{e}_{5\ell} + \mathbf{z}_{\ell} \mathbf{a}_{4\ell}^{\top}
$$
$$
\frac{\partial^2 \mathcal{L}}{ \partial b_5 \partial b_1}  = \frac{1}{N}\sum_{\ell=1}^N\mathbf{z}_{\ell}
,\quad
\frac{\partial^2 \mathcal{L}}{ \partial W_2^2} = \frac{1}{N}\sum_{\ell=1}^{N}\left(\widetilde{\mathbf{z}}_{6\ell} \otimes \mathbf{a}_{1\ell}\right) \left(\widetilde{\mathbf{z}}_{6\ell} \otimes \mathbf{a}_{1\ell}\right)^{\top}
$$
$$
\frac{\partial^2 \mathcal{L}}{ \partial b_2\partial W_2} = \frac{1}{N}\sum_{\ell=1}^{N}\left(\widetilde{\mathbf{z}}_{6\ell} \otimes \mathbf{a}_{1\ell}\right)\widetilde{\mathbf{z}}_{6\ell}^{\top}
$$
$$
\frac{\partial^2 \mathcal{L}}{ \partial W_3\partial W_2} = \frac{1}{N}\sum_{\ell=1}^{N}\left(\widetilde{\mathbf{z}}_{5\ell}^{\top} \otimes \mathbb{I}_{\mathbf{W}_2, \mathbf{b}_2, \mathbf{a}_{1\ell}} \otimes \mathbf{a}_{1\ell}\right) \mathbf{e}_{5\ell}
+ (\widetilde{\mathbf{z}}_{6\ell} \otimes \mathbf{a}_{1\ell} )( \widetilde{\mathbf{z}}_{5\ell} \otimes \mathbf{a}_{2\ell})^{\top}
$$
$$
\frac{\partial^2 \mathcal{L}}{ \partial b_3\partial W_2} = \frac{1}{N}\sum_{\ell=1}^{N}(\widetilde{\mathbf{z}}_{6\ell} \otimes \mathbf{a}_{1\ell} )\widetilde{\mathbf{z}}_{5\ell}^{\top}
$$
$$
\frac{\partial^2 \mathcal{L}}{ \partial W_4\partial W_2} = \frac{1}{N}\sum_{\ell=1}^{N}\left( \mathbf{z}_{4\ell}^{\top} \otimes \left(\mathbf{z}_{3\ell}\mathbb{I}_{\mathbf{W}_3, \mathbf{b}_3, \mathbf{a}_{2\ell}} \right)\otimes \mathbf{a}_{1\ell}\right) \mathbf{e}_{5\ell}
+ (\widetilde{\mathbf{z}}_{6\ell} \otimes \mathbf{a}_{1\ell} ) \left(\mathbf{z}_{4\ell} \otimes \mathbf{a}_{3\ell}\right)^{\top}
$$
$$
\frac{\partial^2 \mathcal{L}}{ \partial b_4\partial W_2} = \frac{1}{N}\sum_{\ell=1}^{N}(\widetilde{\mathbf{z}}_{6\ell} \otimes \mathbf{a}_{1\ell} ) \mathbf{z}_{4\ell}^{\top}
$$
$$
\frac{\partial^2 \mathcal{L}}{ \partial W_5\partial W_2} = \frac{1}{N}\sum_{\ell=1}^{N}\left( \left(\widetilde{\mathbf{z}}_{9\ell}\mathbb{I}_{\mathbf{W}_4, \mathbf{b}_4, \mathbf{a}_{3\ell}} \right)\otimes \mathbf{a}_{1\ell}\right) \mathbf{e}_{5\ell}
+ (\widetilde{\mathbf{z}}_{6\ell} \otimes \mathbf{a}_{1\ell} ) \mathbf{a}_{4\ell}^{\top}
$$
$$
\frac{\partial^2 \mathcal{L}}{ \partial b_5\partial W_2} = \frac{1}{N}\sum_{\ell=1}^{N} \widetilde{\mathbf{z}}_{6\ell} \otimes \mathbf{a}_{1\ell}, \quad
\frac{\partial^2 \mathcal{L}}{ \partial b_2^2} = \frac{1}{N}\sum_{\ell=1}^{N}\widetilde{\mathbf{z}}_{6\ell}\widetilde{\mathbf{z}}_{6\ell}^{\top}
$$
$$
\frac{\partial^2 \mathcal{L}}{ \partial W_3\partial b_2} = \frac{1}{N}\sum_{\ell=1}^{N}\left(\widetilde{\mathbf{z}}_{5\ell}^{\top} \otimes \mathbb{I}_{\mathbf{W}_2, \mathbf{b}_2, \mathbf{a}_{1\ell}}\right) \mathbf{e}_{5\ell}
+\widetilde{\mathbf{z}}_{6\ell}( \widetilde{\mathbf{z}}_{5\ell} \otimes \mathbf{a}_{2\ell})^{\top},\quad
\frac{\partial^2 \mathcal{L}}{ \partial b_3\partial b_2} = \frac{1}{N}\sum_{\ell=1}^{N}\widetilde{\mathbf{z}}_{6\ell} \widetilde{\mathbf{z}}_{5\ell}^{\top}
$$
$$
\frac{\partial^2 \mathcal{L}}{ \partial W_4\partial b_2} = \frac{1}{N}\sum_{\ell=1}^{N}\left( \mathbf{z}_{4\ell}^{\top} \otimes \left(\mathbf{z}_{3\ell}\mathbb{I}_{\mathbf{W}_3, \mathbf{b}_3, \mathbf{a}_{2\ell}} \right)\right) \mathbf{e}_{5\ell}
+ \widetilde{\mathbf{z}}_{6\ell} \left(\mathbf{z}_{4\ell} \otimes \mathbf{a}_{3\ell}\right)^{\top},\quad
\frac{\partial^2 \mathcal{L}}{ \partial b_4\partial b_2} = \frac{1}{N}\sum_{\ell=1}^{N}\widetilde{\mathbf{z}}_{6\ell}\mathbf{z}_{4\ell}^{\top}
$$
$$
\frac{\partial^2 \mathcal{L}}{ \partial W_5\partial b_2} = \frac{1}{N}\sum_{\ell=1}^{N} \left(\widetilde{\mathbf{z}}_{9\ell}\mathbb{I}_{\mathbf{W}_4, \mathbf{b}_4, \mathbf{a}_{4\ell}} \right)\mathbf{e}_{5\ell}
+ \widetilde{\mathbf{z}}_{6\ell}\mathbf{a}_{4\ell}^{\top}
,\quad
\frac{\partial^2 \mathcal{L}}{ \partial b_5\partial b_2} = \frac{1}{N}\sum_{\ell=1}^{N} \widetilde{\mathbf{z}}_{6\ell}
$$
$$
\frac{\partial^2 \mathcal{L}}{\partial W_3^2}=\frac{1}{N}\sum_{\ell=1}^{N}\left(\widetilde{\mathbf{z}}_{5\ell} \otimes \mathbf{a}_{2\ell} \right) \otimes \left(\widetilde{\mathbf{z}}_{5\ell} \otimes \mathbf{a}_{2\ell} \right)^{\top}
,\quad
\frac{\partial^2 \mathcal{L}}{\partial b_3 \partial W_3} =  \frac{1}{N}\sum_{\ell=1}^{N}\left(\widetilde{\mathbf{z}}_{5\ell} \otimes \mathbf{a}_{2\ell} \right) \otimes \widetilde{\mathbf{z}}_{5\ell}^{\top}
$$
$$
\frac{\partial^2 \mathcal{L}}{\partial W_4 \partial W_3} =  \frac{1}{N}\sum_{\ell=1}^{N}\left( \mathbf{z}_{4\ell}^{\top} \otimes \mathbb{I}_{\mathbf{W}_3, \mathbf{b}_3, \mathbf{a}_{2\ell}} \otimes \mathbf{a}_{2\ell}\right) \mathbf{e}_{5\ell}
+ \left(\widetilde{\mathbf{z}}_{5\ell} \otimes \mathbf{a}_{2\ell} \right)(  \mathbf{z}_{4\ell} \otimes \mathbf{a}_{3\ell})^{\top}
$$
$$
\frac{\partial^2 \mathcal{L}}{\partial b_4 \partial W_3} =  \frac{1}{N}\sum_{\ell=1}^{N} (\widetilde{\mathbf{z}}_{5\ell} \otimes \mathbf{a}_{2\ell} )\mathbf{z}_{4\ell}^{\top}
,\quad
\frac{\partial^2 \mathcal{L}}{\partial W_5 \partial W_3} =  \frac{1}{N}\sum_{\ell=1}^{N}\left( \left(\mathbf{z}_{3\ell}\mathbb{I}_{\mathbf{W}_4, \mathbf{b}_4, \mathbf{a}_{3\ell}}\right) \otimes \mathbf{a}_{2\ell}\right) \mathbf{e}_{5\ell}
+ \left(\widetilde{\mathbf{z}}_{5\ell} \otimes \mathbf{a}_{2\ell} \right) \mathbf{a}_{4\ell}^{\top}
$$
$$
\frac{\partial^2 \mathcal{L}}{\partial b_5 \partial W_3} =  \frac{1}{N}\sum_{\ell=1}^{N}\widetilde{\mathbf{z}}_{5\ell} \otimes \mathbf{a}_{2\ell} 
,\quad
\frac{\partial^2 \mathcal{L}}{\partial b_3^2}=\frac{1}{N}\sum_{\ell=1}^{N}\widetilde{\mathbf{z}}_{5\ell}\widetilde{\mathbf{z}}_{5\ell}^{\top}
$$
$$
\frac{\partial^2 \mathcal{L}}{ \partial W_4\partial b_3} = \frac{1}{N}\sum_{\ell=1}^{N}\left( \mathbf{z}_{4\ell}^{\top} \otimes \mathbb{I}_{\mathbf{W}_3, \mathbf{b}_3, \mathbf{a}_{2\ell}}\right) \mathbf{e}_{5\ell}
+ \widetilde{\mathbf{z}}_{5\ell}(  \mathbf{z}_{4\ell} \otimes \mathbf{a}_{3\ell})^{\top}
,\quad
\frac{\partial^2 \mathcal{L}}{ \partial b_4\partial b_3} = \frac{1}{N}\sum_{\ell=1}^{N}\widetilde{\mathbf{z}}_{5\ell}\mathbf{z}_{4\ell} ^{\top}
$$
$$
\frac{\partial^2 \mathcal{L}}{ \partial W_5\partial b_3} = \frac{1}{N}\sum_{\ell=1}^{N} \left(\mathbf{z}_{3\ell}\mathbb{I}_{\mathbf{W}_4, \mathbf{b}_4, \mathbf{a}_{3\ell}}\right)  \mathbf{e}_{5\ell}
+ \widetilde{\mathbf{z}}_{5\ell}  \mathbf{a}_{4\ell}^{\top}, \quad \frac{\partial^2 \mathcal{L}}{ \partial b_5\partial b_3} = \frac{1}{N}\sum_{\ell=1}^{N} \widetilde{\mathbf{z}}_{5\ell} 
$$
$$
\frac{\partial^2 \mathcal{L}}{\partial W_4^2}=\frac{1}{N}\sum_{\ell=1}^{N}\left(\mathbf{z}_{4\ell} \otimes \mathbf{a}_{3\ell}\right) \left(\mathbf{z}_{4\ell} \otimes \mathbf{a}_{3\ell}\right)^{\top}, \quad \frac{\partial^2 \mathcal{L}}{\partial b_4 \partial W_4} = \frac{1}{N}\sum_{\ell=1}^{N}\left(\mathbf{z}_{4\ell} \otimes \mathbf{a}_{3\ell}\right) \mathbf{z}_{4\ell}^{\top}
$$
$$
\frac{\partial^2 \mathcal{L}}{\partial W_5\partial W_4}=\frac{1}{N}\sum_{\ell=1}^{N}\left( \mathbb{I}_{\mathbf{W}_4, \mathbf{b}_3, \mathbf{a}_{3\ell}} \otimes \mathbf{a}_{3\ell} \right)\mathbf{e}_{5,\ell}+\left(\mathbf{z}_{4\ell}\otimes \mathbf{a}_{3\ell} \right) \mathbf{a}_{4\ell}^{\top}, \quad \frac{\partial^2 \mathcal{L}}{\partial b_5\partial W_4}=\frac{1}{N}\sum_{\ell=1}^{N}\mathbf{z}_{4\ell}\otimes \mathbf{a}_{3\ell} 
,\quad
\frac{\partial^2 \mathcal{L}}{\partial b_4^2}=\frac{1}{N}\sum_{\ell=1}^{N}\mathbf{z}_{4\ell}\mathbf{z}_{4\ell}^{\top}
$$
$$
\frac{\partial^2 \mathcal{L}}{\partial W_5\partial b_4}=\frac{1}{N}\sum_{\ell=1}^{N}\left( \mathbb{I}_{\mathbf{W}_4, \mathbf{b}_3, \mathbf{a}_{3\ell}} \right)\mathbf{e}_{5,\ell}+\mathbf{z}_{4\ell}\mathbf{a}_{4\ell}^{\top}, \quad \frac{\partial^2 \mathcal{L}}{\partial b_5\partial b_4}=\frac{1}{N}\sum_{\ell=1}^{N}\mathbf{z}_{4\ell}
$$
$$
\frac{\partial^2 \mathcal{L}}{\partial W_5^2} = \frac{1}{N}\sum_{\ell=1}^{N}\mathbf{a}_{4\ell}\mathbf{a}_{4\ell}^{\top}
,\quad
\frac{\partial^2 \mathcal{L}}{\partial b_5 \partial W_5} = \frac{1}{N}\sum_{\ell=1}^{N}\mathbf{a}_{4\ell}, \quad
\frac{\partial^2 \mathcal{L}}{\partial b_5^5}=1.
$$}

\textbf{Hessian computation of MSDL}

We consider MSDL with two subnetworks and each with a single hidden layer. The network function is
\begin{equation}\label{eq: MSDL-1}
\mathcal{N}_2^{(2)}\left(\left\{\mathbf{W}_{js} , \mathbf{b}_{js}\right\}_{j,s =1}^{2, 2}; \left\{\alpha_s\right\}_{s=1}^2, \mathbf{x}\right):= \sum_{s=1}^2 \mathbf{W}_{2s}^{\top}\sigma(\mathbf{W}_{1s}^{\top}\alpha_s \mathbf{x} + \mathbf{b}_{1s}) + \mathbf{b}_{2s}.
\end{equation}
The vectorized form of parameter is $W = [W_1^{\top}, W_2^{\top}]^{\top}$ where 
$$
W_1 = [W_{11}^{\top}, b_{11}^{\top}, W_{21}^{\top}, b_{21}^{\top}]^{\top}, \quad W_2 =  [W_{12}^{\top}, b_{12}^{\top}, W_{22}^{\top}, b_{22}^{\top}]^{\top}.
$$
We define
$$
\mathbf{Z}^1 := \mathbb{I}_{\mathbf{W}_{11}, \mathbf{b}_{11}, \alpha_1\mathbf{x}}\mathbf{W}_{21}, \quad \mathbf{Z}^2 := \mathbb{I}_{\mathbf{W}_{12}, \mathbf{b}_{12}, \alpha_2\mathbf{x}}\mathbf{W}_{22}
$$
and 
$$
\mathbf{a}_1^1 := \sigma(\mathbf{W}_{11}^{\top}\alpha_1 \mathbf{x} + \mathbf{b}_{11}), \quad \mathbf{a}_1^2 := \sigma(\mathbf{W}_{12}^{\top}\alpha_2 \mathbf{x} + \mathbf{b}_{12}).
$$

\begin{lemma}
    Let the network function be defined by \eqref{eq: MSDL-1} and the loss function $\mathcal{L}$ be the corresponding loss function for this network. Then, we have that
$$
    \frac{\partial \mathcal{L}}{\partial W_{11}} = \frac{1}{N}\sum_{\ell=1}^N (\mathbf{z}_{\ell}^1 \otimes \alpha_1\mathbf{x}_{\ell} )\mathbf{e}_{2\ell}, 
    \quad
    \frac{\partial \mathcal{L}}{\partial b_{11}} = \frac{1}{N}\sum_{\ell=1}^N \mathbf{z}_{\ell}^1  \mathbf{e}_{2\ell},
    $$
    $$
    \frac{\partial \mathcal{L}}{\partial W_{21}} = \frac{1}{N}\sum_{\ell=1}^N  \mathbf{a}^1_{1, \ell} \mathbf{e}_{2\ell}, \quad
    \frac{\partial \mathcal{L}}{\partial b_{21}} = \frac{1}{N}\sum_{\ell=1}^N   \mathbf{e}_{2\ell}.
    $$
$$
    \frac{\partial \mathcal{L}}{\partial W_{12}} = \frac{1}{N}\sum_{\ell=1}^N (\mathbf{z}_{\ell}^2 \otimes \alpha_2\mathbf{x}_{\ell} )\mathbf{e}_{2\ell}, 
    \quad
    \frac{\partial \mathcal{L}}{\partial b_{12}} = \frac{1}{N}\sum_{\ell=1}^N \mathbf{z}_{\ell}^2  \mathbf{e}_{2\ell},
    $$
    $$
    \frac{\partial \mathcal{L}}{\partial W_{22}} = \frac{1}{N}\sum_{\ell=1}^N  \mathbf{a}^2_{1, \ell} \mathbf{e}_{2\ell}, \quad
    \frac{\partial \mathcal{L}}{\partial b_{22}} = \frac{1}{N}\sum_{\ell=1}^N   \mathbf{e}_{2\ell}.
    $$
\end{lemma}
The hessian of $\mathcal{L}$ at $W$ is 
$$
\mathbf{H}_{\mathcal{L}}(W) = \begin{bmatrix}
    \frac{\partial^2 \mathcal{L}}{\partial W_1^2} & \frac{\partial^2 \mathcal{L}}{\partial W_2 \partial W_1} \\ 
     \frac{\partial^2 \mathcal{L}}{\partial W_1 \partial W_2} & \frac{\partial^2 \mathcal{L}}{\partial W_2^2}
\end{bmatrix}.
$$
The form $\frac{\partial^2 \mathcal{L}}{\partial W_1^2}$ and $\frac{\partial^2 \mathcal{L}}{\partial W_2^2}$ are same with Lemma \ref{lemma: hessian}. We focus on $\frac{\partial^2 \mathcal{L}}{\partial W_2\partial W_1}$.
$$
\frac{\partial^2 \mathcal{L}}{ \partial W_{12} \partial W_{11}} = \frac{1}{N}\sum_{\ell=1}^{N}(\mathbf{z}_{\ell}^1 \otimes \alpha_1\mathbf{x}_{\ell} ) (\mathbf{z}_{\ell}^2 \otimes \alpha_2\mathbf{x}_{\ell} )^{\top}, \quad \frac{\partial^2 \mathcal{L}}{ \partial b_{12} \partial W_{11}} = \frac{1}{N}\sum_{\ell=1}^{N}(\mathbf{z}_{\ell}^1 \otimes \alpha_1\mathbf{x}_{\ell} ) (\mathbf{z}_{\ell}^2 )^{\top},
$$
$$
\frac{\partial^2 \mathcal{L}}{ \partial W_{22} \partial W_{11}} = \frac{1}{N}\sum_{\ell=1}^{N}(\mathbf{z}_{\ell}^1 \otimes \alpha_1\mathbf{x}_{\ell} ) ( \mathbf{a}_{1\ell}^2 )^{\top},
\quad \frac{\partial^2 \mathcal{L}}{ \partial b_{22} \partial W_{11}} = \frac{1}{N}\sum_{\ell=1}^{N}\mathbf{z}_{\ell}^1 \otimes \alpha_1\mathbf{x}_{\ell},
$$

$$
\frac{\partial^2 \mathcal{L}}{ \partial W_{12} \partial b_{11}} = \frac{1}{N}\sum_{\ell=1}^{N} \mathbf{z}_{\ell}^1 (\mathbf{z}_{\ell}^2 \otimes \alpha_2\mathbf{x}_{\ell} )^{\top},
\quad \frac{\partial^2 \mathcal{L}}{ \partial b_{12} \partial b_{11}} = \frac{1}{N}\sum_{\ell=1}^{N}\mathbf{z}_{\ell}^1  (\mathbf{z}_{\ell}^2 )^{\top},
$$
$$
\frac{\partial^2 \mathcal{L}}{ \partial W_{22} \partial W_{11}} = \frac{1}{N}\sum_{\ell=1}^{N}\mathbf{z}_{\ell}^1  ( \mathbf{a}_{1\ell}^2 )^{\top}, \quad \frac{\partial^2 \mathcal{L}}{ \partial b_{22} \partial W_{11}} = \frac{1}{N}\sum_{\ell=1}^{N}\mathbf{z}_{\ell}^1,
$$

$$
\frac{\partial^2 \mathcal{L}}{ \partial W_{12} \partial W_{21}} = \frac{1}{N}\sum_{\ell=1}^{N} \mathbf{a}_{1\ell}^1 (\mathbf{z}_{\ell}^2 \otimes \alpha_2\mathbf{x}_{\ell} )^{\top}, \quad \frac{\partial^2 \mathcal{L}}{ \partial b_{12} \partial W_{21}} = \frac{1}{N}\sum_{\ell=1}^{N}\mathbf{a}_{1\ell}^1 (\mathbf{z}_{\ell}^2 )^{\top},
$$
$$
\frac{\partial^2 \mathcal{L}}{ \partial W_{22} \partial W_{21}} = \frac{1}{N}\sum_{\ell=1}^{N}\mathbf{a}_{1\ell}^1 ( \mathbf{a}_{1\ell}^2 )^{\top},
\quad \frac{\partial^2 \mathcal{L}}{ \partial b_{22} \partial W_{21}} = \frac{1}{N}\sum_{\ell=1}^{N}\mathbf{a}_{1\ell}^1,
$$

$$
\frac{\partial^2 \mathcal{L}}{ \partial W_{12} \partial b_{21}} = \frac{1}{N}\sum_{\ell=1}^{N}(\mathbf{z}_{\ell}^2 \otimes \alpha_2\mathbf{x}_{\ell} )^{\top}, \quad \frac{\partial^2 \mathcal{L}}{ \partial b_{12} \partial b_{21}} = \frac{1}{N}\sum_{\ell=1}^{N}(\mathbf{z}_{\ell}^2 )^{\top},
$$
$$
\frac{\partial^2 \mathcal{L}}{ \partial W_{22} \partial b_{21}} = \frac{1}{N}\sum_{\ell=1}^{N}( \mathbf{a}_{1\ell}^2 )^{\top},
\quad \frac{\partial^2 \mathcal{L}}{ \partial b_{22} \partial b_{21}} = 1.
$$

\textbf{Hessian computation of image denoising problem}

To compute the hessian matrix of $\mathcal{L}$, we write $\mathcal{L}$ in the elementwise form
\small{
\begin{equation}
\begin{aligned}
&\mathcal{L}(\Theta, \mathbf{u}) = \frac{1}{2}\sum_{s, t=1}^{n}\left(  \mathcal{N}_D(\Theta; \mathbf{x}_{st}) - \mathbf{\hat{f}}_{st}\right)^2 + \lambda \sum_{s, t=1}^n(|\mathbf{u}^1_{st}|+|\mathbf{u}^2_{st}|) + \\
&\frac{\beta}{2}\sum_{s,t=1}^{n}\left(\left( \mathcal{N}_D(\Theta; \mathbf{x}_{st}) - \mathcal{N}_D(\Theta; \mathbf{x}_{s(t-1)})\right) - \mathbf{u}^1_{st}\right)^2 + \left(\left( \mathcal{N}_D(\Theta; \mathbf{x}_{st}) - \mathcal{N}_D(\Theta; \mathbf{x}_{(s-1)t})\right) - \mathbf{u}^2_{st}\right)^2
\end{aligned}
\end{equation}}
where we set $\mathcal{N}_D(\Theta; \mathbf{x}_{s0}) = \mathcal{N}_D(\Theta; \mathbf{x}_{s1})$ and $\mathcal{N}_D(\Theta; \mathbf{x}_{0t}) = \mathcal{N}_D(\Theta; \mathbf{x}_{1t})$ for $s, t \in \mathbb{N}_n$.

\textbf{One hidden layer}

We consider the case with one hidden layer and the network function is given by \eqref{layer 2 network function}. Let 
\small{
$$
\mathbf{z}_{st} =  
\mathbb{I}_{\mathbf{W}_1, \mathbf{b}_1, \mathbf{x}_{st}}\mathbf{W}_2, \quad \mathbf{a}_{1st} = \sigma(\mathbf{W}_1^{\top} \mathbf{x}_{st} + \mathbf{b}_1)
,\quad
\mathbf{e}_{1st} = \mathcal{N}_{2}\left(\left\{\mathbf{W}_j, \mathbf{b}_j\right\}_{j=1}^2; \mathbf{x}_{st}\right) - \mathbf{\hat{f}}_{st} 
$$
$$
\mathbf{e}_{2st} = \mathcal{N}_{2}\left(\left\{\mathbf{W}_j, \mathbf{b}_j\right\}_{j=1}^2; \mathbf{x}_{st}\right) - \mathcal{N}_{2}\left(\left\{\mathbf{W}_j, \mathbf{b}_j\right\}_{j=1}^2; \mathbf{x}_{s(t-1)}\right) - \mathbf{u}_{st}^1
$$
$$
\mathbf{e}_{3st} = \mathcal{N}_{2}\left(\left\{\mathbf{W}_j, \mathbf{b}_j\right\}_{j=1}^2; \mathbf{x}_{st}\right) - \mathcal{N}_{2}\left(\left\{\mathbf{W}_j, \mathbf{b}_j\right\}_{j=1}^2; \mathbf{x}_{(s-1)t}\right) - \mathbf{u}_{st}^2.
$$}
We first consider the gradient of $\mathcal{L}$.
\small{
$$
\begin{aligned}
    \frac{\partial \mathcal{L}}{\partial W_1} = \sum_{s,t=1}^n &(\mathbf{z}_{st} \otimes \mathbf{x}_{st} )\mathbf{e}_{1st}+\big( (\mathbf{z}_{st} \otimes \mathbf{x}_{st} - \mathbf{z}_{s(t-1)} \otimes \mathbf{x}_{s(t-1)} )\mathbf{e}_{2st} +\\
    &(\mathbf{z}_{st} \otimes \mathbf{x}_{st} - \mathbf{z}_{(s-1)t} \otimes \mathbf{x}_{(s-1)t} )\mathbf{e}_{3st}\big)
\end{aligned}
   $$
$$
\begin{aligned}
    \frac{\partial \mathcal{L}}{\partial b_1} = \sum_{s,t=1}^n &\mathbf{z}_{st}\mathbf{e}_{1st}+\beta\left( (\mathbf{z}_{st}  - \mathbf{z}_{s(t-1)} )\mathbf{e}_{2st} +(\mathbf{z}_{st}- \mathbf{z}_{(s-1)t} )\mathbf{e}_{3st}\right)
\end{aligned}
   $$
    $$
    \frac{\partial \mathcal{L}}{\partial W_2} = \sum_{s, t=1}^n  \mathbf{a}_{1st} \mathbf{e}_{1st} + \beta \left((\mathbf{a}_{1st} - \mathbf{a}_{1s(t-1)})\mathbf{e}_{2st} + (\mathbf{a}_{1st} - \mathbf{a}_{1(s-1)t})\mathbf{e}_{3st}\right), \quad 
    \frac{\partial \mathcal{L}}{\partial b_2} = \sum_{s, t=1}^n   \mathbf{e}_{1st}.
    $$
    }

We then consider Hessian matrix of $\mathcal{L}$.

\small{
\begin{align*}
    \frac{\partial^2 \mathcal{L}}{\partial W_1^2} =& \sum_{s, t=1}^n(\mathbf{z}_{st} \otimes \mathbf{x}_{st} )(\mathbf{z}_{st} \otimes \mathbf{x}_{st} )^{\top}&
    \\ 
    & +\beta \big(
    (\mathbf{z}_{st} \otimes \mathbf{x}_{st} - \mathbf{z}_{s(t-1)} \otimes \mathbf{x}_{s(t-1)} )(\mathbf{z}_{st} \otimes \mathbf{x}_{st} - \mathbf{z}_{s(t-1)} \otimes \mathbf{x}_{s(t-1)} )^{\top}  \\
    &+(\mathbf{z}_{st} \otimes \mathbf{x}_{st} - \mathbf{z}_{(s-1)t} \otimes \mathbf{x}_{(s-1)t} )(\mathbf{z}_{st} \otimes \mathbf{x}_{st} - \mathbf{z}_{(s-1)t} \otimes \mathbf{x}_{(s-1)t} )^{\top}
    \big)
\end{align*}

\begin{align*}
    \frac{\partial^2 \mathcal{L}}{\partial b_1\partial W_1}=& \sum_{s, t=1}^n(\mathbf{z}_{st} \otimes \mathbf{x}_{st} )\mathbf{z}_{st}^{\top} +\beta \big(
    (\mathbf{z}_{st} \otimes \mathbf{x}_{st} - \mathbf{z}_{s(t-1)} \otimes \mathbf{x}_{s(t-1)} )(\mathbf{z}_{st} - \mathbf{z}_{s(t-1)})^{\top}  \\
    &+(\mathbf{z}_{st} \otimes \mathbf{x}_{st} - \mathbf{z}_{(s-1)t} \otimes \mathbf{x}_{(s-1)t} )(\mathbf{z}_{st}  - \mathbf{z}_{(s-1)t} )^{\top}
    \big)
\end{align*}

\begin{align*}
    \frac{\partial^2 \mathcal{L}}{\partial W_2 \partial W_1} = \sum_{s, t=1}^n& \left(\mathbb{I}_{\mathbf{W}_1, \mathbf{b}_1, \mathbf{x}_{st}} \otimes \mathbf{x}_{st}\right)\mathbf{e}_{1st}+
 (\mathbf{z}_{st} \otimes \mathbf{x}_{st} )\mathbf{a}_{1st}^{\top}\\
 & +\beta \big((\mathbb{I}_{\mathbf{W}_1, \mathbf{b}_1, \mathbf{x}_{st}} \otimes \mathbf{x}_{st} - \mathbb{I}_{\mathbf{W}_1, \mathbf{b}_1, \mathbf{x}_{s(t-1)}} \otimes \mathbf{x}_{s(t-1)} )\mathbf{e}_{2st} \\
    &+
    (\mathbf{z}_{st} \otimes \mathbf{x}_{st} - \mathbf{z}_{s(t-1)} \otimes \mathbf{x}_{s(t-1)} )(\mathbf{a}_{1st} - \mathbf{a}_{1s(t-1)})^{\top}  \\
    &+(\mathbb{I}_{\mathbf{W}_1, \mathbf{b}_1, \mathbf{x}_{st}} \otimes \mathbf{x}_{st} - \mathbb{I}_{\mathbf{W}_1, \mathbf{b}_1, \mathbf{x}_{(s-1)t}} \otimes \mathbf{x}_{(s-1)t} )\mathbf{e}_{3st} \\
    &+
    (\mathbf{z}_{st} \otimes \mathbf{x}_{st} - \mathbf{z}_{(s-1)t} \otimes \mathbf{x}_{(s-1)t} )(\mathbf{a}_{1st} - \mathbf{a}_{1(s-1)t})^{\top}
    \big)
\end{align*}

\begin{align*}
    \frac{\partial^2 \mathcal{L}}{\partial b_2 \partial W_1} = \sum_{s,t=1}^n (\mathbf{z}_{st} \otimes \mathbf{x}_{st} )
\end{align*}

\begin{align*}
    \frac{\partial^2 \mathcal{L}}{\partial b_1^2}=& \sum_{s, t=1}^n\mathbf{z}_{st} \mathbf{z}_{st}^{\top} +\beta \big(
    (\mathbf{z}_{st}  - \mathbf{z}_{s(t-1)} )(\mathbf{z}_{st} - \mathbf{z}_{s(t-1)})^{\top} + (\mathbf{z}_{st}- \mathbf{z}_{(s-1)t} )(\mathbf{z}_{st}  - \mathbf{z}_{(s-1)t} )^{\top}
    \big)
\end{align*}

$$
\begin{aligned}
    \frac{\partial^2 \mathcal{L}}{\partial W_2 \partial b_1} = &\sum_{s,t=1}^n \mathbb{I}_{\mathbf{W}_1, \mathbf{b}_1, \mathbf{x}_{st}}\mathbf{e}_{1st} + \mathbf{z}_{st}\mathbf{a}_{1st}^{\top}\\
    &+\beta\big( (\mathbb{I}_{\mathbf{W}_1, \mathbf{b}_1, \mathbf{x}_{st}}  - \mathbb{I}_{\mathbf{W}_1, \mathbf{b}_1, \mathbf{x}_{s(t-1)}} )\mathbf{e}_{2st} + (\mathbf{z}_{st}  - \mathbf{z}_{s(t-1)} )(\mathbf{a}_{1st} - \mathbf{a}_{1s(t-1)})^{\top} \\
    &+(\mathbb{I}_{\mathbf{W}_1, \mathbf{b}_1, \mathbf{x}_{st}}- \mathbb{I}_{\mathbf{W}_1, \mathbf{b}_1, \mathbf{x}_{(s-1)t}} )\mathbf{e}_{3st} + (\mathbf{z}_{st}- \mathbf{z}_{(s-1)t} )(\mathbf{a}_{1st} - \mathbf{a}_{1(s-1)t})^{\top}\big)
\end{aligned}
$$

$$
\frac{\partial^2 \mathcal{L}}{\partial b_2 \partial b_1} = \sum_{s,t=1}^n \mathbf{z}_{st}
$$

\begin{align*}
\frac{\partial^2 \mathcal{L}}{\partial W_2^2} = \sum_{s, t=1}^n & \mathbf{a}_{1st} \mathbf{a}_{1st}^{\top} + \beta \big((\mathbf{a}_{1st} - \mathbf{a}_{1s(t-1)})(\mathbf{a}_{1st} - \mathbf{a}_{1s(t-1)})^{\top} + \\
&(\mathbf{a}_{1st} - \mathbf{a}_{1(s-1)t})(\mathbf{a}_{1st} - \mathbf{a}_{1(s-1)t})^{\top}\big)
\end{align*}

$$
    \frac{\partial^2 \mathcal{L}}{\partial b_2 \partial W_2} = \sum_{s, t=1}^n  \mathbf{a}_{1st}, \quad 
    \frac{\partial^2 \mathcal{L}}{\partial b_2^2} = n^2.
    $$
}

\textbf{Four hidden layers}

We consider the case with four hidden layers and the network function is given by \eqref{layer 5 network function}.

Let
$$
\mathbf{z}_{st} := \mathbb{I}_{\mathbf{W}_1, \mathbf{b}_1, \mathbf{x}_{st}} 
 \mathbf{W}_2 \mathbb{I}_{\mathbf{W}_2, \mathbf{b}_2, \mathbf{a}_{1st}}\mathbf{W}_3 \mathbb{I}_{\mathbf{W}_3, \mathbf{b}_3, \mathbf{a}_{2st}}\mathbf{W}_4\mathbb{I}_{\mathbf{W}_4, \mathbf{b}_4, \mathbf{a}_{3st}}\mathbf{W}_5
 $$
 and
 $$
\mathbf{z}_{1st}:=\mathbb{I}_{\mathbf{W}_1, \mathbf{b}_1, \mathbf{x}_{st}}\mathbf{W}_2,\quad \mathbf{z}_{2st}:=\mathbb{I}_{\mathbf{W}_2, \mathbf{b}_2, \mathbf{a}_{1st}}\mathbf{W}_3
,\quad
\mathbf{z}_{3st}:=\mathbb{I}_{\mathbf{W}_3, \mathbf{b}_3, \mathbf{a}_{2st}}\mathbf{W}_4, \quad 
\mathbf{z}_{4st}:=\mathbb{I}_{\mathbf{W}_4, \mathbf{b}_4, \mathbf{a}_{3st}}\mathbf{W}_5
$$
and 
$$
\widetilde{\mathbf{z}}_{5st} = \mathbf{z}_{3st}\mathbf{z}_{4st}, \quad \widetilde{\mathbf{z}}_{6st} = \mathbf{z}_{2st}\mathbf{z}_{3st}\mathbf{z}_{4st}, \quad \widetilde{\mathbf{z}}_{7st} = \mathbf{z}_{1st}\mathbf{z}_{2st}, \quad \widetilde{\mathbf{z}}_{8st} = \mathbf{z}_{1st}\mathbf{z}_{2st}\mathbf{z}_{3st}, \quad \widetilde{\mathbf{z}}_{9st} = \mathbf{z}_{2st}\mathbf{z}_{3st}.
$$
\small{
$$
\mathbf{e}_{1st} = \mathcal{N}_{4}\left(\left\{\mathbf{W}_j, \mathbf{b}_j\right\}_{j=1}^4; \mathbf{x}_{st}\right) - \mathbf{\hat{f}}_{st} 
$$
$$
\mathbf{e}_{2st} = \mathcal{N}_{4}\left(\left\{\mathbf{W}_j, \mathbf{b}_j\right\}_{j=1}^4; \mathbf{x}_{st}\right) - \mathcal{N}_{4}\left(\left\{\mathbf{W}_j, \mathbf{b}_j\right\}_{j=1}^4; \mathbf{x}_{s(t-1)}\right) - \mathbf{u}_{st}^1
$$
$$
\mathbf{e}_{3st} = \mathcal{N}_{4}\left(\left\{\mathbf{W}_j, \mathbf{b}_j\right\}_{j=1}^4; \mathbf{x}_{st}\right) - \mathcal{N}_{4}\left(\left\{\mathbf{W}_j, \mathbf{b}_j\right\}_{j=1}^4; \mathbf{x}_{(s-1)t}\right) - \mathbf{u}_{st}^2.
$$
}

We first compute the gradient of $\mathcal{L}$.
\small{
$$
\begin{aligned}
    \frac{\partial \mathcal{L}}{\partial W_1} = \sum_{s,t=1}^n &(\mathbf{z}_{st} \otimes \mathbf{x}_{st} )\mathbf{e}_{1st}+\beta\big( (\mathbf{z}_{st} \otimes \mathbf{x}_{st} - \mathbf{z}_{s(t-1)} \otimes \mathbf{x}_{s(t-1)} )\mathbf{e}_{2st} +\\
    &(\mathbf{z}_{st} \otimes \mathbf{x}_{st} - \mathbf{z}_{(s-1)t} \otimes \mathbf{x}_{(s-1)t} )\mathbf{e}_{3st}\big)
\end{aligned}
$$
$$
\begin{aligned}
    \frac{\partial \mathcal{L}}{\partial b_1} = \sum_{s,t=1}^n &\mathbf{z}_{st}\mathbf{e}_{1st}+\beta\left( (\mathbf{z}_{st}  - \mathbf{z}_{s(t-1)} )\mathbf{e}_{2st} +(\mathbf{z}_{st}- \mathbf{z}_{(s-1)t} )\mathbf{e}_{3st}\right)
\end{aligned}
   $$
$$
\begin{aligned}
    \frac{\partial \mathcal{L}}{\partial W_2} = \sum_{s,t=1}^n &(\widetilde{\mathbf{z}}_{6st} \otimes \mathbf{a}_{1st} )\mathbf{e}_{1st}+\beta\big( (\widetilde{\mathbf{z}}_{6st} \otimes \mathbf{a}_{1st} - \widetilde{\mathbf{z}}_{6s(t-1)} \otimes \mathbf{a}_{1s(t-1)} )\mathbf{e}_{2st} +\\
    &(\widetilde{\mathbf{z}}_{6st} \otimes \mathbf{a}_{1st} - \widetilde{\mathbf{z}}_{6(s-1)t} \otimes \mathbf{a}_{1(s-1)t} )\mathbf{e}_{3st}\big)
\end{aligned}
$$

$$
\begin{aligned}
    \frac{\partial \mathcal{L}}{\partial b_2} = \sum_{s,t=1}^n &\widetilde{\mathbf{z}}_{6st}\mathbf{e}_{1st}+\beta\big( (\widetilde{\mathbf{z}}_{6st} - \widetilde{\mathbf{z}}_{6s(t-1)} )\mathbf{e}_{2st} +(\widetilde{\mathbf{z}}_{6st}  - \widetilde{\mathbf{z}}_{6(s-1)t}  )\mathbf{e}_{3st}\big)
\end{aligned}
$$

$$
\begin{aligned}
    \frac{\partial \mathcal{L}}{\partial W_3} = \sum_{s,t=1}^n &(\widetilde{\mathbf{z}}_{5st} \otimes \mathbf{a}_{2st} )\mathbf{e}_{1st}+\beta\big( (\widetilde{\mathbf{z}}_{5st} \otimes \mathbf{a}_{2st} - \widetilde{\mathbf{z}}_{5s(t-1)} \otimes \mathbf{a}_{2s(t-1)} )\mathbf{e}_{2st} +\\
    &(\widetilde{\mathbf{z}}_{5st} \otimes \mathbf{a}_{2st} - \widetilde{\mathbf{z}}_{5(s-1)t} \otimes \mathbf{a}_{2(s-1)t} )\mathbf{e}_{3st}\big)
\end{aligned}
$$

$$
\begin{aligned}
    \frac{\partial \mathcal{L}}{\partial b_3} = \sum_{s,t=1}^n &\widetilde{\mathbf{z}}_{5st}\mathbf{e}_{1st}+\beta\big( (\widetilde{\mathbf{z}}_{5st} - \widetilde{\mathbf{z}}_{5s(t-1)} )\mathbf{e}_{2st} +(\widetilde{\mathbf{z}}_{5st}  - \widetilde{\mathbf{z}}_{5(s-1)t}  )\mathbf{e}_{3st}\big)
\end{aligned}
$$

$$
\begin{aligned}
    \frac{\partial \mathcal{L}}{\partial W_4} = \sum_{s,t=1}^n &(\widetilde{\mathbf{z}}_{4st} \otimes \mathbf{a}_{3st} )\mathbf{e}_{1st}+\beta\big( (\widetilde{\mathbf{z}}_{4st} \otimes \mathbf{a}_{3st} - \widetilde{\mathbf{z}}_{4s(t-1)} \otimes \mathbf{a}_{3s(t-1)} )\mathbf{e}_{2st} +\\
    &(\widetilde{\mathbf{z}}_{4st} \otimes \mathbf{a}_{3st} - \widetilde{\mathbf{z}}_{4(s-1)t} \otimes \mathbf{a}_{3(s-1)t} )\mathbf{e}_{3st}\big)
\end{aligned}
$$

$$
\begin{aligned}
    \frac{\partial \mathcal{L}}{\partial b_4} = \sum_{s,t=1}^n &\widetilde{\mathbf{z}}_{4st}\mathbf{e}_{1st}+\beta\big( (\widetilde{\mathbf{z}}_{4st} - \widetilde{\mathbf{z}}_{4s(t-1)} )\mathbf{e}_{2st} +(\widetilde{\mathbf{z}}_{4st}  - \widetilde{\mathbf{z}}_{4(s-1)t}  )\mathbf{e}_{3st}\big)
\end{aligned}
$$

$$
    \frac{\partial \mathcal{L}}{\partial W_5} = \sum_{s,t=1}^n  \mathbf{a}_{4st} \mathbf{e}_{1st}+\beta\big( (\mathbf{a}_{4st} -  \mathbf{a}_{4s(t-1)} )\mathbf{e}_{2st} +(\mathbf{a}_{4st} - \mathbf{a}_{4(s-1)t} )\mathbf{e}_{3st}\big)
,\quad
    \frac{\partial \mathcal{L}}{\partial b_5} = \sum_{s,t=1}^n\mathbf{e}_{1st}
$$
}

We then consider Hessian matrix of $\mathcal{L}$.
\small{

\begin{align*}
    \frac{\partial^2 \mathcal{L}}{\partial W_1^2} =& \sum_{s, t=1}^n(\mathbf{z}_{st} \otimes \mathbf{x}_{st} )(\mathbf{z}_{st} \otimes \mathbf{x}_{st} )^{\top}&
    \\ 
    & +\beta \big(
    (\mathbf{z}_{st} \otimes \mathbf{x}_{st} - \mathbf{z}_{s(t-1)} \otimes \mathbf{x}_{s(t-1)} )(\mathbf{z}_{st} \otimes \mathbf{x}_{st} - \mathbf{z}_{s(t-1)} \otimes \mathbf{x}_{s(t-1)} )^{\top}  \\
    &+(\mathbf{z}_{st} \otimes \mathbf{x}_{st} - \mathbf{z}_{(s-1)t} \otimes \mathbf{x}_{(s-1)t} )(\mathbf{z}_{st} \otimes \mathbf{x}_{st} - \mathbf{z}_{(s-1)t} \otimes \mathbf{x}_{(s-1)t} )^{\top}
    \big)
\end{align*}

\begin{align*}
    \frac{\partial^2 \mathcal{L}}{\partial b_1\partial W_1}=& \sum_{s, t=1}^n(\mathbf{z}_{st} \otimes \mathbf{x}_{st} )\mathbf{z}_{st}^{\top} +\beta \big(
    (\mathbf{z}_{st} \otimes \mathbf{x}_{st} - \mathbf{z}_{s(t-1)} \otimes \mathbf{x}_{s(t-1)} )(\mathbf{z}_{st} - \mathbf{z}_{s(t-1)})^{\top}  \\
    &+(\mathbf{z}_{st} \otimes \mathbf{x}_{st} - \mathbf{z}_{(s-1)t} \otimes \mathbf{x}_{(s-1)t} )(\mathbf{z}_{st}  - \mathbf{z}_{(s-1)t} )^{\top}
    \big)
\end{align*}

\begin{align*}
    \frac{\partial^2 \mathcal{L}}{\partial W_2 \partial W_1} = \sum_{s, t=1}^n& \left(\widetilde{\mathbf{z}}_{6st}^{\top} \otimes 
    \mathbb{I}_{\mathbf{W}_1, \mathbf{b}_1, \mathbf{x}_{st}} \otimes \mathbf{x}_{st}\right)\mathbf{e}_{1st}+
 (\mathbf{z}_{st} \otimes \mathbf{x}_{st})(\widetilde{\mathbf{z}}_{6st}\otimes \mathbf{a}_{1st})^{\top}\\
 & +\beta \big((\widetilde{\mathbf{z}}_{6st}^{\top} \otimes 
    \mathbb{I}_{\mathbf{W}_1, \mathbf{b}_1, \mathbf{x}_{st}} \otimes \mathbf{x}_{st} - \widetilde{\mathbf{z}}_{6s(t-1)}^{\top} \otimes 
    \mathbb{I}_{\mathbf{W}_1, \mathbf{b}_1, \mathbf{x}_{s(t-1)}} \otimes \mathbf{x}_{s(t-1)})\mathbf{e}_{2st} \\
    &+
    (\mathbf{z}_{st} \otimes \mathbf{x}_{st} - \mathbf{z}_{s(t-1)} \otimes \mathbf{x}_{s(t-1)} )(\widetilde{\mathbf{z}}_{6st}\otimes \mathbf{a}_{1st} - \widetilde{\mathbf{z}}_{6s(t-1)}\otimes\mathbf{a}_{1s(t-1)})^{\top}  \\
    &+(\widetilde{\mathbf{z}}_{6st}^{\top} \otimes 
    \mathbb{I}_{\mathbf{W}_1, \mathbf{b}_1, \mathbf{x}_{st}} \otimes \mathbf{x}_{st} - \widetilde{\mathbf{z}}_{6(s-1)t}^{\top} \otimes 
    \mathbb{I}_{\mathbf{W}_1, \mathbf{b}_1, \mathbf{x}_{(s-1)t}} \otimes \mathbf{x}_{(s-1)t})\mathbf{e}_{3st} \\
    &+
    (\mathbf{z}_{st} \otimes \mathbf{x}_{st} - \mathbf{z}_{(s-1)t} \otimes \mathbf{x}_{(s-1)t} )(\widetilde{\mathbf{z}}_{6st}\otimes \mathbf{a}_{1st} - \widetilde{\mathbf{z}}_{6(s-1)t}\otimes\mathbf{a}_{1(s-1)t})^{\top}\big)
\end{align*}

\begin{align*}
    \frac{\partial^2 \mathcal{L}}{\partial b_2 \partial W_1} = \sum_{s, t=1}^n& 
 (\mathbf{z}_{st} \otimes \mathbf{x}_{st})\widetilde{\mathbf{z}}_{6st}^{\top} +\beta \big(
    (\mathbf{z}_{st} \otimes \mathbf{x}_{st} - \mathbf{z}_{s(t-1)} \otimes \mathbf{x}_{s(t-1)} )(\widetilde{\mathbf{z}}_{6st} - \widetilde{\mathbf{z}}_{6s(t-1)})^{\top}  \\
    &+
    (\mathbf{z}_{st} \otimes \mathbf{x}_{st} - \mathbf{z}_{(s-1)t} \otimes \mathbf{x}_{(s-1)t} )(\widetilde{\mathbf{z}}_{6st}- \widetilde{\mathbf{z}}_{6(s-1)t})^{\top}\big)
\end{align*}

\begin{align*}
    \frac{\partial^2 \mathcal{L}}{\partial W_3 \partial W_1} = \sum_{s, t=1}^n& \left(\widetilde{\mathbf{z}}_{5st}^{\top} \otimes 
    (\mathbf{z}_{1st}\mathbb{I}_{\mathbf{W}_2, \mathbf{b}_2, \mathbf{a}_{1st}}) \otimes \mathbf{x}_{st}\right)\mathbf{e}_{1st}+
 (\mathbf{z}_{st} \otimes \mathbf{x}_{st})(\widetilde{\mathbf{z}}_{5st}\otimes \mathbf{a}_{2st})^{\top}\\
 & +\beta \big((\widetilde{\mathbf{z}}_{5st}^{\top} \otimes 
    (\mathbf{z}_{1st}\mathbb{I}_{\mathbf{W}_2, \mathbf{b}_2, \mathbf{a}_{1st}}) \otimes \mathbf{x}_{st} - \widetilde{\mathbf{z}}_{5s(t-1)}^{\top} \otimes 
    (\mathbf{z}_{1s(t-1)}\mathbb{I}_{\mathbf{W}_2, \mathbf{b}_2, \mathbf{a}_{1s(t-1)}}) \otimes \mathbf{x}_{s(t-1)})\mathbf{e}_{2st} \\
    &+
    (\mathbf{z}_{st} \otimes \mathbf{x}_{st} - \mathbf{z}_{s(t-1)} \otimes \mathbf{x}_{s(t-1)} )(\widetilde{\mathbf{z}}_{5st}\otimes \mathbf{a}_{2st} - \widetilde{\mathbf{z}}_{5s(t-1)}\otimes\mathbf{a}_{2s(t-1)})^{\top}  \\
    &+(\widetilde{\mathbf{z}}_{5st}^{\top} \otimes 
    (\mathbf{z}_{1st}\mathbb{I}_{\mathbf{W}_2, \mathbf{b}_2, \mathbf{a}_{1st}}) \otimes \mathbf{x}_{st} - \widetilde{\mathbf{z}}_{5(s-1)t}^{\top} \otimes 
    (\mathbf{z}_{1(s-1)t}\mathbb{I}_{\mathbf{W}_2, \mathbf{b}_2, \mathbf{a}_{1(s-1)t}}) \otimes \mathbf{x}_{(s-1)t})
    \mathbf{e}_{3st} \\
    &+
    (\mathbf{z}_{st} \otimes \mathbf{x}_{st} - \mathbf{z}_{(s-1)t} \otimes \mathbf{x}_{(s-1)t} )(\widetilde{\mathbf{z}}_{5st}\otimes \mathbf{a}_{2st} - \widetilde{\mathbf{z}}_{5(s-1)t}\otimes\mathbf{a}_{2(s-1)t})^{\top}\big)
\end{align*}

\begin{align*}
    \frac{\partial^2 \mathcal{L}}{\partial b_3 \partial W_1} = \sum_{s, t=1}^n& 
 (\mathbf{z}_{st} \otimes \mathbf{x}_{st})\widetilde{\mathbf{z}}_{5st}^{\top} +\beta \big(
    (\mathbf{z}_{st} \otimes \mathbf{x}_{st} - \mathbf{z}_{s(t-1)} \otimes \mathbf{x}_{s(t-1)} )(\widetilde{\mathbf{z}}_{5st} - \widetilde{\mathbf{z}}_{5s(t-1)})^{\top}  \\
    &+
    (\mathbf{z}_{st} \otimes \mathbf{x}_{st} - \mathbf{z}_{(s-1)t} \otimes \mathbf{x}_{(s-1)t} )(\widetilde{\mathbf{z}}_{5st}- \widetilde{\mathbf{z}}_{5(s-1)t})^{\top}\big)
\end{align*}

\begin{align*}
    \frac{\partial^2 \mathcal{L}}{\partial W_4 \partial W_1} = \sum_{s, t=1}^n& \left(\mathbf{z}_{4st}^{\top} \otimes 
    (\widetilde{\mathbf{z}}_{7st}\mathbb{I}_{\mathbf{W}_3, \mathbf{b}_3, \mathbf{a}_{2st}}) \otimes \mathbf{x}_{st}\right)\mathbf{e}_{1st}+
 (\mathbf{z}_{st} \otimes \mathbf{x}_{st})(\mathbf{z}_{4st}\otimes \mathbf{a}_{3st})^{\top}\\
 & +\beta \big((\mathbf{z}_{4st}^{\top} \otimes 
    (\widetilde{\mathbf{z}}_{7st}\mathbb{I}_{\mathbf{W}_3, \mathbf{b}_3, \mathbf{a}_{2st}}) \otimes \mathbf{x}_{st} - \mathbf{z}_{4s(t-1)}^{\top} \otimes 
    (\widetilde{\mathbf{z}}_{7s(t-1)}\mathbb{I}_{\mathbf{W}_3, \mathbf{b}_3, \mathbf{a}_{2s(t-1)}}) \otimes \mathbf{x}_{s(t-1)})\mathbf{e}_{2st} \\
    &+
    (\mathbf{z}_{st} \otimes \mathbf{x}_{st} - \mathbf{z}_{s(t-1)} \otimes \mathbf{x}_{s(t-1)} )(\mathbf{z}_{4st}\otimes \mathbf{a}_{3st} - \mathbf{z}_{4s(t-1)}\otimes\mathbf{a}_{3s(t-1)})^{\top}  \\
    &+(\mathbf{z}_{4st}^{\top} \otimes 
    (\widetilde{\mathbf{z}}_{7st}\mathbb{I}_{\mathbf{W}_3, \mathbf{b}_3, \mathbf{a}_{2st}}) \otimes \mathbf{x}_{st} - \mathbf{z}_{4(s-1)t}^{\top} \otimes 
    (\widetilde{\mathbf{z}}_{7(s-1)t}\mathbb{I}_{\mathbf{W}_3, \mathbf{b}_3, \mathbf{a}_{2(s-1)t}}) \otimes \mathbf{x}_{(s-1)t})\mathbf{e}_{3st} \\
    &+
    (\mathbf{z}_{st} \otimes \mathbf{x}_{st} - \mathbf{z}_{(s-1)t} \otimes \mathbf{x}_{(s-1)t} )(\mathbf{z}_{4st}\otimes \mathbf{a}_{3st} - \mathbf{z}_{4(s-1)t}\otimes\mathbf{a}_{3(s-1)t})^{\top}\big)
\end{align*}

\begin{align*}
    \frac{\partial^2 \mathcal{L}}{\partial b_4 \partial W_1} = \sum_{s, t=1}^n& 
 (\mathbf{z}_{st} \otimes \mathbf{x}_{st})\mathbf{z}_{4st}^{\top} +\beta \big(
    (\mathbf{z}_{st} \otimes \mathbf{x}_{st} - \mathbf{z}_{s(t-1)} \otimes \mathbf{x}_{s(t-1)} )(\mathbf{z}_{4st} - \mathbf{z}_{4s(t-1)})^{\top}  \\
    &+
    (\mathbf{z}_{st} \otimes \mathbf{x}_{st} - \mathbf{z}_{(s-1)t} \otimes \mathbf{x}_{(s-1)t} )(\mathbf{z}_{4st}- \mathbf{z}_{4(s-1)t})^{\top}\big)
\end{align*}

\begin{align*}
    \frac{\partial^2 \mathcal{L}}{\partial W_5 \partial W_1} = \sum_{s, t=1}^n& \left(
    (\widetilde{\mathbf{z}}_{8st}\mathbb{I}_{\mathbf{W}_4, \mathbf{b}_4, \mathbf{a}_{3st}}) \otimes \mathbf{x}_{st}\right)\mathbf{e}_{1st}+
 (\mathbf{z}_{st} \otimes \mathbf{x}_{st}) \mathbf{a}_{4st}^{\top}\\
 & +\beta \big( ((\widetilde{\mathbf{z}}_{8st}\mathbb{I}_{\mathbf{W}_4, \mathbf{b}_4, \mathbf{a}_{3st}}) \otimes \mathbf{x}_{st} - (\widetilde{\mathbf{z}}_{8s(t-1)}\mathbb{I}_{\mathbf{W}_4, \mathbf{b}_4, \mathbf{a}_{3s(t-1)}}) \otimes \mathbf{x}_{s(t-1)})\mathbf{e}_{2st} \\
    &+
    (\mathbf{z}_{st} \otimes \mathbf{x}_{st} - \mathbf{z}_{s(t-1)} \otimes \mathbf{x}_{s(t-1)} )(\mathbf{a}_{4st} - \mathbf{a}_{4s(t-1)})^{\top}  \\
    &+((\widetilde{\mathbf{z}}_{8st}\mathbb{I}_{\mathbf{W}_4, \mathbf{b}_4, \mathbf{a}_{3st}}) \otimes \mathbf{x}_{st} - (\widetilde{\mathbf{z}}_{8(s-1)t}\mathbb{I}_{\mathbf{W}_4, \mathbf{b}_4, \mathbf{a}_{3(s-1)t}}) \otimes \mathbf{x}_{(s-1)t})\mathbf{e}_{3st} \\
    &+
    (\mathbf{z}_{st} \otimes \mathbf{x}_{st} - \mathbf{z}_{(s-1)t} \otimes \mathbf{x}_{(s-1)t} )(\mathbf{a}_{4st} - \mathbf{a}_{4(s-1)t})^{\top}\big)
\end{align*}

\begin{align*}
    \frac{\partial^2 \mathcal{L}}{\partial b_5 \partial W_1} = \sum_{s, t=1}^n& 
 \mathbf{z}_{st} \otimes \mathbf{x}_{st}
\end{align*}

\begin{align*}
    \frac{\partial^2 \mathcal{L}}{\partial b_1^2} = \sum_{s, t=1}^n\mathbf{z}_{st} \mathbf{z}_{st}^{\top} +\beta \big(
    (\mathbf{z}_{st}  - \mathbf{z}_{s(t-1)} )(\mathbf{z}_{st} - \mathbf{z}_{s(t-1)} )^{\top} +(\mathbf{z}_{st}  - \mathbf{z}_{(s-1)t} )(\mathbf{z}_{st}  - \mathbf{z}_{(s-1)t} )^{\top}
    \big)
\end{align*}

\begin{align*}
    \frac{\partial^2 \mathcal{L}}{\partial W_2 \partial b_1} = \sum_{s, t=1}^n& \left(\widetilde{\mathbf{z}}_{6st}^{\top} \otimes 
    \mathbb{I}_{\mathbf{W}_1, \mathbf{b}_1, \mathbf{x}_{st}}\right)\mathbf{e}_{1st}+
 \mathbf{z}_{st}(\widetilde{\mathbf{z}}_{6st}\otimes \mathbf{a}_{1st})^{\top}\\
 & +\beta \big((\widetilde{\mathbf{z}}_{6st}^{\top} \otimes 
    \mathbb{I}_{\mathbf{W}_1, \mathbf{b}_1, \mathbf{x}_{st}} - \widetilde{\mathbf{z}}_{6s(t-1)}^{\top} \otimes 
    \mathbb{I}_{\mathbf{W}_1, \mathbf{b}_1, \mathbf{x}_{s(t-1)}} )\mathbf{e}_{2st} \\
    &+
    (\mathbf{z}_{st}- \mathbf{z}_{s(t-1)})(\widetilde{\mathbf{z}}_{6st}\otimes \mathbf{a}_{1st} - \widetilde{\mathbf{z}}_{6s(t-1)}\otimes\mathbf{a}_{1s(t-1)})^{\top}  \\
    &+(\widetilde{\mathbf{z}}_{6st}^{\top} \otimes 
    \mathbb{I}_{\mathbf{W}_1, \mathbf{b}_1, \mathbf{x}_{st}} - \widetilde{\mathbf{z}}_{6(s-1)t}^{\top} \otimes 
    \mathbb{I}_{\mathbf{W}_1, \mathbf{b}_1, \mathbf{x}_{(s-1)t}} )\mathbf{e}_{3st} \\
    &+
    (\mathbf{z}_{st}- \mathbf{z}_{(s-1)t} )(\widetilde{\mathbf{z}}_{6st}\otimes \mathbf{a}_{1st} - \widetilde{\mathbf{z}}_{6(s-1)t}\otimes\mathbf{a}_{1(s-1)t})^{\top}\big)
\end{align*}

\begin{align*}
    \frac{\partial^2 \mathcal{L}}{\partial b_2 \partial b_1} = \sum_{s, t=1}^n& 
 \mathbf{z}_{st} \widetilde{\mathbf{z}}_{6st}^{\top} +\beta \big(
    (\mathbf{z}_{st}- \mathbf{z}_{s(t-1)})(\widetilde{\mathbf{z}}_{6st} - \widetilde{\mathbf{z}}_{6s(t-1)})^{\top} +
    (\mathbf{z}_{st}- \mathbf{z}_{(s-1)t}  )(\widetilde{\mathbf{z}}_{6st}- \widetilde{\mathbf{z}}_{6(s-1)t})^{\top}\big)
\end{align*}

\begin{align*}
    \frac{\partial^2 \mathcal{L}}{\partial W_3 \partial b_1} = \sum_{s, t=1}^n& \left(\widetilde{\mathbf{z}}_{5st}^{\top} \otimes 
    (\mathbf{z}_{1st}\mathbb{I}_{\mathbf{W}_2, \mathbf{b}_2, \mathbf{a}_{1st}})\right)\mathbf{e}_{1st}+
 \mathbf{z}_{st}(\widetilde{\mathbf{z}}_{5st}\otimes \mathbf{a}_{2st})^{\top}\\
 & +\beta \big((\widetilde{\mathbf{z}}_{5st}^{\top} \otimes 
    (\mathbf{z}_{1st}\mathbb{I}_{\mathbf{W}_2, \mathbf{b}_2, \mathbf{a}_{1st}})- \widetilde{\mathbf{z}}_{5s(t-1)}^{\top} \otimes 
    (\mathbf{z}_{1s(t-1)}\mathbb{I}_{\mathbf{W}_2, \mathbf{b}_2, \mathbf{a}_{1s(t-1)}}) )\mathbf{e}_{2st} \\
    &+
    (\mathbf{z}_{st}- \mathbf{z}_{s(t-1)}  )(\widetilde{\mathbf{z}}_{5st}\otimes \mathbf{a}_{2st} - \widetilde{\mathbf{z}}_{5s(t-1)}\otimes\mathbf{a}_{2s(t-1)})^{\top}  \\
    &+(\widetilde{\mathbf{z}}_{5st}^{\top} \otimes 
    (\mathbf{z}_{1st}\mathbb{I}_{\mathbf{W}_2, \mathbf{b}_2, \mathbf{a}_{1st}}) - \widetilde{\mathbf{z}}_{5(s-1)t}^{\top} \otimes 
    (\mathbf{z}_{1(s-1)t}\mathbb{I}_{\mathbf{W}_2, \mathbf{b}_2, \mathbf{a}_{1(s-1)t}}) )
    \mathbf{e}_{3st} \\
    &+
    (\mathbf{z}_{st}  - \mathbf{z}_{(s-1)t} )(\widetilde{\mathbf{z}}_{5st}\otimes \mathbf{a}_{2st} - \widetilde{\mathbf{z}}_{5(s-1)t}\otimes\mathbf{a}_{2(s-1)t})^{\top}\big)
\end{align*}

\begin{align*}
    \frac{\partial^2 \mathcal{L}}{\partial b_3 \partial b_1} = \sum_{s, t=1}^n& 
 \mathbf{z}_{st}\widetilde{\mathbf{z}}_{5st}^{\top} +\beta \big(
    (\mathbf{z}_{st}- \mathbf{z}_{s(t-1)} )(\widetilde{\mathbf{z}}_{5st} - \widetilde{\mathbf{z}}_{5s(t-1)})^{\top} +
    (\mathbf{z}_{st}- \mathbf{z}_{(s-1)t}  )(\widetilde{\mathbf{z}}_{5st}- \widetilde{\mathbf{z}}_{5(s-1)t})^{\top}\big)
\end{align*}

\begin{align*}
    \frac{\partial^2 \mathcal{L}}{\partial W_4 \partial b_1} = \sum_{s, t=1}^n& \left(\mathbf{z}_{4st}^{\top} \otimes 
    (\widetilde{\mathbf{z}}_{7st}\mathbb{I}_{\mathbf{W}_3, \mathbf{b}_3, \mathbf{a}_{2st}})\right)\mathbf{e}_{1st}+
 \mathbf{z}_{st}(\mathbf{z}_{4st}\otimes \mathbf{a}_{3st})^{\top}\\
 & +\beta \big((\mathbf{z}_{4st}^{\top} \otimes 
    (\widetilde{\mathbf{z}}_{7st}\mathbb{I}_{\mathbf{W}_3, \mathbf{b}_3, \mathbf{a}_{2st}}) - \mathbf{z}_{4s(t-1)}^{\top} \otimes 
    (\widetilde{\mathbf{z}}_{7s(t-1)}\mathbb{I}_{\mathbf{W}_3, \mathbf{b}_3, \mathbf{a}_{2s(t-1)}}) )\mathbf{e}_{2st} \\
    &+
    (\mathbf{z}_{st}- \mathbf{z}_{s(t-1)}  )(\mathbf{z}_{4st}\otimes \mathbf{a}_{3st} - \mathbf{z}_{4s(t-1)}\otimes\mathbf{a}_{3s(t-1)})^{\top}  \\
    &+(\mathbf{z}_{4st}^{\top} \otimes 
    (\widetilde{\mathbf{z}}_{7st}\mathbb{I}_{\mathbf{W}_3, \mathbf{b}_3, \mathbf{a}_{2st}})- \mathbf{z}_{4(s-1)t}^{\top} \otimes 
    (\widetilde{\mathbf{z}}_{7(s-1)t}\mathbb{I}_{\mathbf{W}_3, \mathbf{b}_3, \mathbf{a}_{2(s-1)t}}) )\mathbf{e}_{3st} \\
    &+
    (\mathbf{z}_{st}  - \mathbf{z}_{(s-1)t} )(\mathbf{z}_{4st}\otimes \mathbf{a}_{3st} - \mathbf{z}_{4(s-1)t}\otimes\mathbf{a}_{3(s-1)t})^{\top}\big)
\end{align*}

\begin{align*}
    \frac{\partial^2 \mathcal{L}}{\partial b_4 \partial b_1} = \sum_{s, t=1}^n& 
 \mathbf{z}_{st}\mathbf{z}_{4st}^{\top} +\beta \big(
    (\mathbf{z}_{st} - \mathbf{z}_{s(t-1)} )(\mathbf{z}_{4st} - \mathbf{z}_{4s(t-1)})^{\top} +
    (\mathbf{z}_{st} - \mathbf{z}_{(s-1)t}  )(\mathbf{z}_{4st}- \mathbf{z}_{4(s-1)t})^{\top}\big)
\end{align*}

\begin{align*}
    \frac{\partial^2 \mathcal{L}}{\partial W_5 \partial b_1}& = \sum_{s, t=1}^n \widetilde{\mathbf{z}}_{8st}\mathbb{I}_{\mathbf{W}_4, \mathbf{b}_4, \mathbf{a}_{3st}}\mathbf{e}_{1st}+
 \mathbf{z}_{st} \mathbf{a}_{4st}^{\top}\\
 & +\beta \big( ( \widetilde{\mathbf{z}}_{8st}\mathbb{I}_{\mathbf{W}_4, \mathbf{b}_4, \mathbf{a}_{3st}}  - \widetilde{\mathbf{z}}_{8s(t-1)}\mathbb{I}_{\mathbf{W}_4, \mathbf{b}_4, \mathbf{a}_{3s(t-1)}} )\mathbf{e}_{2st} +
    (\mathbf{z}_{st} - \mathbf{z}_{s(t-1)})(\mathbf{a}_{4st} - \mathbf{a}_{4s(t-1)})^{\top}  \\
    &+(\widetilde{\mathbf{z}}_{8st}\mathbb{I}_{\mathbf{W}_4, \mathbf{b}_4, \mathbf{a}_{3st}} - \widetilde{\mathbf{z}}_{8(s-1)t}\mathbb{I}_{\mathbf{W}_4, \mathbf{b}_4, \mathbf{a}_{3(s-1)t}})\mathbf{e}_{3st} +
    (\mathbf{z}_{st} - \mathbf{z}_{(s-1)t} )(\mathbf{a}_{4st} - \mathbf{a}_{4(s-1)t})^{\top}\big)
\end{align*}

\begin{align*}
    \frac{\partial^2 \mathcal{L}}{\partial b_5 \partial b_1} = \sum_{s, t=1}^n 
 \mathbf{z}_{st} 
\end{align*}

\begin{align*}
    \frac{\partial^2 \mathcal{L}}{\partial W_2^2} =& \sum_{s, t=1}^n(\widetilde{\mathbf{z}}_{6st}\otimes \mathbf{a}_{1st} )(\widetilde{\mathbf{z}}_{6st}\otimes \mathbf{a}_{1st})^{\top}&
    \\ 
    & +\beta \big(
    (\widetilde{\mathbf{z}}_{6st}\otimes \mathbf{a}_{1st} - \widetilde{\mathbf{z}}_{6s(t-1)}\otimes \mathbf{a}_{1s(t-1)})(\widetilde{\mathbf{z}}_{6st}\otimes \mathbf{a}_{1st} - \widetilde{\mathbf{z}}_{6s(t-1)}\otimes \mathbf{a}_{1s(t-1)}  )^{\top}  \\
    &+(\widetilde{\mathbf{z}}_{6st}\otimes \mathbf{a}_{1st} - \widetilde{\mathbf{z}}_{6(s-1)t}\otimes \mathbf{a}_{1(s-1)t}  )(\widetilde{\mathbf{z}}_{6st}\otimes \mathbf{a}_{1st} - \widetilde{\mathbf{z}}_{6(s-1)t}\otimes \mathbf{a}_{1(s-1)t} )^{\top}
    \big)
\end{align*}

\begin{align*}
    \frac{\partial^2 \mathcal{L}}{\partial b_2 \partial W_2} =& \sum_{s, t=1}^n(\widetilde{\mathbf{z}}_{6st}\otimes \mathbf{a}_{1st} )\widetilde{\mathbf{z}}_{6st}^{\top}
    +\beta \big(
    (\widetilde{\mathbf{z}}_{6st}\otimes \mathbf{a}_{1st} - \widetilde{\mathbf{z}}_{6s(t-1)}\otimes \mathbf{a}_{1s(t-1)})(\widetilde{\mathbf{z}}_{6st}- \widetilde{\mathbf{z}}_{6s(t-1)}  )^{\top}  \\
    &+(\widetilde{\mathbf{z}}_{6st}\otimes \mathbf{a}_{1st} - \widetilde{\mathbf{z}}_{6(s-1)t}\otimes \mathbf{a}_{1(s-1)t}  )(\widetilde{\mathbf{z}}_{6st} - \widetilde{\mathbf{z}}_{6(s-1)t} )^{\top}
    \big)
\end{align*}

\begin{align*}
    \frac{\partial^2 \mathcal{L}}{\partial W_3 \partial W_2} = \sum_{s, t=1}^n& \left(\widetilde{\mathbf{z}}_{5st}^{\top} \otimes 
    \mathbb{I}_{\mathbf{W}_2, \mathbf{b}_2, \mathbf{a}_{1st}} \otimes \mathbf{a}_{1st}\right)\mathbf{e}_{1st}+
 (\widetilde{\mathbf{z}}_{6st} \otimes \mathbf{a}_{1st})(\widetilde{\mathbf{z}}_{5st}\otimes \mathbf{a}_{2st})^{\top}\\
 & +\beta \big((\widetilde{\mathbf{z}}_{5st}^{\top} \otimes 
    \mathbb{I}_{\mathbf{W}_2, \mathbf{b}_2, \mathbf{a}_{1st}} \otimes \mathbf{a}_{1st} - \widetilde{\mathbf{z}}_{5s(t-1)}^{\top} \otimes 
    \mathbb{I}_{\mathbf{W}_2, \mathbf{b}_2, \mathbf{a}_{1s(t-1)}} \otimes \mathbf{a}_{1s(t-1)})\mathbf{e}_{2st} \\
    &+
    (\widetilde{\mathbf{z}}_{6st} \otimes \mathbf{a}_{1st} - \widetilde{\mathbf{z}}_{6s(t-1)} \otimes \mathbf{a}_{1s(t-1)} )(\widetilde{\mathbf{z}}_{5st}\otimes \mathbf{a}_{2st} - \widetilde{\mathbf{z}}_{5s(t-1)}\otimes\mathbf{a}_{2s(t-1)})^{\top}  \\
    &+(\widetilde{\mathbf{z}}_{5st}^{\top} \otimes 
    \mathbb{I}_{\mathbf{W}_2, \mathbf{b}_2, \mathbf{a}_{1st}} \otimes \mathbf{a}_{1st} - \widetilde{\mathbf{z}}_{5(s-1)t}^{\top} \otimes 
    \mathbb{I}_{\mathbf{W}_2, \mathbf{b}_2, \mathbf{a}_{1(s-1)t}} \otimes \mathbf{a}_{1(s-1)t})
    \mathbf{e}_{3st} \\
    &+
    (\widetilde{\mathbf{z}}_{6st} \otimes \mathbf{a}_{1st} - \widetilde{\mathbf{z}}_{6s(t-1)} \otimes \mathbf{a}_{1s(t-1)} )(\widetilde{\mathbf{z}}_{5st}\otimes \mathbf{a}_{2st} - \widetilde{\mathbf{z}}_{5(s-1)t}\otimes\mathbf{a}_{2(s-1)t})^{\top}\big)
\end{align*}

\begin{align*}
    \frac{\partial^2 \mathcal{L}}{\partial b_3 \partial W_2} = \sum_{s, t=1}^n& 
(\widetilde{\mathbf{z}}_{6st} \otimes \mathbf{a}_{1st})\widetilde{\mathbf{z}}_{5st}^{\top} +\beta \big((\widetilde{\mathbf{z}}_{6st} \otimes \mathbf{a}_{1st} - \widetilde{\mathbf{z}}_{6s(t-1)} \otimes \mathbf{a}_{1s(t-1)} )(\widetilde{\mathbf{z}}_{5st} - \widetilde{\mathbf{z}}_{5s(t-1)})^{\top}  \\
    &+
    (\widetilde{\mathbf{z}}_{6st} \otimes \mathbf{a}_{1st} - \widetilde{\mathbf{z}}_{6s(t-1)} \otimes \mathbf{a}_{1s(t-1)} )(\widetilde{\mathbf{z}}_{5st} - \widetilde{\mathbf{z}}_{5(s-1)t}\big)
\end{align*}

\begin{align*}
    \frac{\partial^2 \mathcal{L}}{\partial W_4 \partial W_2} = \sum_{s, t=1}^n& \left(\mathbf{z}_{4st}^{\top} \otimes 
    (\mathbf{z}_{3st}\mathbb{I}_{\mathbf{W}_3, \mathbf{b}_3, \mathbf{a}_{2st}}) \otimes \mathbf{a}_{1st}\right)\mathbf{e}_{1st}+
 (\widetilde{\mathbf{z}}_{6st} \otimes \mathbf{a}_{1st})(\mathbf{z}_{4st}\otimes \mathbf{a}_{3st})^{\top}\\
 & +\beta \big((\mathbf{z}_{4st}^{\top} \otimes 
    (\mathbf{z}_{3st}\mathbb{I}_{\mathbf{W}_3, \mathbf{b}_3, \mathbf{a}_{2st}}) \otimes \mathbf{a}_{1st}- \mathbf{z}_{4s(t-1)}^{\top} \otimes 
    (\mathbf{z}_{3s(t-1)}\mathbb{I}_{\mathbf{W}_3, \mathbf{b}_3, \mathbf{a}_{2s(t-1)}}) \otimes \mathbf{a}_{1s(t-1)})\mathbf{e}_{2st} \\
    &+
    (\widetilde{\mathbf{z}}_{6st} \otimes \mathbf{a}_{1st} - \widetilde{\mathbf{z}}_{6s(t-1)} \otimes \mathbf{a}_{1s(t-1)} )(\mathbf{z}_{4st}\otimes \mathbf{a}_{3st} - \mathbf{z}_{4s(t-1)}\otimes \mathbf{a}_{3s(t-1)})^{\top}  \\
    &+(\mathbf{z}_{4st}^{\top} \otimes 
    (\mathbf{z}_{3st}\mathbb{I}_{\mathbf{W}_3, \mathbf{b}_3, \mathbf{a}_{2st}}) \otimes \mathbf{a}_{1st} - \mathbf{z}_{4(s-1)t}^{\top} \otimes 
    (\mathbf{z}_{3(s-1)t}\mathbb{I}_{\mathbf{W}_3, \mathbf{b}_3, \mathbf{a}_{3(s-1)t}}) \otimes \mathbf{a}_{1(s-1)t})
    \mathbf{e}_{3st} \\
    &+
    (\widetilde{\mathbf{z}}_{6st} \otimes \mathbf{a}_{1st} - \widetilde{\mathbf{z}}_{6s(t-1)} \otimes \mathbf{a}_{1s(t-1)} )(\mathbf{z}_{4st}\otimes \mathbf{a}_{3st} - \mathbf{z}_{4(s-1)t}\otimes\mathbf{a}_{3(s-1)t})^{\top}\big)
\end{align*}

\begin{align*}
    \frac{\partial^2 \mathcal{L}}{\partial b_4 \partial W_2} = \sum_{s, t=1}^n&(\widetilde{\mathbf{z}}_{6st} \otimes \mathbf{a}_{1st})\mathbf{z}_{4st}^{\top} +\beta \big(
    (\widetilde{\mathbf{z}}_{6st} \otimes \mathbf{a}_{1st} - \widetilde{\mathbf{z}}_{6s(t-1)} \otimes \mathbf{a}_{1s(t-1)} )(\mathbf{z}_{4st} - \mathbf{z}_{4s(t-1)})^{\top}  \\
    &+ 
    (\widetilde{\mathbf{z}}_{6st} \otimes \mathbf{a}_{1st} - \widetilde{\mathbf{z}}_{6s(t-1)} \otimes \mathbf{a}_{1s(t-1)} )(\mathbf{z}_{4st} - \mathbf{z}_{4(s-1)t})^{\top}\big)
\end{align*}

\begin{align*}
    \frac{\partial^2 \mathcal{L}}{\partial W_5 \partial W_2} = \sum_{s, t=1}^n& \left( 
    (\widetilde{\mathbf{z}}_{9st}\mathbb{I}_{\mathbf{W}_4, \mathbf{b}_4, \mathbf{a}_{3st}}) \otimes \mathbf{a}_{1st}\right)\mathbf{e}_{1st}+
 (\widetilde{\mathbf{z}}_{6st} \otimes \mathbf{a}_{1st})\mathbf{a}_{4st}^{\top}\\
 & +\beta \big(( 
    (\widetilde{\mathbf{z}}_{9st}\mathbb{I}_{\mathbf{W}_4, \mathbf{b}_4, \mathbf{a}_{3st}}) \otimes \mathbf{a}_{1st}-  
    (\widetilde{\mathbf{z}}_{9s(t-1)}\mathbb{I}_{\mathbf{W}_3, \mathbf{b}_3, \mathbf{a}_{2s(t-1)}}) \otimes \mathbf{a}_{1s(t-1)})\mathbf{e}_{2st} \\
    &+
    (\widetilde{\mathbf{z}}_{6st} \otimes \mathbf{a}_{1st} - \widetilde{\mathbf{z}}_{6s(t-1)} \otimes \mathbf{a}_{1s(t-1)} )( \mathbf{a}_{4st} -  \mathbf{a}_{4s(t-1)})^{\top}  \\
    &+((\widetilde{\mathbf{z}}_{9st}\mathbb{I}_{\mathbf{W}_4, \mathbf{b}_4, \mathbf{a}_{3st}}) \otimes \mathbf{a}_{1st} - 
    (\widetilde{\mathbf{z}}_{9(s-1)t}\mathbb{I}_{\mathbf{W}_4, \mathbf{b}_4, \mathbf{a}_{4(s-1)t}}) \otimes \mathbf{a}_{1(s-1)t})
    \mathbf{e}_{3st} \\
    &+
    (\widetilde{\mathbf{z}}_{6st} \otimes \mathbf{a}_{1st} - \widetilde{\mathbf{z}}_{6s(t-1)} \otimes \mathbf{a}_{1s(t-1)} )( \mathbf{a}_{4st} - \mathbf{a}_{4(s-1)t})^{\top}\big)
\end{align*}

\begin{align*}
    \frac{\partial^2 \mathcal{L}}{\partial W_5 \partial W_2} = \sum_{s, t=1}^n
 \widetilde{\mathbf{z}}_{6st} \otimes \mathbf{a}_{1st}
\end{align*}

\begin{align*}
    \frac{\partial^2 \mathcal{L}}{\partial b_2^2} = \sum_{s, t=1}^n\widetilde{\mathbf{z}}_{6st} \widetilde{\mathbf{z}}_{6st}^{\top} +\beta \big(
    (\widetilde{\mathbf{z}}_{6st}  - \widetilde{\mathbf{z}}_{6s(t-1)} )(\widetilde{\mathbf{z}}_{6st} - \widetilde{\mathbf{z}}_{6s(t-1)} )^{\top} +(\widetilde{\mathbf{z}}_{6st}  - \widetilde{\mathbf{z}}_{6(s-1)t} )(\widetilde{\mathbf{z}}_{6st}  - \widetilde{\mathbf{z}}_{6(s-1)t} )^{\top}
    \big)
\end{align*}

\begin{align*}
    \frac{\partial^2 \mathcal{L}}{\partial W_3 \partial b_2} = \sum_{s, t=1}^n& \left(\widetilde{\mathbf{z}}_{5st}^{\top} \otimes 
    \mathbb{I}_{\mathbf{W}_2, \mathbf{b}_2, \mathbf{a}_{1st}} \right)\mathbf{e}_{1st}+
 \widetilde{\mathbf{z}}_{6st}(\widetilde{\mathbf{z}}_{5st}\otimes \mathbf{a}_{2st})^{\top}\\
 & +\beta \big((\widetilde{\mathbf{z}}_{5st}^{\top} \otimes 
    \mathbb{I}_{\mathbf{W}_2, \mathbf{b}_2, \mathbf{a}_{1st}}  - \widetilde{\mathbf{z}}_{5s(t-1)}^{\top} \otimes 
    \mathbb{I}_{\mathbf{W}_2, \mathbf{b}_2, \mathbf{a}_{1s(t-1)}} )\mathbf{e}_{2st} \\
    &+
    (\widetilde{\mathbf{z}}_{6st}  - \widetilde{\mathbf{z}}_{6s(t-1)} )(\widetilde{\mathbf{z}}_{5st}\otimes \mathbf{a}_{2st} - \widetilde{\mathbf{z}}_{5s(t-1)}\otimes\mathbf{a}_{2s(t-1)})^{\top}  \\
    &+(\widetilde{\mathbf{z}}_{5st}^{\top} \otimes 
    \mathbb{I}_{\mathbf{W}_2, \mathbf{b}_2, \mathbf{a}_{1st}}  - \widetilde{\mathbf{z}}_{5(s-1)t}^{\top} \otimes 
    \mathbb{I}_{\mathbf{W}_2, \mathbf{b}_2, \mathbf{a}_{1(s-1)t}} )
    \mathbf{e}_{3st} \\
    &+
    (\widetilde{\mathbf{z}}_{6st}  - \widetilde{\mathbf{z}}_{6s(t-1)} )(\widetilde{\mathbf{z}}_{5st}\otimes \mathbf{a}_{2st} - \widetilde{\mathbf{z}}_{5(s-1)t}\otimes\mathbf{a}_{2(s-1)t})^{\top}\big)
\end{align*}

\begin{align*}
    \frac{\partial^2 \mathcal{L}}{\partial b_3 \partial b_2} = \sum_{s, t=1}^n&
 \widetilde{\mathbf{z}}_{6st}\widetilde{\mathbf{z}}_{5st}^{\top}+\beta \big(
    (\widetilde{\mathbf{z}}_{6st}  - \widetilde{\mathbf{z}}_{6s(t-1)} )(\widetilde{\mathbf{z}}_{5st} - \widetilde{\mathbf{z}}_{5s(t-1)})^{\top}  \\
    &+ 
    (\widetilde{\mathbf{z}}_{6st}  - \widetilde{\mathbf{z}}_{6s(t-1)} )(\widetilde{\mathbf{z}}_{5st} - \widetilde{\mathbf{z}}_{5(s-1)t}\big)^{\top}
\end{align*}

\begin{align*}
    \frac{\partial^2 \mathcal{L}}{\partial W_4 \partial b_2} = \sum_{s, t=1}^n& \left(\mathbf{z}_{4st}^{\top} \otimes 
    (\mathbf{z}_{3st}\mathbb{I}_{\mathbf{W}_3, \mathbf{b}_3, \mathbf{a}_{2st}}) \right)\mathbf{e}_{1st}+
 \widetilde{\mathbf{z}}_{6st} (\mathbf{z}_{4st}\otimes \mathbf{a}_{3st})^{\top}\\
 & +\beta \big((\mathbf{z}_{4st}^{\top} \otimes 
    (\mathbf{z}_{3st}\mathbb{I}_{\mathbf{W}_3, \mathbf{b}_3, \mathbf{a}_{2st}}) - \mathbf{z}_{4s(t-1)}^{\top} \otimes 
    (\mathbf{z}_{3s(t-1)}\mathbb{I}_{\mathbf{W}_3, \mathbf{b}_3, \mathbf{a}_{2s(t-1)}}))\mathbf{e}_{2st} \\
    &+
    (\widetilde{\mathbf{z}}_{6st}  - \widetilde{\mathbf{z}}_{6s(t-1)}  )(\mathbf{z}_{4st}\otimes \mathbf{a}_{3st} - \mathbf{z}_{4s(t-1)}\otimes \mathbf{a}_{3s(t-1)})^{\top}  \\
    &+(\mathbf{z}_{4st}^{\top} \otimes 
    (\mathbf{z}_{3st}\mathbb{I}_{\mathbf{W}_3, \mathbf{b}_3, \mathbf{a}_{2st}})  - \mathbf{z}_{4(s-1)t}^{\top} \otimes 
    (\mathbf{z}_{3(s-1)t}\mathbb{I}_{\mathbf{W}_3, \mathbf{b}_3, \mathbf{a}_{3(s-1)t}}) )
    \mathbf{e}_{3st} \\
    &+
    (\widetilde{\mathbf{z}}_{6st}  - \widetilde{\mathbf{z}}_{6s(t-1)} )(\mathbf{z}_{4st}\otimes \mathbf{a}_{3st} - \mathbf{z}_{4(s-1)t}\otimes\mathbf{a}_{3(s-1)t})^{\top}\big)
\end{align*}

\begin{align*}
    \frac{\partial^2 \mathcal{L}}{\partial b_4 \partial b_2} = \sum_{s, t=1}^n&
 \widetilde{\mathbf{z}}_{6st} \mathbf{z}_{4st}^{\top} +\beta \big(
    (\widetilde{\mathbf{z}}_{6st}  - \widetilde{\mathbf{z}}_{6s(t-1)}  )(\mathbf{z}_{4st} - \mathbf{z}_{4s(t-1)})^{\top}  \\
    &+
    (\widetilde{\mathbf{z}}_{6st}  - \widetilde{\mathbf{z}}_{6s(t-1)} )(\mathbf{z}_{4st}- \mathbf{z}_{4(s-1)t})^{\top}\big)
\end{align*}

\begin{align*}
    \frac{\partial^2 \mathcal{L}}{\partial W_5 \partial b_2} = \sum_{s, t=1}^n& \left(\widetilde{\mathbf{z}}_{9st}\mathbb{I}_{\mathbf{W}_4, \mathbf{b}_4, \mathbf{a}_{3st}} \right)\mathbf{e}_{1st}+\widetilde{\mathbf{z}}_{6st}\mathbf{a}_{4st}^{\top}\\
 & +\beta \big((\widetilde{\mathbf{z}}_{9st}\mathbb{I}_{\mathbf{W}_4, \mathbf{b}_4, \mathbf{a}_{3st}}-  
    \widetilde{\mathbf{z}}_{9s(t-1)}\mathbb{I}_{\mathbf{W}_3, \mathbf{b}_3, \mathbf{a}_{2s(t-1)}})\mathbf{e}_{2st} \\
    &+
    (\widetilde{\mathbf{z}}_{6st}  - \widetilde{\mathbf{z}}_{6s(t-1)} )( \mathbf{a}_{4st} -  \mathbf{a}_{4s(t-1)})^{\top}  \\
    &+(\widetilde{\mathbf{z}}_{9st}\mathbb{I}_{\mathbf{W}_4, \mathbf{b}_4, \mathbf{a}_{3st}} - 
\widetilde{\mathbf{z}}_{9(s-1)t}\mathbb{I}_{\mathbf{W}_4, \mathbf{b}_4, \mathbf{a}_{4(s-1)t}})
    \mathbf{e}_{3st} \\
    &+
    (\widetilde{\mathbf{z}}_{6st} - \widetilde{\mathbf{z}}_{6s(t-1)}  )( \mathbf{a}_{4st} - \mathbf{a}_{4(s-1)t})^{\top}\big)
\end{align*}

\begin{align*}
    \frac{\partial^2 \mathcal{L}}{\partial W_5 \partial b_2} = \sum_{s, t=1}^n\widetilde{\mathbf{z}}_{6st}
\end{align*}

\begin{align*}
    \frac{\partial^2 \mathcal{L}}{\partial W_3^2} =& \sum_{s, t=1}^n(\widetilde{\mathbf{z}}_{5st}\otimes \mathbf{a}_{2st} )(\widetilde{\mathbf{z}}_{5st}\otimes \mathbf{a}_{2st})^{\top}&
    \\ 
    & +\beta \big(
    (\widetilde{\mathbf{z}}_{5st}\otimes \mathbf{a}_{2st} - \widetilde{\mathbf{z}}_{5s(t-1)}\otimes \mathbf{a}_{2s(t-1)})(\widetilde{\mathbf{z}}_{5st}\otimes \mathbf{a}_{2st} - \widetilde{\mathbf{z}}_{5s(t-1)}\otimes \mathbf{a}_{2s(t-1)}  )^{\top}  \\
    &+(\widetilde{\mathbf{z}}_{5st}\otimes \mathbf{a}_{2st} - \widetilde{\mathbf{z}}_{5(s-1)t}\otimes \mathbf{a}_{2(s-1)t}  )(\widetilde{\mathbf{z}}_{5st}\otimes \mathbf{a}_{2st} - \widetilde{\mathbf{z}}_{5(s-1)t}\otimes \mathbf{a}_{2(s-1)t} )^{\top}
    \big)
\end{align*}

\begin{align*}
    \frac{\partial^2 \mathcal{L}}{\partial b_3 \partial W_3} = \sum_{s, t=1}^n& 
(\widetilde{\mathbf{z}}_{5st} \otimes \mathbf{a}_{2st})\widetilde{\mathbf{z}}_{5st}^{\top} +\beta \big((\widetilde{\mathbf{z}}_{5st} \otimes \mathbf{a}_{2st} - \widetilde{\mathbf{z}}_{5s(t-1)} \otimes \mathbf{a}_{2s(t-1)} )(\widetilde{\mathbf{z}}_{5st} - \widetilde{\mathbf{z}}_{5s(t-1)})^{\top}  \\
    &+
    (\widetilde{\mathbf{z}}_{5st} \otimes \mathbf{a}_{2st} - \widetilde{\mathbf{z}}_{5s(t-1)} \otimes \mathbf{a}_{2s(t-1)} )(\widetilde{\mathbf{z}}_{5st} - \widetilde{\mathbf{z}}_{5(s-1)t}\big)^{\top}\big)
\end{align*}

\begin{align*}
    \frac{\partial^2 \mathcal{L}}{\partial W_4 \partial W_3} = \sum_{s, t=1}^n& \left(\mathbf{z}_{4st}^{\top}\otimes \mathbb{I}_{\mathbf{W}_3, \mathbf{b}_3, \mathbf{a}_{2st}} \otimes \mathbf{a}_{2st} \right)\mathbf{e}_{1st}+
 (\widetilde{\mathbf{z}}_{5st} \otimes \mathbf{a}_{2st}) (\mathbf{z}_{4st}\otimes \mathbf{a}_{3st})^{\top}\\
 & +\beta \big((\mathbf{z}_{4st}^{\top} \otimes 
    \mathbb{I}_{\mathbf{W}_3, \mathbf{b}_3, \mathbf{a}_{2st}}\otimes \mathbf{a}_{2st} - \mathbf{z}_{4s(t-1)}^{\top} \otimes 
\mathbb{I}_{\mathbf{W}_3, \mathbf{b}_3, \mathbf{a}_{2s(t-1)}} \otimes \mathbf{a}_{2s(t-1)})\mathbf{e}_{2st} \\
    &+
    (\widetilde{\mathbf{z}}_{5st} \otimes \mathbf{a}_{2st}  - \widetilde{\mathbf{z}}_{5s(t-1)} \otimes \mathbf{a}_{2s(t-1)}  )(\mathbf{z}_{4st}\otimes \mathbf{a}_{3st} - \mathbf{z}_{4s(t-1)}\otimes \mathbf{a}_{3s(t-1)})^{\top}  \\
    &+(\mathbf{z}_{4st}^{\top} \otimes 
    \mathbb{I}_{\mathbf{W}_3, \mathbf{b}_3, \mathbf{a}_{2st}}\otimes \mathbf{a}_{2st}  - \mathbf{z}_{4(s-1)t}^{\top} \otimes \mathbb{I}_{\mathbf{W}_3, \mathbf{b}_3, \mathbf{a}_{2(s-1)t}} \otimes \mathbf{a}_{2(s-1)t} )
    \mathbf{e}_{3st} \\
    &+
    (\widetilde{\mathbf{z}}_{5st} \otimes \mathbf{a}_{2st}  - \widetilde{\mathbf{z}}_{5(s-1)t} \otimes \mathbf{a}_{2(s-1)t} )(\mathbf{z}_{4st}\otimes \mathbf{a}_{3st} - \mathbf{z}_{4(s-1)t}\otimes\mathbf{a}_{3(s-1)t})^{\top}\big)
\end{align*}

\begin{align*}
    \frac{\partial^2 \mathcal{L}}{\partial b_4 \partial W_3} = \sum_{s, t=1}^n&(\widetilde{\mathbf{z}}_{5st} \otimes \mathbf{a}_{2st}) \mathbf{z}_{4st}^{\top}+\beta \big(
    (\widetilde{\mathbf{z}}_{5st} \otimes \mathbf{a}_{2st}  - \widetilde{\mathbf{z}}_{5s(t-1)} \otimes \mathbf{a}_{2s(t-1)}  )(\mathbf{z}_{4st}- \mathbf{z}_{4s(t-1)})^{\top}  \\
    &+(\widetilde{\mathbf{z}}_{5st} \otimes \mathbf{a}_{2st}  - \widetilde{\mathbf{z}}_{5(s-1)t} \otimes \mathbf{a}_{2(s-1)t} )(\mathbf{z}_{4st}- \mathbf{z}_{4(s-1)t})^{\top}\big)
\end{align*}

\begin{align*}
    \frac{\partial^2 \mathcal{L}}{\partial W_5 \partial W_3} = \sum_{s, t=1}^n& \left( 
    (\mathbf{z}_{3st}\mathbb{I}_{\mathbf{W}_4, \mathbf{b}_4, \mathbf{a}_{3st}}) \otimes \mathbf{a}_{2st}\right)\mathbf{e}_{1st}+
 (\widetilde{\mathbf{z}}_{5st} \otimes \mathbf{a}_{2st})\mathbf{a}_{4st}^{\top}\\
 & +\beta \big(( 
    (\mathbf{z}_{3st}\mathbb{I}_{\mathbf{W}_4, \mathbf{b}_4, \mathbf{a}_{3st}}) \otimes \mathbf{a}_{2st}-  
    (\mathbf{z}_{3s(t-1)}\mathbb{I}_{\mathbf{W}_4, \mathbf{b}_4, \mathbf{a}_{3s(t-1)}}) \otimes \mathbf{a}_{2s(t-1)})\mathbf{e}_{2st} \\
    &+
    (\widetilde{\mathbf{z}}_{5st} \otimes \mathbf{a}_{2st} - \widetilde{\mathbf{z}}_{5s(t-1)} \otimes \mathbf{a}_{2s(t-1)} )( \mathbf{a}_{4st} -  \mathbf{a}_{4s(t-1)})^{\top}  \\
    &+((\mathbf{z}_{3st}\mathbb{I}_{\mathbf{W}_4, \mathbf{b}_4, \mathbf{a}_{3st}}) \otimes \mathbf{a}_{2st} - 
    (\mathbf{z}_{3(s-1)t}\mathbb{I}_{\mathbf{W}_4, \mathbf{b}_4, \mathbf{a}_{3(s-1)t}}) \otimes \mathbf{a}_{2(s-1)t})
    \mathbf{e}_{3st} \\
    &+
    (\widetilde{\mathbf{z}}_{5st} \otimes \mathbf{a}_{2st} - \widetilde{\mathbf{z}}_{5s(t-1)} \otimes \mathbf{a}_{2s(t-1)} )( \mathbf{a}_{4st} - \mathbf{a}_{4(s-1)t})^{\top}\big)
\end{align*}

$$
    \frac{\partial^2 \mathcal{L}}{\partial b_5 \partial W_3} = \sum_{s, t=1}^n \widetilde{\mathbf{z}}_{5st} \otimes \mathbf{a}_{2st}
$$

\begin{align*}
    \frac{\partial^2 \mathcal{L}}{\partial b_3^2} = \sum_{s, t=1}^n\widetilde{\mathbf{z}}_{5st} \widetilde{\mathbf{z}}_{5st}^{\top} +\beta \big(
    (\widetilde{\mathbf{z}}_{5st}  - \widetilde{\mathbf{z}}_{5s(t-1)} )(\widetilde{\mathbf{z}}_{5st} - \widetilde{\mathbf{z}}_{5s(t-1)} )^{\top} +(\widetilde{\mathbf{z}}_{5st}  - \widetilde{\mathbf{z}}_{5(s-1)t} )(\widetilde{\mathbf{z}}_{5st}  - \widetilde{\mathbf{z}}_{5(s-1)t} )^{\top}
    \big)
\end{align*}

\begin{align*}
    \frac{\partial^2 \mathcal{L}}{\partial W_4 \partial b_3} = \sum_{s, t=1}^n& \left(\mathbf{z}_{4st}^{\top}\otimes \mathbb{I}_{\mathbf{W}_3, \mathbf{b}_3, \mathbf{a}_{2st}}\right)\mathbf{e}_{1st}+
 \widetilde{\mathbf{z}}_{5st} (\mathbf{z}_{4st}\otimes \mathbf{a}_{3st})^{\top}\\
 & +\beta \big((\mathbf{z}_{4st}^{\top} \otimes 
    \mathbb{I}_{\mathbf{W}_3, \mathbf{b}_3, \mathbf{a}_{2st}}- \mathbf{z}_{4s(t-1)}^{\top} \otimes 
\mathbb{I}_{\mathbf{W}_3, \mathbf{b}_3, \mathbf{a}_{2s(t-1)}})\mathbf{e}_{2st} \\
    &+
    (\widetilde{\mathbf{z}}_{5st}  - \widetilde{\mathbf{z}}_{5s(t-1)})(\mathbf{z}_{4st}\otimes \mathbf{a}_{3st} - \mathbf{z}_{4s(t-1)}\otimes \mathbf{a}_{3s(t-1)})^{\top}  \\
    &+(\mathbf{z}_{4st}^{\top} \otimes 
    \mathbb{I}_{\mathbf{W}_3, \mathbf{b}_3, \mathbf{a}_{2st}} - \mathbf{z}_{4(s-1)t}^{\top} \otimes \mathbb{I}_{\mathbf{W}_3, \mathbf{b}_3, \mathbf{a}_{2(s-1)t}} )
    \mathbf{e}_{3st} \\
    &+
    (\widetilde{\mathbf{z}}_{5st}   - \widetilde{\mathbf{z}}_{5(s-1)t} )(\mathbf{z}_{4st}\otimes \mathbf{a}_{3st} - \mathbf{z}_{4(s-1)t}\otimes\mathbf{a}_{3(s-1)t})^{\top}\big)
\end{align*}

\begin{align*}
    \frac{\partial^2 \mathcal{L}}{\partial b_4 \partial b_3} = \sum_{s, t=1}^n& 
 \widetilde{\mathbf{z}}_{5st} \mathbf{z}_{4st}^{\top}+\beta \big(
    (\widetilde{\mathbf{z}}_{5st}  - \widetilde{\mathbf{z}}_{5s(t-1)})(\mathbf{z}_{4st}- \mathbf{z}_{4s(t-1)})^{\top}\\
    &+
    (\widetilde{\mathbf{z}}_{5st}   - \widetilde{\mathbf{z}}_{5(s-1)t} )(\mathbf{z}_{4st}- \mathbf{z}_{4(s-1)t})^{\top}\big)
\end{align*}

\begin{align*}
    \frac{\partial^2 \mathcal{L}}{\partial W_5 \partial b_3} = \sum_{s, t=1}^n& \left(\mathbf{z}_{3st}\mathbb{I}_{\mathbf{W}_4, \mathbf{b}_4, \mathbf{a}_{3st}} \right)\mathbf{e}_{1st}+\widetilde{\mathbf{z}}_{5st}\mathbf{a}_{4st}^{\top}\\
 & +\beta \big((\mathbf{z}_{3st}\mathbb{I}_{\mathbf{W}_4, \mathbf{b}_4, \mathbf{a}_{3st}}-  
    \mathbf{z}_{3s(t-1)}\mathbb{I}_{\mathbf{W}_4, \mathbf{b}_4, \mathbf{a}_{3s(t-1)}})\mathbf{e}_{2st} \\
    &+
    (\widetilde{\mathbf{z}}_{5st}  - \widetilde{\mathbf{z}}_{5s(t-1)} )( \mathbf{a}_{4st} -  \mathbf{a}_{4s(t-1)})^{\top}  \\
    &+(\mathbf{z}_{3st}\mathbb{I}_{\mathbf{W}_4, \mathbf{b}_4, \mathbf{a}_{3st}} - 
\mathbf{z}_{3(s-1)t}\mathbb{I}_{\mathbf{W}_4, \mathbf{b}_4, \mathbf{a}_{3(s-1)t}})
    \mathbf{e}_{3st} \\
    &+
    (\widetilde{\mathbf{z}}_{5st} - \widetilde{\mathbf{z}}_{5s(t-1)}  )( \mathbf{a}_{4st} - \mathbf{a}_{4(s-1)t})^{\top}\big)
\end{align*}

$$
    \frac{\partial^2 \mathcal{L}}{\partial b_5 \partial b_3} = \sum_{s, t=1}^n \widetilde{\mathbf{z}}_{5st}
$$

\begin{align*}
    \frac{\partial^2 \mathcal{L}}{\partial W_4^2} =& \sum_{s, t=1}^n(\mathbf{z}_{4st}\otimes \mathbf{a}_{3st} )(\mathbf{z}_{4st}\otimes \mathbf{a}_{3st})^{\top}&
    \\ 
    & +\beta \big(
    (\mathbf{z}_{4st}\otimes \mathbf{a}_{3st} - \mathbf{z}_{4s(t-1)}\otimes \mathbf{a}_{3s(t-1)})(\mathbf{z}_{4st}\otimes \mathbf{a}_{3st} - \mathbf{z}_{4s(t-1)}\otimes \mathbf{a}_{3s(t-1)}  )^{\top}  \\
    &+(\mathbf{z}_{4st}\otimes \mathbf{a}_{3st} - \mathbf{z}_{4(s-1)t}\otimes \mathbf{a}_{3(s-1)t}  )(\mathbf{z}_{4st}\otimes \mathbf{a}_{3st} - \mathbf{z}_{4(s-1)t}\otimes \mathbf{a}_{3(s-1)t} )^{\top}
    \big)
\end{align*}

\begin{align*}
    \frac{\partial^2 \mathcal{L}}{\partial b_4\partial W_4} =& \sum_{s, t=1}^n(\mathbf{z}_{4st}\otimes \mathbf{a}_{3st} )\mathbf{z}_{4st}^{\top}&
    \\ 
    & +\beta \big(
    (\mathbf{z}_{4st}\otimes \mathbf{a}_{3st} - \mathbf{z}_{4s(t-1)}\otimes \mathbf{a}_{3s(t-1)})(\mathbf{z}_{4st} - \mathbf{z}_{4s(t-1)} )^{\top}  \\
    &+(\mathbf{z}_{4st}\otimes \mathbf{a}_{3st} - \mathbf{z}_{4(s-1)t}\otimes \mathbf{a}_{3(s-1)t}  )(\mathbf{z}_{4st}- \mathbf{z}_{4(s-1)t} )^{\top}
    \big)
\end{align*}

\begin{align*}
    &\frac{\partial^2 \mathcal{L}}{\partial W_5 \partial W_4} = \sum_{s, t=1}^n \left(\mathbb{I}_{\mathbf{W}_4, \mathbf{b}_4, \mathbf{a}_{3st}} \otimes \mathbf{a}_{3st}\right)\mathbf{e}_{1st}+
 (\mathbf{z}_{4st} \otimes \mathbf{a}_{3st})\mathbf{a}_{4st}^{\top}\\
 & +\beta \big((\mathbb{I}_{\mathbf{W}_4, \mathbf{b}_4, \mathbf{a}_{3st}} \otimes \mathbf{a}_{3st}-  
    \mathbb{I}_{\mathbf{W}_4, \mathbf{b}_4, \mathbf{a}_{3s(t-1)}} \otimes \mathbf{a}_{3s(t-1)})\mathbf{e}_{2st} +
    (\mathbf{z}_{4st} \otimes \mathbf{a}_{3st} - \mathbf{z}_{4s(t-1)} \otimes \mathbf{a}_{3s(t-1)} )( \mathbf{a}_{4st} -  \mathbf{a}_{4s(t-1)})^{\top}  \\
    &+(\mathbb{I}_{\mathbf{W}_4, \mathbf{b}_4, \mathbf{a}_{3st}} \otimes \mathbf{a}_{3st} - 
    \mathbb{I}_{\mathbf{W}_4, \mathbf{b}_4, \mathbf{a}_{3(s-1)t}} \otimes \mathbf{a}_{3(s-1)t})
    \mathbf{e}_{3st} +
    (\mathbf{z}_{4st} \otimes \mathbf{a}_{3st} - \mathbf{z}_{4s(t-1)} \otimes \mathbf{a}_{3s(t-1)} )( \mathbf{a}_{4st} - \mathbf{a}_{4(s-1)t})^{\top}\big)
\end{align*}

\begin{align*}
    \frac{\partial^2 \mathcal{L}}{\partial b_5 \partial W_4} = \sum_{s, t=1}^n
 \mathbf{z}_{4st} \otimes \mathbf{a}_{3st}
\end{align*}

\begin{align*}
    \frac{\partial^2 \mathcal{L}}{\partial b_4^2} = \sum_{s, t=1}^n\mathbf{z}_{4st}\mathbf{z}_{4st}^{\top} +\beta \big(
    (\mathbf{z}_{4st}- \mathbf{z}_{4s(t-1)})(\mathbf{z}_{4st} - \mathbf{z}_{4s(t-1)}  )^{\top} 
    +(\mathbf{z}_{4st}- \mathbf{z}_{4(s-1)t}  )(\mathbf{z}_{4st} - \mathbf{z}_{4(s-1)t} )^{\top}
    \big)
\end{align*}

\begin{align*}
    &\frac{\partial^2 \mathcal{L}}{\partial W_5 \partial b_4} = \sum_{s, t=1}^n\mathbb{I}_{\mathbf{W}_4, \mathbf{b}_4, \mathbf{a}_{3st}}\mathbf{e}_{1st}+
 \mathbf{z}_{4st}\mathbf{a}_{4st}^{\top}\\
 & +\beta \big((\mathbb{I}_{\mathbf{W}_4, \mathbf{b}_4, \mathbf{a}_{3st}}-  
    \mathbb{I}_{\mathbf{W}_4, \mathbf{b}_4, \mathbf{a}_{3s(t-1)}} )\mathbf{e}_{2st} +
    (\mathbf{z}_{4st}- \mathbf{z}_{4s(t-1)}  )( \mathbf{a}_{4st} -  \mathbf{a}_{4s(t-1)})^{\top}  \\
    &+(\mathbb{I}_{\mathbf{W}_4, \mathbf{b}_4, \mathbf{a}_{3st}} - 
    \mathbb{I}_{\mathbf{W}_4, \mathbf{b}_4, \mathbf{a}_{3(s-1)t}})
    \mathbf{e}_{3st} +
    (\mathbf{z}_{4st}- \mathbf{z}_{4s(t-1)}  )( \mathbf{a}_{4st} - \mathbf{a}_{4(s-1)t})^{\top}\big)
\end{align*}

$$
\frac{\partial^2 \mathcal{L}}{\partial b_5 \partial b_4} = \sum_{s, t=1}^n \mathbf{z}_{4st}
$$

\begin{align*}
    \frac{\partial^2 \mathcal{L}}{\partial W_5^2} =& \sum_{s, t=1}^n\mathbf{a}_{4st}\mathbf{a}_{4st}^{\top} +\beta \big(
    ( \mathbf{a}_{4st} -  \mathbf{a}_{4s(t-1)})( \mathbf{a}_{4st} - \mathbf{a}_{4s(t-1)}  )^{\top}+( \mathbf{a}_{4st} -  \mathbf{a}_{4(s-1)t}  )( \mathbf{a}_{4st} - \mathbf{a}_{4(s-1)t} )^{\top}
    \big)
\end{align*}

$$
\frac{\partial^2 \mathcal{L}}{ \partial b_5 \partial W_5} = \sum_{s, t=1}^n\mathbf{a}_{4st}
, \quad
\frac{\partial^2 \mathcal{L}}{ \partial b_5^2} = n^2.
$$

}

\bibliographystyle{plainnat} 
\bibliography{ref2025} %